%% file: neurips_2024.tex
\documentclass{article}

\input{math_commands.tex}

\usepackage{xspace}
\usepackage{amsfonts}
\usepackage[table, svgnames, dvipsnames]{xcolor}
\usepackage{enumitem}
\usepackage{mathrsfs}
\usepackage{microtype}
\usepackage{amssymb}
\usepackage{booktabs}
\usepackage{multirow}
\usepackage[textsize=tiny]{todonotes}
\usepackage{tcolorbox}
\usepackage{soul}
\usepackage{wrapfig}
\usepackage{tikz}
\usepackage{minitoc}
\usepackage{makecell}
\usepackage{cellspace}
\usepackage{graphicx}
\usepackage{bbm}
\usepackage{bbding}
\usepackage{caption}
\usepackage{subcaption}
\usepackage{placeins}

\usepackage{algorithm}
\usepackage{algpseudocode}

\newcommand{\ftheta}{f_{\theta}}

\newcommand{\LogSumExp}{\mathrm{{LogSumExp}}}

\PassOptionsToPackage{numbers, compress}{natbib}

\usepackage[final]{neurips_2024}

\usepackage[utf8]{inputenc} 
\usepackage[T1]{fontenc}    
\usepackage{hyperref}       
\hypersetup{
    colorlinks = true,
    linkcolor = blue,
    anchorcolor = blue,
    citecolor = blue,
    filecolor = blue,
    urlcolor = blue
}
\usepackage{url}            
\usepackage{booktabs}       
\usepackage{nicefrac}       
\usepackage{microtype}      
\usepackage[capitalize,noabbrev]{cleveref}

\newcommand{\titlecontent}{TabEBM: A Tabular Data Augmentation Method with Distinct Class-Specific Energy-Based Models\xspace}

\title{\titlecontent}

\author{%
  Andrei Margeloiu\textsuperscript{1}\thanks{Equal contribution.}\space,\space Xiangjian Jiang\textsuperscript{1}$^*$,\space Nikola Simidjievski\textsuperscript{2,1},\space Mateja Jamnik\textsuperscript{1} \\
  \textsuperscript{1}Department of Computer Science and Technology, University of Cambridge, UK \\
  \textsuperscript{2}PBCI, Department of Oncology, University of Cambridge, UK \\
  \texttt{\{am2770, xj265, ns779, mj201\}@cam.ac.uk} \\
}

\begin{document}
\doparttoc
\faketableofcontents

\maketitle

\begin{abstract}
    \looseness-1
    Data collection is often difficult in critical fields such as medicine, physics, and chemistry, yielding typically only small tabular datasets. However, classification methods tend to struggle with these small datasets, leading to poor predictive performance. Increasing the training set with additional synthetic data, similar to data augmentation in images, is commonly believed to improve downstream tabular classification performance. However, current tabular generative methods that learn either the joint distribution $ p(\mathbf{x}, y) $ or the class-conditional distribution $ p(\mathbf{x} \mid y) $ often overfit on small datasets, resulting in poor-quality synthetic data, usually worsening classification performance compared to using real data alone. To solve these challenges, we introduce TabEBM, a novel class-conditional generative method using Energy-Based Models (EBMs). Unlike existing tabular methods that use a shared model to approximate all class-conditional densities, our key innovation is to create distinct EBM generative models for each class, each modelling its class-specific data distribution individually. This approach creates robust energy landscapes, even in ambiguous class distributions. Our experiments show that TabEBM generates synthetic data with higher quality and better statistical fidelity than existing methods. When used for data augmentation, our synthetic data consistently leads to improved classification performance across diverse datasets of various sizes, especially small ones. Code is available at~\url{https://github.com/andreimargeloiu/TabEBM}.
\end{abstract}

\section{Introduction}
\label{sec:intro}

\begin{wrapfigure}[]{r}{0.5\textwidth}
\begin{center}
    \vspace{-23pt}
    \centering
    \includegraphics[width = \linewidth]{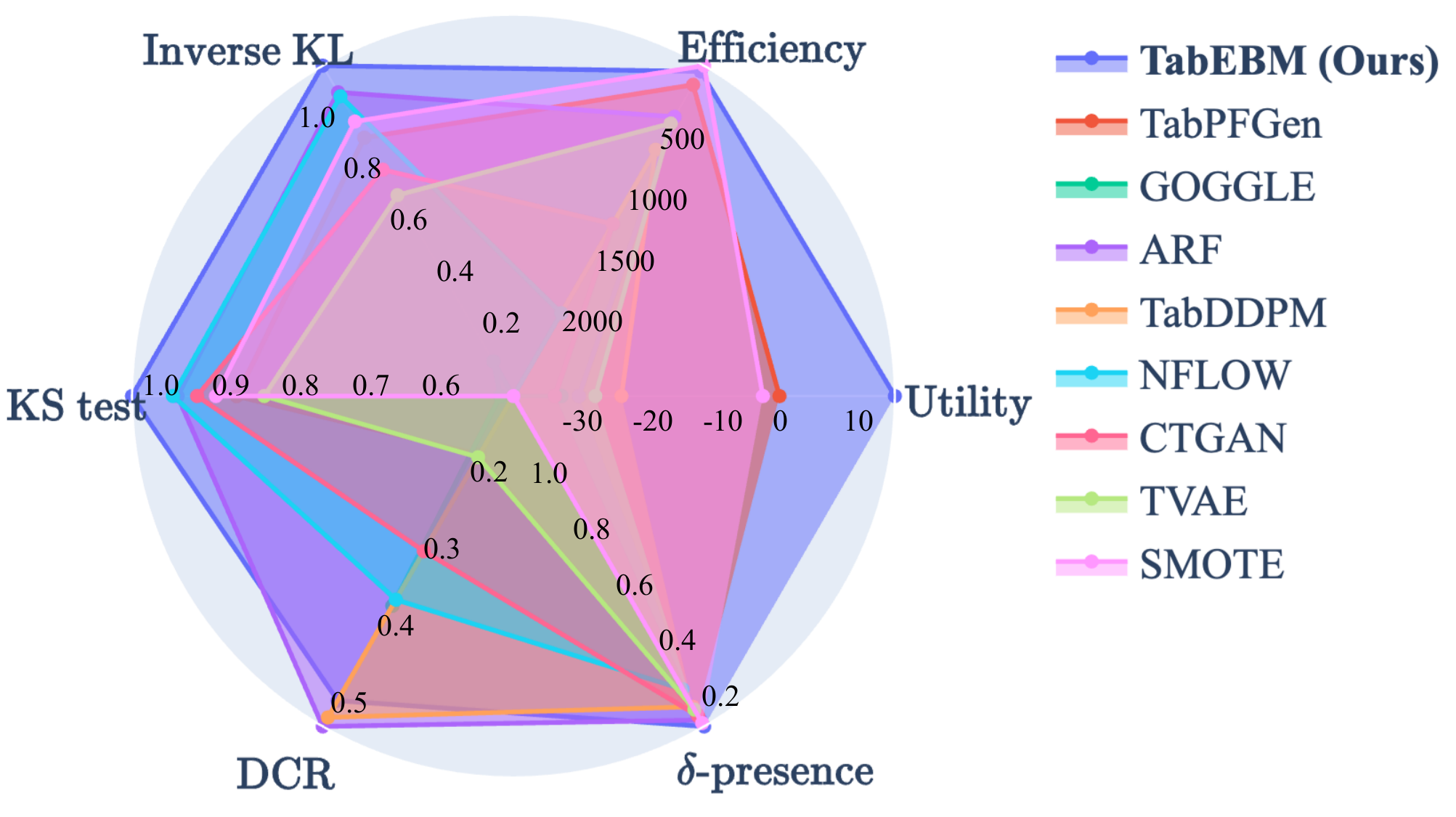}
    \caption{\textbf{Evaluation of TabEBM and other state-of-the-art tabular generative methods across six key metrics} (larger area indicates better performance). The results demonstrate that TabEBM excels in data augmentation (utility), with a larger area than all other methods.}
    \label{fig:radar}
    \vspace{-40pt}
\end{center}
\end{wrapfigure}    


Many scientific domains within medicine, physics, and chemistry often rely on intricate and challenging data acquisition procedures~\citep{baxevanis2020bioinformatics, malin2013biomedical, bansal2022systematic, hernandez2022synthetic, sufi2024addressing, chang2022towards} that typically render small-size tabular datasets~\citep{baxevanis2020bioinformatics, levin2022transfer}. Using these to train machine learning models that can aid in tasks such as disease diagnosis~\citep{margeloiu2022weight, jiang2024protogate}, material property prediction~\citep{jha2019enhancing}, and chemical compound classification~\citep{cai2020transfer}, can lead to poor performance~\citep{shwartz2022tabular, margeloiu2022weight, jiang2024protogate}. In the case of learning tasks which leverage image and text data, a standard remedy to address performance issues due to data scarcity is employing data augmentation techniques~\citep{shorten2019survey, shorten2021text, mumuni2022data, seedat2024curated} that generate additional synthetic samples from existing data. 
However, applying data augmentation to tabular data introduces additional challenges, as tabular datasets are often very diverse and lack explicit symmetries~\citep{borisov2022deep}, such as rotations or translations seen in images. 
Consequently, existing tabular data augmentation methods often yield mixed results and can even degrade model performance~\citep{manousakas2023usefulness, seedat2024curated, TabPFGen}, hindering their widespread adoption. 

\looseness-1
Tabular augmentation typically involves training generative models to approximate either the joint distribution $p(\rvx, y)$~\citep{xu2019modeling, durkan2019neural} or the class-conditional distribution $p(\rvx|y)$~\citep{xu2019modeling, kotelnikov2023tabddpm, watson2023adversarial, liu2023goggle, TabPFGen}. A key challenge of joint distribution methods is maintaining the original training label distribution, as sampling from such generators can produce label distributions that deviate from the original and even fail to generate data for specific classes (see \cref{appendix:limitations-existing-work} for an example). These issues compromise the effectiveness of data augmentation~\citep{manousakas2023usefulness} by undermining the label accuracy and distribution. On the other hand, while class-conditional models that learn $p(\rvx|y)$ preserve the stratification of the original data, they often employ a \textit{shared} model to represent all class-conditional densities. This, however, can lead to overfitting, particularly in imbalanced datasets where the model may prioritise more frequent classes~\citep{douzas2018effective}, ignoring unique features needed for generating label-invariant samples. Additionally, in datasets with limited data, this can lead to mode collapse~\citep{qin2023class, sampath2021survey}, where the model does not effectively capture the diversity of each class~\citep{sampath2021survey}, and thus tends to perform poorly in a multi-class setting.

\begin{figure}[t!]
    \centering
    \includegraphics[width=1\textwidth]{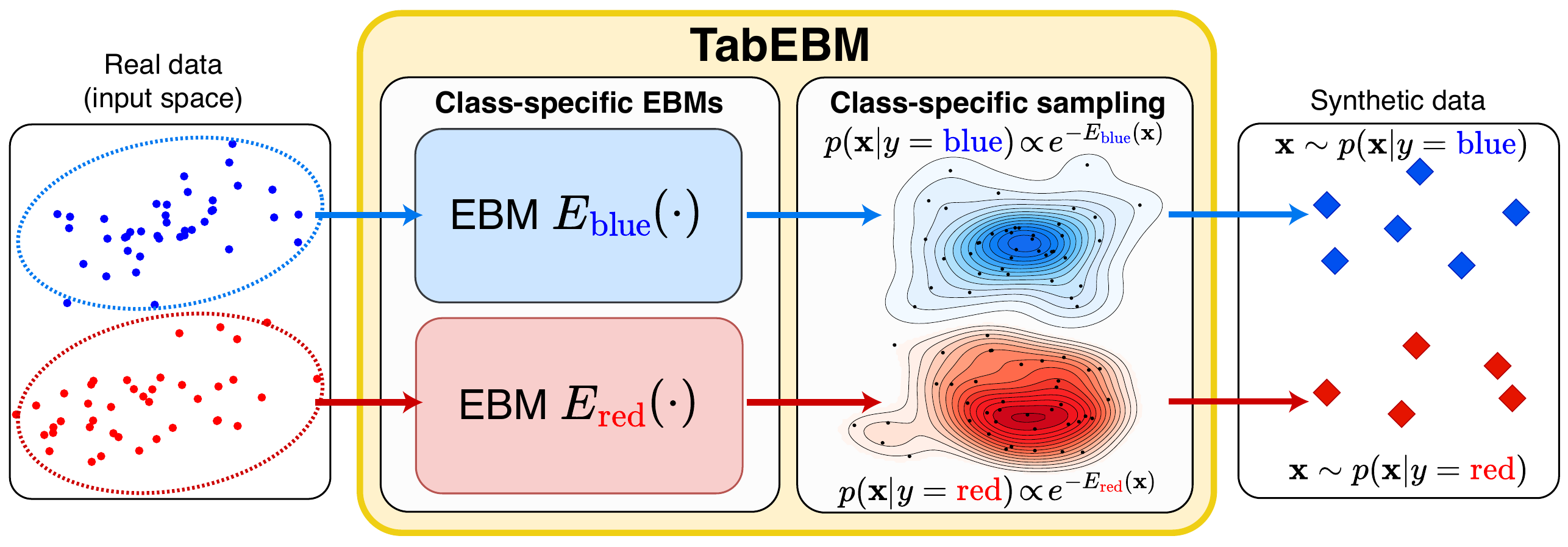}
    \caption{\textbf{An overview of 
    TabEBM}. We learn \textit{distinct} class-specific Energy-Based Models (EBMs) \(E_{\text{\textcolor{blue}{blue}}}(\rvx)\) and \(E_{\textcolor{red}{\text{red}}}(\rvx)\) exclusively on the points of their respective class. Each EBM approximates a class-conditional distribution \(p(\rvx | y)\). TabEBM allows synthetic data generation by sampling from the estimated distributions for each class \(p(\rvx | y=\textcolor{blue}{\text{blue}})\) and \(p(\rvx | y=\textcolor{red}{\text{red}})\).}
    \label{fig:tabebm}
    \vspace{-10pt}
\end{figure}

\looseness-1
To address the challenges of class-conditional tabular generation, we introduce TabEBM (Figure~\ref{fig:tabebm}), a new method for tabular data augmentation utilising Energy-Based Models (EBMs).
Our method introduces two innovations:
(i)~\textit{Distinct class-specific models:} TabEBM constructs a collection of individual models -- one for each class -- which, by design, enables learning distinct marginal distributions for the inputs associated with each class. This, in turn, enables performing data augmentation while maintaining the original label distribution.
(ii)~\textit{Generative models:} we build novel class-specific generators that produce high-quality synthetic data even from extremely few samples. Specifically, we create a surrogate binary classification task for each class and fit it with a pre-trained tabular in-context classifier. We then convert the binary classifier into an EBM, a generative model, without additional training. Using class-specific EBMs makes the energy landscape more robust to class overlaps, compared to using a single shared EBM to approximate the class-conditional distribution.

Our contributions can be summarised as:

\begin{itemize}[topsep=0pt, leftmargin=7pt, itemsep=0pt]
    \item \textbf{Technical:} We propose TabEBM, which is the first generative method to create class-specific EBMs, learning the marginal distribution for each class separately.

    \item \textbf{Empirical:} We present the first comprehensive analysis of tabular data augmentation across different dataset sizes and use cases beyond predictive performance. Our analysis compares TabEBM with eight leading tabular generative models across various datasets, demonstrating that TabEBM consistently improves data augmentation performance on small datasets, while our generated data demonstrates better statistical fidelity and privacy-preserving properties (Figure~\ref{fig:radar}).
    
    \item \textbf{Library:} We release TabEBM as an open-source library, available at~\url{https://github.com/andreimargeloiu/TabEBM}. Our library enables off-the-shelf data generation and data augmentation on any tabular dataset without requiring training. Further details are available in \cref{appendix:tabebm-library}.
\end{itemize}

\section{TabEBM}
\label{sec:method}

\textbf{Notation.} We address classification problems with $C$ classes, denoted by $\gY = \{ 1, 2, \ldots, C\}$. Let $\{(\rvx^{(i)}, y_i) \}_{i=1}^N$ represent a dataset of $N$ samples, each being a $D$-dimensional vector $\rvx^{(i)} \in \sR^D$, with a corresponding label $y_i \in \gY$. For each class $c \in \gY$, we define $\gX_c = \{\rvx^{(i)} \mid y_i = c \}$ as the subset of samples labelled with class $c$. Let $\ftheta(\cdot)$ denote a classifier. The expression $\ftheta(\rvx)[y]$ represents the (unnormalised) logit assigned to the class $y$ for the input $\rvx$.

\subsection{Preliminaries on Energy-Based Models}
An Energy-Based Model (EBM)~\citep{EBM2006LeCun} defines a probability density function $p_{\theta}(\rvx)$ through an energy function $E(\rvx)$. Specifically, the model posits that $p(\rvx) \propto e^{-E(\rvx)}$, where $E(\rvx)$ represents the unnormalised negative log-density of the input $\rvx$. In this framework, lower energy values correspond to higher probability densities. This relationship allows EBMs to model distributions by learning to assign lower energy to more probable configurations of $\rvx$ and higher energy to less probable ones.

An important observation is that energy-based models can utilise the same model architectures as standard classification models~\citep{JEM}. Typically, the logits $\ftheta(\rvx)[y]$ from a classification model define a discriminative distribution through the softmax function, expressed as $p_\theta(y|\rvx) = \mathrm{softmax}(f_\theta(\rvx)[y])$. Intriguingly, these same logits can be reinterpreted to define an energy-based model for the joint distribution \( p(\rvx, y) \). This is achieved by setting the energy function to $E(\rvx, y) = -f_\theta(\rvx)$. Furthermore, the energy function for the marginal distribution $p(\rvx)$ is obtained by marginalising over $p(\rvx, y)$, resulting in $E(\rvx) = -\LogSumExp_{y'} f_\theta(\rvx)[y']$.

\looseness-1
Such an energy-based model, trained with EBM-specific protocols on multiple classes, is typically used as a classifier, as demonstrated on several computer vision tasks in~\citep{JEM}. In contrast, in this work our focus is the opposite: we propose employing trained classifiers, one for each specific class, as a generative energy-based model for the class-conditional distributions $p(\rvx | y)$. We apply our TabEBM method for generative tasks on tabular data.

\subsection{Distinct Class-Specific Energy-Based Models}
\label{sec:tabebm-energy-derivation}

TabEBM is a class-conditional generative model \( p(\rvx | y) \) implemented using a set of EBMs, \(\{E_1(\rvx), E_2(\rvx), \ldots, E_C(\rvx)\}\). Our approach assumes that the class-conditional density \( p(\rvx | y=c) \) is best modelled using its class-specific data \(\mathcal{X}_c\). Thus, for each class \(c\), we construct a class-specific EBM, \(E_c(\rvx)\), using only the data from that class, \(\mathcal{X}_c\), such that \( p(\rvx | y=c) \propto \exp(-E_c(\rvx)) \).

\begin{wrapfigure}[]{r}{0.30\textwidth}
    \centering
    \vspace{-13pt}
    \includegraphics[clip,trim=5pt 5pt 5pt 5pt, width=0.92\linewidth]{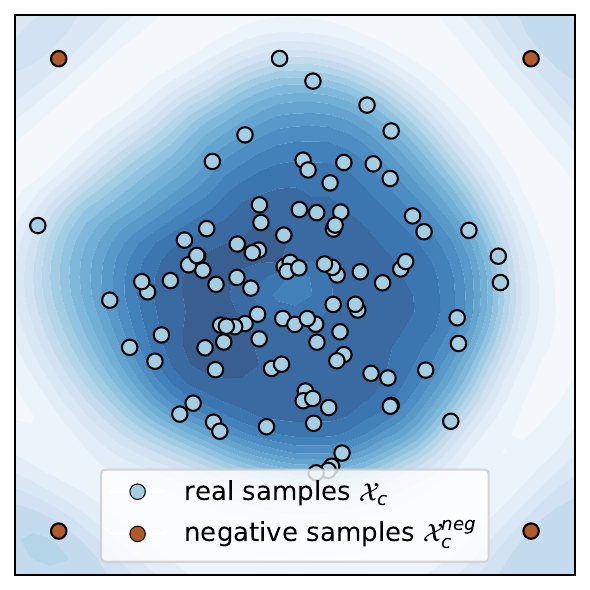}
    \vspace{-2pt}    
    \caption{The class-specific energy function $E_c(\rvx)$ from the surrogate binary task, where the blue region represents low energy (i.e., high data density). Placing the negative samples in a hypercube distant from the data results in an accurate energy function.}

    \label{fig:tabebm-energy-surface-negative-samples-corners}
    
    \vspace{-5pt}
\end{wrapfigure}

We derive each class-specific EBM \(E_c(\rvx)\) by training a classifier on a novel task and reinterpreting its logits. Specifically, for each class~\(c\), we propose a \textit{surrogate binary classification task} to determine if a sample belongs to class $c$ by comparing $\mathcal{X}_c$ against a set of surrogate negative samples $\mathcal{X}_c^{\text{neg}}$, which we show in \cref{fig:tabebm-energy-surface-negative-samples-corners}. Specifically, we generate the negative samples at the corners of a hypercube in~$R^D$. For each dimension $d$, the coordinates of a negative sample are either $\alpha^{\text{neg}}_{\text{dist}} \sigma_d$ or $-\alpha^{\text{neg}}_{\text{dist}} \sigma_d$, where $\alpha^{\text{neg}}_{\text{dist}}$ is a fixed constant and $\sigma_d$ is the standard deviation of dimension $d$. For example, in $R^3$, a negative sample might have coordinates $[\alpha^{\text{neg}}_{\text{dist}} \sigma_1, \alpha^{\text{neg}}_{\text{dist}} \sigma_2, -\alpha^{\text{neg}}_{\text{dist}} \sigma_3]$. Placing the negative samples at the corners of a hypercube ensures they are easily distinguishable from the real data, which is crucial for an accurate energy function (see \cref{appendix:ablation-placing-negative-samples}). This placement is also robust to variations in the number and distance of the negative samples (see \cref{appendix:varying-number-negative-samples,appendix:varying-distance-negative-samples}).

We create the combined dataset \(\mathcal{D}_c\) for the surrogate binary classification task by labelling \(\mathcal{X}_c\) as 1 and \(\mathcal{X}_c^{\text{neg}}\) as 0:

{
    \setlength{\abovedisplayskip}{-5pt}
    \begin{align}
    \mathcal{D}_c = (\mathcal{X}_c \cup \mathcal{X}_c^{\text{neg}}, \{1\}^{|\mathcal{X}_c|} \cup \{0\}^{|\mathcal{X}_c^{\text{neg}}|})
    \end{align}
}

We then train a binary classifier \(f^c_{\theta}(\cdot)\) on $\mathcal{D}_c$ and use it to construct the class-specific energy \(E_c(\rvx)\) for class \(c\). To do this, we reinterpret the logits $\{\ftheta^c(\rvx)[0], \ftheta^c(\rvx)[1]\}$ of the trained binary classifier as components of an approximated joint distribution for the surrogate binary task:


\begin{align}
    \label{main:TabEBM-joint}
    p_c(\rvx, 0) = \frac{\exp(\ftheta^c(\rvx)[0])}{Z}, \quad p_c(\rvx, 1) = \frac{\exp(\ftheta^c(\rvx)[1])}{Z} \quad \text{\small{($Z$ is the normalisation constant)}}
\end{align}

Next, we derive the approximated distribution $p_c(\rvx)$ by marginalisation:
{
\setlength{\abovedisplayskip}{-5pt}
\setlength{\abovedisplayshortskip}{-5pt}

\begin{align}
    p_c(\rvx) &= p_c(\rvx, 0) + p_c(\rvx, 1) && \nonumber\\ 
        &= \frac{\exp(\ftheta^c(\rvx)[0]) + \exp(\ftheta^c(\rvx)[1])}{Z} \nonumber\\ 
        &= \frac{\exp \left( \log \left( \exp(\ftheta^c(\rvx)[0]) + \exp(\ftheta^c(\rvx)[1]) \right) \right)}{Z} \nonumber\\ 
    \label{eq:class-specific-energy}
    \rightarrow E_c(\rvx) &= -\log \left( \exp(\ftheta^c(\rvx)[0]) + \exp(\ftheta^c(\rvx)[1]) \right) && \text{\small{(TabEBM class-specific energy)}}
\end{align}
}

\looseness-1
For the binary classifier \(f^c_{\theta}(\cdot)\) in the surrogate binary classification, we use TabPFN~\citep{TabPFN}, a pre-trained tabular in-context model. Note that TabPFN is intended for inference only, with no updates to its parameters (see \cref{sec:related-work} for more details about TabPFN). In this context, ``training'' the TabPFN classifier is analogous to the \mbox{K-Nearest Neighbour} algorithm, which simply performs inference based on a training dataset provided to the model. We apply TabPFN multiple times on separate datasets \( \{ \mathcal{D}_1, \mathcal{D}_2, \ldots, \mathcal{D}_C \}\) to obtain multiple classifiers $\{ \ftheta^1, \ftheta^2, \ldots, \ftheta^C \}$. In \cref{sec:why-is-tabebm-effective-for-energy-estimation}, we explore why reinterpreting TabPFN's logits, trained on our surrogate binary tasks, can be useful for estimating an energy function. We emphasise that TabEBM is a general method, capable of using any gradient-based classifier that computes logits (using \cref{eq:class-specific-energy}), and is not limited to TabPFN.

\looseness-1
\textbf{Generating data with TabEBM} involves two steps. First, we sample a class $c$ from the empirical distribution $c \sim p(y)$. Then, we sample a data point $\rvx$ from the conditional distribution $\rvx \sim p(\rvx | y=c)$ approximated by the class-specific energy-based model $E_c(\rvx)$, as outlined in~\cref{algorithm:tabebm}. We employ Stochastic Gradient Langevin Dynamics (SGLD)~\citep{SGLD} to perform this sampling. SGLD is an efficient method for high-dimensional data, combining stochastic gradient descent (SGLD) with Langevin dynamics. The update rule for SGLD at each iteration is:
\begin{align}
    \rvx_{t+1} = \rvx_t - \frac{\eta}{2} \nabla E(\rvx_t) + \epsilon_t, \quad \epsilon_t \sim \mathcal{N}(0, \eta \mathbf{I})
\end{align}
where a Gaussian noise term $\epsilon_t$ introduces randomness into the sampling process, enhancing the exploration of the distribution. In practice, the step size and the noise standard deviations are often chosen separately, resulting in a biased sampler that allows for faster training.
\cref{appendix:ablation_sgld} further shows that TabEBM is stable to hyperparameters for the sampling process.

\looseness-1
In our method, SGLD performs iterative augmentation. We start by sampling close to real data and iteratively adjust these synthetic data points, steering them towards regions of higher probability under the learned energy model. TabEBM enables sampling from any specified class distribution, including the original class distribution, which is crucial for data augmentation.

\begin{algorithm}[!h]
\caption{TabEBM sampling from Class-Specific EBM $E_c(\rvx)$}
\label{algorithm:tabebm}
\small
\textbf{Input:} Training data $\mathcal{X}_c$ for class $c$, step size $\alpha_{\text{step}}$, noise scale $\alpha_{\text{noise}}$, initial perturbation $\sigma_{\text{start}}$, number of steps $T$ \\
\textbf{Output:} Set of synthetic samples for class $c$

\begin{algorithmic}[1]  
    \vspace{2pt}
    \Statex \textit{Initialise a surrogate binary classification task and train the model}
    \vspace{2pt}
    \State Assign new labels to the samples $\mathcal{X}_c$ from class $c$, setting them to class 1
    \State Generate a set of surrogate negative samples $\mathcal{X}_c^{\text{neg}}$ and assign them class 0 labels
    \State Train a binary classifier $\ftheta^c$ on the dataset $\mathcal{D}_c = (\mathcal{X}_c \cup \mathcal{X}_c^{\text{neg}}, \{1\}^{|\mathcal{X}_c|} \cup \{0\}^{|\mathcal{X}_c^{\text{neg}}|})$
    \vspace{2pt}
    \Statex \textit{Synthesise samples using Stochastic Gradient Langevin Dynamics (SGLD)}
    \vspace{2pt}
    \State Initialise synthetic data points $\rvx_0^{\text{synth}}$ by sampling from $\mathcal{N}(\mathcal{X}_c, \sigma_{\text{start}}^2 \mathbf{I})$

    \For{each iteration $t = 0, 1, \dots, T-1$}
        \State $E_c(\rvx_t^{\text{synth}}) = -\log\left( \exp(f_{\theta}^c(\rvx_t^{\text{synth}})[0]) + \exp(f_{\theta}^c(\rvx_t^{\text{synth}})[1]) \right)$
        \State $ \rvx_{t+1}^{\text{synth}} = \rvx_{t}^{\text{synth}} - \alpha_{\text{step}} \nabla E_c(\rvx_t^{\text{synth}}) + \mathcal{N}(0, \alpha_{\text{noise}}^2 \mathbf{I})$
    \EndFor
    \State \Return $\rvx_T^{\text{synth}}$ as the generated synthetic data for class $c$
\end{algorithmic}
\end{algorithm}

\section{Experiments}
\label{sec:exp}

We evaluate TabEBM by focusing on four research questions:

\begin{itemize}[topsep=0pt, leftmargin=7pt, itemsep=-2pt]
    \item \textbf{Data Augmentation Improvement (Q1, \cref{sec:exp_utility}):} Can TabEBM generate synthetic data that improves the accuracy of downstream predictors via data augmentation?

    \item \textbf{Statistical Fidelity (Q2, \cref{sec:exp_fidelity}):} Can TabEBM generate synthetic data with high statistical fidelity (i.e., with similar distributions to those of real data)?

    \item \textbf{Privacy Preservation (Q3, \cref{sec:exp_privacy}):} Can TabEBM generate synthetic data that finds a competitive trade-off between downstream performance and privacy preservation?

    \item \textbf{Understanding TabEBM’s energy formulation (Q4, \cref{sec:why-is-tabebm-effective-for-energy-estimation}):} Why is TabEBM’s class-specific energy effective, and how do the proposed surrogate tasks influence this?
\end{itemize}

\looseness-1
\textbf{Datasets.} 
We utilise eight open-source tabular datasets from OpenML \cite{OpenML} across five domains: Medicine, Chemistry, Engineering, Language and Economics. 
As TabPFN utilises many small-size OpenML datasets in its meta-validation~\cite{TabPFN}, it can lead to data leakage when evaluating TabEBM. Therefore, to provide fair comparisons, we select six additional leakage-free datasets from UCI~\cite{dua2019uci}.
These diverse datasets contain 7 to 77 features and 698 to 5500 samples across 2 to 26 classes. Five datasets contain both numerical and categorical features, while the remaining are numerical only. We further enlarge the evaluation scope by varying the degrees of data availability (i.e., $N_{\text{real}}$), leading up to 33 different test cases for the eight OpenML datasets.
\cref{appendix:datasets} provides detailed descriptions.

\textbf{Benchmark generators.}
We compare TabEBM against eight existing tabular data generation methods of eight different categories: 
(i)~a standard interpolation method SMOTE~\cite{chawla2002smote}; 
(ii)~a Variational Autoencoders (VAE) based method TVAE~\cite{xu2019modeling}; 
(iii)~a Generative Adversarial Networks (GAN) method CTGAN~\cite{xu2019modeling}; 
(iv)~a normalising flow model Neural Spine Flows (NFLOW)~\cite{durkan2019neural}; 
(v)~a diffusion model TabDDPM~\cite{kotelnikov2023tabddpm}; 
(vi)~a tree-based method Adversarial Random Forests (ARF)~\cite{watson2023adversarial}; 
(vii)~a Graph Neural Network (GNN) based method GOGGLE~\cite{liu2023goggle}; and 
(viii)~a Prior-Data Fitted Networks (PFN) based method TabPFGen~\cite{TabPFGen}. 
Furthermore, we also include a ``Baseline'' model, where no data augmentation is applied (i.e., only real data is used to train downstream predictors). In \cref{appendix:implementation-generators}, we detail the settings used for TabEBM and all other generators.

\textbf{Downstream predictors.}
We select six representative downstream predictors, including three standard baselines: Logistic Regression (LR)~\cite{cox1958regression}, KNN~\cite{fix1985discriminatory} and MLP~\cite{gorishniy2021revisiting}; two tree-based methods: Random Forest (RF)~\cite{breiman2001random} and XGBoost~\cite{chen2016xgboost}; and a PFN method: TabPFN~\cite{TabPFN}.

\looseness-1
\textbf{General experimental setup.}
For each dataset of $N$ samples, we first split it into stratified train and test sets. We create large test sets to reduce the likelihood that the model's performance is accidentally inflated due to a small, unrepresentative set of samples~\cite{raschka2018model}, and thus the test size is computed via $N_{\text{test}}=\min \left( \frac{N}{2}, 500 \right)$. 
The full train set approximates the upper bound of the quality of synthetic data, and we call this set ``oracle''. We subsample the full train set to simulate different levels of data availability, thus the subset size $N_{\text{real}}$ varies over $\{20, 50, 100, 200, 500\}$. We split each subset into stratified training and validation sets with a ratio of 4:1. 
We provide detailed descriptions of data splitting in \cref{appendix:data-splitting} and preprocessing in \cref{appendix:data-preprocessing}.
We repeat the splitting ten times, summing up to 10 runs per subset size. The reported results are averaged by default over ten runs on the test sets. When aggregating results across datasets, we use the average distance to the minimum (ADTM) metric via affine renormalisation between the top-performing and worse-performing models~\cite{grinsztajn2022tree, mcelfresh2024neural}. We provide the evaluation results averaged over six downstream predictors for a general conclusion, and the fine-grained numerical results for each predictor are in \cref{appendix:numerical_results}.

\textbf{Data augmentation setup.} Given $N_{\text{real}}$ real samples, we first train generators on the real training data and then generate $N_{\text{syn}}$ synthetic samples. For training the downstream predictors, we expand the real training split by adding the synthetic samples. The real validation data is used for early stopping, and the real test set is used for evaluating the predictor's performance. The optimal $N_{\text{syn}}$ remains an open problem for tabular data~\cite{manousakas2023usefulness, seedat2024curated, hansen2023reimagining}. 
Prior works~\cite{liu2023goggle, TabPFGen} mainly use synthetic sets with equivalent sizes to the real sets (i.e., $N_{\text{real}}=N_{\text{syn}}$). However, we observe that $N_{\text{real}}=N_{\text{syn}}$ can lead to highly unstable results, especially on small datasets that we investigate. Recent work has used different $N_{\text{syn}}$ for various generators, such as by applying post-processing~\citep{hansen2023reimagining, seedat2024curated}. In this work, we want to provide a head-to-head comparison of the effect of data augmentation across subsampled datasets of varying sizes $N_{\text{real}} \in \{20, 50, 100, 200, 500\}$. Therefore, we perform data augmentation with a large synthetic set ($N_{\text{syn}}=500$) across all splits, and the synthetic data has the same class distribution as the real training data. We provide an illustrative figure of the data splitting setup in \cref{appendix:data-splitting}. 

\subsection{Data Augmentation Improvement (Q1)}
\label{sec:exp_utility}

We evaluate the effect of using synthetic data for data augmentation by comparing the \textit{balanced accuracy} of downstream predictors before and after augmentation. Typically, higher classification accuracy (i.e., \(\text{ACC}_{\text{Generator}})\) and accuracy improvements (i.e., \(\text{ACC}_{\text{Generator}} - \text{ACC}_{\text{Baseline}} > 0\)) demonstrate the effectiveness of the synthetic data for data augmentation.

\begin{figure}[t!]
    \centering
    \subfloat{\includegraphics[width=0.5\textwidth]{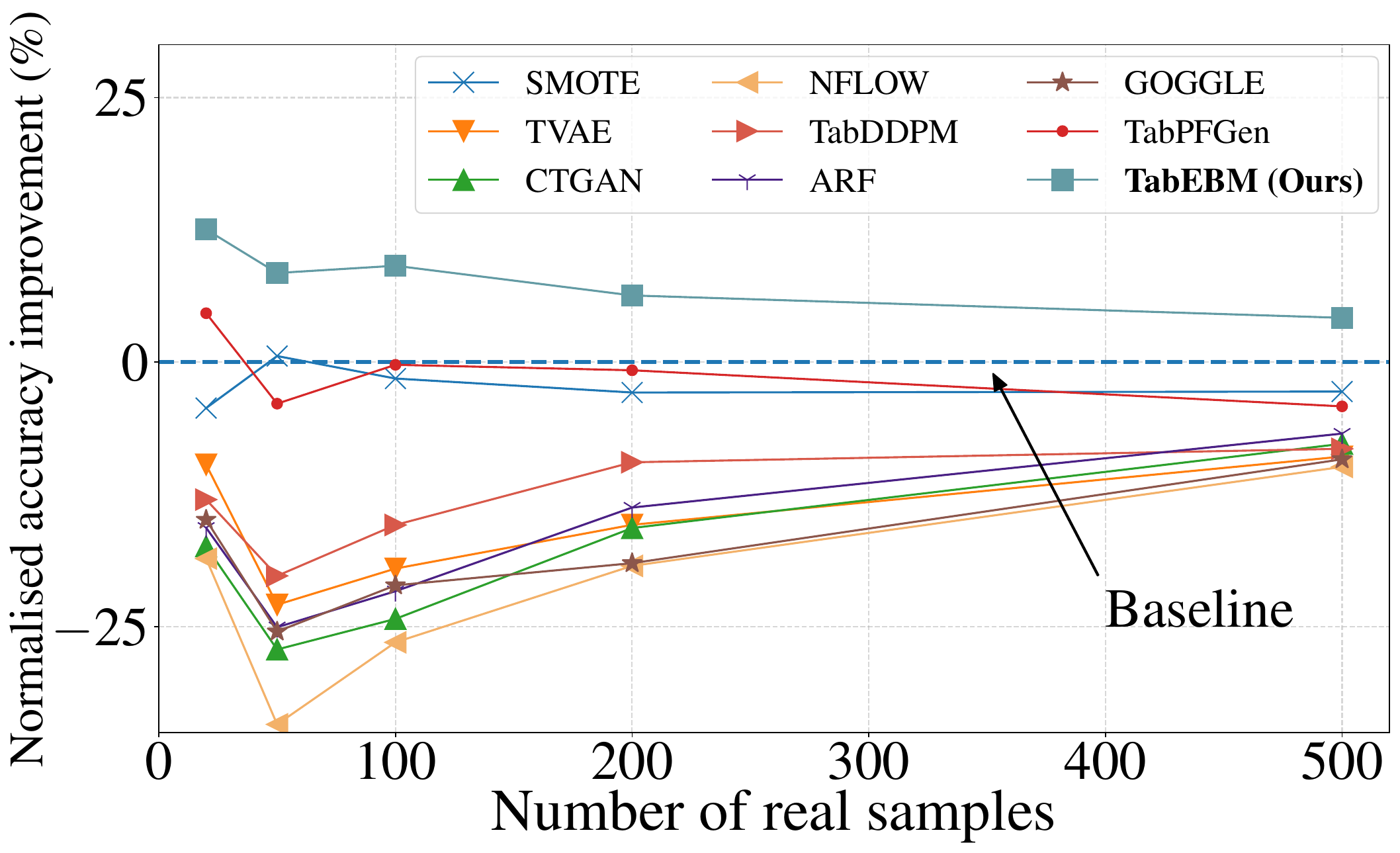}}
    \hfill
    \subfloat{\includegraphics[width=0.475\textwidth]{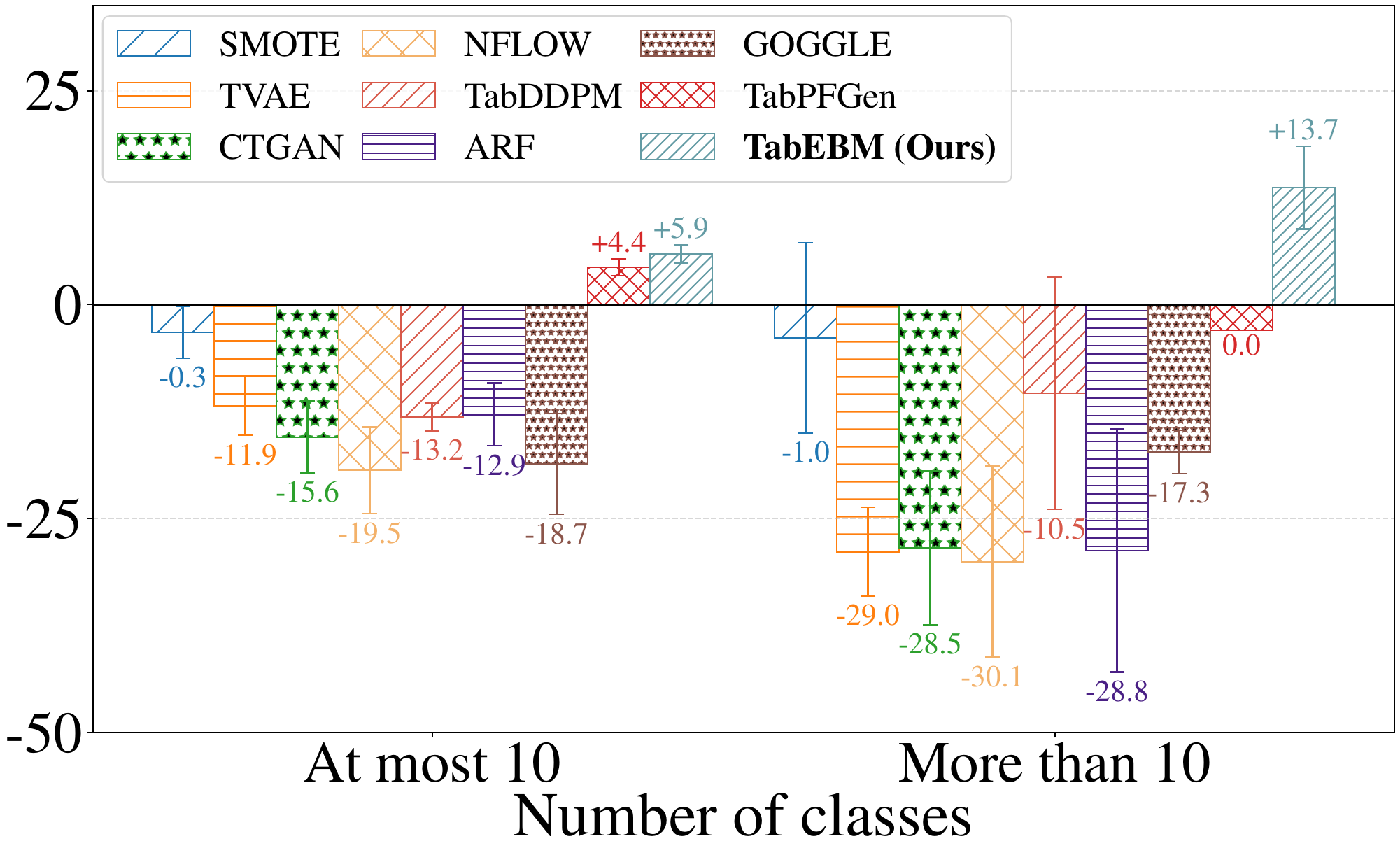}}
    
    \caption{Mean normalised balanced accuracy improvement (\%) across different sample sizes (\textbf{Left}) and across datasets with varying numbers of classes (\textbf{Right}). Because TabPFGen is not applicable for datasets with more than ten classes, we plot short bars at zeros for visual clearance. Positive values indicate that the generator improves downstream classification performance. TabEBM generally outperforms benchmark generators across varying sample sizes and number of classes.}
    \vspace*{-3mm}
\label{fig:acc_vs_sample_size_and_classes_and_time}
\end{figure}

\looseness-1
\textbf{TabEBM effectively improves downstream performance across sample sizes, especially for very low-sample-size regimes.} 
\cref{tab:test_acc_per_dataset_augmentation} and \cref{fig:acc_vs_sample_size_and_classes_and_time} (Left) show that TabEBM exhibits competitive performance in data augmentation, generally achieving the highest downstream accuracy and average rank across most datasets and sample sizes. Notably, TabEBM is the only generator that consistently improves performance across sample sizes. A key observation is that most modern benchmark generators underperform even the Baseline, indicating poor approximated distributions in the low-sample-size regime. Moreover, TabEBM achieves the largest overall performance improvement on six leakage-free UCI datasets, further supporting its effectiveness (see \cref{appendix:res_uci} for details).

\begin{table}[t!]
\centering
\caption{\textbf{Classification accuracy} (\%) aggregated over six downstream predictors, comparing data augmentation on eight real-world tabular datasets with varied real data availability. We report the mean $\pm$ std balanced accuracy and average accuracy rank across datasets. A higher rank implies higher accuracy. Note that ``N/A'' denotes that a specific generator was not applicable, and the rank is computed with the mean balanced accuracy of other methods. We \textbf{bold} the highest accuracy for each dataset of different sample size. Our method, TabEBM, consistently outperforms training on real data alone, and achieves the best overall performance against Baseline and benchmark generators.}
\label{tab:test_acc_per_dataset_augmentation}

\resizebox{\textwidth}{!}{
\setlength{\tabcolsep}{4pt}
\begin{tabular}{m{0.2cm}lr|r|rrrrrrrr|r}

\toprule

\multicolumn{2}{l}{Datasets}  & $N_{\text{real}}$ & \makecell[r]{Baseline\\(Real data)} & SMOTE & TVAE & CTGAN & NFLOW & TabDDPM & ARF & GOGGLE & TabPFGen & \textbf{TabEBM} \\

\midrule

\multirow{24}{*}{\rotatebox{90}{\begin{tabular}{l} \textit{At most 10 classes} \end{tabular}}}& \multirow[c]{5}{*}{protein} & 20 & 28.14$_{\pm\text{6.83}}$ & N/A & 21.18$_{\pm\text{1.48}}$ & 22.00$_{\pm\text{3.43}}$ & 21.30$_{\pm\text{2.84}}$ & 22.12$_{\pm\text{5.30}}$ & 24.82$_{\pm\text{2.88}}$ & 22.40$_{\pm\text{9.28}}$ & 33.25$_{\pm\text{5.01}}$ & \textbf{33.84}$_{\pm\text{4.92}}$ \\

& & 50 & 50.72$_{\pm\text{10.53}}$ & 54.52$_{\pm\text{8.59}}$ & 39.54$_{\pm\text{5.19}}$ & 36.32$_{\pm\text{7.17}}$ & 35.37$_{\pm\text{8.00}}$ & 35.11$_{\pm\text{11.78}}$ & 41.99$_{\pm\text{5.24}}$ & 37.53$_{\pm\text{14.72}}$ & 54.45$_{\pm\text{7.96}}$ & \textbf{55.91}$_{\pm\text{6.41}}$ \\

& & 100 & 67.83$_{\pm\text{11.72}}$ & 73.25$_{\pm\text{7.48}}$ & 59.28$_{\pm\text{7.20}}$ & 57.64$_{\pm\text{9.95}}$ & 52.57$_{\pm\text{9.55}}$ & 56.37$_{\pm\text{9.64}}$ & 57.01$_{\pm\text{8.56}}$ & 51.69$_{\pm\text{16.68}}$ & 71.53$_{\pm\text{9.87}}$ & \textbf{73.31}$_{\pm\text{6.77}}$ \\

& & 200 & 81.66$_{\pm\text{10.18}}$ & 85.65$_{\pm\text{6.24}}$ & 76.42$_{\pm\text{7.71}}$ & 74.88$_{\pm\text{8.20}}$ & 72.10$_{\pm\text{10.04}}$ & 75.86$_{\pm\text{9.30}}$ & 74.07$_{\pm\text{8.74}}$ & 73.57$_{\pm\text{6.74}}$ & 84.95$_{\pm\text{7.47}}$ & \textbf{86.14}$_{\pm\text{5.50}}$ \\

& & 500 & 93.49$_{\pm\text{5.28}}$ & 94.73$_{\pm\text{3.67}}$ & 92.24$_{\pm\text{3.73}}$ & 91.48$_{\pm\text{4.43}}$ & 90.44$_{\pm\text{5.54}}$ & 90.62$_{\pm\text{5.63}}$ & 91.79$_{\pm\text{4.53}}$ & 91.31$_{\pm\text{5.20}}$ & 94.87$_{\pm\text{3.70}}$ & \textbf{95.18}$_{\pm\text{3.10}}$ \\

\cmidrule{2-13}

& \multirow[c]{5}{*}{fourier} & 20 & 28.30$_{\pm\text{12.09}}$ & N/A & 21.32$_{\pm\text{4.06}}$ & 18.19$_{\pm\text{3.90}}$ & 17.30$_{\pm\text{3.03}}$ & 15.35$_{\pm\text{3.26}}$ & 21.75$_{\pm\text{2.76}}$ & 16.70$_{\pm\text{2.91}}$ & 36.72$_{\pm\text{7.30}}$ & \textbf{37.13}$_{\pm\text{6.01}}$ \\

& & 50 & 53.69$_{\pm\text{8.04}}$ & 55.51$_{\pm\text{7.43}}$ & 37.96$_{\pm\text{4.48}}$ & 35.09$_{\pm\text{7.46}}$ & 31.94$_{\pm\text{8.99}}$ & 35.99$_{\pm\text{13.06}}$ & 40.32$_{\pm\text{6.70}}$ & 33.56$_{\pm\text{14.02}}$ & 55.11$_{\pm\text{10.66}}$ & \textbf{56.57}$_{\pm\text{7.12}}$ \\

& & 100 & 63.70$_{\pm\text{6.76}}$ & 64.10$_{\pm\text{6.89}}$ & 50.46$_{\pm\text{8.61}}$ & 49.26$_{\pm\text{9.15}}$ & 44.58$_{\pm\text{8.40}}$ & 52.79$_{\pm\text{10.04}}$ & 51.13$_{\pm\text{6.35}}$ & 41.93$_{\pm\text{15.60}}$ & 63.86$_{\pm\text{7.76}}$ & \textbf{65.21}$_{\pm\text{6.42}}$ \\

& & 200 & 70.99$_{\pm\text{4.88}}$ & 71.43$_{\pm\text{4.47}}$ & 62.17$_{\pm\text{7.29}}$ & 62.92$_{\pm\text{7.87}}$ & 59.15$_{\pm\text{8.33}}$ & 68.05$_{\pm\text{6.91}}$ & 62.53$_{\pm\text{6.97}}$ & 56.44$_{\pm\text{10.13}}$ & 71.81$_{\pm\text{5.35}}$ & \textbf{72.36}$_{\pm\text{3.77}}$ \\

& & 500 & 77.72$_{\pm\text{2.36}}$ & 77.51$_{\pm\text{2.60}}$ & 73.29$_{\pm\text{4.97}}$ & 74.61$_{\pm\text{4.89}}$ & 71.74$_{\pm\text{6.54}}$ & 77.04$_{\pm\text{3.64}}$ & 74.31$_{\pm\text{4.40}}$ & 70.61$_{\pm\text{6.01}}$ & 77.15$_{\pm\text{2.57}}$ & \textbf{78.20}$_{\pm\text{2.87}}$ \\

\cmidrule{2-13}

& \multirow[c]{5}{*}{biodeg} & 20 & 66.20$_{\pm\text{4.26}}$ & 68.59$_{\pm\text{1.17}}$ & 66.77$_{\pm\text{2.64}}$ & 58.03$_{\pm\text{2.47}}$ & 59.37$_{\pm\text{1.74}}$ & 52.72$_{\pm\text{2.38}}$ & 61.17$_{\pm\text{2.00}}$ & 61.39$_{\pm\text{6.39}}$ & 68.99$_{\pm\text{2.54}}$ & \textbf{69.79}$_{\pm\text{2.15}}$ \\

& & 50 & 72.66$_{\pm\text{3.98}}$ & 72.80$_{\pm\text{3.08}}$ & 71.31$_{\pm\text{2.71}}$ & 67.99$_{\pm\text{3.63}}$ & 62.40$_{\pm\text{4.28}}$ & 60.72$_{\pm\text{10.11}}$ & 71.62$_{\pm\text{2.43}}$ & 66.68$_{\pm\text{6.00}}$ & 73.29$_{\pm\text{3.53}}$ & \textbf{73.78}$_{\pm\text{3.42}}$ \\

& & 100 & \textbf{76.69}$_{\pm\text{2.70}}$ & 76.31$_{\pm\text{2.42}}$ & 75.38$_{\pm\text{2.06}}$ & 74.82$_{\pm\text{2.89}}$ & 69.50$_{\pm\text{4.59}}$ & 68.28$_{\pm\text{9.54}}$ & 74.42$_{\pm\text{2.38}}$ & 71.68$_{\pm\text{3.72}}$ & 76.22$_{\pm\text{2.31}}$ & 76.45$_{\pm\text{3.08}}$ \\

& & 200 & 80.01$_{\pm\text{2.66}}$ & 79.67$_{\pm\text{2.56}}$ & 78.11$_{\pm\text{2.68}}$ & 78.19$_{\pm\text{1.78}}$ & 75.05$_{\pm\text{4.68}}$ & 74.43$_{\pm\text{8.09}}$ & 77.97$_{\pm\text{2.32}}$ & 77.13$_{\pm\text{3.01}}$ & 79.76$_{\pm\text{2.63}}$ & \textbf{80.11}$_{\pm\text{2.33}}$ \\

& & 500 & 82.63$_{\pm\text{2.43}}$ & \textbf{82.85}$_{\pm\text{1.93}}$ & 82.13$_{\pm\text{1.94}}$ & 82.42$_{\pm\text{1.58}}$ & 81.11$_{\pm\text{3.23}}$ & 79.19$_{\pm\text{6.60}}$ & 81.92$_{\pm\text{2.28}}$ & 81.24$_{\pm\text{2.30}}$ & 82.35$_{\pm\text{2.21}}$ & 82.29$_{\pm\text{2.15}}$ \\

\cmidrule{2-13}

& \multirow[c]{5}{*}{steel} & 20 & 57.51$_{\pm\text{4.58}}$ & 58.32$_{\pm\text{3.27}}$ & 57.99$_{\pm\text{3.06}}$ & 56.61$_{\pm\text{1.70}}$ & 53.89$_{\pm\text{1.73}}$ & 55.74$_{\pm\text{6.02}}$ & 54.24$_{\pm\text{2.08}}$ & 53.04$_{\pm\text{2.36}}$ & 63.21$_{\pm\text{5.86}}$ & \textbf{63.27}$_{\pm\text{5.45}}$ \\

& & 50 & 75.06$_{\pm\text{10.43}}$ & 65.63$_{\pm\text{4.00}}$ & 64.18$_{\pm\text{3.95}}$ & 63.70$_{\pm\text{6.10}}$ & 58.90$_{\pm\text{6.39}}$ & 65.85$_{\pm\text{14.84}}$ & 61.72$_{\pm\text{3.39}}$ & 56.72$_{\pm\text{3.47}}$ & 78.67$_{\pm\text{11.79}}$ & \textbf{80.50}$_{\pm\text{8.67}}$ \\

& & 100 & 86.87$_{\pm\text{12.49}}$ & 74.61$_{\pm\text{5.99}}$ & 70.12$_{\pm\text{5.76}}$ & 69.89$_{\pm\text{5.58}}$ & 65.67$_{\pm\text{9.10}}$ & 76.01$_{\pm\text{17.54}}$ & 67.33$_{\pm\text{5.15}}$ & 60.56$_{\pm\text{5.37}}$ & 90.58$_{\pm\text{9.50}}$ & \textbf{92.71}$_{\pm\text{7.57}}$ \\

& & 200 & 92.90$_{\pm\text{9.14}}$ & 81.97$_{\pm\text{4.12}}$ & 78.73$_{\pm\text{5.06}}$ & 78.36$_{\pm\text{6.98}}$ & 75.90$_{\pm\text{9.57}}$ & 85.45$_{\pm\text{15.03}}$ & 78.65$_{\pm\text{6.70}}$ & 68.20$_{\pm\text{5.30}}$ & 95.56$_{\pm\text{5.85}}$ & \textbf{96.29}$_{\pm\text{4.64}}$ \\

& & 500 & 97.52$_{\pm\text{3.76}}$ & 92.44$_{\pm\text{4.46}}$ & 92.47$_{\pm\text{3.66}}$ & 92.42$_{\pm\text{4.76}}$ & 88.20$_{\pm\text{8.36}}$ & 96.34$_{\pm\text{4.67}}$ & 90.41$_{\pm\text{5.35}}$ & 84.23$_{\pm\text{10.90}}$ & 98.14$_{\pm\text{2.67}}$ & \textbf{98.47}$_{\pm\text{2.15}}$ \\

\cmidrule{2-13}

& \multirow[c]{4}{*}{stock} & 20 & 78.75$_{\pm\text{4.39}}$ & 82.18$_{\pm\text{2.15}}$ & 74.11$_{\pm\text{3.71}}$ & 64.25$_{\pm\text{6.29}}$ & 72.64$_{\pm\text{2.01}}$ & 78.61$_{\pm\text{3.57}}$ & 69.54$_{\pm\text{1.65}}$ & 76.35$_{\pm\text{5.08}}$ & 82.42$_{\pm\text{2.17}}$ & \textbf{83.49}$_{\pm\text{1.60}}$ \\

& & 50 & 86.10$_{\pm\text{3.62}}$ & 87.82$_{\pm\text{3.41}}$ & 82.81$_{\pm\text{3.51}}$ & 79.63$_{\pm\text{3.93}}$ & 80.14$_{\pm\text{3.90}}$ & 86.72$_{\pm\text{4.29}}$ & 82.48$_{\pm\text{2.95}}$ & 83.36$_{\pm\text{5.23}}$ & 88.14$_{\pm\text{3.01}}$ & \textbf{88.44}$_{\pm\text{3.14}}$ \\

& & 100 & 89.07$_{\pm\text{3.71}}$ & 89.99$_{\pm\text{3.22}}$ & 87.55$_{\pm\text{4.25}}$ & 86.44$_{\pm\text{4.40}}$ & 84.64$_{\pm\text{4.79}}$ & 89.40$_{\pm\text{4.26}}$ & 87.32$_{\pm\text{4.42}}$ & 87.44$_{\pm\text{5.46}}$ & 90.27$_{\pm\text{3.33}}$ & \textbf{90.36}$_{\pm\text{3.51}}$ \\

& & 200 & 90.85$_{\pm\text{4.39}}$ & \textbf{91.75}$_{\pm\text{3.73}}$ & 90.12$_{\pm\text{5.44}}$ & 89.44$_{\pm\text{4.94}}$ & 88.47$_{\pm\text{6.06}}$ & 90.76$_{\pm\text{5.27}}$ & 89.59$_{\pm\text{5.37}}$ & 89.62$_{\pm\text{6.29}}$ & 91.56$_{\pm\text{3.91}}$ & 91.71$_{\pm\text{3.77}}$ \\

\midrule

\multirow{9}{*}{\rotatebox{90}{\begin{tabular}{l} \textit{More than 10 classes} \end{tabular}}}& \multirow[c]{3}{*}{energy} & 50 & 17.77$_{\pm\text{6.15}}$ & N/A & 12.30$_{\pm\text{2.59}}$ & 12.11$_{\pm\text{3.16}}$ & 10.14$_{\pm\text{2.87}}$ & 10.55$_{\pm\text{2.44}}$ & 11.99$_{\pm\text{2.27}}$ & 15.46$_{\pm\text{3.54}}$ & N/A & \textbf{23.98}$_{\pm\text{2.73}}$ \\

& & 100 & 25.94$_{\pm\text{4.86}}$ & N/A & 17.78$_{\pm\text{4.73}}$ & 18.60$_{\pm\text{6.09}}$ & 18.56$_{\pm\text{6.39}}$ & 18.84$_{\pm\text{6.23}}$ & 19.91$_{\pm\text{5.21}}$ & 17.65$_{\pm\text{5.88}}$ & N/A & \textbf{31.24}$_{\pm\text{5.53}}$ \\

& & 200 & 35.99$_{\pm\text{8.92}}$ & N/A & 27.65$_{\pm\text{11.12}}$ & 27.77$_{\pm\text{10.55}}$ & 28.37$_{\pm\text{10.82}}$ & 29.50$_{\pm\text{10.33}}$ & 29.57$_{\pm\text{9.18}}$ & 28.95$_{\pm\text{10.40}}$ & N/A & \textbf{41.28}$_{\pm\text{7.66}}$ \\

\cmidrule{2-13}

& \multirow[c]{2}{*}{collins} & 100 & 11.44$_{\pm\text{2.77}}$ & N/A & 8.38$_{\pm\text{1.52}}$ & 8.11$_{\pm\text{1.00}}$ & 7.93$_{\pm\text{1.40}}$ & 12.67$_{\pm\text{2.16}}$ & 7.53$_{\pm\text{1.10}}$ & 9.21$_{\pm\text{2.35}}$ & N/A & \textbf{13.07}$_{\pm\text{2.51}}$ \\

& & 200 & 15.74$_{\pm\text{3.73}}$ & \textbf{17.45}$_{\pm\text{3.46}}$ & 12.08$_{\pm\text{3.03}}$ & 11.37$_{\pm\text{1.20}}$ & 10.74$_{\pm\text{1.72}}$ & 15.39$_{\pm\text{3.37}}$ & 10.71$_{\pm\text{1.37}}$ & 14.30$_{\pm\text{3.42}}$ & N/A & 17.03$_{\pm\text{3.20}}$ \\

\cmidrule{2-13}

& \multirow[c]{4}{*}{texture} & 50 & 72.40$_{\pm\text{13.07}}$ & 76.40$_{\pm\text{10.50}}$ & 55.32$_{\pm\text{6.20}}$ & 54.80$_{\pm\text{12.97}}$ & 55.39$_{\pm\text{10.65}}$ & 62.27$_{\pm\text{8.01}}$ & 55.65$_{\pm\text{10.58}}$ & 62.94$_{\pm\text{12.06}}$ & N/A & \textbf{78.90}$_{\pm\text{7.96}}$ \\

& & 100 & 82.42$_{\pm\text{10.38}}$ & 84.35$_{\pm\text{9.67}}$ & 66.00$_{\pm\text{7.21}}$ & 69.49$_{\pm\text{10.93}}$ & 71.78$_{\pm\text{9.06}}$ & 76.25$_{\pm\text{7.40}}$ & 70.93$_{\pm\text{9.71}}$ & 76.34$_{\pm\text{9.55}}$ & N/A & \textbf{86.01}$_{\pm\text{7.36}}$ \\

& & 200 & 87.54$_{\pm\text{7.62}}$ & 89.29$_{\pm\text{6.20}}$ & 78.37$_{\pm\text{6.03}}$ & 82.44$_{\pm\text{7.15}}$ & 81.94$_{\pm\text{6.30}}$ & 84.67$_{\pm\text{4.79}}$ & 83.29$_{\pm\text{6.32}}$ & 82.53$_{\pm\text{7.99}}$ & N/A & \textbf{89.77}$_{\pm\text{5.77}}$ \\

& & 500 & 92.96$_{\pm\text{4.07}}$ & 93.69$_{\pm\text{3.83}}$ & 90.09$_{\pm\text{3.56}}$ & 91.48$_{\pm\text{3.50}}$ & 90.50$_{\pm\text{2.71}}$ & 91.53$_{\pm\text{3.29}}$ & 91.76$_{\pm\text{3.98}}$ & 91.24$_{\pm\text{3.56}}$ & N/A & \textbf{93.76}$_{\pm\text{3.64}}$ \\

\midrule
\rowcolor{Gainsboro!60}
\multicolumn{3}{l|}{\textbf{Average rank}} & 3.30$_{\pm\text{1.02}}$ & 3.03$_{\pm\text{1.25}}$ & 6.79$_{\pm\text{1.80}}$ & 7.48$_{\pm\text{1.50}}$ & 8.94$_{\pm\text{0.70}}$ & 6.39$_{\pm\text{2.41}}$ & 6.94$_{\pm\text{1.50}}$ & 7.76$_{\pm\text{2.03}}$ & 3.15$_{\pm\text{1.27}}$ & \textbf{1.21}$_{\pm\text{0.74}}$ \\
\bottomrule

\end{tabular}
}
\end{table}

Furthermore, TabEBM is the most widely applicable method among the top three competitive generators on the considered datasets: (i)~SMOTE requires at least two samples per class for interpolation, and thus it is not applicable for some datasets, such as the ``protein'' dataset ($N_{\text{real}}=20$); (ii)~TabPFGen cannot scale up to more than ten classes, such as the ``collins'' dataset. In addition, TabEBM can stabilise downstream performance, especially when real data is very scarce ($N_{\text{real}}=20$): TabEBM leads to smaller standard deviations than Baseline on seven out of eight datasets.

\textbf{TabEBM effectively improves downstream performance across any number of classes, especially for more than ten classes.}
\cref{fig:acc_vs_sample_size_and_classes_and_time} (Right) shows that TabEBM consistently outperforms the Baseline with notable improvements, particularly in datasets with more than ten classes. In contrast, an increased number of classes tends to cause a performance degradation in the benchmark generators.

\begin{figure}[!h]
    \vspace{-5pt}
    \centering
    \begin{minipage}[b]{0.53\textwidth}
        \centering
        \includegraphics[width=0.965\linewidth]{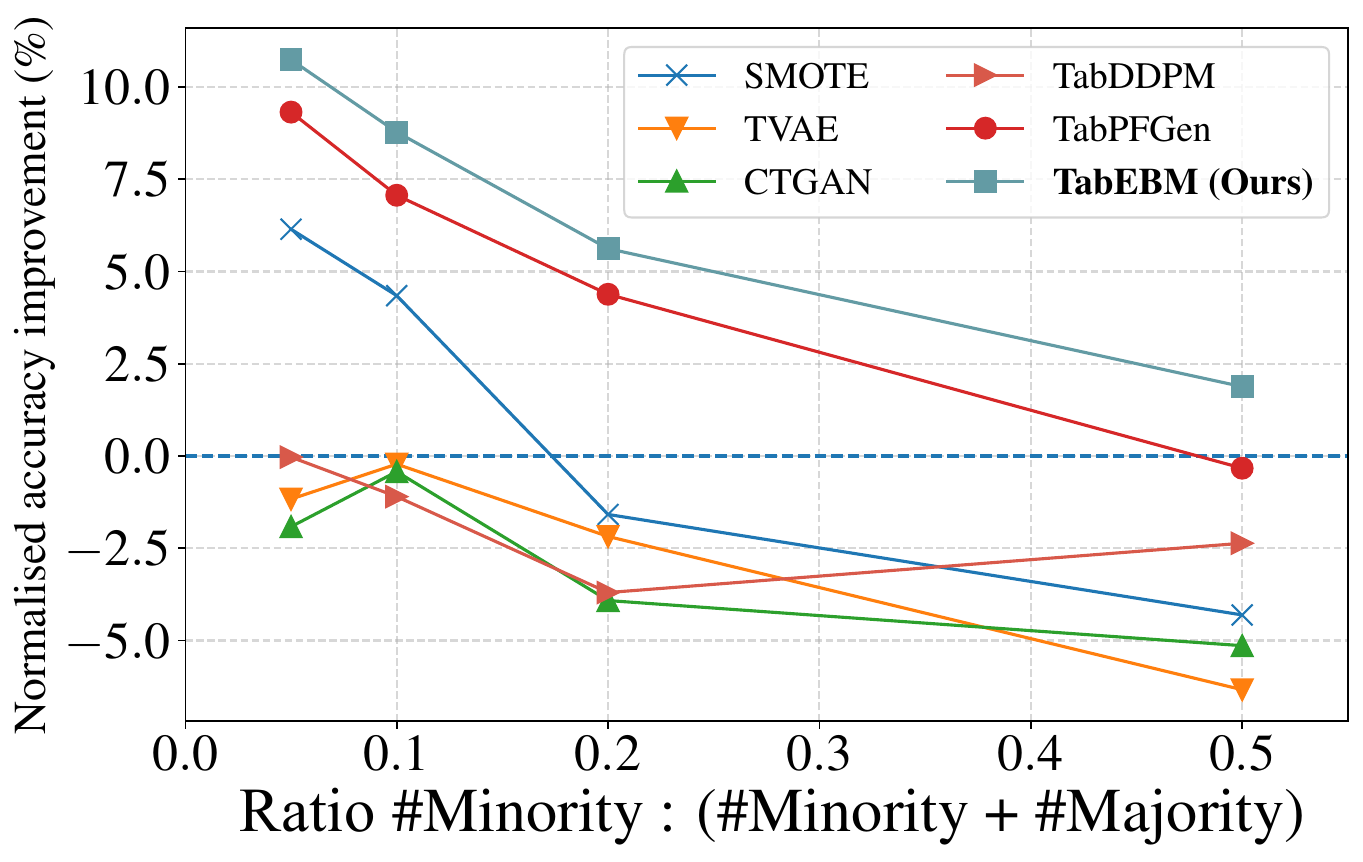}
        \caption{\textbf{Mean normalised balanced accuracy improvement (\%) on imbalanced datasets.} TabEBM consistently outperforms the Baseline and other generators across different levels of data imbalance.}
        \label{fig:acc_imbalanced}
    \end{minipage}
    \hfill
    \begin{minipage}[b]{0.42\textwidth}
        \centering
        \includegraphics[width=0.725\linewidth]{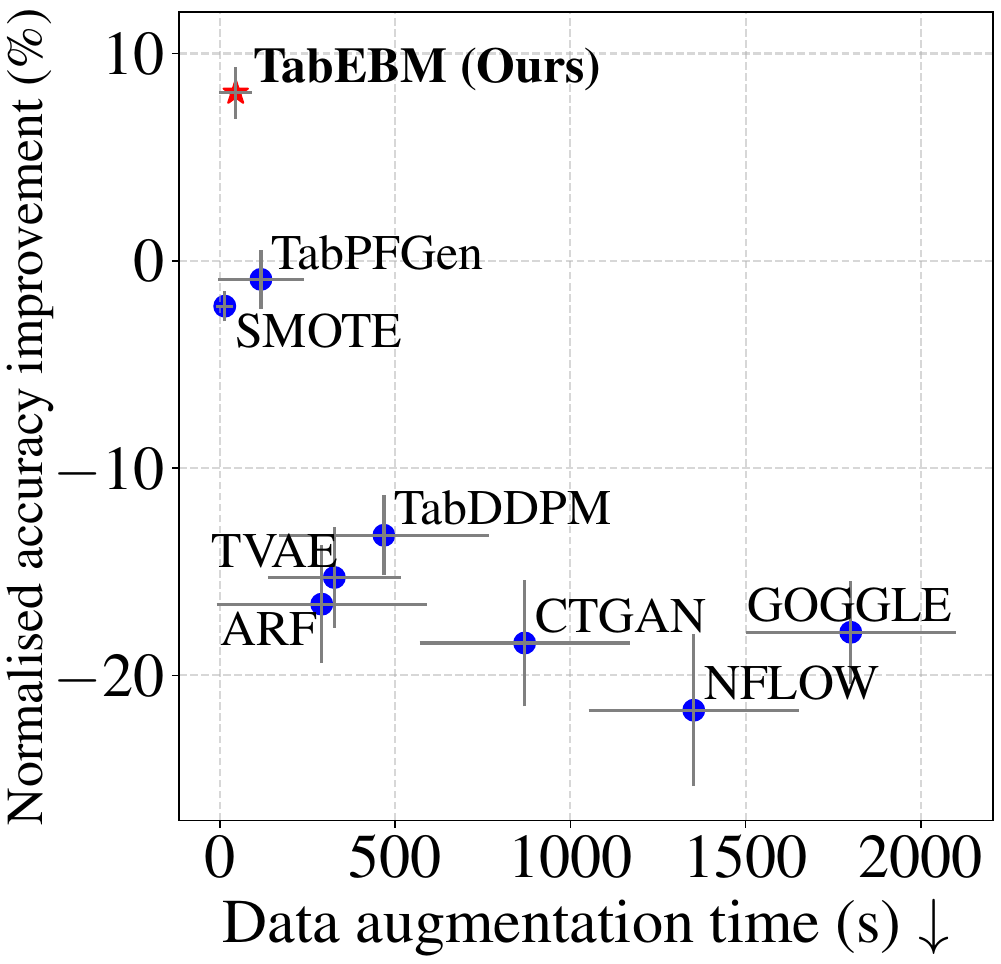}
        \caption{\textbf{Median data augmentation time vs. mean normalised balanced accuracy.} TabEBM achieves higher downstream accuracy while typically operating 3-30 times faster than most other methods.}
        \label{fig:acc_vs_time}
    \end{minipage}
    \vspace{-4pt}
\end{figure}

\textbf{TabEBM is robust on imbalanced datasets.} For the three binary OpenML datasets (i.e., ``biodeg'', ``steel'' and ``stock''), we adjust the class distribution in the training data to vary the class imbalance, while keeping the test data fixed. \cref{fig:acc_imbalanced} shows that TabEBM consistently outperforms Baseline, while the other generators exhibit performance degradation as data imbalance increases.

\textbf{TabEBM is computationally efficient.} \cref{fig:acc_vs_time} shows the trade-off between accuracy and the time needed for generating stratified synthetic data (for data augmentation). We measure the total duration of (i)~training the model and (ii)~generating 500 synthetic samples. The results show that TabEBM is practical, as it achieves higher downstream accuracy with lower time costs.

\vspace*{-3mm}
\subsection{Statistical Fidelity (Q2)}
\label{sec:exp_fidelity}

\looseness-1
We evaluate the fidelity of synthetic data by 
measuring
the similarity of synthetic data to real \textit{train} data and to real \textit{test} data (\cref{fig:acc_vs_fidelity_and_privacy}). 
We evaluate this similarity 
via (i)~\textit{average inverse of the Kullback–Leibler Divergence} (inverse KL)~\cite{csiszar1975divergence}, (ii)~p-value of \textit{Kolmogorov–Smirnov test} (KS test)~\cite{karson1968handbook} and (iii)~\mbox{p-value} of \textit{Chi-squared test} ($\chi^2$ test)~\cite{mchugh2013chi}. 
For full numerical results, including $\chi^2$ test, 
see  \cref{appendix:res_fidelity}. 
For all three metrics, a bigger value denotes that synthetic data is more likely to have the same distribution as real data.

In \cref{fig:acc_vs_fidelity_and_privacy} (a1\&a2), TabEBM consistently exhibits the highest accuracy and distribution similarity between real train data and synthetic data, indicating that TabEBM learns the distributions of real train data better than benchmark generators. In \cref{appendix:res_fidelity}, we further show that TabEBM remains the most competitive method in similarity between real test data and synthetic data. This indicates that TabEBM can extrapolate beyond real train data and thus generate synthetic data from a more general distribution that aligns with both train and test data. This extrapolation ability also explains why TabEBM can outperform Baseline via data augmentation (\cref{sec:exp_utility}).

\subsection{Privacy Preservation (Q3)}
\label{sec:exp_privacy}

More broadly, data privacy is a critical concern for organisations and governments handling sensitive data~\cite{stadler2022search}. Privacy-preserving synthetic data allows researchers and practitioners to bypass ethical and logistical issues while enabling model training and testing~\cite{jin2019review}. We further explore the use of TabEBM-generated data for data sharing, where only synthetic data is accessible for downstream users~\cite{stadler2022search, zheng2020privacy, zhang2018towards, dunning2012privacy, kotelnikov2023tabddpm}. In this case, downstream models are trained exclusively on synthetic data.

Specifically, we evaluate synthetic data via three metrics: (i)~\textit{balanced accuracy} of downstream predictors trained with only synthetic data (i.e., train-on-synthetic, test-on-real~\cite{xu2019modeling, kotelnikov2023tabddpm, zhang2023mixed}); (ii)~{median Distance to Closest Record} (DCR)~\cite{zhao2021ctab}, where a greater DCR denotes synthetic data is less likely to be copied from real data; and (iii)~\({\delta}\)-presense~\cite{nergiz2019delta}, where a smaller value denotes a lower re-identification risk for real data from synthetic data. Full numerical results are in \cref{appendix:res_privacy}.

\cref{fig:acc_vs_fidelity_and_privacy} (b1\&b2) shows that TabEBM consistently finds a better trade-off between accuracy and privacy preservation. Notably, the ``train-on-synthetic, test-on-real'' scenario poses a greater challenge for generators in achieving high accuracy because real data is inaccessible for model training and data augmentation. Despite this difficulty, TabEBM is the only generator that surpasses the overall performance of training on real data (i.e., Baseline). The relatively high DCR for TabEBM indicates that it can extrapolate beyond real train data, aligning with the finding that TabEBM's synthetic data is statistically similar to real test data (\cref{sec:exp_fidelity}). These results further suggest that TabEBM learns the general distribution of real data, and can generate high-quality synthetic data suitable for various purposes, including privacy preservation.

\begin{figure}[!t]
    \centering
    \subfloat{\includegraphics[width=0.245\textwidth]{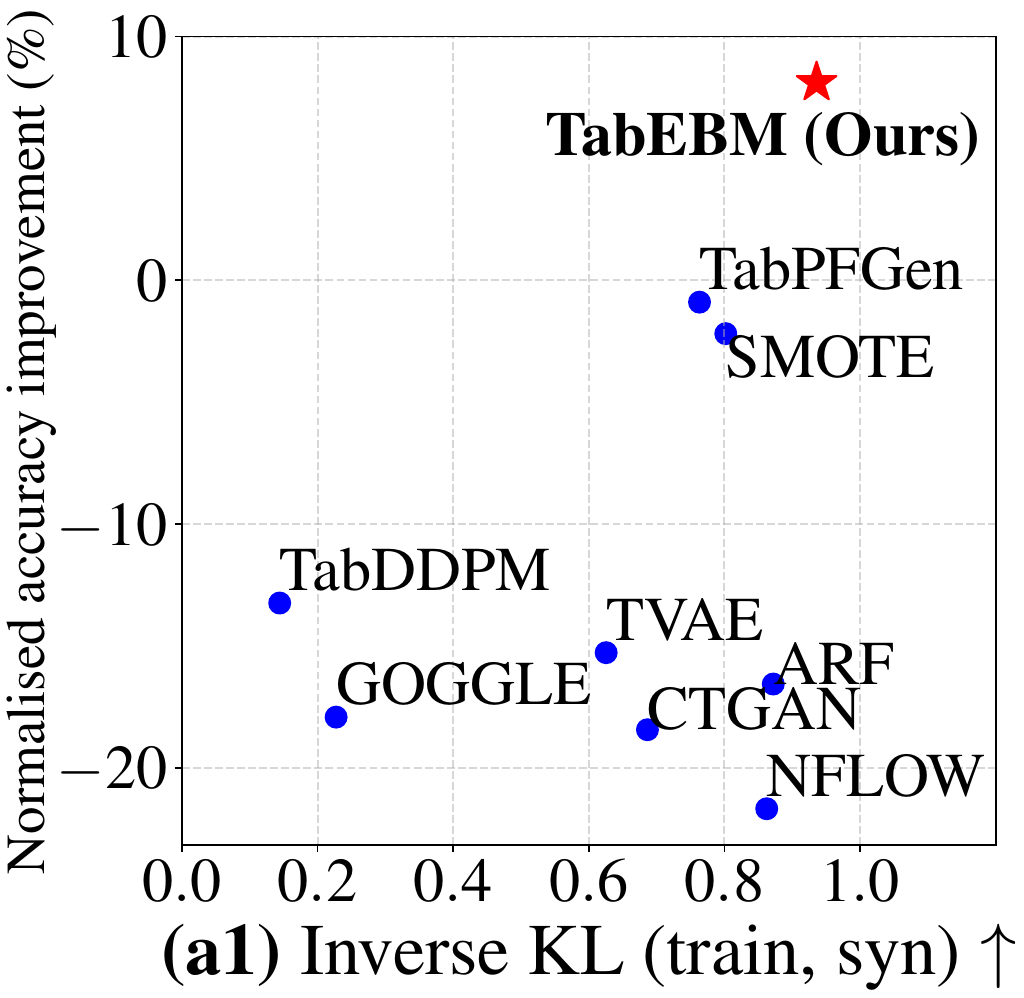}}
    \subfloat{\includegraphics[width=0.2288\textwidth]{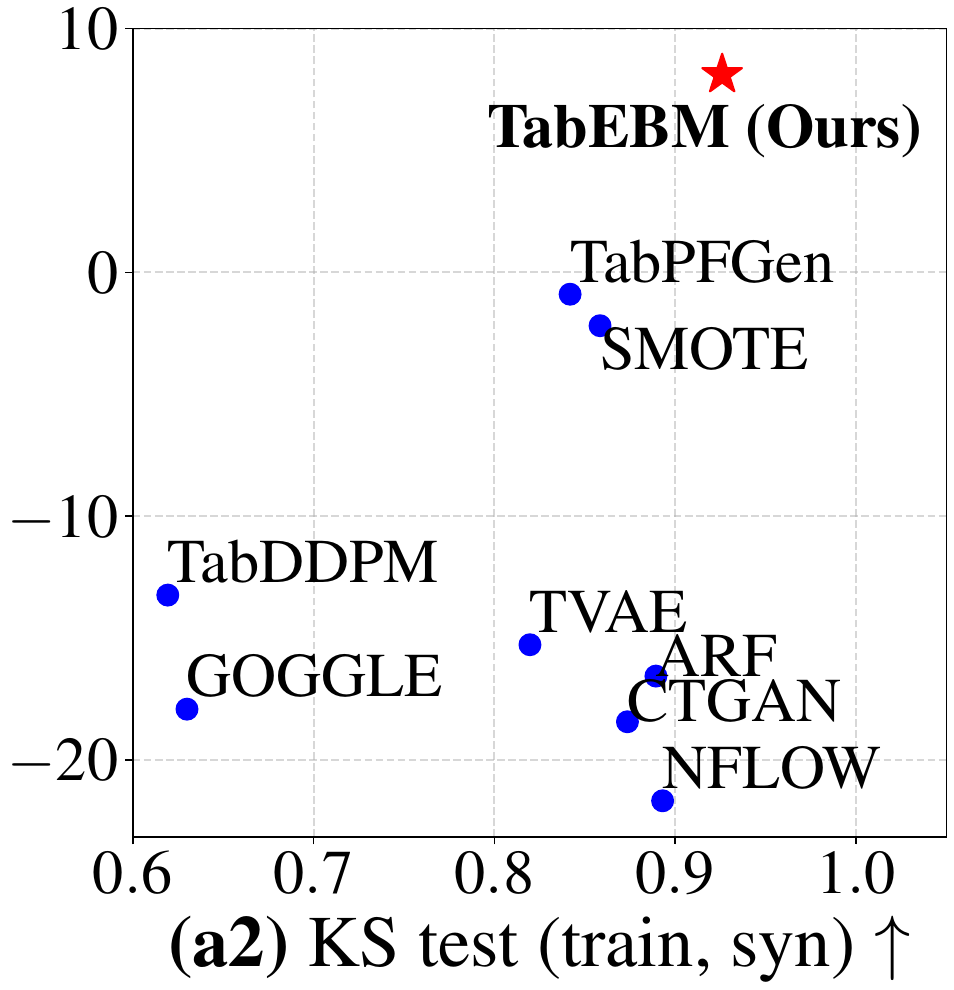}}\hfill
    \subfloat{\includegraphics[width=0.2519\textwidth]{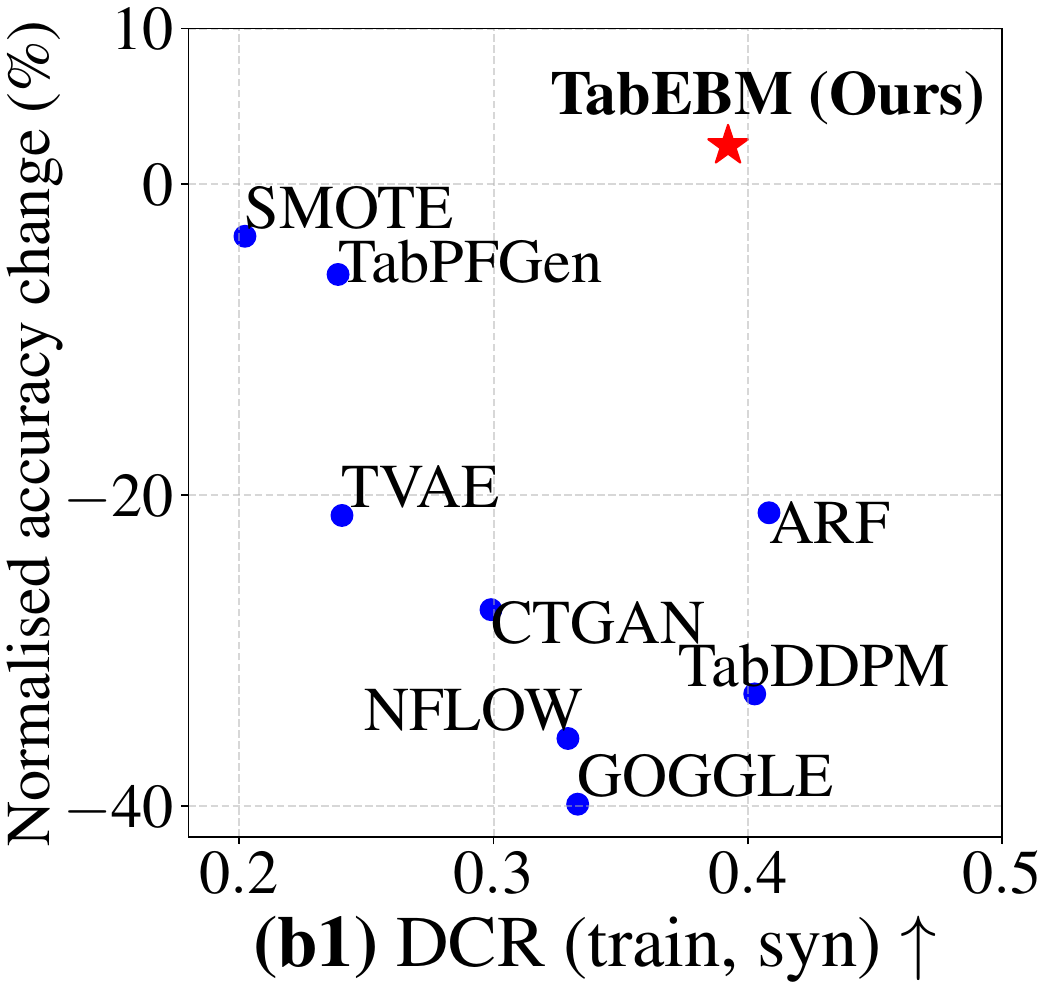}}
    \subfloat{\includegraphics[width=0.2337\textwidth]{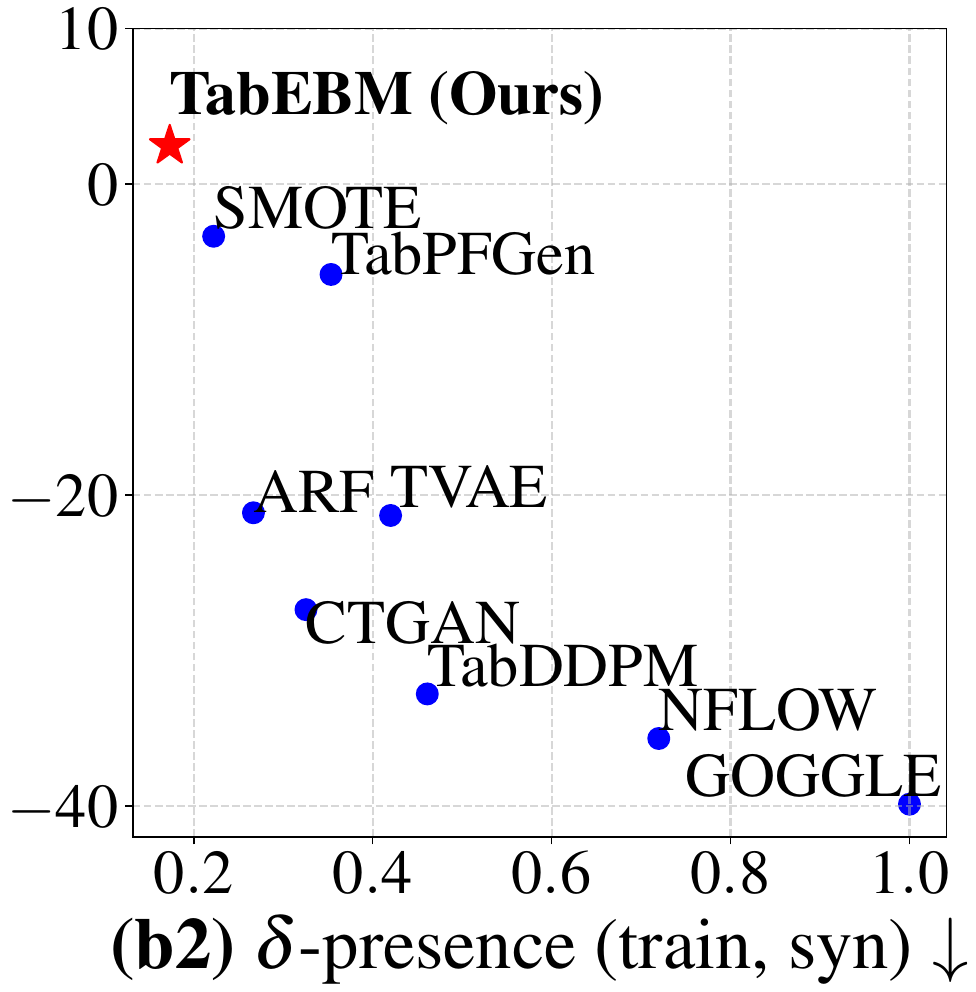}}
    
    \caption{ \textbf{(a1\&a2):} Median inverse KL and KS test vs.\ mean normalised balanced accuracy improvement (\%) between real train data and synthetic data. \textbf{(b1\&b2):} Median DCR and \(\delta\)-presence vs.\ mean normalised balanced accuracy change (\%) between real train data and synthetic data. Note that ``accuracy improvement'' is for data augmentation, and ``accuracy change'' is for data sharing. Complete results with standard deviations are in \cref{appendix:tabebm-tradeoff-figures-with-error-bars}. TabEBM generates high-fidelity synthetic data that can also be used for privacy preservation.}
    \label{fig:acc_vs_fidelity_and_privacy}
    \vspace{-10pt}
\end{figure}

\subsection{Why is TabEBM effective for estimating Energy-Based Models? (Q4)}
\label{sec:why-is-tabebm-effective-for-energy-estimation}

\begin{wrapfigure}[]{r}{0.55\textwidth}
    \centering
    \vspace{-10pt}
    \includegraphics[clip,trim=5pt 5pt 5pt 5pt, width=0.97\linewidth]{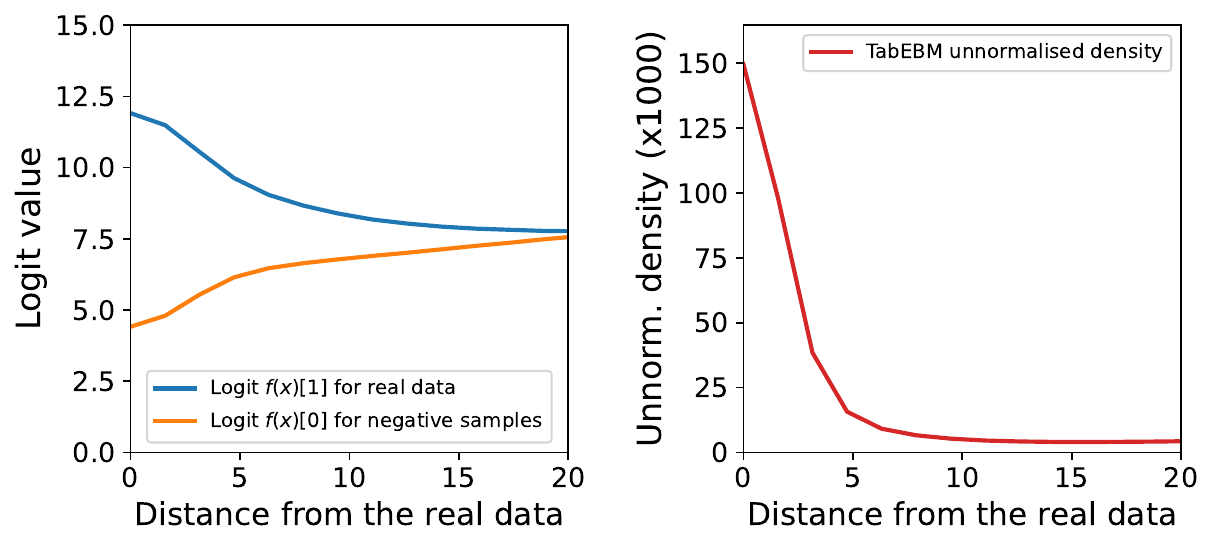}
    
    \caption{\textbf{(Left)} Logit distribution of TabPFN trained on our surrogate binary tasks at increasing distances from the real data (on ``steel''). \textbf{(Right)} The corresponding unnormalised density approximated by TabEBM. TabEBM assigns higher density closer to the real data.}

    \label{fig:tabpfn-logits-mini}
    
    \vspace{-20pt}
\end{wrapfigure}

Having established that TabEBM excels in data augmentation, we explore why classifier logits can be useful when reinterpreted as a class-conditional energy function. \cref{fig:tabpfn-logits-mini} shows the logit distribution of TabPFN trained on surrogate binary tasks and the corresponding energy function of TabEBM (with TabPFN as the binary classifier) as the Euclidean distance from the real data increases.

\looseness-1
We found it essential to place the negative samples far from the real data, since TabPFN, which is pre-trained to approximate Bayesian inference~\citep{TabPFN}, has its confidence influenced by the distance from the training data~\citep{mccarter2024whatexactlyhas}. \cref{fig:tabpfn-logits-mini} (left) shows that TabPFN outputs high logit values near the real data. As the distance from the real data increases, the logit $f(\rvx)[1]$ decreases smoothly until the two logits become similar, making the classifier uncertain (because the class probabilities become equal). \cref{fig:tabpfn-logits-mini} (right) shows that TabEBM's inferred density drops significantly as the maximum logit decreases, because $p_c(\rvx) \propto (\exp(f(\rvx)[0]) + \exp(f(\rvx)[1]))$ from \cref{eq:class-specific-energy}. Since SGLD sampling performs gradient ascent on the density, the TabEBM-generated samples will be close to the real data. These findings are consistent across datasets (see \cref{appendix:distribution-logits-tabebm}), where TabPFN's logits remain positive, with similar ranges and a relatively constant sum as distance increases, warranting further investigation. Overall, TabPFN's distance-based uncertainty is useful for inferring accurate energy functions within our TabEBM framework. Since TabEBM can be paired with any other gradient-based classifier that produces logits, we leave these extensions for future work.

\section{Discussion \& Related Work}
\label{sec:related-work}
\looseness-1
\cref{sec:exp} showed that TabEBM efficiently generates high-fidelity data that can effectively improve the downstream performance via data augmentation. In \cref{tab:Comparison}, we further provide a summary of tabular data generative models analysed from three important perspectives: 
(i)~\emph{Training:} the type of distribution that the generators learn (crucial for preserving the original training label distribution), and the training costs associated with learning;
(ii)~\emph{Generation:} do the generators employ class-specific models (reflecting their capability to capture unique features essential for label-invariant generation), and do models support stratified generation (crucial for effective data augmentation);
(iii)~\emph{Practicability:} the scalability of the generators with respect to the number of classes (a common requirement in real-world multi-class tasks), and consistent downstream performance improvement across different class sizes.\looseness-1

\looseness-1
\textbf{Generative Models for Tabular Data.}
The common paradigm for tabular data generation is to adapt Generative Adversarial Networks (GANs) and Variational Autoencoders (VAEs)~\cite{xu2019modeling, park2018data}. For instance, TableGAN employs a convolutional neural network to optimise the label quality~\cite{park2018data}, and TVAE is introduced in~\cite{xu2019modeling} as a variant of VAE for tabular data. However, these methods learn the joint distribution and thus cannot preserve the stratification of the original data (\cref{appendix:limitations-existing-work}). CTGAN~\cite{xu2019modeling} refines the generation to be class-conditional. The recent ARF~\citep{watson2023adversarial} is an adversarial variant of random forest for density estimation, and GOGGLE~\cite{liu2023goggle} enhances VAE by learning relational structure with a Graph Neural Network (GNN). Some recent work focuses on generation with denoising diffusion models~\cite{kotelnikov2023tabddpm, zhang2023mixed, kim2022stasy, lee2023codi}. For instance, TabDDPM~\cite{kotelnikov2023tabddpm} demonstrates that diffusion models can approximate typical distributions of tabular data. Although these class-conditional models can preserve the label distribution, they struggle to outperform Baseline and standard SMOTE in data augmentation~\cite{seedat2024curated, TabPFGen}. 

We attribute the performance degradation in current class-conditional models to their reliance on a single shared model to approximate all class-conditional densities. For instance, another promising generative approach uses pre-trained models like Prior-Data Fitted Networks (PFNs), and the recent TabPFGen~\cite{TabPFGen} adapts such models into one shared class-conditional generator. However, TabPFGen’s shared generator can lead to inaccurate density estimates, particularly in high-noise and class-imbalance situations (see examples in \cref{appendix:limitations-existing-work}). As noise increases, TabPFGen's inferred densities fluctuate significantly and diverge from the true data distributions. In contrast, TabEBM uses class-specific EBMs to model each class’s marginal distributions, and the results in \cref{appendix:limitations-existing-work} reveal that our design choice reduces the impact of noise and data imbalance. TabEBM focuses on approximating and generating for one class at a time, remaining unaffected by noise from other classes. Overall, our results demonstrate that TabEBM consistently improves performance across different datasets and sample sizes, outperforming TabPFGen. Moreover, TabPFGen is limited in usability (e.g., it supports only up to ten classes), while TabEBM scales to any number of classes.

In a broader context, some recent work attempts to adapt Large Language Models (LLMs) for tabular data generation~\cite{fang2024large, seedat2024curated, borisov2022language}. However, data contamination is an inherent issue with such LLM-based models~\cite{deng2024investigating, jiang2024investigating, dekoninck2024evading, magar2022data}. As the pre-training data is not typically open-source, these models can have unfair advantages in downstream tasks (i.e., the full real dataset, including the real test data, may have been used for pre-training). Therefore, in this paper, we focus on models without support from LLMs, thus avoiding potential biases from data contamination.

%
\begin{table}[t]
\centering
\caption{\textbf{Comparison of the properties between TabEBM and prior tabular generative methods.} TabEBM has novel design rationales of training-free class-specific models, and TabEBM is highly practicable with wide applicability and consistent accuracy improvement.}
\label{tab:Comparison}
\resizebox{\textwidth}{!}{
\begin{tabular}{lc|cc|cc|ccc}

\toprule

\multirow[b]{2}{*}{Methods} & \multirow[b]{2}{*}{Category} & \multicolumn{2}{c|}{Training}                                   & \multicolumn{2}{c|}{Generation}                                                    & \multicolumn{3}{c}{Practicability}                                                                                                     \\

\cmidrule{3-9}

                         &                           & \makecell[c]{Learned\\distribution}  & Training-free                           & \makecell[c]{Class-specific\\models}                    & \makecell[c]{Stratified\\generation}                  & \makecell[c]{Unlimited\\classes}              & \makecell[c]{ACC improve\\($\leq$ 10 classes)}  & \makecell[c]{ACC improve\\($>$ 10 classes)} \\

\midrule

SMOTE~\citep{chawla2002smote}                    & Interpolation             & N/A                  & {\color[HTML]{3A7B21}   \CheckmarkBold}                                     & N/A                                     & {\color[HTML]{3A7B21}   \CheckmarkBold} & {\color[HTML]{3A7B21}   \CheckmarkBold} & {\color[HTML]{E6341C}   \XSolidBrush}       & {\color[HTML]{E6341C}   \XSolidBrush}       \\
TVAE~\citep{xu2019modeling}                     & VAE                       & $p(\rvx, y)$         & {\color[HTML]{E6341C}   \XSolidBrush}   & {\color[HTML]{E6341C}   \XSolidBrush}   & {\color[HTML]{E6341C}   \XSolidBrush} & {\color[HTML]{3A7B21}   \CheckmarkBold} & {\color[HTML]{E6341C}   \XSolidBrush}       & {\color[HTML]{E6341C}   \XSolidBrush}       \\
CTGAN~\citep{xu2019modeling}                    & GAN                       & $p(\rvx \mid y)$     & {\color[HTML]{E6341C}   \XSolidBrush}   & {\color[HTML]{E6341C}   \XSolidBrush}   & {\color[HTML]{3A7B21}   \CheckmarkBold} & {\color[HTML]{3A7B21}   \CheckmarkBold} & {\color[HTML]{E6341C}   \XSolidBrush}       & {\color[HTML]{E6341C}   \XSolidBrush}       \\
NFLOW~\citep{durkan2019neural}                    & Normal. Flows         & $p(\rvx, y)$         & {\color[HTML]{E6341C}   \XSolidBrush}   & {\color[HTML]{E6341C}   \XSolidBrush}   & {\color[HTML]{E6341C}   \XSolidBrush} & {\color[HTML]{3A7B21}   \CheckmarkBold} & {\color[HTML]{E6341C}   \XSolidBrush}       & {\color[HTML]{E6341C}   \XSolidBrush}       \\
TabDDPM~\citep{kotelnikov2023tabddpm}                  & Diffusion                 & $p(\rvx \mid y)$     & {\color[HTML]{E6341C}   \XSolidBrush}   & {\color[HTML]{E6341C}   \XSolidBrush}   & {\color[HTML]{3A7B21}   \CheckmarkBold} & {\color[HTML]{3A7B21}   \CheckmarkBold} & {\color[HTML]{E6341C}   \XSolidBrush}       & {\color[HTML]{E6341C}   \XSolidBrush}       \\
ARF~\citep{watson2023adversarial}                      & Random Forest             & $p(\rvx, y)$     & {\color[HTML]{E6341C}   \XSolidBrush}   & {\color[HTML]{E6341C}   \XSolidBrush}   & {\color[HTML]{E6341C}   \XSolidBrush} & {\color[HTML]{3A7B21}   \CheckmarkBold} & {\color[HTML]{E6341C}   \XSolidBrush}       & {\color[HTML]{E6341C}   \XSolidBrush}       \\
GOGGLE~\citep{liu2023goggle}                   & GNN                       & $p(\rvx \mid y)$     & {\color[HTML]{E6341C}   \XSolidBrush}   & {\color[HTML]{E6341C}   \XSolidBrush}   & {\color[HTML]{3A7B21}   \CheckmarkBold} & {\color[HTML]{3A7B21}   \CheckmarkBold} & {\color[HTML]{E6341C}   \XSolidBrush}       & {\color[HTML]{E6341C}   \XSolidBrush}       \\
TabPFGen~\citep{TabPFGen}                 & PFN                       & $p(\rvx \mid y)$     & {\color[HTML]{3A7B21}   \CheckmarkBold} & {\color[HTML]{E6341C}   \XSolidBrush}   & {\color[HTML]{3A7B21}   \CheckmarkBold} & {\color[HTML]{E6341C}   \XSolidBrush}   & {\color[HTML]{3A7B21}   \CheckmarkBold}       & {\color[HTML]{E6341C}   \XSolidBrush}         \\

\midrule
\rowcolor{Gainsboro!60}
\textbf{TabEBM (Ours)}          & PFN                       & $p(\rvx \mid y)$     & {\color[HTML]{3A7B21}   \CheckmarkBold} & {\color[HTML]{3A7B21}   \CheckmarkBold} & {\color[HTML]{3A7B21}   \CheckmarkBold} & {\color[HTML]{3A7B21}   \CheckmarkBold} & {\color[HTML]{3A7B21}   \CheckmarkBold}     & {\color[HTML]{3A7B21}   \CheckmarkBold}       \\

\bottomrule

\end{tabular}
}
\vspace{-6pt}
\end{table}

\looseness-1
\textbf{Data Augmentation (DA) for Tabular Data.}
DA is an omnipresent technique in computer vision and natural language processing~\cite{van2001art, shorten2021text, shorten2019survey, mumuni2022data, feng2021survey, antoniou2017data}. However, DA for tabular data remains underexplored, and existing methods often perform poorly in real-world tasks, sometimes even reduce performance~\cite{manousakas2023usefulness}. Recent studies show that using the same transformations across all classes leads to varied performance impacts~\citep{balestriero2022effects, kirichenko2024understanding}, indicating that data augmentation effects are class-specific and suggesting that different classes may require distinct augmentations. Given the lack of symmetries in tabular data, we believe this class-dependent effect is even more pronounced. Therefore, we propose TabEBM as a class-specific generative model to produce tailored augmentations for each class.

\looseness-1
\textbf{Prior-fitted Networks (PFNs) for Tabular Data.}
Recent work proposes to approximate the posterior predictive distribution with transformers~\cite{PFN, TabPFN,nagler2023statistical, ubbens2023gpfn, dooley2024forecastpfn}. PFNs can be adapted for various purposes by pre-training the transformer with corresponding ``prior data'', and then it can make in-context predictions with unseen downstream data. For instance, TabPFN is a variant that is pre-trained on a prior designed for tabular data~\cite{TabPFN}. We note that prior data is different to synthetic data in this paper. Specifically, prior data refers to manually crafted fake data (e.g., $y=2x$) with no real-world semantics. In contrast, synthetic data from generators is expected to have the same semantics as real data. Inspired by TabPFN’s success in small-size classification tasks, TabEBM converts TabPFN into multiple EBMs that learn the marginal distribution for each class. The training-free nature of TabPFN enables TabEBM to generate high-quality tabular data without introducing extra training costs. Additionally, our class-specific design lets TabEBM surpass TabPFN’s limits and scale to more than ten classes.

\looseness-1
\textbf{Limitations and Future Work.} TabEBM is a general method that relies on an underlying binary classifier, and as such, its strengths and weaknesses are directly tied to this classifier. We used TabPFN because it is a well-established open-source pre-trained model for tabular data. Therefore, TabEBM inherits some of TabPFN's limitations, particularly in scaling to a larger number of features. TabEBM can handle datasets with over 1000 samples, overcoming TabPFN's limitation, as it processes one class at a time. In \cref{appendix:results-larger-sample-sizes}, we show that TabEBM outperforms other generators on larger datasets, though the performance gains decrease as the sample size increases. 
Although we implement TabEBM with TabPFN in this paper, we stress that TabEBM is compatible with any classifier that can be adapted into EBMs, as described in \cref{sec:method}. As foundational models for tabular data evolve \citep{van2024tabular}, new models capable of handling more features and samples are expected. Integrating them into TabEBM will enhance its ability to manage high-dimensional datasets, increasing its versatility and utility.
Finally, note that, generators that are limited in modelling multivariate distributions may still perform well on univariate fidelity metrics, which is a standard approach to evaluating such models. However, evaluating their ability to learn more complex, high-order, relationships between features remains an open research question~\cite{tu2024causality}, which we leave for future work. 

\section{Conclusion} 
We introduced TabEBM, the first tabular data augmentation method that creates class-specific EBM generators, learning the marginal distribution for each class separately. We also provide the first comprehensive analysis of tabular data augmentation across various dataset sizes. Our results demonstrate that TabEBM improves downstream performance through data augmentation on real-world datasets, outperforming other benchmark generators. The statistical evaluation confirms that TabEBM generates high-fidelity synthetic data, particularly for small datasets. We release our method as an open-source library, allowing users to generate data immediately without additional training.


\clearpage

\begin{ack}
The authors would like to thank Francisco Vargas, Randall Balestriero, and Otilia Stretcu for their insightful discussions and valuable input early in the project. NS and MJ acknowledge the support of the U.S. Army Medical Research and Development Command of the Department of Defense; through the FY22 Breast Cancer Research Program of the Congressionally Directed Medical Research Programs, Clinical Research Extension Award GRANT13769713. Opinions, interpretations, conclusions, and recommendations are those of the authors and are not necessarily endorsed by the Department of Defense. \looseness -1
\end{ack}

\bibliography{references}
\bibliographystyle{plain}
\setcitestyle{square}

\clearpage
\appendix
\input{appendix.tex}
\input{checklist.tex}

\end{document}

%% file: math_commands.tex
\usepackage{amsmath,amsfonts,bm}

\def\eqref#1{equation~\ref{#1}}

\def\1{\bm{1}}

\def\rvx{{\mathbf{x}}}

\DeclareMathAlphabet{\mathsfit}{\encodingdefault}{\sfdefault}{m}{sl}
\SetMathAlphabet{\mathsfit}{bold}{\encodingdefault}{\sfdefault}{bx}{n}

\def\gX{{\mathcal{X}}}
\def\gY{{\mathcal{Y}}}

\def\sR{{\mathbb{R}}}



%% file: appendix.tex
\hypersetup{linkcolor=black}
\addcontentsline{toc}{section}{Appendix}
\part{Appendix}

{\Large{\textbf{\titlecontent}}}
\mtcsetdepth{parttoc}{3} 
\parttoc
\hypersetup{linkcolor=blue}

\clearpage
\FloatBarrier
\section{Broader Impact Statement}
\label{sec:border_impacts}
This paper introduces a novel data augmentation approach, TabEBM, that aims to advance the field of machine learning by addressing challenges in the low-sample-size regime. Furthermore, TabEBM offers an elegant solution to learning the unique features in generating samples for each class, leading to high-fidelity synthetic data that can effectively improve downstream performance. These characteristics can be particularly useful in data-scarce domains like healthcare (e.g., pre-clinical drug evaluation in early-stage clinical trials~\cite{bespalov2016failed, morford2011preclinical}). Moreover, we also demonstrate that TabEBM is readily applicable for privacy-preserving data sharing in high-stake tasks~\cite{zheng2020privacy, sun2023private}.

TabEBM's impact further extends to enabling broader machine learning applications in data-scarce domains, for instance, facilitating data analysis in clinical scenarios with limited access to data collection techniques. Improving the performance of machine learning models in such applications can further foster the uptake of more sophisticated ML approaches and, ultimately, help improve the quality of healthcare~\cite{alami2020artificial, ciecierski2022artificial, mollura2020artificial}. TabEBM can further facilitate research and enhance machine learning accessibility in various communities across societal and scientific domains. To this end, our work has only been evaluated in a strictly research setting. Further applications of our work in scenarios with sensitive data bear some risks. As TabEBM is a generative model, training models with the resulting generated samples can bias the downstream model. Therefore, this risk, together with other data privacy risks during downstream deployment, must be carefully managed.

\FloatBarrier
\section{Reproducibility}
\label{appendix:reproducibility}

\FloatBarrier
\subsection{Datasets}
\label{appendix:datasets}
\looseness-1
All eight datasets are publicly available on OpenML~\cite{OpenML}, and their details are listed in Table~\ref{tab:datasets}. To ensure consistent stratified data-splitting across all datasets, we remove classes with fewer than 10 samples. For example, the original ``energy'' dataset contains 14 classes with fewer than 10 samples, which could result in a validation set lacking samples from these classes, leading to unstratified data splitting.

\begin{table}[htbp]
\centering
\caption{Details of the eight real-world tabular datasets.}
\label{tab:datasets}
\resizebox{\textwidth}{!}{%
\begin{tabular}{lrrrrrrrr}

\toprule
Dataset       & OpenML ID & \makecell[r]{Not evaluated in\\TabPFN~\cite{TabPFN}} & \# Samples ($N$) & \# Features ($D$) & \# Classes & $N/D$    & \makecell[r]{\# Samples per class\\(Min)} & \makecell[r]{\# Samples per class\\(Max)} \\

\midrule

\multicolumn{9}{c}{\cellcolor[HTML]{EFEFEF}{At most 10 classes}} \\

\midrule
protein & 40966 & \XSolidBrush          & 1,080          & 77              & 8          & 14.03  & 105                      & 150                      \\
fourier & 14 & \XSolidBrush          & 2,000          & 76              & 10         & 26.32  & 200                      & 200                      \\
biodeg  & 1494 & \XSolidBrush          & 1,055          & 41              & 2          & 25.73  & 356                      & 699                      \\
steel   & 1504 & \XSolidBrush          & 1,941          & 33              & 2          & 58.82  & 673                      & 1,268                    \\
stock   & 841 & \XSolidBrush          & 950            & 9               & 2          & 105.56 & 462                      & 488                      \\

\midrule

\multicolumn{9}{c}{\cellcolor[HTML]{EFEFEF}{More than 10 classes}} \\

\midrule

energy & 1472 & \XSolidBrush          & 698            & 9               & 23         & 77.56  & 10                       & 74                       \\
collins & 40971 & \CheckmarkBold          & 970            & 19              & 26         & 51.05  & 17                       & 80                       \\
texture & 40499 & \CheckmarkBold           & 5,500          & 40              & 11         & 137.5  & 500                      & 500   \\
\bottomrule
\end{tabular}

}
\end{table}

\FloatBarrier
\subsection{Data Splitting}
\label{appendix:data-splitting}

\begin{figure}[!htbp]
    \centering
    \includegraphics[width=0.75\textwidth]{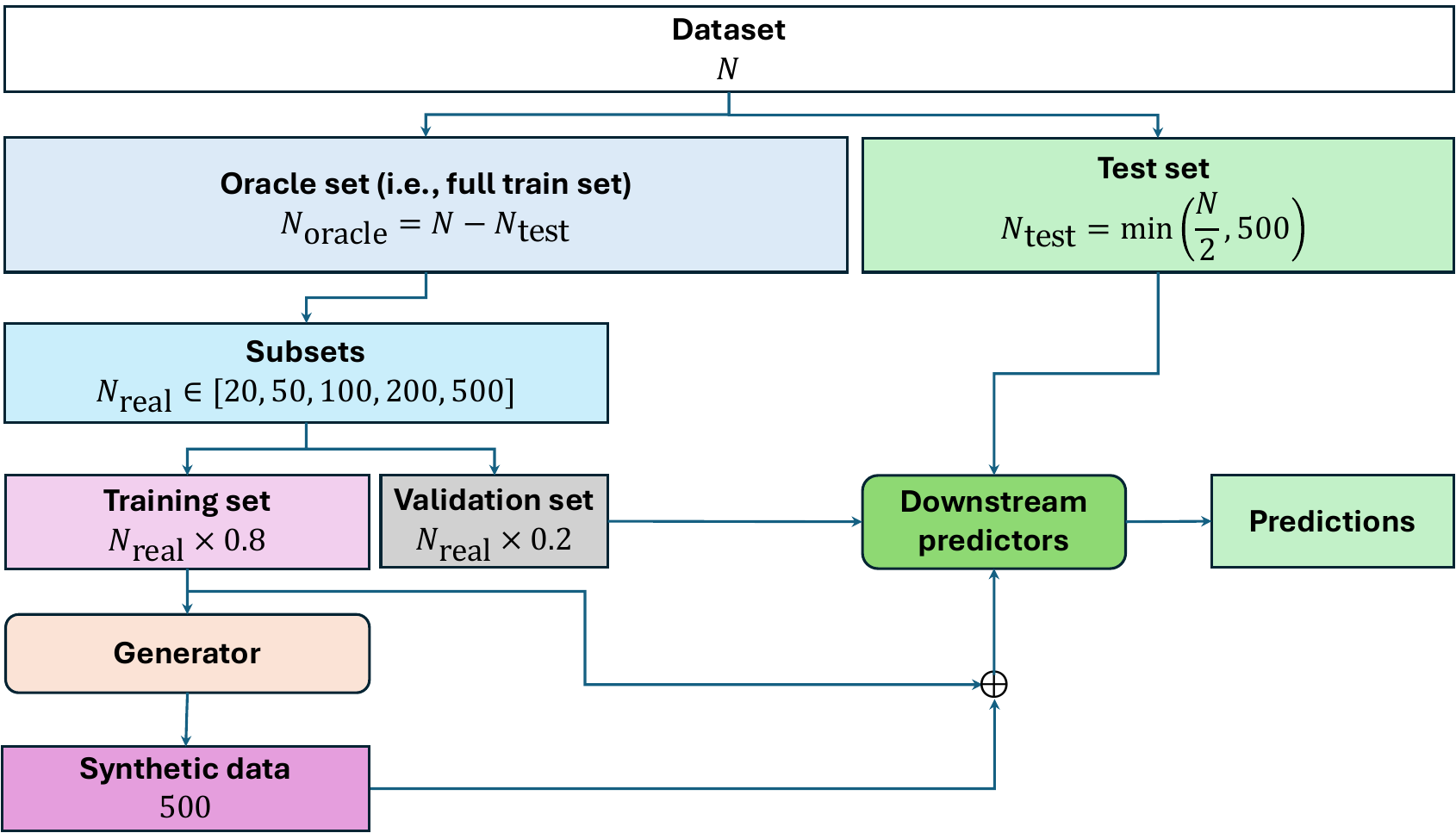}
    \caption{Data splitting strategies for data augmentation for all datasets.}
    \label{fig:data_split}
\end{figure}

\cref{fig:data_split} shows the data splitting setup used across all datasets. Note that data sharing (\cref{sec:exp_privacy}) shares the same data splitting as data augmentation, except that the ``Training set'' and ``Validation set'' containing real data are no longer used for training the downstream predictors.

\FloatBarrier
\subsection{Data Preprocessing}
\label{appendix:data-preprocessing}

Following the procedures presented in prior work~\cite{mcelfresh2024neural, grinsztajn2022tree}, we perform preprocessing in two steps. We first compute the required statistics with training data and then transform it. Firstly, we impute the missing values with the mean value for numerical features and the most mode value for categorical features. Secondly, we convert the categorical features into numerical features equal to Leave-one-out Target Statistic~\cite{prokhorenkova2017catboost, micci2001preprocessing}. Next, we perform Z-score normalisation for each feature. Specifically, we compute each feature's mean and standard deviation in the training data and then transform the training samples to have a mean of zero and a variance of one for each feature. Finally, we apply the same transformation to the validation and test data before conducting evaluations.

\FloatBarrier
\subsection{Software and Computing Resources}
\label{appendix:software-and-computing}

\looseness-1
\textbf{Software implementation.} \textit{(i) For generators:}  We implemented TabEBM using PyTorch 1.13 \citep{paszke2019pytorch}, an open-source deep learning library with a BSD licence. We implemented SMOTE with Imbalanced-learn~\cite{imbalanced-learn}, an open-source Python library for imbalanced datasets with an MIT licence. For other benchmark generators, we used their open-source implementations in Synthcity~\cite{qian2024synthcity}, a library for generating and evaluating synthetic tabular data with an Apache-2.0 license. \textit{(ii) For downstream predictors:} We implemented TabPFN with its open-source implementation (\url{https://github.com/automl/TabPFN}). We implemented the other five downstream predictors (i.e., Logistic Regression, KNN, MLP, Random Forest and XGBoost) with their open-source implementation in scikit-learn~\cite{scikit-learn}, an open-source Python library under the 3-Clause BSD license. \textit{(iii) For result analysis and visualisation:} All numerical plots and graphics have been generated using Matplotlib 3.7~\citep{matplotlib}, a Python-based plotting library with a BSD licence. The model architecture was generated using draw.io (\url{https://github.com/jgraph/drawio}), a free drawing software under Apache License 2.0.

We ensure the consistency and reproducibility of experimental results by implementing a uniform pipeline using PyTorch Lightning, an open-source library under an Apache-2.0 licence. We further fixed the random seeds for data loading and evaluation throughout the training and evaluation process. This ensured that TabEBM and all benchmark models were trained and evaluated on the same set of samples. The experimental environment settings, including library dependencies, are specified in the open-source library for reference and reproduction purposes.

\looseness-1
\textbf{Computing Resources.}
We trained 140,000 models for evaluations (including over 35,000 of generators and over 10,500 for downstream predictors). All our experiments are run on a single machine from an internal cluster with a GPU Nvidia Quadro RTX 8000 with 48GB memory and an Intel(R) Xeon(R) Gold 5218 CPU with 16 cores (at 2.30GHz). The operating system was Ubuntu 20.4.4 LTS.

\FloatBarrier
\subsection{TabEBM open-source library}
\label{appendix:tabebm-library}

We implemented TabEBM as an extensible library, and the code is available on \texttt{https://github.com/andreimargeloiu/TabEBM}. For practitioners, it offers an easy-to-use, domain-agnostic tool that requires no training, making it particularly suitable for data augmentation, especially in small datasets. For researchers, the library includes the complete implementation of TabEBM, facilitating future extensions and investigations into class-specific energy-based models.

The library has two core functionalities:

\begin{enumerate}[topsep=2pt, leftmargin=15pt, itemsep=0pt]

\item \textbf{Generate synthetic data}: The library can generate data for augmentation.

\begin{verbatim}
from tabebm.TabEBM import TabEBM

tabebm = TabEBM()
augmented_data = tabebm.generate(X_train, y_train, num_samples=100)
% augmented_data[class_id] = numpy.ndarray of generated data 
%                            for a specific ''class_id``
\end{verbatim}

\item \textbf{Compute and visualise the energy function}: The library allows computation of TabEBM's energy function and the unnormalised data density. The demo notebook, \texttt{TabEBM\_approximated\_density.ipynb}, shows the TabEBM-inferred densities under conditions of data noise and class imbalance (thus recreating the plots from \cref{appendix:limitations-existing-work}).
\end{enumerate}

\FloatBarrier
\subsection{Implementation of Generators}
\label{appendix:implementation-generators}

\textbf{TabEBM.} In all our experiments, the surrogate binary classifier in TabEBM is a pretrained in-context model, TabPFN~\citep{TabPFN}, using the official model weights released by the authors (\url{https://github.com/automl/TabPFN/raw/main/tabpfn/models_diff/prior_diff_real_checkpoint_n_0_epoch_42.cpkt}). We use TabPFN with three ensembles. We use four surrogate negative samples, \(\mathcal{X}_c^{\text{neg}}\), positioned at $\alpha^{\text{neg}}_{\text{dist}} = 5$ standard deviations from zero, in random corners of a hypercube in $\sR^D$ (as explained in \cref{sec:tabebm-energy-derivation}), distant from any real data. In \cref{appendix:ablations-negative-samples}, we show that TabEBM is robust to the distribution of the negative samples.

We use SGLD~\citep{SGLD} for sampling from TabEBM, where the starting points \(\rvx_0^{\text{synth}}\) are initialised by adding Gaussian noise with zero mean and standard deviation \(\sigma_{\text{start}} = 0.01\) to a randomly selected sample of the specific class, i.e., \(\rvx_0^{\text{synth}} \sim \mathcal{N}(\mathcal{X}_c, \sigma_{\text{start}}^2 \mathbf{I})\). For SGLD, we used the following parameters: step size \(\alpha_{\text{step}}=0.1\), noise scale \(\alpha_{\text{noise}}=0.01\) and number of steps \(T=200\). We found TabEBM to be robust to the SGLD settings (see \cref{appendix:ablation_sgld}).

\textbf{TabPFGen.} We re-implemented TabPFGen~\citep{TabPFGen} by closely following the original paper since no official implementation is available. As recommended in~\citep{TabPFGen}, the starting points are initialised by adding Gaussian noise with zero mean and standard deviation of 0.01 to the training points.

\textbf{SMOTE.} We use the open-source implementation of SMOTE from Imbalanced-learn~\cite{imbalanced-learn}, and the number neighbours $k$ is set within the range of $\{1, 3, 5\}$. When applicable, we always set the maximum value for nearest neighbours (i.e., $k=5$). However, very low-sample-size datasets may not contain sufficient samples for large $k$. For instance, the ``fourier'' dataset ($N_{\text{real}}=20$) only has two samples per class. We set $k=1$ to generate synthetic data with SMOTE in these cases.

For the other six benchmark generators, we use their open-source implementations in Synthcity~\cite{qian2024synthcity}. Following prior studies~\cite{zhang2023mixed, van2023synthetic, seedat2024curated, TabPFGen}, we use the default settings for all generators.

\FloatBarrier
\subsection{Implementation of Downstream Predictors}
We implemented TabPFN with its official implementation~\cite{TabPFN} and the other five downstream predictors with the scikit-learn library~\cite{scikit-learn}. Following prior studies~\cite{van2023synthetic, seedat2024curated}, we use the default settings for all downstream predictors.

\FloatBarrier
\section{Limitations of Existing Generative Methods}
\label{appendix:limitations-existing-work}

\begin{figure}[!htbp]
    \centering
    \includegraphics[width=0.8\textwidth]{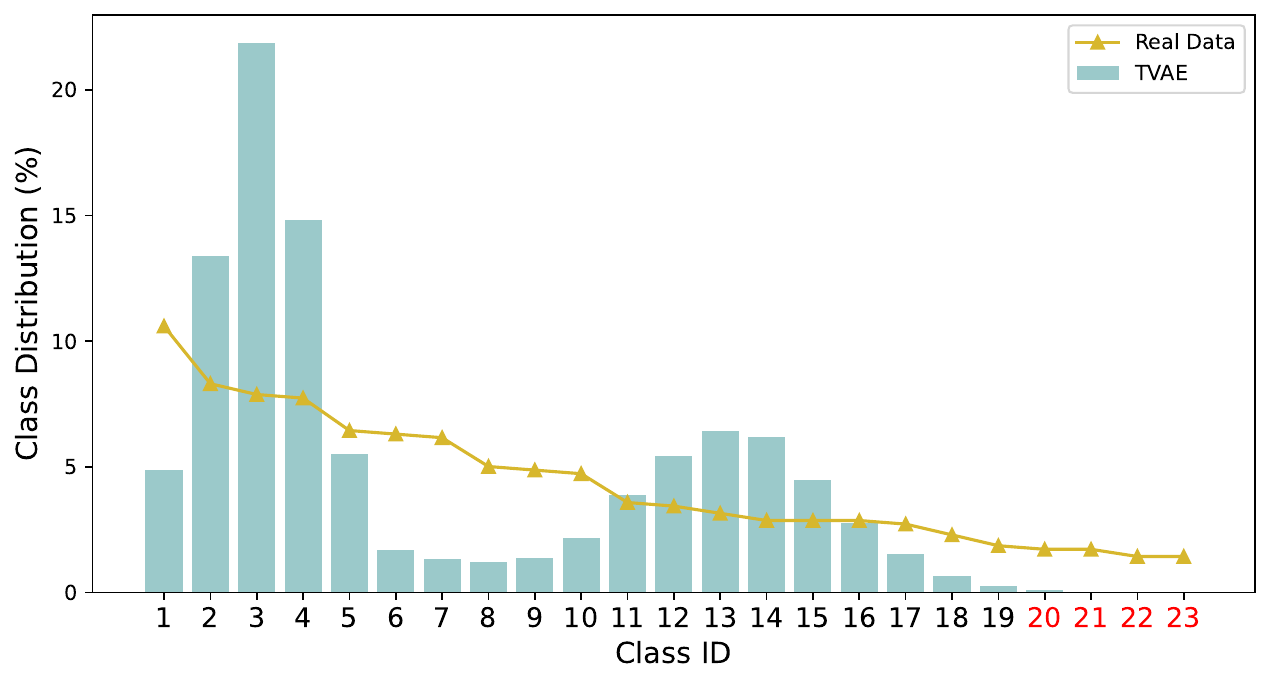}
    \label{fig:tabddpm-labeldistribution}
    \caption{Comparison of class distribution between real data and synthetic data from TVAE. We first train TVAE on the ``energy-efficiency'' dataset and then randomly generate 10,000 samples with it. We {\color[HTML]{FF0000}{highlight}} the classes where no synthetic samples are generated. TVAE fails to generate samples for 4 of 23 classes, showing the impracticability to preserve stratification by generative methods that learn joint distribution $p(\rvx,y)$.}
\end{figure}

\begin{figure}[htp]
\centering

\subfloat[Noise level 0.1]{
  \includegraphics[width=\textwidth, trim={0 0 0 40pt}, clip]{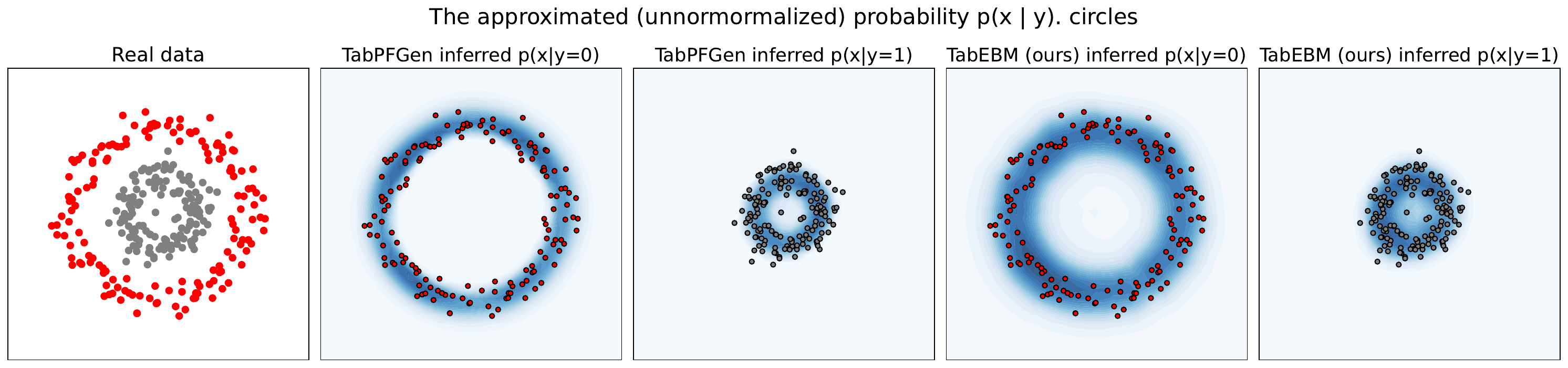}
}

\subfloat[Noise level 0.25]{
  \includegraphics[width=\textwidth, trim={0 0 0 40pt}, clip]{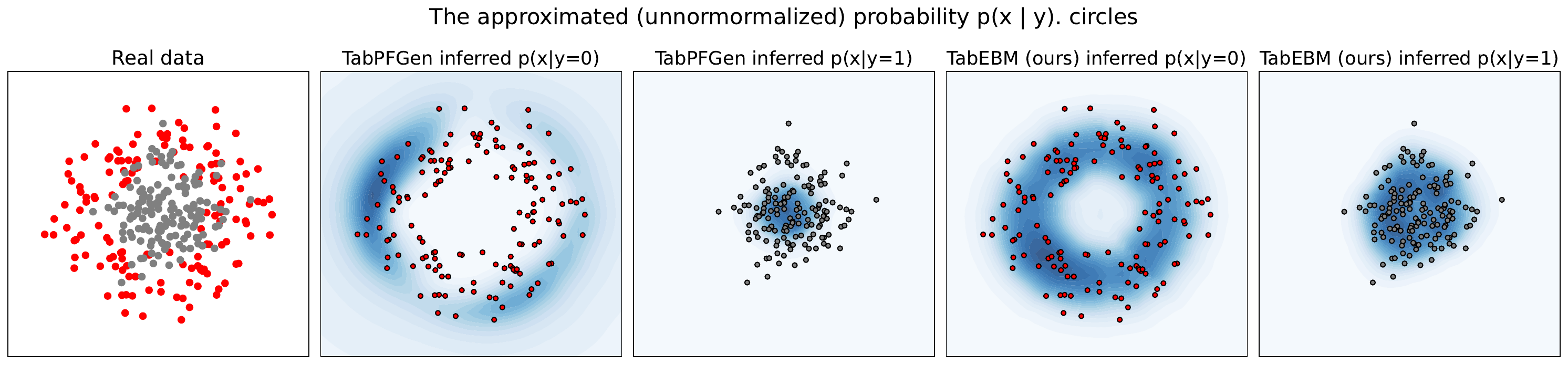}
}

\subfloat[Noise level 0.5]{
  \includegraphics[width=\textwidth, trim={0 0 0 40pt}, clip]{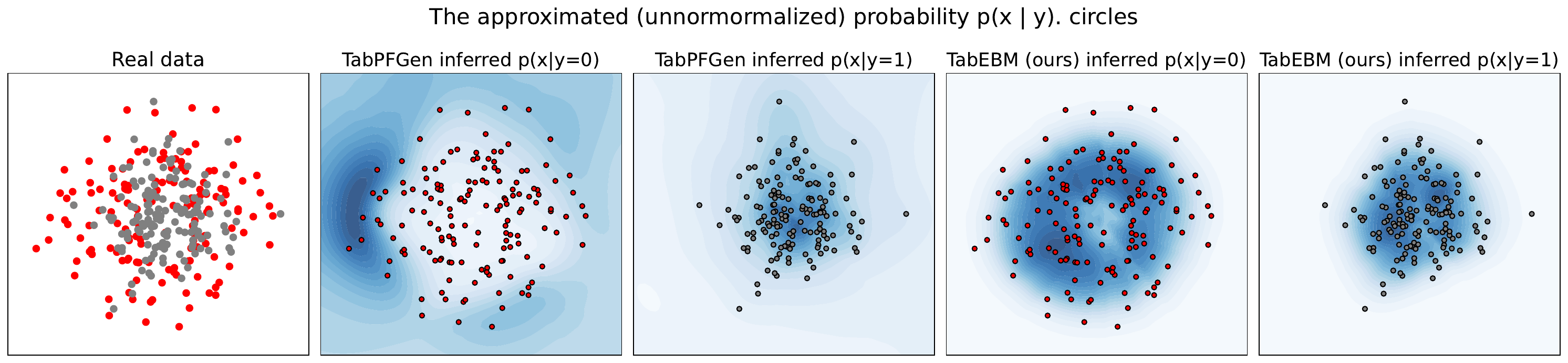}
}

\subfloat[Noise level 1]{
  \includegraphics[width=\textwidth, trim={0 0 0 40pt}, clip]{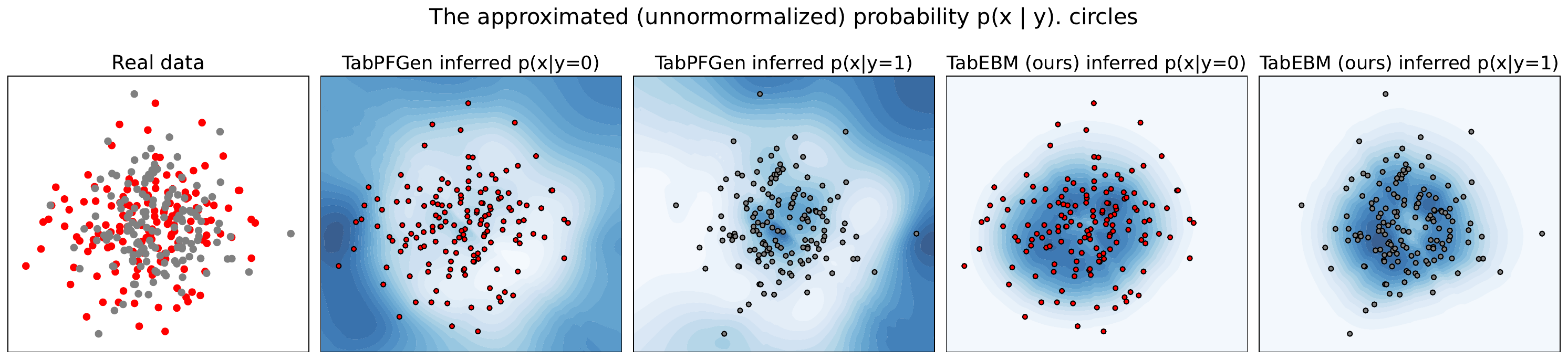}
}

\subfloat[Noise level 2]{
  \includegraphics[width=\textwidth, trim={0 0 0 40pt}, clip]{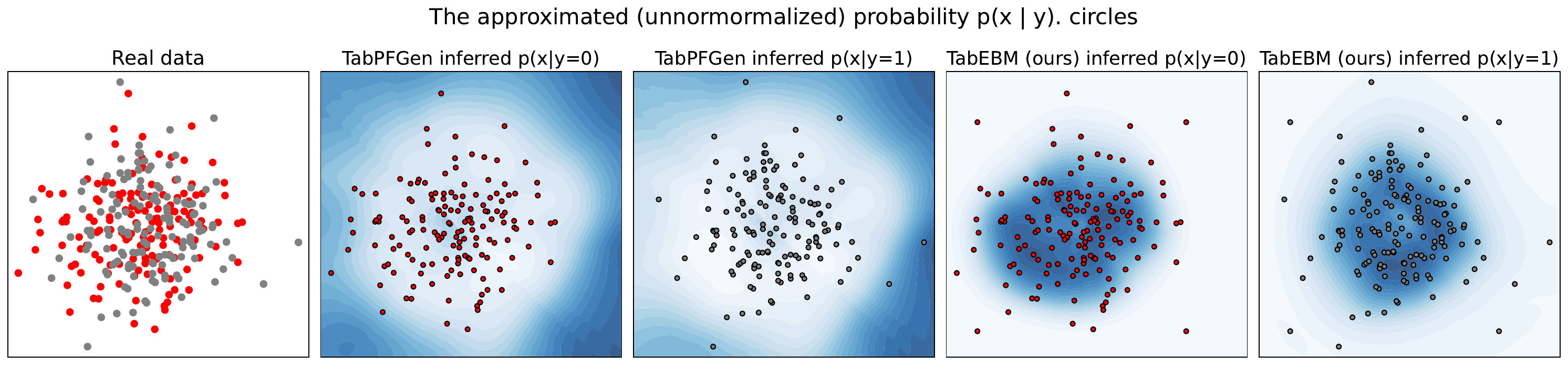}
}

\label{fig:tabpfgen-increasing-noise}
\caption{Evaluating the approximated class-conditional distributions on data with increasing noise levels. Darker blue indicates a higher assigned probability. TabPFGen uses a single shared energy-based model to infer the class-conditional distribution \( p(\rvx | y) \). As noise increases, TabPFGen's probability assignments vary significantly and end up assigning very high probabilities that are far from the real data. For instance, the areas of assigned probability for \( p(\rvx | y=1) \) completely flip when noise increases from 0.5 to 1. In contrast, our TabEBM uses class-specific energy models, resulting in robust inferred conditionals. TabEBM performs well even under very high noise (see \( p(\rvx | y=0) \) for noise level 2), while TabPFGen struggles.}
\end{figure}

\begin{figure}[htp]
\centering

\subfloat[Class ratio 150:150]{
  \includegraphics[width=\textwidth, trim={0 0 0 40pt}, clip]{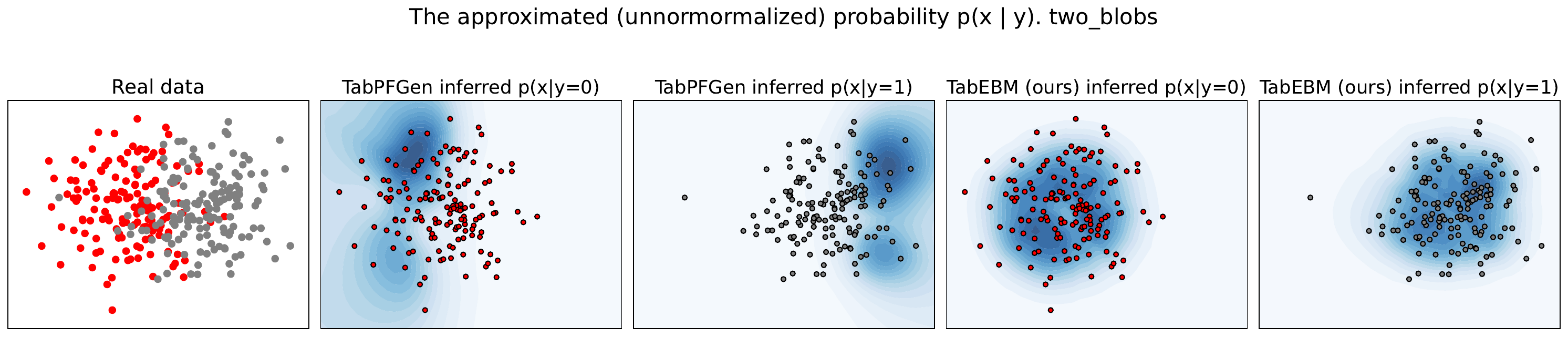}
}

\subfloat[Class ratio 50:250]{
  \includegraphics[width=\textwidth, trim={0 0 0 40pt}, clip]{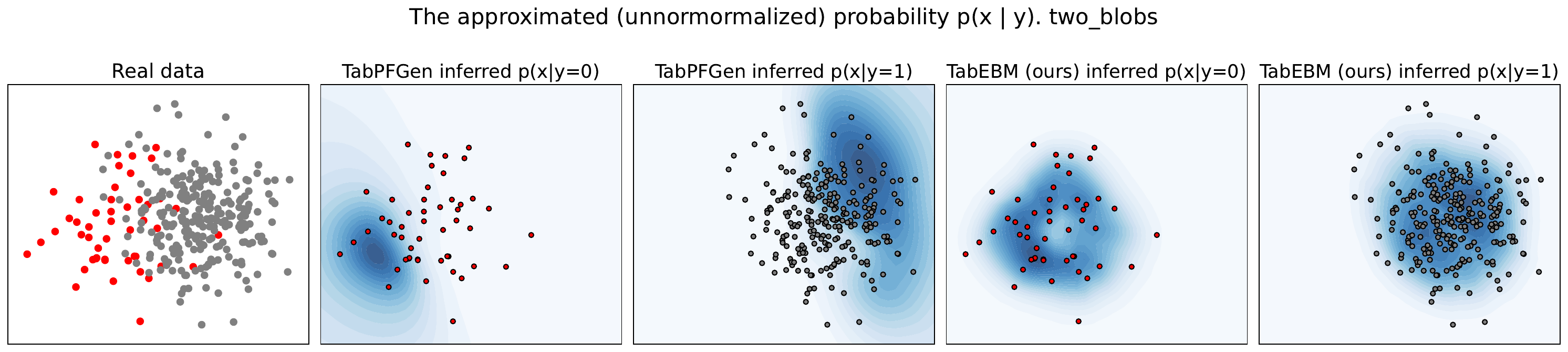}
}

\subfloat[Class ratio 25:275]{
  \includegraphics[width=\textwidth, trim={0 0 0 40pt}, clip]{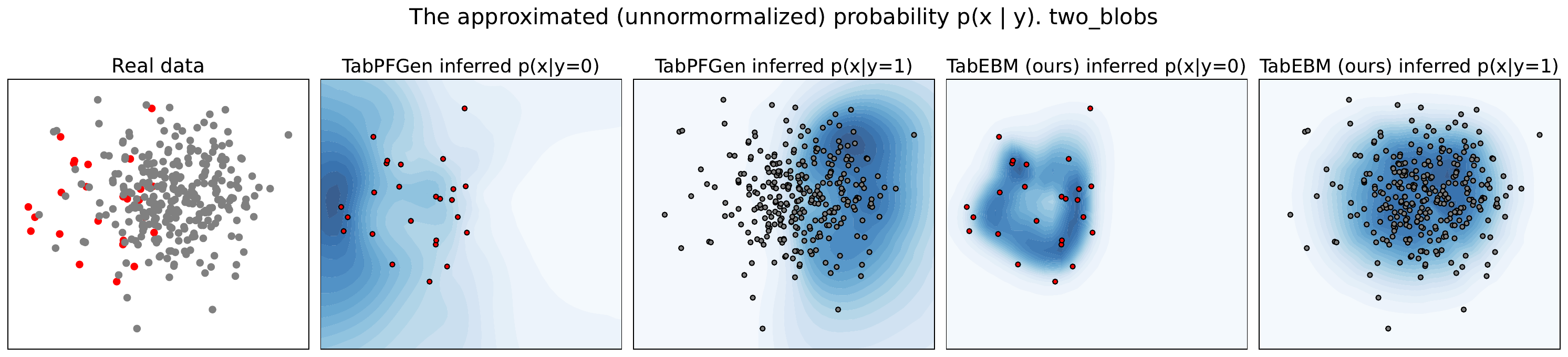}
}

\subfloat[Class ratio 10:290]{
  \includegraphics[width=\textwidth, trim={0 0 0 40pt}, clip]{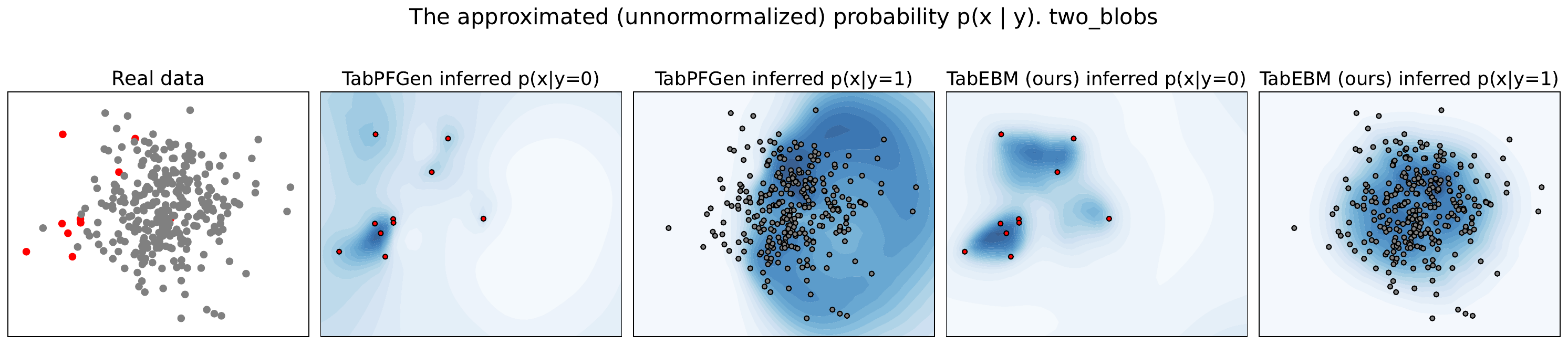}
}

\label{fig:tabpfgen-increasing-class-imbalance}
\caption{Evaluating the approximated class-conditional distributions on a toy dataset of 300 samples with varying class imbalances. The two clusters maintain their positions. Darker blue indicates a higher assigned probability. TabPFGen uses a single shared energy-based model to infer the class-conditional distribution \( p(\rvx | y) \). As class imbalance increases, TabPFGen starts assigning high probability in areas far from the real data, for instance, in the case of \( p(\rvx | y=1) \) for class ratio 10:290. In contrast, our TabEBM fits class-specific energy models only on the class-wise data \(\mathcal{X}_c = \{\rvx^{(i)} \mid y_i = c \}\). This results in very robust inferred conditional distributions even under heavy class imbalance (e.g., see that \( p(\rvx | y=1) \) remains relatively constant).}
\end{figure}

We showcase three limitations of current generative models: (1) \cref{fig:tabddpm-labeldistribution} shows that models approximating the joint distribution \( p(\rvx, y) \) may fail to preserve the stratification of the real data and even fail to generate samples from specific classes. (2) \cref{fig:tabpfgen-increasing-noise} evaluates the approximated class-conditional distributions \( p(\rvx \mid y) \) on data with increasing noise levels, and (3) \cref{fig:tabpfgen-increasing-class-imbalance} evaluates the approximated class-conditional distributions \( p(\rvx \mid y) \) on data with increasing class imbalance.

\FloatBarrier
\section{Extended Experimental Results}
\label{appendix:numerical_results}

\FloatBarrier
\subsection{Ablations on the distribution of the surrogate negative samples}
\label{appendix:ablations-negative-samples}

\FloatBarrier
\subsubsection{Ablations on placing the negative samples}
\label{appendix:ablation-placing-negative-samples}

\begin{figure}[!htbp]
    \centering
    \includegraphics[width=\linewidth, trim=0 0 0 2.5cm, clip]{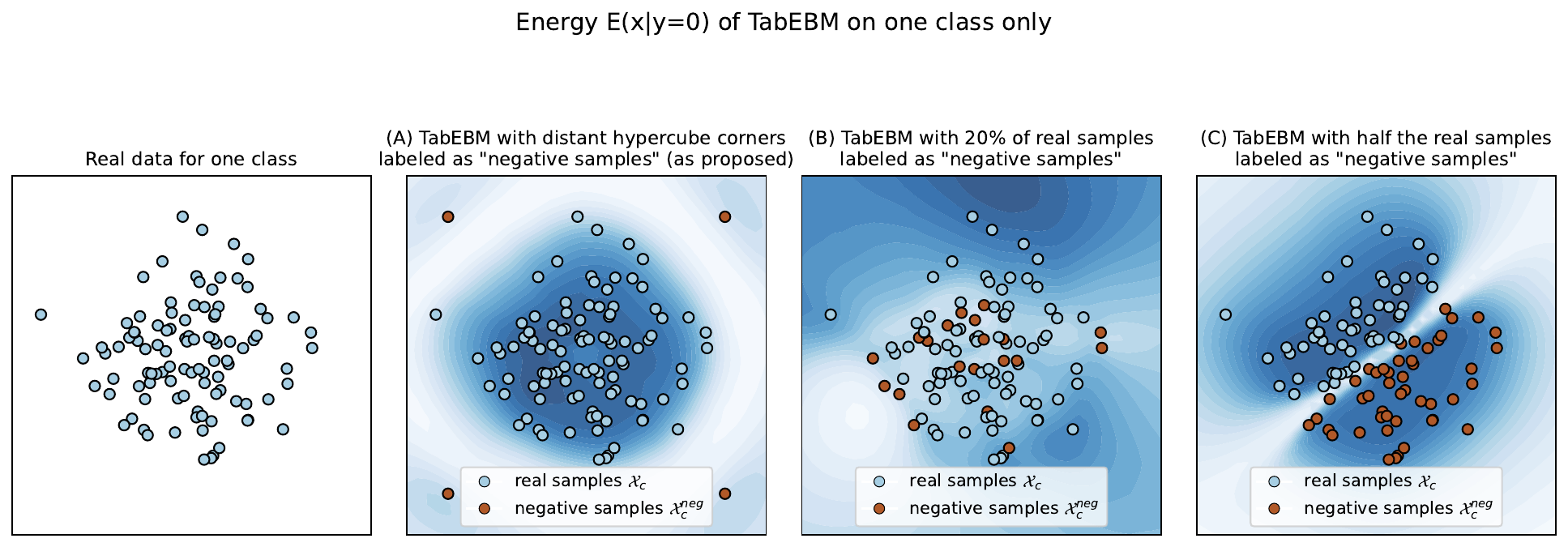}
    \label{fig:tabpfn-negative-samples-distribution}
    \caption{TabEBM energy \(E_c(x)\) for different choices of negative samples. The \textcolor{blue}{blue} region represents low energy, indicating high data density. In (A), TabEBM, with the proposed negative samples placed in a hypercube far from the data, infers an accurate energy surface, resulting in generated data close to the real points. In (B), labelling a random subset of the real data as negative samples leads to a completely inaccurate energy surface. In (C), labelling half of the real points as negative samples reduces density near the decision boundary, as TabPFN assigns low maximal logit due to the high uncertainty. In conclusion, placing negative samples far from the real data results in a robust energy surface.}
\end{figure}

\cref{fig:tabpfn-negative-samples-distribution} shows TabEBM's energy \(E_c(x)\) when varying the selection of the negative samples. TabEBM infers an accurate energy surface with distant negative samples, and the energy surface becomes inaccurate when negative samples resemble real samples. This occurs because TabPFN is uncertain when points of different classes are close, affecting its logits magnitude and making them unsuitable for density estimation.

\FloatBarrier
\subsubsection{Varying the number of negative samples}
\label{appendix:varying-number-negative-samples}

We evaluate the impact of the ratio $|\mathcal{X}^{\text{neg}}_c|:|\mathcal{X}_c|$ between the negative samples $\mathcal{X}^{\text{neg}}_c$ and the real samples $|\mathcal{X}_c|$. We vary $|\mathcal{X}^{\text{neg}}_c|$ while keeping $|\mathcal{X}_c|$ fixed, simulating both balanced and highly imbalanced scenarios. The negative samples are placed in random corners of the hypercube (as described in \cref{sec:method}), at five standard deviations in each direction (i.e., $\alpha^{\text{neg}}_{\text{dist}} = 5$). To ensure reliable outcomes, we maintained a consistent ratio across all classes, keeping the same proportion of negative samples for each class.

\cref{table-appendix:number-negative-samples} shows the results across six datasets with $N_{\text{real}} = 100$ real samples, demonstrating that TabEBM is robust to imbalances in the surrogate binary tasks. The column with $|\mathcal{X}^{\text{neg}}_c| = 4$ represents the TabEBM results from the main paper, where four negative samples were placed in the corners (as described in \cref{sec:method}). There are negligible differences in performance, and TabEBM consistently outperforms both the baseline and other generators (as shown in \cref{tab:test_acc_per_dataset_augmentation}).

\begin{table}[!htbp]
    \centering
    \caption{Evaluating the impact of varying the ratio $|\mathcal{X}^{\text{neg}}_c|:|\mathcal{X}_c|$. We show the test classification accuracy performance (\%) of TabEBM on data augmentation averaged over six datasets and ten repeats. TabEBM shows consistent performance and outperforms the baseline, regardless of the number of negative samples.}
    \label{table-appendix:number-negative-samples}
    \resizebox{\textwidth}{!}{
        \begin{tabular}{lcccccc}
        \toprule
         & \multicolumn{5}{c}{\textbf{TabEBM}} & \makecell[c]{Baseline\\(Real data)} \\
        \cmidrule(lr){2-6}
        \textbf{Ratio $|\mathcal{X}^{\text{neg}}_c| : |\mathcal{X}_c|$} & 0.1 & 0.2 & 0.5 & 1 & Fixed $|\mathcal{X}^{\text{neg}}_c| = 4$ & - \\
        \midrule
        biodeg  & 76.59$_{\pm\text{3.95}}$ & 76.54$_{\pm\text{3.95}}$ & 76.47$_{\pm\text{4.05}}$ & 76.81$_{\pm\text{3.58}}$ & 76.45$_{\pm\text{3.08}}$ & 76.69$_{\pm\text{2.70}}$ \\
        steel   & 92.71$_{\pm\text{7.46}}$ & 92.60$_{\pm\text{7.45}}$ & 92.79$_{\pm\text{7.50}}$ & 92.63$_{\pm\text{7.59}}$ & 92.71$_{\pm\text{7.57}}$ & 86.87$_{\pm\text{12.4}}$ \\
        stock   & 90.46$_{\pm\text{3.49}}$ & 90.41$_{\pm\text{3.65}}$ & 90.52$_{\pm\text{3.52}}$ & 90.31$_{\pm\text{3.63}}$ & 90.36$_{\pm\text{3.14}}$ & 89.07$_{\pm\text{3.71}}$ \\
        energy  & 31.20$_{\pm\text{6.22}}$ & 31.20$_{\pm\text{6.22}}$ & 30.89$_{\pm\text{5.83}}$ & 30.90$_{\pm\text{6.09}}$ & 31.24$_{\pm\text{5.53}}$ & 25.94$_{\pm\text{4.86}}$ \\
        collins & 13.06$_{\pm\text{2.88}}$ & 13.02$_{\pm\text{2.85}}$ & 13.05$_{\pm\text{2.89}}$ & 12.97$_{\pm\text{2.79}}$ & 13.07$_{\pm\text{2.51}}$ & 11.44$_{\pm\text{2.77}}$ \\
        texture & 85.91$_{\pm\text{6.92}}$ & 85.91$_{\pm\text{6.92}}$ & 85.94$_{\pm\text{6.76}}$ & 86.26$_{\pm\text{6.72}}$ & 86.01$_{\pm\text{7.36}}$ & 82.42$_{\pm\text{10.38}}$ \\
        \midrule
        \textbf{Average accuracy} & 64.99 & 64.95 & 64.94 & 64.98 & 64.97 & 62.07 \\
        \bottomrule
        \end{tabular}%
    }
\end{table}

\FloatBarrier
\subsubsection{Varying the distance of the negative samples}
\label{appendix:varying-distance-negative-samples}

We assess the effect of varying the distance of negative samples. We use TabEBM with four negative samples positioned randomly at the corners of the hypercube, as outlined in \cref{sec:method} (this corresponds to the experimental setup from the main paper). The distance of the negative samples, denoted as $\alpha^{\text{neg}}_{\text{dist}}$, is varied. \cref{table-appendix:distance-negative-samples} demonstrates that TabEBM remains generally robust to changes in this distance, with only small performance variations across different datasets. Importantly, using TabEBM for data augmentation consistently improves performance by approximately 3\% compared to the Baseline, regardless of the distance used.

\begin{table}[!htbp]
    \centering
    \caption{Evaluating the impact of varying the distance of the negative samples $\alpha^{\text{neg}}_{\text{dist}}$ across various datasets. We show the test classification accuracy performance (\%) of TabEBM on data augmentation averaged over six datasets and ten repeats. TabEBM is robust, and optional tuning of the negative samples could slightly improve performance.}
    \label{table-appendix:distance-negative-samples}
    \resizebox{\textwidth}{!}{
        \begin{tabular}{lccccccc}
        \toprule
         & \multicolumn{6}{c}{\textbf{TabEBM}} & \makecell[c]{Baseline\\(Real data)} \\
        \cmidrule(lr){2-7}
        \textbf{\makecell[l]{Per-dimension distance $\alpha_d$\\of the negative samples}} & 0.1 & 0.2 & 0.5 & 1 & 2 & 5 & - \\
        \midrule
        biodeg  & 76.72$_{\pm\text{3.33}}$ & 76.62$_{\pm\text{3.40}}$ & 77.12$_{\pm\text{2.60}}$ & 76.85$_{\pm\text{3.14}}$ & 76.50$_{\pm\text{3.93}}$ & 76.45$_{\pm\text{3.08}}$ & 76.69$_{\pm\text{2.70}}$ \\
        steel   & 93.97$_{\pm\text{5.76}}$ & 93.46$_{\pm\text{6.24}}$ & 93.00$_{\pm\text{6.92}}$ & 92.60$_{\pm\text{7.31}}$ & 92.68$_{\pm\text{7.38}}$ & 92.71$_{\pm\text{7.57}}$ & 86.87$_{\pm\text{12.4}}$ \\
        stock   & 90.42$_{\pm\text{3.46}}$ & 90.29$_{\pm\text{3.61}}$ & 90.56$_{\pm\text{3.46}}$ & 90.38$_{\pm\text{3.64}}$ & 90.43$_{\pm\text{3.56}}$ & 90.36$_{\pm\text{3.14}}$ & 89.07$_{\pm\text{3.71}}$ \\
        energy  & 31.73$_{\pm\text{6.21}}$ & 31.42$_{\pm\text{6.08}}$ & 31.86$_{\pm\text{6.12}}$ & 32.53$_{\pm\text{5.96}}$ & 31.65$_{\pm\text{6.06}}$ & 31.24$_{\pm\text{5.53}}$ & 25.94$_{\pm\text{4.86}}$ \\
        collins & 13.03$_{\pm\text{2.59}}$ & 12.92$_{\pm\text{2.60}}$ & 12.97$_{\pm\text{2.69}}$ & 13.03$_{\pm\text{2.84}}$ & 13.08$_{\pm\text{2.93}}$ & 13.07$_{\pm\text{2.51}}$ & 11.44$_{\pm\text{2.77}}$ \\
        texture & 85.62$_{\pm\text{7.41}}$ & 85.58$_{\pm\text{7.49}}$ & 85.50$_{\pm\text{7.65}}$ & 85.05$_{\pm\text{8.21}}$ & 85.20$_{\pm\text{7.95}}$ & 86.01$_{\pm\text{7.36}}$ & 82.42$_{\pm\text{10.38}}$ \\
        \midrule
        \textbf{Average accuracy} & 65.25 & 65.05 & 65.17 & 65.07 & 64.92 & 64.97 & 62.07 \\
        \bottomrule
        \end{tabular}%
    }
\end{table}

\FloatBarrier
\subsection{Ablations on the sensitivity to the hyperparameters of SGLD sampling}
\label{appendix:ablation_sgld}

We vary two key hyperparameters of SGLD on the ``biodeg'' binary dataset with $N_{\text{real}}=100$: the step size $\alpha_{\text{step}}$ and the noise scale $\alpha_{\text{noise}}$. \cref{table-appendix:sgld-sensitivity} shows that TabEBM remains stable with respect to these hyperparameters. Note that smaller values of $\alpha_{\text{noise}}$ are expected to perform better because SGLD sampling adds noise at each iteration (see Line 7 in \cref{algorithm:tabebm}), thus larger values of $\alpha_{\text{noise}}$ will hinder convergence of the SGLD sampler.

\begin{table}[!htbp]
    \centering
    \caption{Test classification accuracy (\%) of TabEBM (averaged over six downstream predictors) with different SGLD settings. Increasing $\alpha_{\text{noise}}$ (added at each SGLD step) is expected to degrade performance, as it causes the sampling to diverge further from the real data.}

    \label{table-appendix:sgld-sensitivity}
    \begin{tabular}{m{0.6cm}|c|c|c|c}
    \toprule
    \multirow{2}{*}{\rotatebox{90}{\begin{tabular}{c} $\alpha_{\text{noise}}$ \end{tabular}}} & \multicolumn{4}{c}{$\alpha_{\text{step}}$} \\
    \cmidrule(r){2-5}
    & 0.1 & 0.3 & 0.5 & 1.0 \\
    \midrule
    0.01 & 76.45 & \textbf{77.09} & 77.04 & 76.58 \\
    0.02 & 76.86 & 76.96 & 76.77 & 76.26 \\
    0.05 & 75.93 & 75.89 & 75.94 & \textbf{75.70} \\
    \bottomrule
    \end{tabular}
\end{table}

\FloatBarrier
\subsection{Distribution of Logits and Unnormalized Density in TabEBM}
\label{appendix:distribution-logits-tabebm}

\begin{figure}[!htbp]
    \centering
    \includegraphics[width=\linewidth]{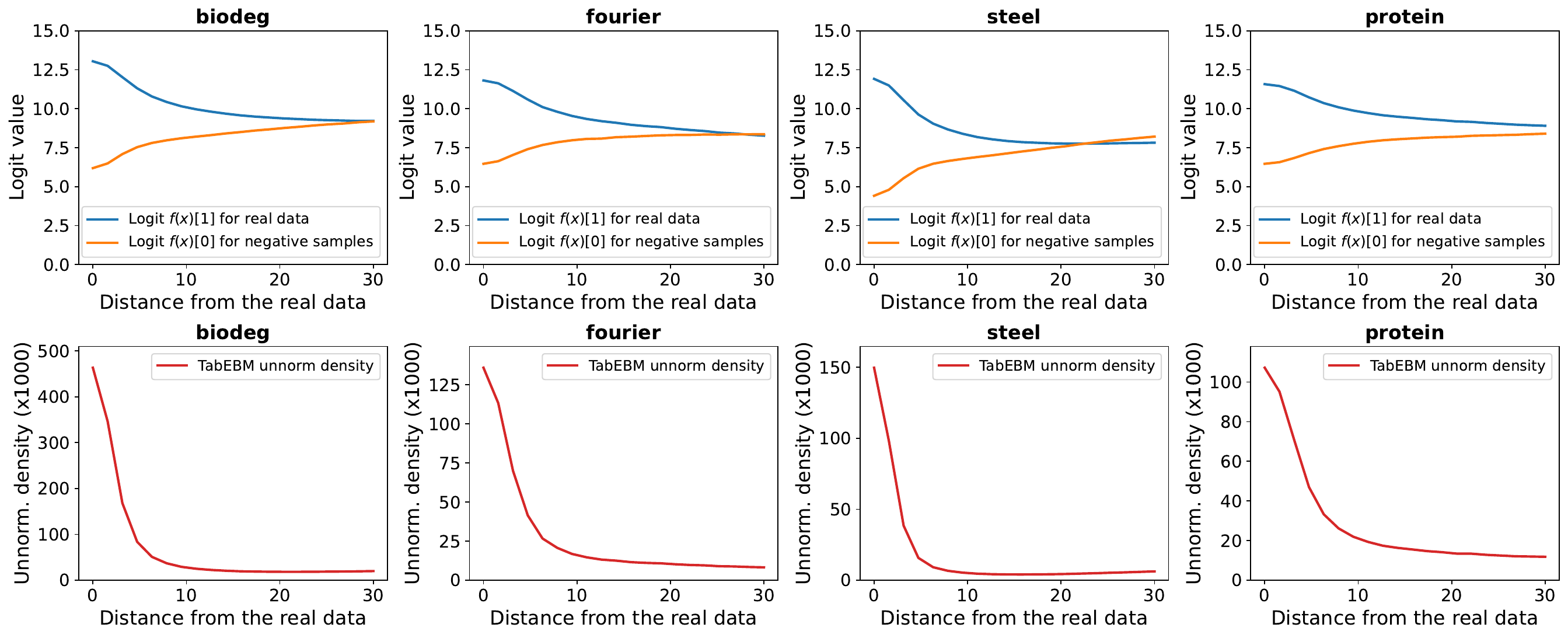}
    \caption{Additional results for \cref{sec:why-is-tabebm-effective-for-energy-estimation}. The logit distribution of TabPFN trained on our surrogate binary tasks across four datasets. Starting from the real samples, random points are selected at increasing distances (shown on the x-axis). The \textbf{top row} shows the logit distributions for the surrogate task. Close to the real data, TabPFN outputs a high logit value. As the distance increases, the logits converge due to increased predictive uncertainty, leading to equal class probabilities after applying softmax. Notably, across datasets, TabPFN's logits are always positive, have similar ranges, and maintain a relatively constant sum as distance increases. The \textbf{bottom row} TabEBM's unnormalized density, \( p_c(x) \propto \exp(-E_c(x)) \rightarrow p_c(x) \propto (\exp(f(x)[0]) + \exp(f(x)[1])) \). The density decreases significantly far from the data, becoming negligible. Because sampling using SGLD perform gradient ascent on the density, the TabEBM-generated samples will be similar when using one or both logits.}
\end{figure}

\newpage
\subsection{Complete Trade-off Figures with Error Bars}
\label{appendix:tabebm-tradeoff-figures-with-error-bars}

\begin{figure}[!htbp]
    \centering
    \subfloat{\includegraphics[width=0.4\textwidth]{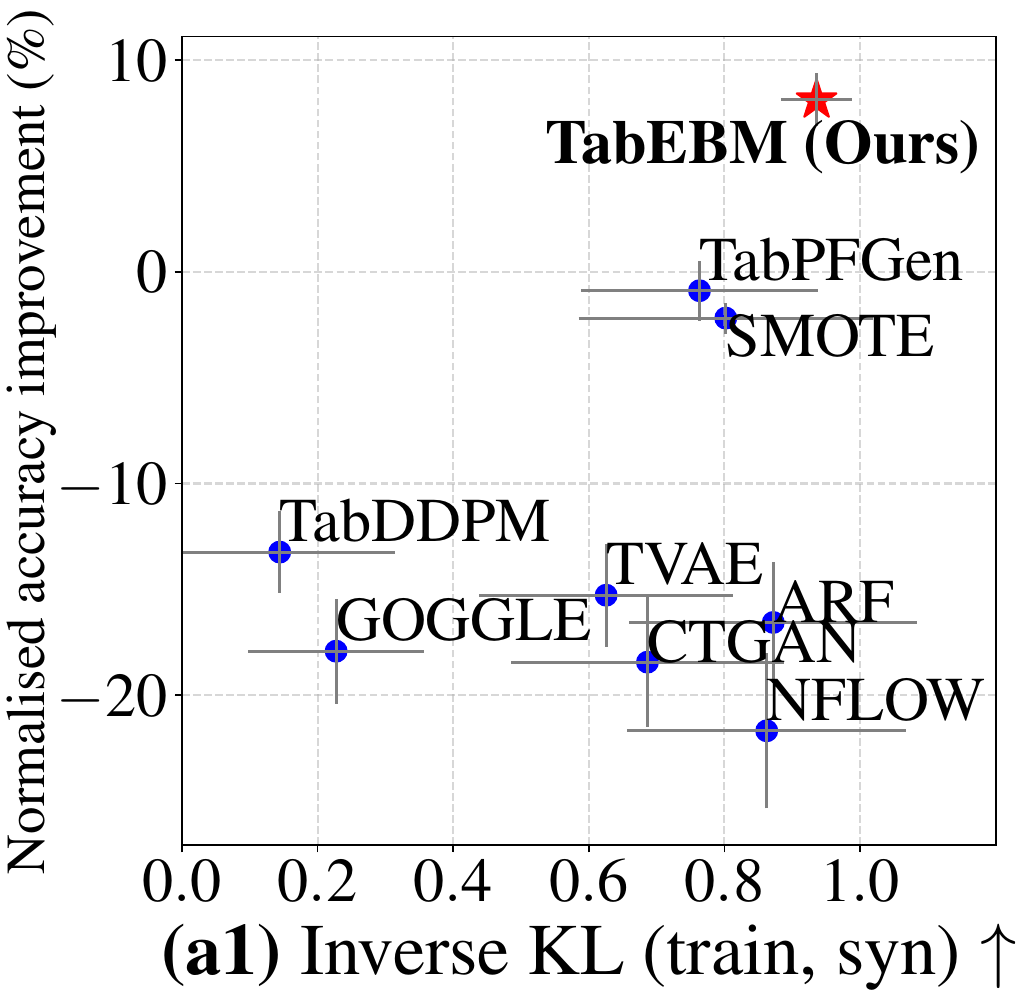}}
    \subfloat{\includegraphics[width=0.373\textwidth]{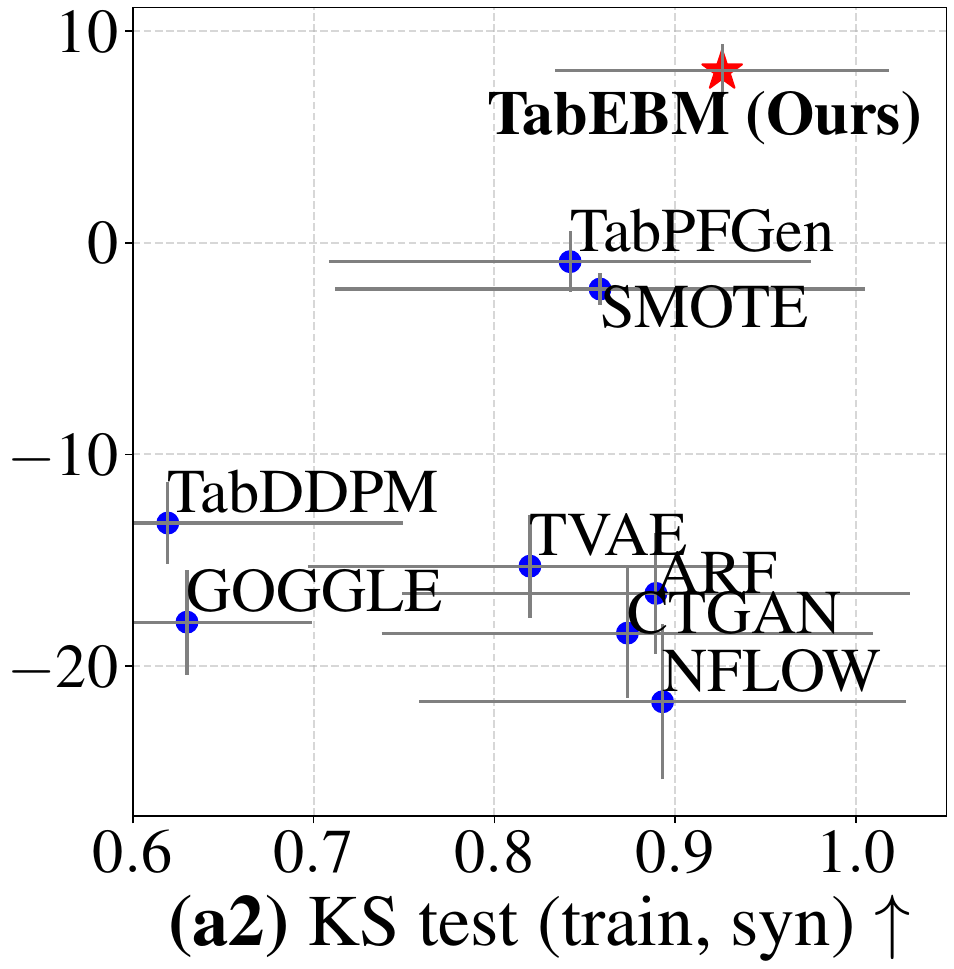}}
    \\
    \subfloat{\includegraphics[width=0.428\textwidth]{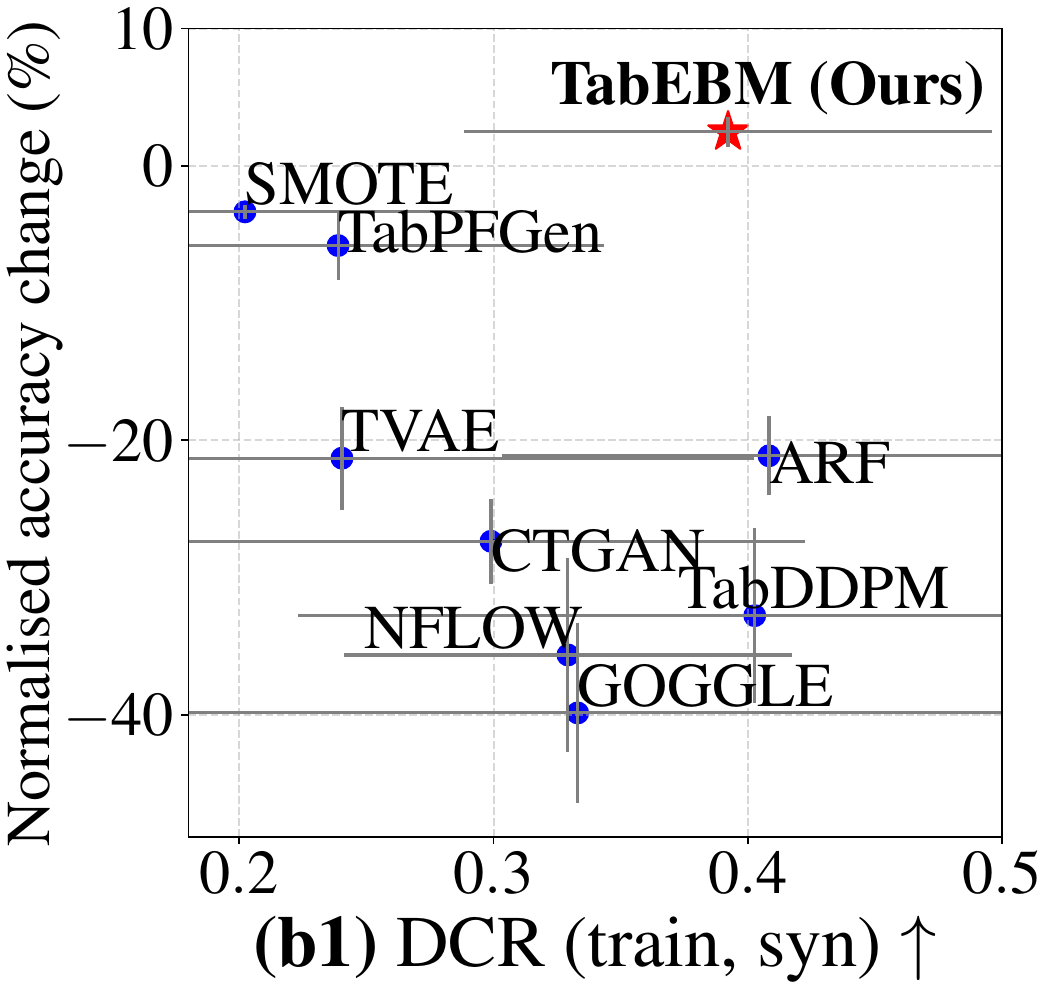}}
    \subfloat{\includegraphics[width=0.4\textwidth]{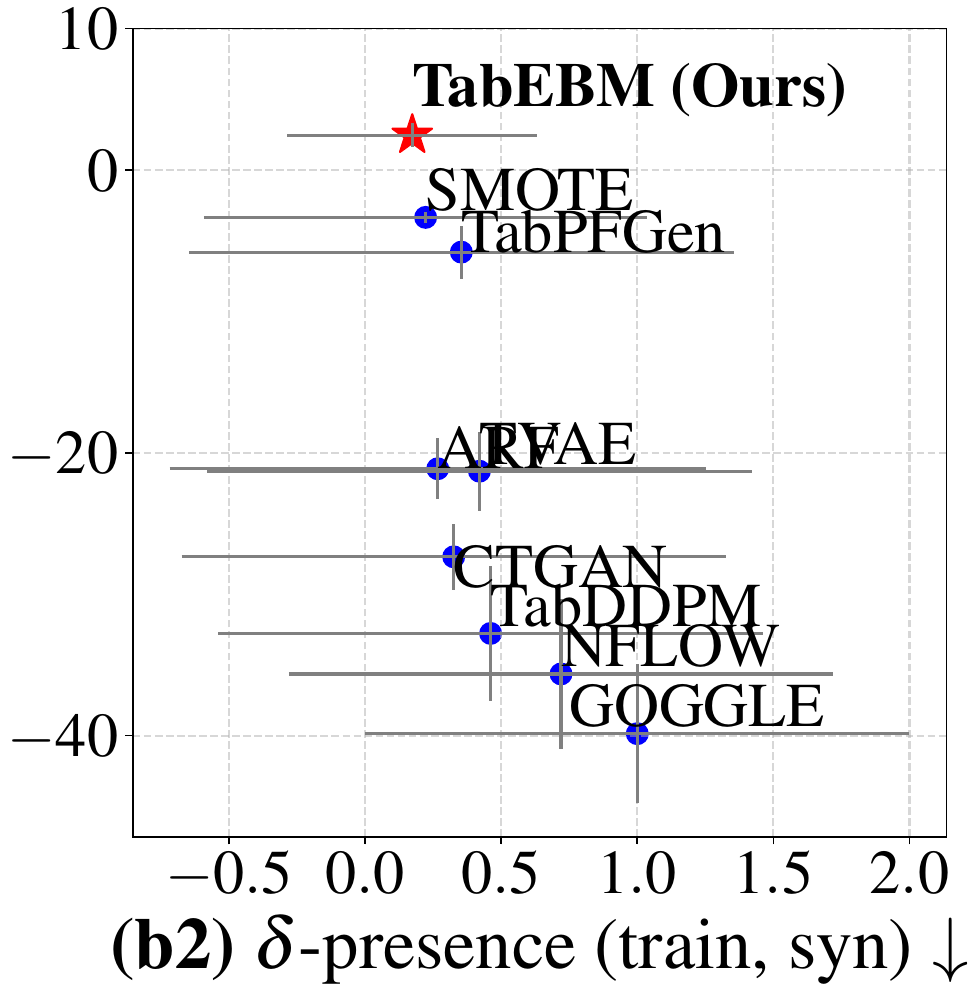}}
    
    \label{fig:acc_vs_fidelity_and_privacy_complete}
    \caption{ \textbf{(a1\&a2):} Median inverse KL and KS test vs.\ mean normalised balanced accuracy improvement (\%) between real train data and synthetic data. \textbf{(b1\&b2):} Median DCR and \( \delta-presence \) vs.\ mean normalised balanced accuracy change (\%) between real train data and synthetic data. Note that ``accuracy improvement'' is for data augmentation, and ``accuracy change'' is for data sharing. TabEBM generates high-fidelity synthetic data that can also be used for privacy preservation.}
    \vspace{-10pt}
\end{figure}
\clearpage

\FloatBarrier
\subsection{Results on Data Augmentation}
\label{appendix:res_utility}

\FloatBarrier
\subsubsection{Results on eight OpenML datasets.}
\begin{table}[hp]
\vspace{-10pt}
\centering
\caption{\textbf{Classification accuracy} (\%) of LR, comparing data augmentation on eight real-world tabular datasets with varied real data availability. We report the mean $\pm$ std balanced accuracy and average accuracy rank across datasets. A higher rank implies higher accuracy. Note that ``N/A'' denotes that a specific generator was not applicable or the downstream predictor failed to converge, and the rank is computed with the mean balanced accuracy of other methods. We \textbf{bold} the highest accuracy for each dataset of different sample sizes. TabEBM achieves the best overall performance against Baseline and benchmark generators.}
\resizebox{\textwidth}{!}{

\setlength{\tabcolsep}{5pt}
\begin{tabular}{m{0.2cm}lr|r|rrrrrrrr|r}

\toprule

\multicolumn{2}{l}{Datasets}  & $N_{\text{real}}$ & \makecell[r]{Baseline\\(Real data)} & SMOTE & TVAE & CTGAN & NFLOW & TabDDPM & ARF & GOGGLE & TabPFGen & \textbf{TabEBM} \\

\midrule

\multirow{24}{*}{\rotatebox{90}{\begin{tabular}{l} \textit{At most 10 classes} \end{tabular}}}& \multirow[c]{5}{*}{protein} & 20 & 36.33$_{\pm\text{3.04}}$ & N/A & 22.02$_{\pm\text{2.91}}$ & 21.04$_{\pm\text{4.76}}$ & 18.40$_{\pm\text{4.82}}$ & 18.77$_{\pm\text{3.84}}$ & 25.92$_{\pm\text{4.30}}$ & 36.61$_{\pm\text{2.53}}$ & \textbf{38.07}$_{\pm\text{1.25}}$ & 38.01$_{\pm\text{2.38}}$ \\

& & 50 & 62.14$_{\pm\text{3.77}}$ & 61.43$_{\pm\text{4.34}}$ & 37.04$_{\pm\text{2.79}}$ & 33.10$_{\pm\text{5.99}}$ & 31.25$_{\pm\text{4.21}}$ & 23.98$_{\pm\text{2.75}}$ & 43.64$_{\pm\text{5.07}}$ & 54.95$_{\pm\text{3.28}}$ & 63.00$_{\pm\text{3.69}}$ & \textbf{63.05}$_{\pm\text{3.84}}$ \\

& & 100 & 79.97$_{\pm\text{3.24}}$ & 79.53$_{\pm\text{3.37}}$ & 61.07$_{\pm\text{5.06}}$ & 55.44$_{\pm\text{1.92}}$ & 46.37$_{\pm\text{4.10}}$ & 45.55$_{\pm\text{4.24}}$ & 56.77$_{\pm\text{3.06}}$ & 67.25$_{\pm\text{4.50}}$ & \textbf{80.54}$_{\pm\text{3.27}}$ & 80.32$_{\pm\text{3.12}}$ \\

& & 200 & 91.53$_{\pm\text{1.58}}$ & 90.92$_{\pm\text{1.81}}$ & 77.43$_{\pm\text{2.75}}$ & 71.27$_{\pm\text{3.07}}$ & 66.16$_{\pm\text{4.31}}$ & 66.37$_{\pm\text{3.42}}$ & 70.52$_{\pm\text{2.17}}$ & 76.30$_{\pm\text{3.70}}$ & \textbf{91.69}$_{\pm\text{1.66}}$ & 91.34$_{\pm\text{1.77}}$ \\

& & 500 & 97.86$_{\pm\text{0.83}}$ & 97.69$_{\pm\text{0.80}}$ & 90.77$_{\pm\text{0.93}}$ & 89.05$_{\pm\text{1.52}}$ & 85.09$_{\pm\text{1.99}}$ & 83.58$_{\pm\text{2.22}}$ & 88.55$_{\pm\text{1.54}}$ & 90.64$_{\pm\text{0.81}}$ & \textbf{97.97}$_{\pm\text{0.61}}$ & 97.88$_{\pm\text{0.86}}$ \\

\cmidrule{2-13}

& \multirow[c]{5}{*}{fourier} & 20 & 42.90$_{\pm\text{5.30}}$ & N/A & 22.46$_{\pm\text{5.88}}$ & 16.00$_{\pm\text{4.70}}$ & 15.48$_{\pm\text{3.79}}$ & 13.58$_{\pm\text{4.30}}$ & 22.04$_{\pm\text{4.42}}$ & 15.80$_{\pm\text{4.15}}$ & \textbf{44.67}$_{\pm\text{8.85}}$ & 43.02$_{\pm\text{5.14}}$ \\

& & 50 & \textbf{60.62}$_{\pm\text{1.64}}$ & 58.40$_{\pm\text{1.95}}$ & 33.42$_{\pm\text{2.98}}$ & 31.18$_{\pm\text{5.47}}$ & 28.70$_{\pm\text{3.74}}$ & 26.18$_{\pm\text{3.80}}$ & 39.04$_{\pm\text{3.12}}$ & 40.00$_{\pm\text{4.97}}$ & 60.07$_{\pm\text{2.14}}$ & 60.36$_{\pm\text{1.55}}$ \\

& & 100 & \textbf{67.76}$_{\pm\text{2.49}}$ & 65.84$_{\pm\text{2.35}}$ & 41.36$_{\pm\text{2.85}}$ & 40.32$_{\pm\text{3.49}}$ & 40.32$_{\pm\text{5.82}}$ & 41.44$_{\pm\text{5.02}}$ & 47.90$_{\pm\text{3.74}}$ & 39.78$_{\pm\text{3.99}}$ & 67.40$_{\pm\text{1.51}}$ & 67.44$_{\pm\text{2.46}}$ \\

& & 200 & \textbf{73.13}$_{\pm\text{2.41}}$ & 71.56$_{\pm\text{2.67}}$ & 54.76$_{\pm\text{3.46}}$ & 55.00$_{\pm\text{3.72}}$ & 52.40$_{\pm\text{3.18}}$ & 58.08$_{\pm\text{3.52}}$ & 58.48$_{\pm\text{2.08}}$ & 50.98$_{\pm\text{2.68}}$ & 70.30$_{\pm\text{2.91}}$ & 72.38$_{\pm\text{3.01}}$ \\

& & 500 & 77.44$_{\pm\text{1.20}}$ & 76.42$_{\pm\text{1.28}}$ & 68.28$_{\pm\text{2.12}}$ & 70.18$_{\pm\text{1.89}}$ & 68.12$_{\pm\text{1.62}}$ & 72.36$_{\pm\text{1.65}}$ & 71.54$_{\pm\text{1.95}}$ & 69.48$_{\pm\text{1.71}}$ & 76.52$_{\pm\text{1.69}}$ & \textbf{77.50}$_{\pm\text{2.14}}$ \\

\cmidrule{2-13}

& \multirow[c]{5}{*}{biodeg} & 20 & \textbf{71.34}$_{\pm\text{5.63}}$ & 70.10$_{\pm\text{5.49}}$ & 70.16$_{\pm\text{5.75}}$ & 58.17$_{\pm\text{8.00}}$ & 58.05$_{\pm\text{9.91}}$ & 49.99$_{\pm\text{5.88}}$ & 62.61$_{\pm\text{6.45}}$ & 69.47$_{\pm\text{6.00}}$ & 70.76$_{\pm\text{3.95}}$ & 71.24$_{\pm\text{4.85}}$ \\

& & 50 & 76.35$_{\pm\text{2.88}}$ & 75.69$_{\pm\text{3.03}}$ & 73.63$_{\pm\text{2.64}}$ & 67.44$_{\pm\text{3.83}}$ & 62.87$_{\pm\text{7.30}}$ & 49.44$_{\pm\text{2.63}}$ & 74.44$_{\pm\text{2.77}}$ & 71.75$_{\pm\text{5.27}}$ & 75.68$_{\pm\text{2.31}}$ & \textbf{76.41}$_{\pm\text{2.93}}$ \\

& & 100 & \textbf{78.91}$_{\pm\text{1.40}}$ & 78.39$_{\pm\text{1.53}}$ & 77.09$_{\pm\text{2.80}}$ & 74.89$_{\pm\text{2.54}}$ & 68.62$_{\pm\text{5.21}}$ & 55.61$_{\pm\text{3.56}}$ & 75.62$_{\pm\text{2.77}}$ & 72.45$_{\pm\text{3.31}}$ & 77.92$_{\pm\text{2.41}}$ & 78.34$_{\pm\text{2.18}}$ \\

& & 200 & \textbf{82.00}$_{\pm\text{1.47}}$ & 81.42$_{\pm\text{1.39}}$ & 80.07$_{\pm\text{1.82}}$ & 78.56$_{\pm\text{3.43}}$ & 72.35$_{\pm\text{1.72}}$ & 59.06$_{\pm\text{4.65}}$ & 78.03$_{\pm\text{1.95}}$ & 73.73$_{\pm\text{2.09}}$ & 81.24$_{\pm\text{1.71}}$ & 81.43$_{\pm\text{1.78}}$ \\

& & 500 & \textbf{83.83}$_{\pm\text{0.57}}$ & 83.74$_{\pm\text{0.90}}$ & 81.69$_{\pm\text{0.82}}$ & 82.12$_{\pm\text{1.17}}$ & 78.06$_{\pm\text{2.13}}$ & 66.86$_{\pm\text{5.43}}$ & 81.47$_{\pm\text{0.93}}$ & 77.98$_{\pm\text{1.27}}$ & 83.43$_{\pm\text{0.82}}$ & 83.10$_{\pm\text{0.98}}$ \\

\cmidrule{2-13}

& \multirow[c]{5}{*}{steel} & 20 & 63.66$_{\pm\text{8.98}}$ & 57.88$_{\pm\text{5.72}}$ & 60.27$_{\pm\text{7.47}}$ & 57.90$_{\pm\text{4.45}}$ & 53.10$_{\pm\text{7.28}}$ & 54.20$_{\pm\text{6.99}}$ & 55.41$_{\pm\text{4.92}}$ & 53.29$_{\pm\text{4.31}}$ & 66.81$_{\pm\text{9.74}}$ & \textbf{67.03}$_{\pm\text{9.35}}$ \\

& & 50 & 87.91$_{\pm\text{5.88}}$ & 69.01$_{\pm\text{6.60}}$ & 66.22$_{\pm\text{3.63}}$ & 66.22$_{\pm\text{5.77}}$ & 57.05$_{\pm\text{5.51}}$ & 57.46$_{\pm\text{8.48}}$ & 64.81$_{\pm\text{4.64}}$ & 57.20$_{\pm\text{5.19}}$ & \textbf{93.63}$_{\pm\text{4.78}}$ & 92.20$_{\pm\text{4.81}}$ \\

& & 100 & 98.85$_{\pm\text{1.20}}$ & 82.67$_{\pm\text{4.30}}$ & 74.33$_{\pm\text{3.85}}$ & 70.49$_{\pm\text{5.35}}$ & 65.09$_{\pm\text{7.30}}$ & 52.77$_{\pm\text{7.06}}$ & 67.85$_{\pm\text{4.94}}$ & 61.62$_{\pm\text{4.05}}$ & \textbf{99.24}$_{\pm\text{0.82}}$ & 99.21$_{\pm\text{0.86}}$ \\

& & 200 & 99.43$_{\pm\text{0.58}}$ & 87.18$_{\pm\text{3.06}}$ & 82.77$_{\pm\text{3.21}}$ & 80.34$_{\pm\text{2.93}}$ & 70.49$_{\pm\text{5.27}}$ & 72.99$_{\pm\text{13.98}}$ & 80.27$_{\pm\text{7.32}}$ & 64.52$_{\pm\text{2.16}}$ & 99.45$_{\pm\text{0.69}}$ & \textbf{99.51}$_{\pm\text{0.69}}$ \\

& & 500 & 99.75$_{\pm\text{0.29}}$ & 96.63$_{\pm\text{2.11}}$ & 94.59$_{\pm\text{2.98}}$ & 96.32$_{\pm\text{1.52}}$ & 84.15$_{\pm\text{2.69}}$ & 98.07$_{\pm\text{1.37}}$ & 95.35$_{\pm\text{2.06}}$ & 70.11$_{\pm\text{2.58}}$ & 99.84$_{\pm\text{0.20}}$ & 99.84$_{\pm\text{0.20}}$ \\

\cmidrule{2-13}

& \multirow[c]{4}{*}{stock} & 20 & 77.99$_{\pm\text{4.40}}$ & 80.45$_{\pm\text{3.98}}$ & 74.21$_{\pm\text{6.36}}$ & 59.20$_{\pm\text{12.69}}$ & 72.50$_{\pm\text{7.92}}$ & 72.09$_{\pm\text{9.75}}$ & 69.04$_{\pm\text{6.25}}$ & \textbf{80.59}$_{\pm\text{3.59}}$ & 79.54$_{\pm\text{4.46}}$ & 80.39$_{\pm\text{3.42}}$ \\

& & 50 & 80.68$_{\pm\text{2.65}}$ & 81.49$_{\pm\text{2.95}}$ & 76.41$_{\pm\text{3.95}}$ & 72.95$_{\pm\text{2.17}}$ & 75.41$_{\pm\text{6.00}}$ & 78.44$_{\pm\text{4.40}}$ & 76.91$_{\pm\text{2.36}}$ & 75.49$_{\pm\text{5.31}}$ & \textbf{82.37}$_{\pm\text{3.20}}$ & 82.21$_{\pm\text{2.60}}$ \\

& & 100 & 82.11$_{\pm\text{1.11}}$ & \textbf{83.86}$_{\pm\text{1.97}}$ & 79.85$_{\pm\text{2.79}}$ & 78.47$_{\pm\text{2.71}}$ & 76.99$_{\pm\text{3.49}}$ & 80.82$_{\pm\text{3.57}}$ & 78.89$_{\pm\text{2.36}}$ & 77.65$_{\pm\text{2.60}}$ & 83.67$_{\pm\text{1.60}}$ & 83.52$_{\pm\text{1.76}}$ \\

& & 200 & 82.18$_{\pm\text{0.81}}$ & \textbf{84.29}$_{\pm\text{1.19}}$ & 79.24$_{\pm\text{2.82}}$ & 79.86$_{\pm\text{2.42}}$ & 76.49$_{\pm\text{1.37}}$ & 80.21$_{\pm\text{2.13}}$ & 78.87$_{\pm\text{2.46}}$ & 76.91$_{\pm\text{1.04}}$ & 83.75$_{\pm\text{1.53}}$ & 84.17$_{\pm\text{1.42}}$ \\

\midrule

\multirow{9}{*}{\rotatebox{90}{\begin{tabular}{l} \textit{More than 10 classes} \end{tabular}}}& \multirow[c]{3}{*}{energy} & 50 & \textbf{22.22}$_{\pm\text{2.36}}$ & N/A & 10.11$_{\pm\text{2.20}}$ & 9.58$_{\pm\text{3.15}}$ & 7.70$_{\pm\text{1.83}}$ & 8.20$_{\pm\text{2.01}}$ & 10.51$_{\pm\text{1.28}}$ & 17.10$_{\pm\text{5.03}}$ & N/A & 21.66$_{\pm\text{1.54}}$ \\

& & 100 & 24.00$_{\pm\text{2.30}}$ & N/A & 13.80$_{\pm\text{2.23}}$ & 13.01$_{\pm\text{1.71}}$ & 12.14$_{\pm\text{1.87}}$ & 10.79$_{\pm\text{3.19}}$ & 15.65$_{\pm\text{2.40}}$ & 14.45$_{\pm\text{2.90}}$ & N/A & \textbf{28.10}$_{\pm\text{2.19}}$ \\

& & 200 & 29.37$_{\pm\text{2.63}}$ & N/A & 16.39$_{\pm\text{2.68}}$ & 16.56$_{\pm\text{3.58}}$ & 16.78$_{\pm\text{3.15}}$ & 18.11$_{\pm\text{1.71}}$ & 20.10$_{\pm\text{2.48}}$ & 20.92$_{\pm\text{2.79}}$ & N/A & \textbf{34.38}$_{\pm\text{2.60}}$ \\

\cmidrule{2-13}

& \multirow[c]{2}{*}{collins} & 100 & \textbf{14.28}$_{\pm\text{1.63}}$ & N/A & 10.57$_{\pm\text{1.72}}$ & 8.69$_{\pm\text{1.17}}$ & 9.59$_{\pm\text{1.35}}$ & 13.31$_{\pm\text{1.67}}$ & 8.69$_{\pm\text{1.80}}$ & 12.08$_{\pm\text{1.56}}$ & N/A & 14.01$_{\pm\text{2.55}}$ \\

& & 200 & 19.20$_{\pm\text{1.71}}$ & \textbf{19.39}$_{\pm\text{1.88}}$ & 16.03$_{\pm\text{1.74}}$ & 11.64$_{\pm\text{1.76}}$ & 10.97$_{\pm\text{1.46}}$ & 17.06$_{\pm\text{1.51}}$ & 11.31$_{\pm\text{1.58}}$ & 17.80$_{\pm\text{1.21}}$ & N/A & 19.33$_{\pm\text{1.55}}$ \\

\cmidrule{2-13}

& \multirow[c]{4}{*}{texture} & 50 & 86.56$_{\pm\text{2.96}}$ & 86.93$_{\pm\text{2.77}}$ & 55.01$_{\pm\text{5.77}}$ & 42.17$_{\pm\text{6.36}}$ & 44.63$_{\pm\text{5.41}}$ & 60.07$_{\pm\text{10.11}}$ & 44.46$_{\pm\text{6.63}}$ & 77.68$_{\pm\text{4.33}}$ & N/A & \textbf{88.54}$_{\pm\text{2.88}}$ \\

& & 100 & 94.07$_{\pm\text{1.70}}$ & 93.87$_{\pm\text{1.82}}$ & 65.36$_{\pm\text{4.49}}$ & 60.07$_{\pm\text{6.81}}$ & 60.76$_{\pm\text{5.18}}$ & 73.16$_{\pm\text{5.11}}$ & 64.69$_{\pm\text{4.79}}$ & 84.13$_{\pm\text{1.97}}$ & N/A & \textbf{94.38}$_{\pm\text{1.24}}$ \\

& & 200 & \textbf{96.65}$_{\pm\text{1.24}}$ & 96.53$_{\pm\text{1.33}}$ & 75.91$_{\pm\text{5.58}}$ & 80.02$_{\pm\text{5.13}}$ & 77.07$_{\pm\text{3.89}}$ & 86.24$_{\pm\text{3.62}}$ & 85.90$_{\pm\text{2.78}}$ & 85.94$_{\pm\text{2.88}}$ & N/A & 96.53$_{\pm\text{1.27}}$ \\

& & 500 & 98.03$_{\pm\text{0.36}}$ & \textbf{98.05}$_{\pm\text{0.23}}$ & 91.87$_{\pm\text{0.93}}$ & 92.93$_{\pm\text{1.78}}$ & 90.01$_{\pm\text{1.80}}$ & 93.92$_{\pm\text{0.81}}$ & 94.83$_{\pm\text{0.89}}$ & 91.72$_{\pm\text{1.49}}$ & N/A & 97.75$_{\pm\text{0.42}}$ \\

\midrule
\rowcolor{Gainsboro!60}
\multicolumn{3}{l|}{\textbf{Average rank}} & 2.36$_{\pm\text{1.14}}$ & 3.45$_{\pm\text{1.35}}$ & 6.52$_{\pm\text{1.48}}$ & 7.53$_{\pm\text{1.42}}$ & 9.08$_{\pm\text{0.77}}$ & 7.61$_{\pm\text{2.33}}$ & 6.70$_{\pm\text{1.47}}$ & 6.67$_{\pm\text{2.53}}$ & 3.17$_{\pm\text{1.81}}$ & \textbf{1.92}$_{\pm\text{0.75}}$ \\

\bottomrule

\end{tabular}
}
\end{table}
\clearpage

\begin{table}[p]
\centering
\caption{\textbf{Classification accuracy} (\%) of KNN, comparing data augmentation on eight real-world tabular datasets with varied real data availability. We report the mean $\pm$ std balanced accuracy and average accuracy rank across datasets. A higher rank implies higher accuracy. Note that ``N/A'' denotes that a specific generator was not applicable or the downstream predictor failed to converge, and the rank is computed with the mean balanced accuracy of other methods. We \textbf{bold} the highest accuracy for each dataset of different sample sizes. TabEBM achieves the best overall performance against Baseline and benchmark generators.}
\resizebox{\textwidth}{!}{

\setlength{\tabcolsep}{5pt}
\begin{tabular}{m{0.2cm}lr|r|rrrrrrrr|r}

\toprule

\multicolumn{2}{l}{Datasets}  & $N_{\text{real}}$ & \makecell[r]{Baseline\\(Real data)} & SMOTE & TVAE & CTGAN & NFLOW & TabDDPM & ARF & GOGGLE & TabPFGen & \textbf{TabEBM} \\

\midrule

\multirow{24}{*}{\rotatebox{90}{\begin{tabular}{l} \textit{At most 10 classes} \end{tabular}}}& \multirow[c]{5}{*}{protein} & 20 & 21.34$_{\pm\text{2.93}}$ & N/A & 21.78$_{\pm\text{2.06}}$ & 21.18$_{\pm\text{4.22}}$ & 21.30$_{\pm\text{1.90}}$ & 22.00$_{\pm\text{2.70}}$ & 22.69$_{\pm\text{3.86}}$ & 16.99$_{\pm\text{3.45}}$ & \textbf{35.78}$_{\pm\text{4.46}}$ & 35.76$_{\pm\text{4.37}}$ \\

& & 50 & 36.41$_{\pm\text{4.33}}$ & \textbf{55.24}$_{\pm\text{3.81}}$ & 35.85$_{\pm\text{2.50}}$ & 36.13$_{\pm\text{4.24}}$ & 35.40$_{\pm\text{4.27}}$ & 36.77$_{\pm\text{4.06}}$ & 36.84$_{\pm\text{4.05}}$ & 31.02$_{\pm\text{4.11}}$ & 53.38$_{\pm\text{3.53}}$ & 53.49$_{\pm\text{3.30}}$ \\

& & 100 & 50.17$_{\pm\text{3.11}}$ & \textbf{70.11}$_{\pm\text{2.82}}$ & 51.97$_{\pm\text{2.84}}$ & 50.61$_{\pm\text{3.15}}$ & 50.62$_{\pm\text{3.27}}$ & 50.63$_{\pm\text{3.55}}$ & 50.36$_{\pm\text{3.44}}$ & 44.70$_{\pm\text{2.22}}$ & 67.99$_{\pm\text{2.43}}$ & 68.27$_{\pm\text{2.51}}$ \\

& & 200 & 65.84$_{\pm\text{2.78}}$ & 80.43$_{\pm\text{2.44}}$ & 65.52$_{\pm\text{2.96}}$ & 66.05$_{\pm\text{2.74}}$ & 66.14$_{\pm\text{2.42}}$ & 67.50$_{\pm\text{2.57}}$ & 66.52$_{\pm\text{3.16}}$ & 63.92$_{\pm\text{3.26}}$ & 79.94$_{\pm\text{2.26}}$ & \textbf{80.55}$_{\pm\text{2.02}}$ \\

& & 500 & 85.63$_{\pm\text{1.41}}$ & 90.92$_{\pm\text{1.42}}$ & 87.08$_{\pm\text{1.86}}$ & 85.77$_{\pm\text{1.43}}$ & 85.51$_{\pm\text{1.50}}$ & 86.47$_{\pm\text{1.40}}$ & 85.87$_{\pm\text{1.63}}$ & 85.64$_{\pm\text{1.94}}$ & 91.32$_{\pm\text{1.09}}$ & \textbf{91.67}$_{\pm\text{1.11}}$ \\

\cmidrule{2-13}

& \multirow[c]{5}{*}{fourier} & 20 & 18.06$_{\pm\text{3.30}}$ & N/A & 26.56$_{\pm\text{4.92}}$ & 24.88$_{\pm\text{3.66}}$ & 19.80$_{\pm\text{3.77}}$ & 19.30$_{\pm\text{3.53}}$ & 23.42$_{\pm\text{3.45}}$ & 18.78$_{\pm\text{2.17}}$ & 41.08$_{\pm\text{6.56}}$ & \textbf{42.78}$_{\pm\text{5.83}}$ \\

& & 50 & 48.00$_{\pm\text{2.47}}$ & \textbf{60.38}$_{\pm\text{1.67}}$ & 39.86$_{\pm\text{3.73}}$ & 46.82$_{\pm\text{3.52}}$ & 43.56$_{\pm\text{3.45}}$ & 49.54$_{\pm\text{2.78}}$ & 42.98$_{\pm\text{2.80}}$ & 28.12$_{\pm\text{2.75}}$ & 59.50$_{\pm\text{1.99}}$ & 58.54$_{\pm\text{1.86}}$ \\

& & 100 & 58.36$_{\pm\text{3.26}}$ & \textbf{66.96}$_{\pm\text{2.47}}$ & 48.44$_{\pm\text{4.14}}$ & 53.94$_{\pm\text{3.47}}$ & 53.50$_{\pm\text{2.54}}$ & 60.80$_{\pm\text{4.28}}$ & 52.74$_{\pm\text{3.15}}$ & 35.70$_{\pm\text{2.40}}$ & 63.88$_{\pm\text{2.53}}$ & 65.08$_{\pm\text{2.47}}$ \\

& & 200 & 68.60$_{\pm\text{2.55}}$ & \textbf{71.90}$_{\pm\text{2.02}}$ & 59.66$_{\pm\text{3.31}}$ & 66.54$_{\pm\text{2.75}}$ & 65.16$_{\pm\text{2.71}}$ & 70.22$_{\pm\text{2.26}}$ & 64.52$_{\pm\text{2.44}}$ & 51.24$_{\pm\text{7.29}}$ & 70.32$_{\pm\text{1.94}}$ & 71.08$_{\pm\text{1.87}}$ \\

& & 500 & 76.90$_{\pm\text{1.30}}$ & 77.64$_{\pm\text{1.07}}$ & 73.20$_{\pm\text{1.68}}$ & 76.22$_{\pm\text{1.62}}$ & 75.72$_{\pm\text{1.40}}$ & \textbf{78.88}$_{\pm\text{1.58}}$ & 76.54$_{\pm\text{0.77}}$ & 63.66$_{\pm\text{2.49}}$ & 74.30$_{\pm\text{1.51}}$ & 75.35$_{\pm\text{1.34}}$ \\

\cmidrule{2-13}

& \multirow[c]{5}{*}{biodeg} & 20 & 65.23$_{\pm\text{5.01}}$ & 68.99$_{\pm\text{3.31}}$ & 66.63$_{\pm\text{7.83}}$ & 56.99$_{\pm\text{5.55}}$ & 59.91$_{\pm\text{6.09}}$ & 55.85$_{\pm\text{4.94}}$ & 58.77$_{\pm\text{5.93}}$ & 56.62$_{\pm\text{7.29}}$ & 67.79$_{\pm\text{4.64}}$ & \textbf{69.76}$_{\pm\text{4.43}}$ \\

& & 50 & 71.26$_{\pm\text{3.13}}$ & 73.19$_{\pm\text{2.46}}$ & 70.80$_{\pm\text{2.14}}$ & 70.00$_{\pm\text{5.92}}$ & 65.90$_{\pm\text{3.57}}$ & 73.50$_{\pm\text{4.43}}$ & 70.23$_{\pm\text{3.35}}$ & 65.29$_{\pm\text{4.57}}$ & 72.08$_{\pm\text{3.84}}$ & \textbf{73.58}$_{\pm\text{3.57}}$ \\

& & 100 & 76.12$_{\pm\text{1.98}}$ & 76.07$_{\pm\text{1.74}}$ & 74.02$_{\pm\text{2.78}}$ & 75.36$_{\pm\text{2.18}}$ & 73.24$_{\pm\text{2.61}}$ & \textbf{77.34}$_{\pm\text{2.19}}$ & 74.28$_{\pm\text{2.02}}$ & 72.26$_{\pm\text{2.46}}$ & 74.56$_{\pm\text{1.58}}$ & 75.60$_{\pm\text{1.55}}$ \\

& & 200 & 78.86$_{\pm\text{2.19}}$ & \textbf{79.67}$_{\pm\text{1.68}}$ & 77.31$_{\pm\text{2.93}}$ & 78.05$_{\pm\text{3.07}}$ & 77.64$_{\pm\text{2.71}}$ & 77.84$_{\pm\text{2.62}}$ & 78.81$_{\pm\text{2.66}}$ & 76.82$_{\pm\text{2.29}}$ & 77.46$_{\pm\text{1.68}}$ & 78.46$_{\pm\text{1.69}}$ \\

& & 500 & 82.59$_{\pm\text{1.17}}$ & \textbf{83.07}$_{\pm\text{1.50}}$ & 82.13$_{\pm\text{1.21}}$ & 82.17$_{\pm\text{1.32}}$ & 82.80$_{\pm\text{1.28}}$ & 81.06$_{\pm\text{1.22}}$ & 82.15$_{\pm\text{1.33}}$ & 82.10$_{\pm\text{0.79}}$ & 79.99$_{\pm\text{1.76}}$ & 81.01$_{\pm\text{1.66}}$ \\

\cmidrule{2-13}

& \multirow[c]{5}{*}{steel} & 20 & 56.40$_{\pm\text{4.48}}$ & 63.95$_{\pm\text{3.14}}$ & 59.45$_{\pm\text{8.27}}$ & 57.04$_{\pm\text{5.05}}$ & 54.59$_{\pm\text{5.81}}$ & 65.46$_{\pm\text{6.10}}$ & 56.97$_{\pm\text{5.43}}$ & 52.90$_{\pm\text{3.76}}$ & \textbf{70.68}$_{\pm\text{3.87}}$ & 69.31$_{\pm\text{4.02}}$ \\

& & 50 & 73.95$_{\pm\text{4.76}}$ & 70.24$_{\pm\text{3.44}}$ & 67.60$_{\pm\text{4.10}}$ & 68.77$_{\pm\text{2.85}}$ & 67.00$_{\pm\text{4.58}}$ & \textbf{85.14}$_{\pm\text{8.76}}$ & 64.02$_{\pm\text{3.71}}$ & 57.54$_{\pm\text{2.34}}$ & 82.09$_{\pm\text{3.09}}$ & 80.47$_{\pm\text{3.48}}$ \\

& & 100 & 84.70$_{\pm\text{5.57}}$ & 77.46$_{\pm\text{3.67}}$ & 71.87$_{\pm\text{2.98}}$ & 72.94$_{\pm\text{4.62}}$ & 77.09$_{\pm\text{2.63}}$ & \textbf{94.05}$_{\pm\text{3.84}}$ & 72.62$_{\pm\text{5.21}}$ & 61.08$_{\pm\text{1.93}}$ & 87.77$_{\pm\text{3.13}}$ & 87.67$_{\pm\text{3.22}}$ \\

& & 200 & 90.44$_{\pm\text{2.80}}$ & 82.46$_{\pm\text{1.43}}$ & 80.83$_{\pm\text{2.65}}$ & 82.73$_{\pm\text{3.88}}$ & 85.49$_{\pm\text{3.59}}$ & \textbf{98.99}$_{\pm\text{0.75}}$ & 83.38$_{\pm\text{2.67}}$ & 69.12$_{\pm\text{2.56}}$ & 92.01$_{\pm\text{1.73}}$ & 92.06$_{\pm\text{1.48}}$ \\

& & 500 & 94.99$_{\pm\text{1.09}}$ & 89.97$_{\pm\text{0.88}}$ & 91.34$_{\pm\text{1.69}}$ & 92.42$_{\pm\text{1.36}}$ & 93.37$_{\pm\text{1.13}}$ & \textbf{99.71}$_{\pm\text{0.21}}$ & 92.02$_{\pm\text{2.03}}$ & 80.79$_{\pm\text{1.93}}$ & 95.08$_{\pm\text{1.30}}$ & 95.50$_{\pm\text{1.49}}$ \\

\cmidrule{2-13}

& \multirow[c]{4}{*}{stock} & 20 & 71.89$_{\pm\text{4.37}}$ & 84.41$_{\pm\text{5.28}}$ & 73.80$_{\pm\text{4.68}}$ & 66.38$_{\pm\text{9.10}}$ & 68.93$_{\pm\text{10.49}}$ & 81.82$_{\pm\text{8.38}}$ & 67.53$_{\pm\text{8.58}}$ & 71.80$_{\pm\text{4.99}}$ & 84.41$_{\pm\text{4.22}}$ & \textbf{84.69}$_{\pm\text{4.16}}$ \\

& & 50 & 85.03$_{\pm\text{3.39}}$ & \textbf{89.77}$_{\pm\text{1.99}}$ & 84.32$_{\pm\text{3.97}}$ & 83.49$_{\pm\text{3.67}}$ & 84.43$_{\pm\text{2.04}}$ & 89.34$_{\pm\text{1.59}}$ & 84.33$_{\pm\text{3.22}}$ & 83.64$_{\pm\text{2.53}}$ & 89.67$_{\pm\text{1.88}}$ & 89.68$_{\pm\text{1.87}}$ \\

& & 100 & 89.66$_{\pm\text{1.39}}$ & 92.32$_{\pm\text{0.99}}$ & 89.58$_{\pm\text{1.22}}$ & 89.61$_{\pm\text{1.36}}$ & 89.66$_{\pm\text{1.01}}$ & 91.40$_{\pm\text{1.41}}$ & 89.66$_{\pm\text{2.10}}$ & 89.44$_{\pm\text{1.41}}$ & 92.02$_{\pm\text{0.81}}$ & \textbf{92.47}$_{\pm\text{0.83}}$ \\

& & 200 & 91.65$_{\pm\text{1.08}}$ & 93.46$_{\pm\text{0.82}}$ & 92.37$_{\pm\text{1.18}}$ & 91.55$_{\pm\text{1.19}}$ & 91.43$_{\pm\text{1.34}}$ & 92.92$_{\pm\text{1.00}}$ & 91.14$_{\pm\text{1.58}}$ & 91.53$_{\pm\text{1.05}}$ & 93.15$_{\pm\text{0.72}}$ & \textbf{93.62}$_{\pm\text{1.14}}$ \\

\midrule

\multirow{9}{*}{\rotatebox{90}{\begin{tabular}{l} \textit{More than 10 classes} \end{tabular}}}& \multirow[c]{3}{*}{energy} & 50 & 10.85$_{\pm\text{1.76}}$ & N/A & 10.64$_{\pm\text{2.36}}$ & 8.22$_{\pm\text{2.03}}$ & 8.83$_{\pm\text{1.53}}$ & 8.92$_{\pm\text{2.51}}$ & 9.14$_{\pm\text{1.95}}$ & 11.86$_{\pm\text{2.33}}$ & N/A & \textbf{25.36}$_{\pm\text{2.27}}$ \\

& & 100 & 18.60$_{\pm\text{1.83}}$ & N/A & 13.71$_{\pm\text{1.66}}$ & 15.81$_{\pm\text{1.50}}$ & 14.67$_{\pm\text{1.55}}$ & 16.18$_{\pm\text{1.75}}$ & 15.71$_{\pm\text{2.79}}$ & 17.64$_{\pm\text{2.68}}$ & N/A & \textbf{29.82}$_{\pm\text{2.74}}$ \\

& & 200 & 26.45$_{\pm\text{1.49}}$ & N/A & 20.71$_{\pm\text{1.02}}$ & 21.71$_{\pm\text{3.23}}$ & 23.40$_{\pm\text{2.15}}$ & 23.95$_{\pm\text{2.94}}$ & 23.09$_{\pm\text{2.56}}$ & 27.35$_{\pm\text{2.28}}$ & N/A & \textbf{35.93}$_{\pm\text{2.85}}$ \\

\cmidrule{2-13}

& \multirow[c]{2}{*}{collins} & 100 & 10.59$_{\pm\text{1.48}}$ & N/A & 7.58$_{\pm\text{0.74}}$ & 7.95$_{\pm\text{1.12}}$ & 7.55$_{\pm\text{1.32}}$ & 14.24$_{\pm\text{1.48}}$ & 7.42$_{\pm\text{1.17}}$ & 8.79$_{\pm\text{0.93}}$ & N/A & \textbf{15.16}$_{\pm\text{1.92}}$ \\

& & 200 & 15.84$_{\pm\text{1.74}}$ & \textbf{19.81}$_{\pm\text{1.73}}$ & 9.79$_{\pm\text{1.14}}$ & 11.21$_{\pm\text{1.45}}$ & 12.24$_{\pm\text{1.65}}$ & 16.30$_{\pm\text{1.54}}$ & 10.96$_{\pm\text{1.43}}$ & 12.86$_{\pm\text{1.50}}$ & N/A & 18.05$_{\pm\text{1.65}}$ \\

\cmidrule{2-13}

& \multirow[c]{4}{*}{texture} & 50 & 62.96$_{\pm\text{2.49}}$ & \textbf{78.80}$_{\pm\text{2.75}}$ & 55.51$_{\pm\text{3.69}}$ & 61.86$_{\pm\text{4.48}}$ & 62.08$_{\pm\text{3.17}}$ & 61.91$_{\pm\text{2.24}}$ & 62.67$_{\pm\text{2.29}}$ & 56.81$_{\pm\text{2.98}}$ & N/A & 75.57$_{\pm\text{2.67}}$ \\

& & 100 & 77.16$_{\pm\text{1.25}}$ & \textbf{86.15}$_{\pm\text{2.62}}$ & 69.54$_{\pm\text{2.66}}$ & 76.53$_{\pm\text{2.22}}$ & 76.85$_{\pm\text{1.56}}$ & 77.77$_{\pm\text{1.80}}$ & 76.70$_{\pm\text{2.05}}$ & 72.64$_{\pm\text{1.81}}$ & N/A & 84.83$_{\pm\text{1.67}}$ \\

& & 200 & 85.34$_{\pm\text{1.18}}$ & 89.07$_{\pm\text{1.74}}$ & 81.70$_{\pm\text{1.32}}$ & 85.46$_{\pm\text{1.28}}$ & 84.62$_{\pm\text{1.09}}$ & 85.94$_{\pm\text{1.35}}$ & 85.11$_{\pm\text{1.20}}$ & 84.72$_{\pm\text{0.80}}$ & N/A & \textbf{89.48}$_{\pm\text{2.01}}$ \\

& & 500 & 91.40$_{\pm\text{1.60}}$ & 93.14$_{\pm\text{1.28}}$ & 89.88$_{\pm\text{1.44}}$ & 91.40$_{\pm\text{1.55}}$ & 91.34$_{\pm\text{1.60}}$ & 92.31$_{\pm\text{1.60}}$ & 91.46$_{\pm\text{1.51}}$ & 91.91$_{\pm\text{1.63}}$ & N/A & \textbf{93.46}$_{\pm\text{0.66}}$ \\

\midrule
\rowcolor{Gainsboro!60}
\multicolumn{3}{l|}{\textbf{Average rank}} & 5.15$_{\pm\text{2.06}}$ & 2.70$_{\pm\text{1.97}}$ & 7.67$_{\pm\text{2.10}}$ & 7.03$_{\pm\text{1.55}}$ & 7.27$_{\pm\text{1.68}}$ & 4.12$_{\pm\text{2.34}}$ & 6.82$_{\pm\text{1.76}}$ & 8.42$_{\pm\text{2.33}}$ & 3.67$_{\pm\text{1.96}}$ & \textbf{2.15}$_{\pm\text{1.75}}$ \\

\bottomrule

\end{tabular}
}
\end{table}
\clearpage

\begin{table}[p]
\centering
\caption{\textbf{Classification accuracy} (\%) of MLP, comparing data augmentation on eight real-world tabular datasets with varied real data availability. We report the mean $\pm$ std balanced accuracy and average accuracy rank across datasets. A higher rank implies higher accuracy. Note that ``N/A'' denotes that a specific generator was not applicable or the downstream predictor failed to converge, and the rank is computed with the mean balanced accuracy of other methods. We \textbf{bold} the highest accuracy for each dataset of different sample sizes. TabEBM achieves the best overall performance against Baseline and benchmark generators.}
\resizebox{\textwidth}{!}{

\setlength{\tabcolsep}{5pt}
\begin{tabular}{m{0.2cm}lr|r|rrrrrrrr|r}

\toprule

\multicolumn{2}{l}{Datasets}  & $N_{\text{real}}$ & \makecell[r]{Baseline\\(Real data)} & SMOTE & TVAE & CTGAN & NFLOW & TabDDPM & ARF & GOGGLE & TabPFGen & \textbf{TabEBM} \\

\midrule

\multirow{24}{*}{\rotatebox{90}{\begin{tabular}{l} \textit{At most 10 classes} \end{tabular}}}& \multirow[c]{5}{*}{protein} & 20 & 35.12$_{\pm\text{2.59}}$ & N/A & 21.89$_{\pm\text{3.62}}$ & 26.95$_{\pm\text{4.13}}$ & 24.56$_{\pm\text{5.06}}$ & 27.30$_{\pm\text{3.23}}$ & 27.96$_{\pm\text{4.51}}$ & 27.09$_{\pm\text{3.47}}$ & 36.19$_{\pm\text{2.84}}$ & \textbf{36.26}$_{\pm\text{2.65}}$ \\

& & 50 & 58.11$_{\pm\text{4.13}}$ & 57.24$_{\pm\text{4.60}}$ & 40.77$_{\pm\text{3.59}}$ & 43.04$_{\pm\text{5.43}}$ & 44.78$_{\pm\text{3.92}}$ & 49.43$_{\pm\text{2.51}}$ & 46.84$_{\pm\text{5.08}}$ & 44.78$_{\pm\text{2.67}}$ & 58.62$_{\pm\text{4.41}}$ & \textbf{58.75}$_{\pm\text{4.48}}$ \\

& & 100 & 76.82$_{\pm\text{3.33}}$ & 76.78$_{\pm\text{3.10}}$ & 62.00$_{\pm\text{3.21}}$ & 64.14$_{\pm\text{3.19}}$ & 63.24$_{\pm\text{4.05}}$ & 69.20$_{\pm\text{2.75}}$ & 65.08$_{\pm\text{2.67}}$ & 62.45$_{\pm\text{3.22}}$ & \textbf{77.84}$_{\pm\text{3.49}}$ & 77.63$_{\pm\text{3.69}}$ \\

& & 200 & 89.53$_{\pm\text{2.34}}$ & 90.28$_{\pm\text{2.13}}$ & 80.28$_{\pm\text{3.49}}$ & 81.94$_{\pm\text{2.56}}$ & 81.85$_{\pm\text{2.46}}$ & 85.48$_{\pm\text{2.07}}$ & 82.57$_{\pm\text{2.19}}$ & 78.04$_{\pm\text{2.66}}$ & \textbf{90.74}$_{\pm\text{2.13}}$ & 90.48$_{\pm\text{2.06}}$ \\

& & 500 & 98.23$_{\pm\text{0.91}}$ & 98.25$_{\pm\text{0.78}}$ & 95.08$_{\pm\text{1.34}}$ & 95.74$_{\pm\text{0.95}}$ & 96.15$_{\pm\text{1.09}}$ & 96.01$_{\pm\text{0.86}}$ & 96.50$_{\pm\text{0.87}}$ & 96.23$_{\pm\text{1.67}}$ & \textbf{98.52}$_{\pm\text{0.80}}$ & 98.50$_{\pm\text{0.70}}$ \\

\cmidrule{2-13}

& \multirow[c]{5}{*}{fourier} & 20 & 33.66$_{\pm\text{3.92}}$ & N/A & 23.20$_{\pm\text{5.54}}$ & 17.08$_{\pm\text{2.75}}$ & 19.40$_{\pm\text{4.03}}$ & 18.32$_{\pm\text{3.82}}$ & 23.26$_{\pm\text{4.21}}$ & 19.64$_{\pm\text{2.40}}$ & \textbf{37.00}$_{\pm\text{2.85}}$ & 35.02$_{\pm\text{3.77}}$ \\

& & 50 & 53.72$_{\pm\text{1.67}}$ & 53.02$_{\pm\text{1.96}}$ & 37.16$_{\pm\text{3.08}}$ & 37.60$_{\pm\text{4.52}}$ & 35.14$_{\pm\text{2.44}}$ & 40.90$_{\pm\text{2.80}}$ & 42.82$_{\pm\text{2.83}}$ & 32.66$_{\pm\text{5.19}}$ & \textbf{55.40}$_{\pm\text{2.23}}$ & 55.34$_{\pm\text{1.40}}$ \\

& & 100 & 62.78$_{\pm\text{1.60}}$ & 61.44$_{\pm\text{2.74}}$ & 43.68$_{\pm\text{3.15}}$ & 48.80$_{\pm\text{2.66}}$ & 46.18$_{\pm\text{3.96}}$ & 56.52$_{\pm\text{5.04}}$ & 52.50$_{\pm\text{2.66}}$ & 37.74$_{\pm\text{2.99}}$ & 63.00$_{\pm\text{1.95}}$ & \textbf{63.54}$_{\pm\text{1.83}}$ \\

& & 200 & 70.18$_{\pm\text{1.85}}$ & 70.06$_{\pm\text{2.10}}$ & 58.90$_{\pm\text{2.56}}$ & 62.36$_{\pm\text{2.86}}$ & 58.40$_{\pm\text{2.53}}$ & 70.08$_{\pm\text{1.90}}$ & 62.14$_{\pm\text{2.00}}$ & 50.92$_{\pm\text{5.13}}$ & \textbf{71.49}$_{\pm\text{1.41}}$ & 71.36$_{\pm\text{1.36}}$ \\

& & 500 & 77.94$_{\pm\text{1.65}}$ & 77.18$_{\pm\text{1.35}}$ & 72.14$_{\pm\text{1.79}}$ & 74.30$_{\pm\text{1.65}}$ & 71.38$_{\pm\text{1.54}}$ & 77.78$_{\pm\text{1.26}}$ & 74.32$_{\pm\text{1.56}}$ & 67.28$_{\pm\text{2.91}}$ & 78.34$_{\pm\text{1.72}}$ & \textbf{79.30}$_{\pm\text{0.99}}$ \\

\cmidrule{2-13}

& \multirow[c]{5}{*}{biodeg} & 20 & 71.31$_{\pm\text{5.13}}$ & 68.84$_{\pm\text{5.95}}$ & 66.64$_{\pm\text{8.27}}$ & 62.11$_{\pm\text{4.95}}$ & 62.61$_{\pm\text{6.78}}$ & 52.96$_{\pm\text{4.22}}$ & 62.06$_{\pm\text{3.69}}$ & 65.81$_{\pm\text{7.24}}$ & 72.04$_{\pm\text{5.12}}$ & \textbf{72.09}$_{\pm\text{4.81}}$ \\

& & 50 & 76.73$_{\pm\text{3.16}}$ & 74.97$_{\pm\text{2.51}}$ & 72.02$_{\pm\text{4.74}}$ & 71.83$_{\pm\text{3.17}}$ & 67.86$_{\pm\text{6.02}}$ & 69.92$_{\pm\text{4.83}}$ & 74.03$_{\pm\text{3.05}}$ & 71.01$_{\pm\text{2.78}}$ & \textbf{77.17}$_{\pm\text{2.93}}$ & 77.11$_{\pm\text{3.20}}$ \\

& & 100 & \textbf{79.13}$_{\pm\text{1.91}}$ & 78.20$_{\pm\text{1.68}}$ & 76.78$_{\pm\text{2.79}}$ & 77.85$_{\pm\text{2.73}}$ & 76.01$_{\pm\text{2.88}}$ & 76.74$_{\pm\text{3.62}}$ & 76.08$_{\pm\text{2.39}}$ & 76.24$_{\pm\text{2.45}}$ & 78.23$_{\pm\text{2.29}}$ & 79.08$_{\pm\text{2.03}}$ \\

& & 200 & \textbf{82.39}$_{\pm\text{1.48}}$ & 81.70$_{\pm\text{1.22}}$ & 80.43$_{\pm\text{2.02}}$ & 79.96$_{\pm\text{2.35}}$ & 79.92$_{\pm\text{1.55}}$ & 80.51$_{\pm\text{1.26}}$ & 79.59$_{\pm\text{1.72}}$ & 80.34$_{\pm\text{1.93}}$ & 81.74$_{\pm\text{1.36}}$ & 82.24$_{\pm\text{1.54}}$ \\

& & 500 & \textbf{84.50}$_{\pm\text{0.61}}$ & 84.50$_{\pm\text{0.81}}$ & 83.78$_{\pm\text{1.51}}$ & 83.67$_{\pm\text{0.81}}$ & 84.13$_{\pm\text{1.20}}$ & 84.09$_{\pm\text{0.84}}$ & 83.76$_{\pm\text{1.27}}$ & 82.97$_{\pm\text{1.21}}$ & 84.37$_{\pm\text{0.48}}$ & 84.14$_{\pm\text{0.59}}$ \\

\cmidrule{2-13}

& \multirow[c]{5}{*}{steel} & 20 & 62.35$_{\pm\text{6.30}}$ & 60.34$_{\pm\text{5.73}}$ & 61.63$_{\pm\text{8.82}}$ & 59.09$_{\pm\text{4.25}}$ & 56.99$_{\pm\text{7.13}}$ & 60.67$_{\pm\text{9.18}}$ & 55.23$_{\pm\text{3.92}}$ & 55.78$_{\pm\text{3.01}}$ & \textbf{64.49}$_{\pm\text{6.03}}$ & 64.22$_{\pm\text{5.89}}$ \\

& & 50 & 79.65$_{\pm\text{5.53}}$ & 68.18$_{\pm\text{3.16}}$ & 69.01$_{\pm\text{3.48}}$ & 70.30$_{\pm\text{4.77}}$ & 66.96$_{\pm\text{5.12}}$ & \textbf{84.04}$_{\pm\text{8.03}}$ & 64.79$_{\pm\text{4.07}}$ & 58.95$_{\pm\text{1.66}}$ & 82.72$_{\pm\text{6.02}}$ & 82.15$_{\pm\text{5.78}}$ \\

& & 100 & 92.18$_{\pm\text{2.93}}$ & 78.44$_{\pm\text{3.67}}$ & 76.37$_{\pm\text{3.06}}$ & 76.92$_{\pm\text{3.63}}$ & 76.50$_{\pm\text{3.18}}$ & \textbf{95.83}$_{\pm\text{1.90}}$ & 71.85$_{\pm\text{3.20}}$ & 67.35$_{\pm\text{2.51}}$ & 95.16$_{\pm\text{3.13}}$ & 95.41$_{\pm\text{3.23}}$ \\

& & 200 & 97.31$_{\pm\text{1.63}}$ & 83.93$_{\pm\text{1.83}}$ & 82.42$_{\pm\text{2.75}}$ & 84.70$_{\pm\text{2.54}}$ & 84.06$_{\pm\text{4.46}}$ & 98.75$_{\pm\text{0.68}}$ & 79.66$_{\pm\text{3.26}}$ & 78.36$_{\pm\text{3.86}}$ & 98.83$_{\pm\text{0.80}}$ & \textbf{98.84}$_{\pm\text{0.67}}$ \\

& & 500 & 99.78$_{\pm\text{0.30}}$ & 91.37$_{\pm\text{2.26}}$ & 93.08$_{\pm\text{1.82}}$ & 94.82$_{\pm\text{1.54}}$ & 93.99$_{\pm\text{1.83}}$ & 99.47$_{\pm\text{0.44}}$ & 90.34$_{\pm\text{2.19}}$ & 96.06$_{\pm\text{1.03}}$ & \textbf{99.81}$_{\pm\text{0.24}}$ & 99.81$_{\pm\text{0.24}}$ \\

\cmidrule{2-13}

& \multirow[c]{4}{*}{stock} & 20 & 83.56$_{\pm\text{3.89}}$ & 83.62$_{\pm\text{4.06}}$ & 77.25$_{\pm\text{5.02}}$ & 69.90$_{\pm\text{9.27}}$ & 72.85$_{\pm\text{7.77}}$ & 80.60$_{\pm\text{5.73}}$ & 69.30$_{\pm\text{6.76}}$ & 83.34$_{\pm\text{3.38}}$ & 83.81$_{\pm\text{3.92}}$ & \textbf{83.89}$_{\pm\text{4.05}}$ \\

& & 50 & 89.57$_{\pm\text{2.01}}$ & 89.71$_{\pm\text{2.21}}$ & 82.62$_{\pm\text{3.49}}$ & 79.35$_{\pm\text{2.71}}$ & 81.52$_{\pm\text{1.99}}$ & 88.48$_{\pm\text{2.18}}$ & 83.36$_{\pm\text{2.30}}$ & 88.38$_{\pm\text{2.41}}$ & 90.23$_{\pm\text{2.02}}$ & \textbf{90.38}$_{\pm\text{2.17}}$ \\

& & 100 & 90.63$_{\pm\text{0.83}}$ & 91.17$_{\pm\text{0.83}}$ & 88.37$_{\pm\text{2.40}}$ & 86.60$_{\pm\text{3.27}}$ & 83.19$_{\pm\text{3.77}}$ & 90.65$_{\pm\text{1.39}}$ & 88.64$_{\pm\text{1.15}}$ & 91.61$_{\pm\text{0.70}}$ & 91.70$_{\pm\text{1.17}}$ & \textbf{91.75}$_{\pm\text{0.97}}$ \\

& & 200 & 91.25$_{\pm\text{0.74}}$ & 92.58$_{\pm\text{0.91}}$ & 91.27$_{\pm\text{0.95}}$ & 90.34$_{\pm\text{1.60}}$ & 89.32$_{\pm\text{2.45}}$ & 92.19$_{\pm\text{0.78}}$ & 90.37$_{\pm\text{1.32}}$ & 91.89$_{\pm\text{1.32}}$ & \textbf{92.91}$_{\pm\text{0.62}}$ & 92.47$_{\pm\text{0.63}}$ \\

\midrule

\multirow{9}{*}{\rotatebox{90}{\begin{tabular}{l} \textit{More than 10 classes} \end{tabular}}}& \multirow[c]{3}{*}{energy} & 50 & \textbf{24.79}$_{\pm\text{1.76}}$ & N/A & 12.51$_{\pm\text{2.89}}$ & 12.61$_{\pm\text{3.45}}$ & 8.43$_{\pm\text{2.11}}$ & 10.91$_{\pm\text{2.11}}$ & 12.45$_{\pm\text{1.87}}$ & 20.42$_{\pm\text{4.41}}$ & N/A & 24.04$_{\pm\text{1.39}}$ \\

& & 100 & 26.86$_{\pm\text{1.51}}$ & N/A & 16.20$_{\pm\text{2.12}}$ & 15.78$_{\pm\text{2.50}}$ & 15.70$_{\pm\text{2.44}}$ & 17.75$_{\pm\text{3.18}}$ & 18.16$_{\pm\text{2.46}}$ & 20.72$_{\pm\text{2.99}}$ & N/A & \textbf{29.30}$_{\pm\text{2.32}}$ \\

& & 200 & 33.36$_{\pm\text{2.98}}$ & N/A & 22.30$_{\pm\text{2.44}}$ & 23.00$_{\pm\text{4.24}}$ & 23.53$_{\pm\text{1.84}}$ & 26.12$_{\pm\text{1.78}}$ & 26.28$_{\pm\text{3.03}}$ & 33.03$_{\pm\text{3.38}}$ & N/A & \textbf{41.27}$_{\pm\text{2.93}}$ \\

\cmidrule{2-13}

& \multirow[c]{2}{*}{collins} & 100 & \textbf{14.16}$_{\pm\text{1.31}}$ & N/A & 9.24$_{\pm\text{1.71}}$ & 9.16$_{\pm\text{1.57}}$ & 9.04$_{\pm\text{1.79}}$ & 14.03$_{\pm\text{1.24}}$ & 8.59$_{\pm\text{1.84}}$ & 10.81$_{\pm\text{1.68}}$ & N/A & 14.07$_{\pm\text{1.58}}$ \\

& & 200 & 19.35$_{\pm\text{1.24}}$ & 19.06$_{\pm\text{1.49}}$ & 14.62$_{\pm\text{2.00}}$ & 13.17$_{\pm\text{0.94}}$ & 12.38$_{\pm\text{1.56}}$ & 18.63$_{\pm\text{1.56}}$ & 12.48$_{\pm\text{1.91}}$ & 17.65$_{\pm\text{1.76}}$ & N/A & \textbf{19.53}$_{\pm\text{1.44}}$ \\

\cmidrule{2-13}

& \multirow[c]{4}{*}{texture} & 50 & 84.50$_{\pm\text{2.81}}$ & 84.12$_{\pm\text{3.02}}$ & 62.92$_{\pm\text{4.09}}$ & 67.69$_{\pm\text{5.49}}$ & 61.99$_{\pm\text{3.58}}$ & 69.69$_{\pm\text{4.62}}$ & 64.45$_{\pm\text{7.84}}$ & 69.68$_{\pm\text{3.53}}$ & N/A & \textbf{85.51}$_{\pm\text{2.89}}$ \\

& & 100 & 91.50$_{\pm\text{1.34}}$ & 91.57$_{\pm\text{1.59}}$ & 74.53$_{\pm\text{3.39}}$ & 79.96$_{\pm\text{4.37}}$ & 80.36$_{\pm\text{4.58}}$ & 85.42$_{\pm\text{2.74}}$ & 80.23$_{\pm\text{2.18}}$ & 85.59$_{\pm\text{1.49}}$ & N/A & \textbf{92.17}$_{\pm\text{1.31}}$ \\

& & 200 & 93.81$_{\pm\text{1.35}}$ & 94.18$_{\pm\text{1.26}}$ & 86.57$_{\pm\text{2.33}}$ & 90.68$_{\pm\text{1.55}}$ & 88.97$_{\pm\text{2.12}}$ & 90.10$_{\pm\text{2.26}}$ & 89.14$_{\pm\text{1.85}}$ & 91.66$_{\pm\text{1.43}}$ & N/A & \textbf{94.35}$_{\pm\text{1.57}}$ \\

& & 500 & 96.55$_{\pm\text{0.63}}$ & \textbf{97.21}$_{\pm\text{0.40}}$ & 94.66$_{\pm\text{1.17}}$ & 96.27$_{\pm\text{0.74}}$ & 94.34$_{\pm\text{1.36}}$ & 94.72$_{\pm\text{0.61}}$ & 95.83$_{\pm\text{1.03}}$ & 96.49$_{\pm\text{0.48}}$ & N/A & 97.13$_{\pm\text{0.53}}$ \\

\midrule
\rowcolor{Gainsboro!60}
\multicolumn{3}{l|}{\textbf{Average rank}} & 3.00$_{\pm\text{1.32}}$ & 4.06$_{\pm\text{1.62}}$ & 7.82$_{\pm\text{1.63}}$ & 7.48$_{\pm\text{1.50}}$ & 8.45$_{\pm\text{1.33}}$ & 5.55$_{\pm\text{2.14}}$ & 7.48$_{\pm\text{1.72}}$ & 6.82$_{\pm\text{2.57}}$ & 2.67$_{\pm\text{1.69}}$ & \textbf{1.67}$_{\pm\text{0.74}}$ \\

\bottomrule

\end{tabular}
}
\end{table}
\clearpage

\begin{table}[p]
\centering
\caption{\textbf{Classification accuracy} (\%) of RF, comparing data augmentation on eight real-world tabular datasets with varied real data availability. We report the mean $\pm$ std balanced accuracy and average accuracy rank across datasets. A higher rank implies higher accuracy. Note that ``N/A'' denotes that a specific generator was not applicable or the downstream predictor failed to converge, and the rank is computed with the mean balanced accuracy of other methods. We \textbf{bold} the highest accuracy for each dataset of different sample sizes. TabEBM achieves the best overall performance against Baseline and benchmark generators.}
\resizebox{\textwidth}{!}{

\setlength{\tabcolsep}{5pt}
\begin{tabular}{m{0.2cm}lr|r|rrrrrrrr|r}

\toprule

\multicolumn{2}{l}{Datasets}  & $N_{\text{real}}$ & \makecell[r]{Baseline\\(Real data)} & SMOTE & TVAE & CTGAN & NFLOW & TabDDPM & ARF & GOGGLE & TabPFGen & \textbf{TabEBM} \\

\midrule

\multirow{24}{*}{\rotatebox{90}{\begin{tabular}{l} \textit{At most 10 classes} \end{tabular}}}& \multirow[c]{5}{*}{protein} & 20 & 28.52$_{\pm\text{2.19}}$ & N/A & 22.74$_{\pm\text{4.23}}$ & 24.94$_{\pm\text{5.15}}$ & 24.61$_{\pm\text{2.46}}$ & 29.62$_{\pm\text{4.06}}$ & 27.69$_{\pm\text{3.73}}$ & 25.76$_{\pm\text{1.62}}$ & 32.04$_{\pm\text{2.40}}$ & \textbf{34.19}$_{\pm\text{2.21}}$ \\

& & 50 & 53.40$_{\pm\text{3.26}}$ & 55.69$_{\pm\text{2.61}}$ & 46.95$_{\pm\text{3.13}}$ & 43.91$_{\pm\text{4.98}}$ & 43.28$_{\pm\text{4.07}}$ & 47.93$_{\pm\text{4.49}}$ & 44.48$_{\pm\text{3.35}}$ & 47.25$_{\pm\text{4.81}}$ & 54.29$_{\pm\text{2.57}}$ & \textbf{56.85}$_{\pm\text{2.49}}$ \\

& & 100 & 68.13$_{\pm\text{3.19}}$ & \textbf{72.89}$_{\pm\text{2.60}}$ & 63.24$_{\pm\text{1.78}}$ & 61.19$_{\pm\text{2.10}}$ & 59.64$_{\pm\text{3.48}}$ & 65.19$_{\pm\text{3.22}}$ & 60.05$_{\pm\text{2.88}}$ & 65.05$_{\pm\text{3.00}}$ & 71.47$_{\pm\text{3.61}}$ & 72.57$_{\pm\text{2.50}}$ \\

& & 200 & 80.34$_{\pm\text{2.35}}$ & 83.60$_{\pm\text{2.71}}$ & 78.51$_{\pm\text{2.58}}$ & 75.84$_{\pm\text{1.61}}$ & 76.84$_{\pm\text{2.35}}$ & 78.74$_{\pm\text{2.24}}$ & 75.61$_{\pm\text{2.90}}$ & 79.44$_{\pm\text{2.73}}$ & 83.36$_{\pm\text{2.40}}$ & \textbf{84.30}$_{\pm\text{1.97}}$ \\

& & 500 & 93.01$_{\pm\text{1.12}}$ & 93.82$_{\pm\text{0.67}}$ & 92.86$_{\pm\text{1.66}}$ & 91.37$_{\pm\text{1.25}}$ & 93.00$_{\pm\text{1.08}}$ & 92.93$_{\pm\text{0.97}}$ & 92.38$_{\pm\text{0.90}}$ & 92.95$_{\pm\text{0.92}}$ & \textbf{94.49}$_{\pm\text{1.16}}$ & 93.94$_{\pm\text{1.23}}$ \\

\cmidrule{2-13}

& \multirow[c]{5}{*}{fourier} & 20 & 35.10$_{\pm\text{4.56}}$ & N/A & 19.06$_{\pm\text{3.91}}$ & 17.52$_{\pm\text{2.84}}$ & 20.78$_{\pm\text{2.54}}$ & 16.98$_{\pm\text{2.31}}$ & 23.78$_{\pm\text{3.12}}$ & 19.00$_{\pm\text{2.93}}$ & 34.88$_{\pm\text{5.93}}$ & \textbf{38.60}$_{\pm\text{5.66}}$ \\

& & 50 & 64.10$_{\pm\text{3.80}}$ & 64.76$_{\pm\text{4.00}}$ & 37.20$_{\pm\text{3.35}}$ & 32.82$_{\pm\text{4.56}}$ & 37.78$_{\pm\text{3.11}}$ & 51.76$_{\pm\text{3.50}}$ & 47.22$_{\pm\text{4.35}}$ & 53.86$_{\pm\text{3.61}}$ & \textbf{66.92}$_{\pm\text{3.05}}$ & 66.26$_{\pm\text{3.16}}$ \\

& & 100 & 73.86$_{\pm\text{3.06}}$ & 73.78$_{\pm\text{3.22}}$ & 64.40$_{\pm\text{2.51}}$ & 60.82$_{\pm\text{3.71}}$ & 51.64$_{\pm\text{4.16}}$ & 66.14$_{\pm\text{1.91}}$ & 58.62$_{\pm\text{3.73}}$ & 68.16$_{\pm\text{3.12}}$ & 73.13$_{\pm\text{2.70}}$ & \textbf{74.84}$_{\pm\text{3.10}}$ \\

& & 200 & 78.54$_{\pm\text{2.15}}$ & 79.18$_{\pm\text{1.92}}$ & 74.86$_{\pm\text{1.60}}$ & 74.26$_{\pm\text{2.20}}$ & 69.36$_{\pm\text{2.61}}$ & 76.42$_{\pm\text{1.95}}$ & 72.88$_{\pm\text{1.22}}$ & 76.64$_{\pm\text{1.99}}$ & \textbf{82.20}$_{\pm\text{0.85}}$ & 79.18$_{\pm\text{2.08}}$ \\

& & 500 & 81.84$_{\pm\text{1.01}}$ & 82.14$_{\pm\text{1.49}}$ & 81.02$_{\pm\text{1.59}}$ & 81.18$_{\pm\text{1.43}}$ & 80.08$_{\pm\text{1.62}}$ & 81.26$_{\pm\text{1.40}}$ & 80.28$_{\pm\text{1.54}}$ & 80.62$_{\pm\text{1.52}}$ & 81.45$_{\pm\text{1.45}}$ & \textbf{83.40}$_{\pm\text{1.24}}$ \\

\cmidrule{2-13}

& \multirow[c]{5}{*}{biodeg} & 20 & 61.11$_{\pm\text{7.87}}$ & \textbf{68.38}$_{\pm\text{5.90}}$ & 65.44$_{\pm\text{8.89}}$ & 56.29$_{\pm\text{7.96}}$ & 58.19$_{\pm\text{6.60}}$ & 52.90$_{\pm\text{4.74}}$ & 62.33$_{\pm\text{6.14}}$ & 63.52$_{\pm\text{7.29}}$ & 67.15$_{\pm\text{5.74}}$ & 67.82$_{\pm\text{5.13}}$ \\

& & 50 & 68.38$_{\pm\text{4.82}}$ & 70.64$_{\pm\text{3.44}}$ & 71.77$_{\pm\text{2.99}}$ & 66.78$_{\pm\text{4.89}}$ & 61.39$_{\pm\text{4.94}}$ & 63.98$_{\pm\text{3.65}}$ & 68.78$_{\pm\text{5.22}}$ & 70.34$_{\pm\text{3.39}}$ & 71.38$_{\pm\text{3.60}}$ & \textbf{72.12}$_{\pm\text{3.29}}$ \\

& & 100 & 73.19$_{\pm\text{2.46}}$ & 75.36$_{\pm\text{2.56}}$ & 74.98$_{\pm\text{2.58}}$ & 72.68$_{\pm\text{2.98}}$ & 69.62$_{\pm\text{3.53}}$ & 73.11$_{\pm\text{2.39}}$ & 72.16$_{\pm\text{2.58}}$ & 74.22$_{\pm\text{2.32}}$ & \textbf{75.85}$_{\pm\text{1.56}}$ & 75.65$_{\pm\text{1.53}}$ \\

& & 200 & 77.85$_{\pm\text{2.72}}$ & 78.86$_{\pm\text{1.97}}$ & 76.42$_{\pm\text{2.25}}$ & 76.68$_{\pm\text{2.77}}$ & 73.43$_{\pm\text{3.01}}$ & 76.16$_{\pm\text{2.00}}$ & 75.79$_{\pm\text{2.49}}$ & 77.42$_{\pm\text{2.24}}$ & \textbf{79.68}$_{\pm\text{1.74}}$ & 79.22$_{\pm\text{1.70}}$ \\

& & 500 & 81.42$_{\pm\text{0.73}}$ & 82.03$_{\pm\text{1.02}}$ & 81.88$_{\pm\text{0.87}}$ & 81.71$_{\pm\text{1.54}}$ & 80.50$_{\pm\text{1.21}}$ & 81.43$_{\pm\text{1.26}}$ & 81.34$_{\pm\text{1.58}}$ & 81.94$_{\pm\text{0.85}}$ & \textbf{82.38}$_{\pm\text{1.35}}$ & 82.10$_{\pm\text{1.31}}$ \\

\cmidrule{2-13}

& \multirow[c]{5}{*}{steel} & 20 & 52.77$_{\pm\text{1.60}}$ & 56.16$_{\pm\text{4.50}}$ & 57.23$_{\pm\text{3.97}}$ & 54.65$_{\pm\text{3.40}}$ & 53.75$_{\pm\text{3.49}}$ & 51.70$_{\pm\text{1.66}}$ & 54.09$_{\pm\text{4.36}}$ & 55.50$_{\pm\text{2.97}}$ & 57.04$_{\pm\text{3.07}}$ & \textbf{57.41}$_{\pm\text{2.67}}$ \\

& & 50 & 59.75$_{\pm\text{3.11}}$ & 62.12$_{\pm\text{2.46}}$ & 60.65$_{\pm\text{1.96}}$ & 58.09$_{\pm\text{1.75}}$ & 54.69$_{\pm\text{2.44}}$ & 58.14$_{\pm\text{4.21}}$ & 57.67$_{\pm\text{2.52}}$ & 60.34$_{\pm\text{2.90}}$ & 65.07$_{\pm\text{3.11}}$ & \textbf{67.74}$_{\pm\text{3.36}}$ \\

& & 100 & 64.97$_{\pm\text{2.05}}$ & 69.08$_{\pm\text{3.62}}$ & 64.46$_{\pm\text{4.17}}$ & 61.62$_{\pm\text{1.98}}$ & 58.43$_{\pm\text{2.46}}$ & 60.53$_{\pm\text{3.64}}$ & 62.71$_{\pm\text{3.43}}$ & 63.07$_{\pm\text{2.23}}$ & 73.28$_{\pm\text{3.39}}$ & \textbf{79.63}$_{\pm\text{3.41}}$ \\

& & 200 & 75.45$_{\pm\text{3.26}}$ & 74.71$_{\pm\text{3.79}}$ & 71.45$_{\pm\text{2.18}}$ & 68.52$_{\pm\text{3.80}}$ & 62.15$_{\pm\text{2.80}}$ & 68.10$_{\pm\text{3.59}}$ & 67.61$_{\pm\text{1.81}}$ & 67.36$_{\pm\text{1.63}}$ & 85.12$_{\pm\text{4.44}}$ & \textbf{88.85}$_{\pm\text{5.10}}$ \\

& & 500 & 90.93$_{\pm\text{2.83}}$ & 85.37$_{\pm\text{2.36}}$ & 85.63$_{\pm\text{3.14}}$ & 84.51$_{\pm\text{3.22}}$ & 76.12$_{\pm\text{2.70}}$ & 89.19$_{\pm\text{3.20}}$ & 81.44$_{\pm\text{2.64}}$ & 80.35$_{\pm\text{3.54}}$ & 94.35$_{\pm\text{1.34}}$ & \textbf{95.90}$_{\pm\text{1.06}}$ \\

\cmidrule{2-13}

& \multirow[c]{4}{*}{stock} & 20 & 79.47$_{\pm\text{5.83}}$ & 81.99$_{\pm\text{4.49}}$ & 77.94$_{\pm\text{5.21}}$ & 72.53$_{\pm\text{7.16}}$ & 73.20$_{\pm\text{9.98}}$ & 80.99$_{\pm\text{7.01}}$ & 72.57$_{\pm\text{8.76}}$ & 78.10$_{\pm\text{5.91}}$ & 83.96$_{\pm\text{5.57}}$ & \textbf{84.73}$_{\pm\text{3.46}}$ \\

& & 50 & 87.57$_{\pm\text{2.60}}$ & 89.69$_{\pm\text{1.99}}$ & 86.62$_{\pm\text{3.44}}$ & 83.75$_{\pm\text{4.32}}$ & 84.28$_{\pm\text{2.98}}$ & 88.69$_{\pm\text{2.11}}$ & 84.92$_{\pm\text{2.09}}$ & 88.65$_{\pm\text{2.55}}$ & 89.35$_{\pm\text{2.18}}$ & \textbf{89.99}$_{\pm\text{2.63}}$ \\

& & 100 & 91.44$_{\pm\text{1.59}}$ & 91.47$_{\pm\text{2.16}}$ & 91.07$_{\pm\text{2.08}}$ & 89.82$_{\pm\text{2.69}}$ & 89.33$_{\pm\text{1.92}}$ & 91.33$_{\pm\text{2.07}}$ & 90.48$_{\pm\text{2.38}}$ & 92.00$_{\pm\text{2.36}}$ & 92.07$_{\pm\text{1.22}}$ & \textbf{92.17}$_{\pm\text{1.24}}$ \\

& & 200 & 93.52$_{\pm\text{0.80}}$ & \textbf{93.94}$_{\pm\text{1.09}}$ & 93.35$_{\pm\text{1.05}}$ & 92.62$_{\pm\text{1.02}}$ & 92.77$_{\pm\text{1.25}}$ & 93.65$_{\pm\text{1.08}}$ & 93.08$_{\pm\text{0.53}}$ & 93.87$_{\pm\text{1.25}}$ & 93.65$_{\pm\text{1.02}}$ & 93.67$_{\pm\text{1.07}}$ \\

\midrule

\multirow{9}{*}{\rotatebox{90}{\begin{tabular}{l} \textit{More than 10 classes} \end{tabular}}}& \multirow[c]{3}{*}{energy} & 50 & 18.96$_{\pm\text{1.40}}$ & N/A & 16.63$_{\pm\text{2.27}}$ & 15.66$_{\pm\text{2.43}}$ & 14.81$_{\pm\text{3.16}}$ & 14.49$_{\pm\text{1.26}}$ & 15.05$_{\pm\text{3.06}}$ & 15.58$_{\pm\text{3.26}}$ & N/A & \textbf{27.74}$_{\pm\text{3.71}}$ \\

& & 100 & 30.85$_{\pm\text{2.19}}$ & N/A & 24.59$_{\pm\text{2.27}}$ & 28.59$_{\pm\text{2.63}}$ & 27.59$_{\pm\text{2.86}}$ & 27.23$_{\pm\text{2.39}}$ & 27.99$_{\pm\text{2.18}}$ & 25.43$_{\pm\text{2.46}}$ & N/A & \textbf{41.03}$_{\pm\text{2.24}}$ \\

& & 200 & 45.80$_{\pm\text{2.32}}$ & N/A & 42.10$_{\pm\text{2.57}}$ & 41.69$_{\pm\text{3.84}}$ & 44.41$_{\pm\text{2.51}}$ & 44.58$_{\pm\text{1.37}}$ & 41.33$_{\pm\text{3.90}}$ & 44.64$_{\pm\text{2.54}}$ & N/A & \textbf{53.87}$_{\pm\text{2.81}}$ \\

\cmidrule{2-13}

& \multirow[c]{2}{*}{collins} & 100 & 10.41$_{\pm\text{1.61}}$ & N/A & 6.75$_{\pm\text{0.69}}$ & 8.23$_{\pm\text{1.76}}$ & 7.34$_{\pm\text{1.46}}$ & 12.84$_{\pm\text{1.61}}$ & 6.73$_{\pm\text{1.36}}$ & 8.43$_{\pm\text{0.94}}$ & N/A & \textbf{13.35}$_{\pm\text{1.49}}$ \\

& & 200 & 13.75$_{\pm\text{1.12}}$ & \textbf{17.56}$_{\pm\text{1.79}}$ & 10.51$_{\pm\text{1.41}}$ & 11.00$_{\pm\text{1.37}}$ & 9.85$_{\pm\text{1.38}}$ & 15.15$_{\pm\text{1.22}}$ & 9.90$_{\pm\text{0.72}}$ & 13.40$_{\pm\text{1.09}}$ & N/A & 16.51$_{\pm\text{1.53}}$ \\

\cmidrule{2-13}

& \multirow[c]{4}{*}{texture} & 50 & 71.27$_{\pm\text{1.99}}$ & 71.17$_{\pm\text{3.89}}$ & 57.41$_{\pm\text{3.33}}$ & 62.78$_{\pm\text{4.21}}$ & 65.24$_{\pm\text{4.52}}$ & 69.45$_{\pm\text{2.15}}$ & 62.93$_{\pm\text{4.84}}$ & 64.33$_{\pm\text{3.57}}$ & N/A & \textbf{75.79}$_{\pm\text{3.07}}$ \\

& & 100 & 80.40$_{\pm\text{2.45}}$ & 80.38$_{\pm\text{2.67}}$ & 65.63$_{\pm\text{4.21}}$ & 75.38$_{\pm\text{3.99}}$ & 77.67$_{\pm\text{2.62}}$ & 79.31$_{\pm\text{1.88}}$ & 75.98$_{\pm\text{2.56}}$ & 77.30$_{\pm\text{2.44}}$ & N/A & \textbf{82.30}$_{\pm\text{2.21}}$ \\

& & 200 & 84.00$_{\pm\text{1.56}}$ & 85.12$_{\pm\text{3.07}}$ & 76.98$_{\pm\text{2.25}}$ & 84.44$_{\pm\text{2.41}}$ & 85.30$_{\pm\text{1.96}}$ & 84.00$_{\pm\text{1.20}}$ & 83.70$_{\pm\text{2.05}}$ & 80.02$_{\pm\text{1.60}}$ & N/A & \textbf{85.92}$_{\pm\text{2.18}}$ \\

& & 500 & 89.43$_{\pm\text{0.80}}$ & 90.17$_{\pm\text{1.25}}$ & 88.97$_{\pm\text{1.44}}$ & 90.00$_{\pm\text{1.66}}$ & 89.99$_{\pm\text{1.06}}$ & 90.17$_{\pm\text{1.32}}$ & \textbf{91.01}$_{\pm\text{1.32}}$ & 88.98$_{\pm\text{1.26}}$ & N/A & 90.77$_{\pm\text{1.10}}$ \\

\midrule
\rowcolor{Gainsboro!60}
\multicolumn{3}{l|}{\textbf{Average rank}} & 4.36$_{\pm\text{1.95}}$ & 3.02$_{\pm\text{1.14}}$ & 6.88$_{\pm\text{2.25}}$ & 7.82$_{\pm\text{1.61}}$ & 8.45$_{\pm\text{1.95}}$ & 6.12$_{\pm\text{2.23}}$ & 7.85$_{\pm\text{1.77}}$ & 6.00$_{\pm\text{1.75}}$ & 3.12$_{\pm\text{1.75}}$ & \textbf{1.38}$_{\pm\text{0.57}}$ \\

\bottomrule

\end{tabular}
}
\end{table}
\clearpage

\begin{table}[p]
\centering
\caption{\textbf{Classification accuracy} (\%) of XGBoost, comparing data augmentation on eight real-world tabular datasets with varied real data availability. We report the mean $\pm$ std balanced accuracy and average accuracy rank across datasets. A higher rank implies higher accuracy. Note that ``N/A'' denotes that a specific generator was not applicable or the downstream predictor failed to converge, and the rank is computed with the mean balanced accuracy of other methods. We \textbf{bold} the highest accuracy for each dataset of different sample sizes. TabEBM achieves the best overall performance against Baseline and benchmark generators.}
\resizebox{\textwidth}{!}{

\setlength{\tabcolsep}{5pt}
\begin{tabular}{m{0.2cm}lr|r|rrrrrrrr|r}

\toprule

\multicolumn{2}{l}{Datasets}  & $N_{\text{real}}$ & \makecell[r]{Baseline\\(Real data)} & SMOTE & TVAE & CTGAN & NFLOW & TabDDPM & ARF & GOGGLE & TabPFGen & \textbf{TabEBM} \\

\midrule

\multirow{24}{*}{\rotatebox{90}{\begin{tabular}{l} \textit{At most 10 classes} \end{tabular}}}& \multirow[c]{5}{*}{protein} & 20 & 19.70$_{\pm\text{6.33}}$ & N/A & 19.44$_{\pm\text{4.11}}$ & 17.32$_{\pm\text{2.75}}$ & 18.11$_{\pm\text{3.07}}$ & 16.15$_{\pm\text{3.80}}$ & 20.71$_{\pm\text{5.24}}$ & 17.40$_{\pm\text{4.89}}$ & 24.00$_{\pm\text{3.64}}$ & \textbf{24.18}$_{\pm\text{3.05}}$ \\

& & 50 & 39.01$_{\pm\text{4.92}}$ & 37.68$_{\pm\text{5.40}}$ & 33.07$_{\pm\text{4.18}}$ & 24.38$_{\pm\text{3.45}}$ & 23.09$_{\pm\text{4.55}}$ & 30.87$_{\pm\text{5.70}}$ & 34.13$_{\pm\text{6.45}}$ & 33.62$_{\pm\text{3.67}}$ & 39.78$_{\pm\text{6.03}}$ & \textbf{44.46}$_{\pm\text{4.97}}$ \\

& & 100 & 57.59$_{\pm\text{3.69}}$ & 60.16$_{\pm\text{5.75}}$ & 49.23$_{\pm\text{5.51}}$ & 43.33$_{\pm\text{7.92}}$ & 37.69$_{\pm\text{5.96}}$ & 48.36$_{\pm\text{4.08}}$ & 43.97$_{\pm\text{5.45}}$ & 47.00$_{\pm\text{3.29}}$ & 53.74$_{\pm\text{7.94}}$ & \textbf{62.77}$_{\pm\text{5.85}}$ \\

& & 200 & 74.05$_{\pm\text{2.92}}$ & 76.90$_{\pm\text{4.96}}$ & 69.71$_{\pm\text{4.28}}$ & 67.46$_{\pm\text{4.39}}$ & 58.29$_{\pm\text{7.96}}$ & 69.68$_{\pm\text{4.23}}$ & 63.69$_{\pm\text{6.32}}$ & 66.09$_{\pm\text{4.78}}$ & 73.19$_{\pm\text{6.06}}$ & \textbf{79.25}$_{\pm\text{3.83}}$ \\

& & 500 & 88.89$_{\pm\text{1.71}}$ & 90.02$_{\pm\text{1.51}}$ & 90.10$_{\pm\text{1.80}}$ & 89.37$_{\pm\text{1.81}}$ & 86.03$_{\pm\text{2.31}}$ & 87.29$_{\pm\text{2.08}}$ & 90.05$_{\pm\text{2.70}}$ & 85.04$_{\pm\text{2.07}}$ & 89.66$_{\pm\text{1.17}}$ & \textbf{91.81}$_{\pm\text{1.44}}$ \\

\cmidrule{2-13}

& \multirow[c]{5}{*}{fourier} & 20 & 10.00$_{\pm\text{0.00}}$ & N/A & 14.64$_{\pm\text{3.13}}$ & 13.58$_{\pm\text{2.57}}$ & 13.82$_{\pm\text{4.14}}$ & 11.72$_{\pm\text{4.19}}$ & 16.38$_{\pm\text{3.86}}$ & 12.34$_{\pm\text{3.59}}$ & 23.50$_{\pm\text{1.56}}$ & \textbf{26.78}$_{\pm\text{4.82}}$ \\

& & 50 & 42.10$_{\pm\text{6.19}}$ & 43.40$_{\pm\text{5.22}}$ & 34.32$_{\pm\text{3.98}}$ & 24.68$_{\pm\text{6.47}}$ & 17.66$_{\pm\text{4.64}}$ & 24.82$_{\pm\text{6.35}}$ & 27.74$_{\pm\text{5.86}}$ & 35.42$_{\pm\text{7.51}}$ & 35.60$_{\pm\text{3.11}}$ & \textbf{45.08}$_{\pm\text{6.47}}$ \\

& & 100 & 54.84$_{\pm\text{2.78}}$ & 52.92$_{\pm\text{5.69}}$ & 48.22$_{\pm\text{3.28}}$ & 36.90$_{\pm\text{5.15}}$ & 30.36$_{\pm\text{3.94}}$ & 42.46$_{\pm\text{4.13}}$ & 40.28$_{\pm\text{3.41}}$ & 48.78$_{\pm\text{4.36}}$ & 49.80$_{\pm\text{1.98}}$ & \textbf{54.94}$_{\pm\text{5.72}}$ \\

& & 200 & 63.88$_{\pm\text{3.35}}$ & 65.34$_{\pm\text{3.57}}$ & 58.36$_{\pm\text{3.27}}$ & 53.20$_{\pm\text{5.26}}$ & 46.96$_{\pm\text{4.58}}$ & 61.40$_{\pm\text{4.12}}$ & 52.10$_{\pm\text{3.32}}$ & 56.66$_{\pm\text{2.67}}$ & 66.60$_{\pm\text{4.24}}$ & \textbf{67.68}$_{\pm\text{3.19}}$ \\

& & 500 & 74.56$_{\pm\text{1.97}}$ & 74.18$_{\pm\text{2.10}}$ & 68.28$_{\pm\text{2.82}}$ & 67.98$_{\pm\text{2.07}}$ & 61.24$_{\pm\text{2.35}}$ & 72.78$_{\pm\text{2.54}}$ & 67.50$_{\pm\text{2.57}}$ & 68.28$_{\pm\text{3.43}}$ & N/A & \textbf{76.25}$_{\pm\text{3.18}}$ \\

\cmidrule{2-13}

& \multirow[c]{5}{*}{biodeg} & 20 & 62.95$_{\pm\text{7.95}}$ & 66.51$_{\pm\text{5.84}}$ & 62.72$_{\pm\text{5.69}}$ & 55.24$_{\pm\text{6.28}}$ & 59.20$_{\pm\text{7.83}}$ & 54.65$_{\pm\text{5.56}}$ & 62.78$_{\pm\text{5.98}}$ & 61.09$_{\pm\text{10.49}}$ & 65.52$_{\pm\text{6.08}}$ & \textbf{66.64}$_{\pm\text{6.71}}$ \\

& & 50 & 67.96$_{\pm\text{3.45}}$ & 67.69$_{\pm\text{4.42}}$ & 66.22$_{\pm\text{5.70}}$ & 61.64$_{\pm\text{6.73}}$ & 60.72$_{\pm\text{5.73}}$ & 57.48$_{\pm\text{8.28}}$ & \textbf{69.48}$_{\pm\text{5.35}}$ & 65.93$_{\pm\text{4.98}}$ & 67.76$_{\pm\text{4.90}}$ & 67.90$_{\pm\text{3.27}}$ \\

& & 100 & \textbf{73.88}$_{\pm\text{2.55}}$ & 72.05$_{\pm\text{4.75}}$ & 72.11$_{\pm\text{3.17}}$ & 70.41$_{\pm\text{3.60}}$ & 66.02$_{\pm\text{6.25}}$ & 69.35$_{\pm\text{4.66}}$ & 71.11$_{\pm\text{3.88}}$ & 69.03$_{\pm\text{4.33}}$ & 72.58$_{\pm\text{2.91}}$ & 71.05$_{\pm\text{5.70}}$ \\

& & 200 & 76.38$_{\pm\text{4.85}}$ & 74.98$_{\pm\text{3.15}}$ & 73.93$_{\pm\text{3.29}}$ & 75.68$_{\pm\text{4.15}}$ & 67.82$_{\pm\text{3.91}}$ & 72.58$_{\pm\text{5.07}}$ & 74.74$_{\pm\text{2.24}}$ & 73.84$_{\pm\text{3.82}}$ & 75.85$_{\pm\text{1.80}}$ & \textbf{76.74}$_{\pm\text{2.44}}$ \\

& & 500 & 78.45$_{\pm\text{3.37}}$ & 79.38$_{\pm\text{1.99}}$ & 78.88$_{\pm\text{3.42}}$ & \textbf{80.15}$_{\pm\text{1.87}}$ & 76.72$_{\pm\text{3.44}}$ & 77.10$_{\pm\text{2.96}}$ & 78.14$_{\pm\text{2.65}}$ & 78.83$_{\pm\text{2.21}}$ & 79.40$_{\pm\text{1.49}}$ & 78.80$_{\pm\text{3.76}}$ \\

\cmidrule{2-13}

& \multirow[c]{5}{*}{steel} & 20 & 53.12$_{\pm\text{5.62}}$ & 55.64$_{\pm\text{4.76}}$ & 53.32$_{\pm\text{7.25}}$ & 55.36$_{\pm\text{6.24}}$ & 52.38$_{\pm\text{3.55}}$ & 52.44$_{\pm\text{4.08}}$ & 51.34$_{\pm\text{4.15}}$ & 50.74$_{\pm\text{2.53}}$ & 55.43$_{\pm\text{5.57}}$ & \textbf{55.78}$_{\pm\text{4.53}}$ \\

& & 50 & 66.73$_{\pm\text{9.11}}$ & 60.79$_{\pm\text{5.52}}$ & 59.51$_{\pm\text{4.15}}$ & 54.82$_{\pm\text{4.23}}$ & 54.79$_{\pm\text{4.69}}$ & 59.71$_{\pm\text{6.94}}$ & 57.66$_{\pm\text{5.19}}$ & 55.89$_{\pm\text{4.50}}$ & 63.78$_{\pm\text{7.20}}$ & \textbf{74.18}$_{\pm\text{13.67}}$ \\

& & 100 & 83.17$_{\pm\text{9.36}}$ & 66.95$_{\pm\text{6.51}}$ & 61.72$_{\pm\text{6.80}}$ & 65.12$_{\pm\text{3.02}}$ & 60.56$_{\pm\text{4.37}}$ & 72.02$_{\pm\text{12.47}}$ & 59.67$_{\pm\text{4.77}}$ & 59.04$_{\pm\text{4.76}}$ & 90.52$_{\pm\text{7.47}}$ & \textbf{96.55}$_{\pm\text{2.66}}$ \\

& & 200 & 95.94$_{\pm\text{2.73}}$ & 81.21$_{\pm\text{5.01}}$ & 73.14$_{\pm\text{5.45}}$ & 70.64$_{\pm\text{10.67}}$ & 70.26$_{\pm\text{9.25}}$ & 74.50$_{\pm\text{23.57}}$ & 74.57$_{\pm\text{9.36}}$ & 65.41$_{\pm\text{6.70}}$ & 99.14$_{\pm\text{1.19}}$ & \textbf{99.54}$_{\pm\text{0.62}}$ \\

& & 500 & 99.95$_{\pm\text{0.10}}$ & 97.04$_{\pm\text{2.14}}$ & 95.27$_{\pm\text{2.88}}$ & 89.46$_{\pm\text{6.88}}$ & 83.25$_{\pm\text{8.10}}$ & 91.72$_{\pm\text{15.34}}$ & 87.59$_{\pm\text{6.72}}$ & 79.54$_{\pm\text{15.29}}$ & 100.00$_{\pm\text{0.00}}$ & 100.00$_{\pm\text{0.00}}$ \\

\cmidrule{2-13}

& \multirow[c]{4}{*}{stock} & 20 & 76.42$_{\pm\text{4.34}}$ & 78.92$_{\pm\text{5.21}}$ & 67.46$_{\pm\text{13.93}}$ & 60.56$_{\pm\text{9.69}}$ & 73.36$_{\pm\text{9.57}}$ & 77.45$_{\pm\text{9.80}}$ & 69.15$_{\pm\text{9.35}}$ & 70.88$_{\pm\text{8.52}}$ & 79.82$_{\pm\text{4.52}}$ & \textbf{83.44}$_{\pm\text{3.74}}$ \\

& & 50 & 83.71$_{\pm\text{3.40}}$ & 86.23$_{\pm\text{2.54}}$ & 84.65$_{\pm\text{4.44}}$ & 79.31$_{\pm\text{6.58}}$ & 76.27$_{\pm\text{3.89}}$ & 85.70$_{\pm\text{3.96}}$ & 81.61$_{\pm\text{1.97}}$ & 84.98$_{\pm\text{4.44}}$ & 87.28$_{\pm\text{3.65}}$ & \textbf{88.21}$_{\pm\text{3.31}}$ \\

& & 100 & 88.19$_{\pm\text{3.04}}$ & 89.01$_{\pm\text{2.07}}$ & 85.66$_{\pm\text{6.01}}$ & 84.68$_{\pm\text{2.87}}$ & 82.50$_{\pm\text{3.73}}$ & \textbf{90.07}$_{\pm\text{3.41}}$ & 86.09$_{\pm\text{4.08}}$ & 84.67$_{\pm\text{7.29}}$ & 90.01$_{\pm\text{3.46}}$ & 89.66$_{\pm\text{3.28}}$ \\

& & 200 & \textbf{92.32}$_{\pm\text{1.35}}$ & 92.26$_{\pm\text{2.33}}$ & 90.94$_{\pm\text{1.98}}$ & 89.01$_{\pm\text{2.53}}$ & 88.92$_{\pm\text{2.67}}$ & 91.36$_{\pm\text{3.79}}$ & 91.04$_{\pm\text{1.46}}$ & 91.42$_{\pm\text{2.66}}$ & 91.72$_{\pm\text{2.77}}$ & 92.17$_{\pm\text{1.51}}$ \\

\midrule

\multirow{9}{*}{\rotatebox{90}{\begin{tabular}{l} \textit{More than 10 classes} \end{tabular}}}& \multirow[c]{3}{*}{energy} & 50 & 12.05$_{\pm\text{2.42}}$ & N/A & 11.60$_{\pm\text{3.83}}$ & 14.47$_{\pm\text{5.32}}$ & 10.95$_{\pm\text{4.68}}$ & 10.21$_{\pm\text{5.11}}$ & 12.81$_{\pm\text{2.51}}$ & 12.34$_{\pm\text{3.55}}$ & N/A & \textbf{21.07}$_{\pm\text{3.99}}$ \\

& & 100 & \textbf{29.37}$_{\pm\text{1.72}}$ & N/A & 20.61$_{\pm\text{5.39}}$ & 19.81$_{\pm\text{4.52}}$ & 22.71$_{\pm\text{6.15}}$ & 22.27$_{\pm\text{2.12}}$ & 22.02$_{\pm\text{3.54}}$ & 10.01$_{\pm\text{3.40}}$ & N/A & 27.93$_{\pm\text{4.16}}$ \\

& & 200 & \textbf{44.96}$_{\pm\text{3.31}}$ & N/A & 36.73$_{\pm\text{6.03}}$ & 35.92$_{\pm\text{8.45}}$ & 33.71$_{\pm\text{6.54}}$ & 34.73$_{\pm\text{5.89}}$ & 37.06$_{\pm\text{5.26}}$ & 18.81$_{\pm\text{7.27}}$ & N/A & 40.95$_{\pm\text{5.59}}$ \\

\cmidrule{2-13}

& \multirow[c]{2}{*}{collins} & 100 & 7.77$_{\pm\text{2.21}}$ & N/A & 7.76$_{\pm\text{0.95}}$ & 6.52$_{\pm\text{1.16}}$ & 6.11$_{\pm\text{1.09}}$ & \textbf{8.95}$_{\pm\text{1.90}}$ & 6.21$_{\pm\text{1.14}}$ & 5.96$_{\pm\text{1.07}}$ & N/A & 8.73$_{\pm\text{1.64}}$ \\

& & 200 & 10.58$_{\pm\text{2.57}}$ & 11.46$_{\pm\text{2.11}}$ & 9.43$_{\pm\text{2.20}}$ & 9.84$_{\pm\text{1.56}}$ & 8.26$_{\pm\text{1.75}}$ & 9.80$_{\pm\text{1.96}}$ & 8.90$_{\pm\text{0.83}}$ & 9.79$_{\pm\text{0.80}}$ & N/A & \textbf{11.72}$_{\pm\text{1.34}}$ \\

\cmidrule{2-13}

& \multirow[c]{4}{*}{texture} & 50 & 56.72$_{\pm\text{6.12}}$ & 60.99$_{\pm\text{4.35}}$ & 45.76$_{\pm\text{6.50}}$ & 39.50$_{\pm\text{6.46}}$ & 43.02$_{\pm\text{6.12}}$ & 50.22$_{\pm\text{6.28}}$ & 43.71$_{\pm\text{5.98}}$ & 46.21$_{\pm\text{7.95}}$ & N/A & \textbf{69.11}$_{\pm\text{3.27}}$ \\

& & 100 & 68.96$_{\pm\text{2.59}}$ & 69.77$_{\pm\text{4.63}}$ & 54.95$_{\pm\text{5.99}}$ & 55.52$_{\pm\text{7.80}}$ & 63.23$_{\pm\text{4.80}}$ & 65.59$_{\pm\text{3.62}}$ & 57.04$_{\pm\text{6.59}}$ & 62.06$_{\pm\text{6.11}}$ & N/A & \textbf{76.35}$_{\pm\text{2.64}}$ \\

& & 200 & 77.91$_{\pm\text{1.98}}$ & 81.55$_{\pm\text{2.22}}$ & 70.70$_{\pm\text{4.40}}$ & 71.60$_{\pm\text{4.19}}$ & 73.76$_{\pm\text{5.69}}$ & 77.06$_{\pm\text{2.17}}$ & 72.56$_{\pm\text{4.09}}$ & 70.31$_{\pm\text{6.55}}$ & N/A & \textbf{82.59}$_{\pm\text{2.15}}$ \\

& & 500 & 89.37$_{\pm\text{1.11}}$ & \textbf{89.87}$_{\pm\text{1.24}}$ & 85.06$_{\pm\text{2.40}}$ & 86.80$_{\pm\text{2.25}}$ & 86.83$_{\pm\text{1.89}}$ & 86.52$_{\pm\text{1.66}}$ & 85.70$_{\pm\text{2.75}}$ & 87.07$_{\pm\text{2.43}}$ & N/A & 89.69$_{\pm\text{1.10}}$ \\

\midrule
\rowcolor{Gainsboro!60}
\multicolumn{3}{l|}{\textbf{Average rank}} & 3.64$_{\pm\text{2.09}}$ & 3.45$_{\pm\text{1.48}}$ & 6.32$_{\pm\text{1.86}}$ & 7.33$_{\pm\text{2.19}}$ & 8.64$_{\pm\text{1.82}}$ & 6.30$_{\pm\text{2.44}}$ & 6.64$_{\pm\text{2.18}}$ & 7.62$_{\pm\text{1.84}}$ & 3.44$_{\pm\text{1.54}}$ & \textbf{1.62}$_{\pm\text{1.29}}$ \\

\bottomrule

\end{tabular}
}
\end{table}
\clearpage

\begin{table}[p]
\centering
\caption{\textbf{Classification accuracy} (\%) of TabPFN, comparing data augmentation on eight real-world tabular datasets with varied real data availability. We report the mean $\pm$ std balanced accuracy and average accuracy rank across datasets. A higher rank implies higher accuracy. Note that ``N/A'' denotes that a specific generator was not applicable or the downstream predictor failed to converge, and the rank is computed with the mean balanced accuracy of other methods. We \textbf{bold} the highest accuracy for each dataset of different sample sizes. TabEBM achieves the best overall performance against Baseline and benchmark generators.}
\resizebox{\textwidth}{!}{

\setlength{\tabcolsep}{5pt}
\begin{tabular}{m{0.2cm}lr|r|rrrrrrrr|r}

\toprule

\multicolumn{2}{l}{Datasets}  & $N_{\text{real}}$ & \makecell[r]{Baseline\\(Real data)} & SMOTE & TVAE & CTGAN & NFLOW & TabDDPM & ARF & GOGGLE & TabPFGen & \textbf{TabEBM} \\

\midrule

\multirow{24}{*}{\rotatebox{90}{\begin{tabular}{l} \textit{At most 10 classes} \end{tabular}}}& \multirow[c]{5}{*}{protein} & 20 & 27.80$_{\pm\text{4.37}}$ & N/A & 19.21$_{\pm\text{3.80}}$ & 20.58$_{\pm\text{4.63}}$ & 20.80$_{\pm\text{4.34}}$ & 18.89$_{\pm\text{4.37}}$ & 23.97$_{\pm\text{3.05}}$ & 10.55$_{\pm\text{1.61}}$ & 33.42$_{\pm\text{5.95}}$ & \textbf{34.63}$_{\pm\text{5.78}}$ \\

& & 50 & 55.24$_{\pm\text{3.46}}$ & \textbf{59.85}$_{\pm\text{3.87}}$ & 43.58$_{\pm\text{6.20}}$ & 37.37$_{\pm\text{8.01}}$ & 34.42$_{\pm\text{6.65}}$ & 21.70$_{\pm\text{7.43}}$ & 46.02$_{\pm\text{2.60}}$ & 13.54$_{\pm\text{4.22}}$ & 57.63$_{\pm\text{2.82}}$ & 58.88$_{\pm\text{3.99}}$ \\

& & 100 & 74.31$_{\pm\text{3.49}}$ & \textbf{80.05}$_{\pm\text{3.16}}$ & 68.15$_{\pm\text{5.54}}$ & 71.10$_{\pm\text{2.55}}$ & 57.89$_{\pm\text{6.13}}$ & 59.28$_{\pm\text{8.23}}$ & 65.84$_{\pm\text{3.03}}$ & 23.69$_{\pm\text{10.40}}$ & 77.60$_{\pm\text{4.03}}$ & 78.26$_{\pm\text{3.75}}$ \\

& & 200 & 88.67$_{\pm\text{1.53}}$ & \textbf{91.79}$_{\pm\text{1.42}}$ & 87.05$_{\pm\text{2.85}}$ & 86.69$_{\pm\text{2.85}}$ & 83.29$_{\pm\text{2.42}}$ & 87.39$_{\pm\text{2.95}}$ & 85.49$_{\pm\text{2.49}}$ & 77.63$_{\pm\text{6.11}}$ & 90.77$_{\pm\text{1.37}}$ & 90.94$_{\pm\text{1.46}}$ \\

& & 500 & 97.31$_{\pm\text{0.69}}$ & \textbf{97.69}$_{\pm\text{0.77}}$ & 97.51$_{\pm\text{0.85}}$ & 97.58$_{\pm\text{0.85}}$ & 96.89$_{\pm\text{0.62}}$ & 97.44$_{\pm\text{0.85}}$ & 97.40$_{\pm\text{0.60}}$ & 97.35$_{\pm\text{0.61}}$ & 97.24$_{\pm\text{0.80}}$ & 97.28$_{\pm\text{0.62}}$ \\

\cmidrule{2-13}

& \multirow[c]{5}{*}{fourier} & 20 & 30.06$_{\pm\text{6.85}}$ & N/A & 22.00$_{\pm\text{4.62}}$ & 20.10$_{\pm\text{4.31}}$ & 14.52$_{\pm\text{3.96}}$ & 12.22$_{\pm\text{2.40}}$ & 21.64$_{\pm\text{5.91}}$ & 14.64$_{\pm\text{3.94}}$ & N/A & 36.56$_{\pm\text{4.96}}$ \\

& & 50 & 53.62$_{\pm\text{4.71}}$ & 53.08$_{\pm\text{3.34}}$ & 45.82$_{\pm\text{4.29}}$ & 37.46$_{\pm\text{5.82}}$ & 28.78$_{\pm\text{2.78}}$ & 22.74$_{\pm\text{5.11}}$ & 42.14$_{\pm\text{3.09}}$ & 11.30$_{\pm\text{1.50}}$ & 53.15$_{\pm\text{3.50}}$ & \textbf{53.82}$_{\pm\text{3.92}}$ \\

& & 100 & 64.62$_{\pm\text{4.14}}$ & 63.66$_{\pm\text{3.92}}$ & 56.68$_{\pm\text{3.02}}$ & 54.78$_{\pm\text{2.80}}$ & 45.50$_{\pm\text{4.50}}$ & 49.36$_{\pm\text{8.51}}$ & 54.74$_{\pm\text{2.78}}$ & 21.40$_{\pm\text{4.29}}$ & \textbf{65.95}$_{\pm\text{3.49}}$ & 65.40$_{\pm\text{3.61}}$ \\

& & 200 & 71.62$_{\pm\text{2.59}}$ & 70.56$_{\pm\text{3.61}}$ & 66.48$_{\pm\text{3.82}}$ & 66.14$_{\pm\text{4.02}}$ & 62.64$_{\pm\text{2.60}}$ & 72.12$_{\pm\text{2.64}}$ & 65.04$_{\pm\text{3.20}}$ & 52.18$_{\pm\text{7.35}}$ & 69.93$_{\pm\text{3.91}}$ & \textbf{72.48}$_{\pm\text{3.08}}$ \\

& & 500 & 77.66$_{\pm\text{1.61}}$ & 77.50$_{\pm\text{1.08}}$ & 76.80$_{\pm\text{1.34}}$ & 77.82$_{\pm\text{1.24}}$ & 73.90$_{\pm\text{1.76}}$ & \textbf{79.16}$_{\pm\text{2.05}}$ & 75.70$_{\pm\text{2.11}}$ & 74.36$_{\pm\text{2.53}}$ & 77.30$_{\pm\text{0.42}}$ & 77.40$_{\pm\text{1.28}}$ \\

\cmidrule{2-13}

& \multirow[c]{5}{*}{biodeg} & 20 & 65.26$_{\pm\text{8.01}}$ & 68.72$_{\pm\text{4.50}}$ & 69.02$_{\pm\text{5.37}}$ & 59.39$_{\pm\text{6.25}}$ & 58.28$_{\pm\text{8.30}}$ & 50.00$_{\pm\text{0.00}}$ & 58.45$_{\pm\text{8.16}}$ & 51.80$_{\pm\text{4.07}}$ & 70.68$_{\pm\text{4.94}}$ & \textbf{71.18}$_{\pm\text{5.25}}$ \\

& & 50 & 75.27$_{\pm\text{2.63}}$ & 74.65$_{\pm\text{3.28}}$ & 73.44$_{\pm\text{4.02}}$ & 70.21$_{\pm\text{3.61}}$ & 55.68$_{\pm\text{9.27}}$ & 50.00$_{\pm\text{0.00}}$ & 72.74$_{\pm\text{3.74}}$ & 55.75$_{\pm\text{7.45}}$ & \textbf{75.69}$_{\pm\text{2.44}}$ & 75.56$_{\pm\text{3.22}}$ \\

& & 100 & 78.92$_{\pm\text{1.98}}$ & 77.78$_{\pm\text{2.65}}$ & 77.27$_{\pm\text{3.15}}$ & 77.71$_{\pm\text{1.81}}$ & 63.50$_{\pm\text{10.77}}$ & 57.50$_{\pm\text{6.27}}$ & 77.25$_{\pm\text{1.66}}$ & 65.87$_{\pm\text{6.72}}$ & 78.15$_{\pm\text{1.45}}$ & \textbf{79.00}$_{\pm\text{1.99}}$ \\

& & 200 & \textbf{82.59}$_{\pm\text{1.84}}$ & 81.42$_{\pm\text{1.27}}$ & 80.48$_{\pm\text{1.82}}$ & 80.19$_{\pm\text{2.48}}$ & 79.16$_{\pm\text{2.49}}$ & 80.45$_{\pm\text{1.48}}$ & 80.88$_{\pm\text{1.68}}$ & 80.66$_{\pm\text{1.49}}$ & 82.56$_{\pm\text{1.68}}$ & 82.58$_{\pm\text{1.90}}$ \\

& & 500 & \textbf{85.00}$_{\pm\text{0.70}}$ & 84.37$_{\pm\text{0.75}}$ & 84.40$_{\pm\text{0.68}}$ & 84.67$_{\pm\text{0.98}}$ & 84.45$_{\pm\text{0.91}}$ & 84.58$_{\pm\text{0.70}}$ & 84.68$_{\pm\text{1.06}}$ & 83.66$_{\pm\text{0.67}}$ & 84.56$_{\pm\text{0.98}}$ & 84.55$_{\pm\text{0.92}}$ \\

\cmidrule{2-13}

& \multirow[c]{5}{*}{steel} & 20 & 56.77$_{\pm\text{4.17}}$ & 55.95$_{\pm\text{4.30}}$ & 56.03$_{\pm\text{4.37}}$ & 55.62$_{\pm\text{4.80}}$ & 52.52$_{\pm\text{4.64}}$ & 50.00$_{\pm\text{0.00}}$ & 52.39$_{\pm\text{3.13}}$ & 50.05$_{\pm\text{0.17}}$ & 64.80$_{\pm\text{5.66}}$ & \textbf{65.87}$_{\pm\text{6.14}}$ \\

& & 50 & 82.34$_{\pm\text{8.38}}$ & 63.42$_{\pm\text{3.93}}$ & 62.08$_{\pm\text{2.69}}$ & 63.98$_{\pm\text{4.08}}$ & 52.92$_{\pm\text{4.72}}$ & 50.64$_{\pm\text{2.01}}$ & 61.32$_{\pm\text{4.55}}$ & 50.36$_{\pm\text{1.09}}$ & 84.70$_{\pm\text{7.84}}$ & \textbf{86.30}$_{\pm\text{6.73}}$ \\

& & 100 & 97.37$_{\pm\text{1.37}}$ & 73.06$_{\pm\text{4.46}}$ & 71.96$_{\pm\text{5.40}}$ & 72.23$_{\pm\text{4.15}}$ & 56.34$_{\pm\text{6.30}}$ & 80.87$_{\pm\text{20.44}}$ & 69.29$_{\pm\text{5.70}}$ & 51.18$_{\pm\text{3.24}}$ & 97.49$_{\pm\text{1.21}}$ & \textbf{97.81}$_{\pm\text{1.49}}$ \\

& & 200 & 98.84$_{\pm\text{0.70}}$ & 82.32$_{\pm\text{2.88}}$ & 81.78$_{\pm\text{3.36}}$ & 83.24$_{\pm\text{2.68}}$ & 82.92$_{\pm\text{6.21}}$ & \textbf{99.35}$_{\pm\text{0.70}}$ & 86.40$_{\pm\text{4.22}}$ & 64.42$_{\pm\text{11.35}}$ & 98.80$_{\pm\text{0.73}}$ & 98.96$_{\pm\text{0.71}}$ \\

& & 500 & 99.74$_{\pm\text{0.29}}$ & 94.27$_{\pm\text{2.39}}$ & 94.93$_{\pm\text{1.89}}$ & 96.98$_{\pm\text{1.34}}$ & 98.32$_{\pm\text{1.19}}$ & \textbf{99.88}$_{\pm\text{0.15}}$ & 95.70$_{\pm\text{1.50}}$ & 98.56$_{\pm\text{0.52}}$ & 99.77$_{\pm\text{0.30}}$ & 99.74$_{\pm\text{0.29}}$ \\

\cmidrule{2-13}

& \multirow[c]{4}{*}{stock} & 20 & 83.18$_{\pm\text{4.37}}$ & 83.69$_{\pm\text{3.10}}$ & 74.01$_{\pm\text{5.09}}$ & 56.92$_{\pm\text{16.52}}$ & 74.99$_{\pm\text{6.60}}$ & 78.73$_{\pm\text{12.25}}$ & 69.64$_{\pm\text{6.88}}$ & 73.40$_{\pm\text{4.88}}$ & 82.95$_{\pm\text{4.44}}$ & \textbf{83.81}$_{\pm\text{4.94}}$ \\

& & 50 & 90.01$_{\pm\text{2.07}}$ & 90.01$_{\pm\text{2.43}}$ & 82.27$_{\pm\text{4.30}}$ & 78.91$_{\pm\text{4.14}}$ & 78.94$_{\pm\text{8.78}}$ & 89.68$_{\pm\text{1.92}}$ & 83.72$_{\pm\text{2.50}}$ & 79.00$_{\pm\text{6.87}}$ & 89.95$_{\pm\text{2.08}}$ & \textbf{90.15}$_{\pm\text{1.76}}$ \\

& & 100 & 92.39$_{\pm\text{1.06}}$ & 92.09$_{\pm\text{1.45}}$ & 90.75$_{\pm\text{2.20}}$ & 89.43$_{\pm\text{3.29}}$ & 86.16$_{\pm\text{3.83}}$ & 92.12$_{\pm\text{1.16}}$ & 90.17$_{\pm\text{1.92}}$ & 89.30$_{\pm\text{1.33}}$ & 92.12$_{\pm\text{1.12}}$ & \textbf{92.57}$_{\pm\text{1.27}}$ \\

& & 200 & 94.16$_{\pm\text{0.92}}$ & 93.99$_{\pm\text{0.70}}$ & 93.57$_{\pm\text{1.10}}$ & 93.28$_{\pm\text{1.59}}$ & 91.92$_{\pm\text{2.00}}$ & \textbf{94.22}$_{\pm\text{1.10}}$ & 93.05$_{\pm\text{1.35}}$ & 92.07$_{\pm\text{1.76}}$ & 94.17$_{\pm\text{0.89}}$ & 94.16$_{\pm\text{1.07}}$ \\

\midrule
\rowcolor{Gainsboro!60}
\multicolumn{3}{l|}{\textbf{Average rank}} & 3.08$_{\pm\text{1.22}}$ & 4.23$_{\pm\text{2.32}}$ & 6.12$_{\pm\text{1.57}}$ & 6.29$_{\pm\text{2.07}}$ & 8.42$_{\pm\text{1.32}}$ & 6.12$_{\pm\text{3.38}}$ & 6.54$_{\pm\text{1.61}}$ & 8.83$_{\pm\text{1.46}}$ & 3.12$_{\pm\text{1.80}}$ & \textbf{2.23}$_{\pm\text{1.83}}$ \\

\bottomrule

\end{tabular}
}
\end{table}
\clearpage

\FloatBarrier
\subsubsection{Results on six UCI Datasets}
\label{appendix:res_uci}

\begin{table}[htbp]
\centering
\caption{Details of the six real-world tabular datasets from UCI.}
\label{tab:uci_datasets}
\resizebox{\textwidth}{!}{%
\begin{tabular}{lrrrrrrrr}
\toprule
Dataset  & UCI ID & \makecell[r]{Not evaluated   in\\TabPFN~\cite{TabPFN}} & \# Samples ($N$) & \# Features ($D$) & \# Classes & N/D    & \makecell[r]{\# Samples per   class\\(Min)} & \makecell[r]{\# Samples per   class\\(Max)} \\

\midrule

clinical & 890    & \CheckmarkBold                                         & 2,139           & 23               & 2         & 93     & 521                                         & 1,618                                       \\
support2 & 880    & \CheckmarkBold                                         & 9,105           & 42               & 2         & 217 & 2,904                                       & 6,201                                       \\
mushroom & 73     & \CheckmarkBold                                         & 8,124           & 22               & 2         & 369 & 3,916                                       & 4,208                                       \\
auction  & 713    & \CheckmarkBold                                         & 2,043           & 7                & 2         & 292 & 262                                         & 1,781                                       \\
abalone  & 1      & \CheckmarkBold                                         & 4,153           & 8                & 19        & 519 & 14                                          & 689                                         \\
statlog  & 144    & \CheckmarkBold                                         & 1,000           & 20               & 2         & 50     & 300                                         & 700   \\

\bottomrule
\end{tabular}%
}
\end{table}

\begin{table}[htbp]
\centering
\caption{Test classification accuracy (\%) aggregated over six downstream predictors, comparing data augmentation on six leakage-free UCI datasets. Note that ``N/A'' denotes that a specific generator was not applicable. TabEBM still achieves the best overall performance against benchmark methods.}
\label{tab:eval_uci}
\resizebox{\textwidth}{!}{%
\begin{tabular}{l|r|rrrrrr}
\toprule
 \makecell[l]{Datasets \\ ($N_{\text{real}}=100$)} & Baseline                 & SMOTE                    & TVAE                     & CTGAN                    & TabDDPM                  & TabPFGen                 & \textbf{TabEBM (Ours)}            \\

\midrule

clinical & 68.63$_{\pm\text{5.81}}$ & 71.07$_{\pm\text{4.67}}$ & 61.80$_{\pm\text{2.76}}$ & 65.21$_{\pm\text{5.77}}$ & 54.03$_{\pm\text{5.36}}$ & 69.66$_{\pm\text{3.65}}$ & \textbf{71.20}$_{\pm\text{3.54}}$ \\
support2 & 64.23$_{\pm\text{1.89}}$ & \textbf{65.60}$_{\pm\text{1.52}}$ & 60.70$_{\pm\text{0.90}}$ & 59.14$_{\pm\text{1.88}}$ & 58.31$_{\pm\text{1.74}}$ & 64.34$_{\pm\text{1.19}}$ & 65.28$_{\pm\text{1.15}}$ \\
mushroom & 95.51$_{\pm\text{2.48}}$ & 95.84$_{\pm\text{1.99}}$ & 93.75$_{\pm\text{1.18}}$ & 93.26$_{\pm\text{2.46}}$ & 79.87$_{\pm\text{2.29}}$ & \textbf{97.05}$_{\pm\text{1.56}}$ & 96.82$_{\pm\text{1.51}}$ \\
auction  & 51.90$_{\pm\text{1.91}}$ & 57.35$_{\pm\text{1.53}}$ & 53.09$_{\pm\text{0.91}}$ & 52.35$_{\pm\text{1.90}}$ & 51.14$_{\pm\text{1.76}}$ & 56.82$_{\pm\text{1.20}}$ & \textbf{57.97}$_{\pm\text{1.16}}$ \\
abalone  & 11.59$_{\pm\text{2.69}}$ & N/A   & 8.49$_{\pm\text{1.28}}$  & 7.72$_{\pm\text{2.67}}$  & 9.95$_{\pm\text{2.48}}$  & N/A                      & \textbf{13.56}$_{\pm\text{1.64}}$ \\
statlog  & 56.22$_{\pm\text{3.20}}$ & 57.30$_{\pm\text{2.57}}$ & 53.12$_{\pm\text{1.52}}$ & 55.55$_{\pm\text{3.18}}$ & 53.07$_{\pm\text{2.95}}$ & 57.65$_{\pm\text{2.01}}$ & \textbf{57.85}$_{\pm\text{1.95}}$ \\
\bottomrule
\end{tabular}%
}
\end{table}

\FloatBarrier
\subsubsection{Results on larger sample sizes}
\label{appendix:results-larger-sample-sizes}

\begin{table}[h]
    \centering
    \caption{Test classification accuracy (\%) aggregated over six downstream predictors, comparing data augmentation with increased real data availability of the ``texture'' dataset. Note that ``N/A'' denotes that a specific generator was not applicable. On larger datasets, TabEBM still outperforms other generators, but training on real data alone appears sufficient. This highlights TabEBM's usefulness in fields with limited training samples.}
    \resizebox{\linewidth}{!}{%
    \begin{tabular}{l|r|rrrrrr|c}
    \toprule
        $N_{\text{real}}$                   & Baseline & SMOTE & TVAE & CTGAN & TabDDPM & TabPFGen & \textbf{TabEBM (Ours)} & \makecell[c]{\textbf{Accuracy improvements}\\\textbf{by TabEBM (\%)}}\\
    \midrule                            
        50                                  & 72.40$_{\pm\text{13.07}}$         & 76.40$_{\pm\text{10.50}}$         & 55.33$_{\pm\text{6.20}}$ & 54.80$_{\pm\text{12.97}}$ & 62.94$_{\pm\text{12.06}}$ & N/A               & \textbf{78.90}$_{\pm\text{7.96}}$ & +6.50 \\
        100                                 & 82.42$_{\pm\text{10.38}}$         & 84.35$_{\pm\text{9.67}}$          & 66.00$_{\pm\text{7.21}}$ & 69.49$_{\pm\text{10.93}}$ & 76.34$_{\pm\text{9.55}}$  & N/A               & \textbf{86.01}$_{\pm\text{7.36}}$ & +3.59 \\
        200                                 & 87.54$_{\pm\text{7.62}}$          & 89.29$_{\pm\text{6.20}}$          & 78.37$_{\pm\text{6.03}}$ & 82.44$_{\pm\text{7.15}}$  & 82.53$_{\pm\text{7.99}}$  & N/A               & \textbf{89.77}$_{\pm\text{5.77}}$ & +2.23 \\
        500                                 & 92.96$_{\pm\text{4.07}}$          & 93.69$_{\pm\text{3.83}}$          & 90.09$_{\pm\text{3.56}}$ & 91.48$_{\pm\text{3.50}}$  & 91.24$_{\pm\text{3.56}}$  & N/A               & \textbf{93.76}$_{\pm\text{3.64}}$ & +0.80 \\
        1000                                & \textbf{96.37}$_{\pm\text{2.17}}$ & 96.21$_{\pm\text{2.37}}$          & 93.61$_{\pm\text{2.10}}$ & 95.36$_{\pm\text{1.71}}$  & 94.56$_{\pm\text{1.59}}$  & N/A               & 96.30$_{\pm\text{2.30}}$          & -0.07 \\
        2000                                & 97.76$_{\pm\text{1.16}}$          & 96.84$_{\pm\text{1.46}}$          & 96.62$_{\pm\text{1.24}}$ & 97.10$_{\pm\text{0.84}}$  & 97.13$_{\pm\text{0.71}}$  & N/A               & \textbf{97.83}$_{\pm\text{1.45}}$ & +0.07 \\
        3000                                & 98.20$_{\pm\text{0.62}}$          & 98.28$_{\pm\text{0.90}}$          & 97.60$_{\pm\text{0.73}}$ & 97.60$_{\pm\text{0.41}}$  & 97.73$_{\pm\text{0.31}}$  & N/A               & \textbf{98.35}$_{\pm\text{0.91}}$ & +0.15 \\
        4000                                & 98.51$_{\pm\text{0.33}}$          & \textbf{98.59}$_{\pm\text{0.56}}$ & 98.11$_{\pm\text{0.43}}$ & 98.00$_{\pm\text{0.20}}$  & 98.46$_{\pm\text{0.14}}$  & N/A               & 98.55$_{\pm\text{0.58}}$          & +0.04 \\
    \bottomrule
    \end{tabular}%
    }
    \label{appendix-table:increasing-sample-size}
\end{table}

\FloatBarrier
\subsection{Results on Statistical Fidelity}
\label{appendix:res_fidelity}
We aim to provide a fair and coherent comparison between TabEBM and existing methods and thus we follow the widely adopted evaluation process in prior studies. Specifically, we compute the three statistical fidelity metrics with the open-source implementations from the well-established benchmark, Synthcity. We note that the previous studies~\cite{zhang2023mixed, qian2024synthcity} often operate under the assumption that the issues associated with multiple comparisons are less pronounced in generating low-dimensional tabular data, hence correction methods for multiple hypothesis testing are seldom employed. Following such assumptions, correction methods are not employed in this work. In addition, we would like to point out the imperfection of widely adopted univariate metrics (i.e., Inverse KL, KS test and test) in existing work. However, evaluating generators' ability to capture the joining feature relationships remains an open research question~\cite{tu2024causality}. We leave this for future work to explore.

\newpage
\subsubsection{Similarity between Real Train Data and Synthetic Data}
\begin{table}[hp]
\centering
\caption{\textbf{Inverse KL between real train data and synthetic data} on eight real-world tabular datasets with varied real data availability. We report the mean $\pm$ std balanced accuracy and average accuracy rank across datasets. A higher rank implies higher fidelity. Note that ``N/A'' denotes that a specific generator was not applicable, and the rank is computed with the mean result of other methods. We \textbf{bold} the highest result for each dataset of different sample sizes. TabEBM achieves the best overall performance against benchmark generators.}
\resizebox{\textwidth}{!}{

\setlength{\tabcolsep}{5pt}
\begin{tabular}{m{0.2cm}lr|rrrrrrrr|r}

\toprule

\multicolumn{2}{l}{Datasets}  & $N_{\text{real}}$ & SMOTE & TVAE & CTGAN & NFLOW & TabDDPM & ARF & GOGGLE & TabPFGen & \textbf{TabEBM} \\

\midrule

\multirow{24}{*}{\rotatebox{90}{\begin{tabular}{l} \textit{At most 10 classes} \end{tabular}}}& \multirow[c]{5}{*}{protein} & 20 & N/A & 0.11$_{\pm\text{0.01}}$ & 0.20$_{\pm\text{0.02}}$ & 0.34$_{\pm\text{0.05}}$ & 0.07$_{\pm\text{0.01}}$ & 0.22$_{\pm\text{0.02}}$ & 0.07$_{\pm\text{0.00}}$ & 0.46$_{\pm\text{0.13}}$ & \textbf{0.77}$_{\pm\text{0.04}}$ \\

& & 50 & 0.88$_{\pm\text{0.01}}$ & 0.80$_{\pm\text{0.02}}$ & 0.66$_{\pm\text{0.05}}$ & 0.87$_{\pm\text{0.01}}$ & 0.07$_{\pm\text{0.00}}$ & 0.87$_{\pm\text{0.01}}$ & 0.50$_{\pm\text{0.04}}$ & 0.82$_{\pm\text{0.06}}$ & \textbf{0.94}$_{\pm\text{0.02}}$ \\

& & 100 & 0.93$_{\pm\text{0.01}}$ & 0.79$_{\pm\text{0.02}}$ & 0.78$_{\pm\text{0.03}}$ & 0.90$_{\pm\text{0.03}}$ & 0.07$_{\pm\text{0.00}}$ & 0.91$_{\pm\text{0.01}}$ & 0.32$_{\pm\text{0.05}}$ & 0.92$_{\pm\text{0.02}}$ & \textbf{0.96}$_{\pm\text{0.01}}$ \\

& & 200 & 0.95$_{\pm\text{0.01}}$ & 0.75$_{\pm\text{0.03}}$ & 0.83$_{\pm\text{0.03}}$ & 0.93$_{\pm\text{0.01}}$ & 0.08$_{\pm\text{0.01}}$ & 0.93$_{\pm\text{0.01}}$ & 0.11$_{\pm\text{0.01}}$ & 0.94$_{\pm\text{0.01}}$ & \textbf{0.96}$_{\pm\text{0.01}}$ \\

& & 500 & 0.96$_{\pm\text{0.00}}$ & 0.70$_{\pm\text{0.02}}$ & 0.87$_{\pm\text{0.03}}$ & 0.94$_{\pm\text{0.00}}$ & 0.09$_{\pm\text{0.00}}$ & 0.95$_{\pm\text{0.01}}$ & 0.13$_{\pm\text{0.01}}$ & 0.96$_{\pm\text{0.01}}$ & \textbf{0.97}$_{\pm\text{0.01}}$ \\

\cmidrule{2-12}

& \multirow[c]{5}{*}{fourier} & 20 & N/A & 0.12$_{\pm\text{0.03}}$ & 0.15$_{\pm\text{0.02}}$ & 0.27$_{\pm\text{0.04}}$ & 0.50$_{\pm\text{0.03}}$ & 0.50$_{\pm\text{0.03}}$ & 0.50$_{\pm\text{0.04}}$ & \textbf{0.97}$_{\pm\text{0.00}}$ & 0.87$_{\pm\text{0.01}}$ \\

& & 50 & 0.93$_{\pm\text{0.01}}$ & 0.79$_{\pm\text{0.02}}$ & 0.66$_{\pm\text{0.05}}$ & 0.90$_{\pm\text{0.01}}$ & 0.07$_{\pm\text{0.00}}$ & 0.90$_{\pm\text{0.00}}$ & 0.47$_{\pm\text{0.05}}$ & 0.87$_{\pm\text{0.02}}$ & \textbf{0.95}$_{\pm\text{0.01}}$ \\

& & 100 & 0.95$_{\pm\text{0.01}}$ & 0.76$_{\pm\text{0.03}}$ & 0.81$_{\pm\text{0.03}}$ & 0.93$_{\pm\text{0.00}}$ & 0.07$_{\pm\text{0.00}}$ & 0.94$_{\pm\text{0.01}}$ & 0.20$_{\pm\text{0.05}}$ & 0.94$_{\pm\text{0.01}}$ & \textbf{0.97}$_{\pm\text{0.01}}$ \\

& & 200 & 0.97$_{\pm\text{0.01}}$ & 0.61$_{\pm\text{0.01}}$ & 0.82$_{\pm\text{0.02}}$ & 0.95$_{\pm\text{0.00}}$ & 0.09$_{\pm\text{0.01}}$ & 0.96$_{\pm\text{0.00}}$ & 0.08$_{\pm\text{0.01}}$ & 0.97$_{\pm\text{0.01}}$ & \textbf{0.98}$_{\pm\text{0.00}}$ \\

& & 500 & 0.97$_{\pm\text{0.00}}$ & 0.52$_{\pm\text{0.03}}$ & 0.90$_{\pm\text{0.02}}$ & 0.95$_{\pm\text{0.01}}$ & 0.10$_{\pm\text{0.01}}$ & 0.97$_{\pm\text{0.00}}$ & 0.09$_{\pm\text{0.00}}$ & 0.98$_{\pm\text{0.00}}$ & \textbf{0.98}$_{\pm\text{0.00}}$ \\

\cmidrule{2-12}

& \multirow[c]{5}{*}{biodeg} & 20 & 0.47$_{\pm\text{0.04}}$ & 0.43$_{\pm\text{0.04}}$ & 0.43$_{\pm\text{0.03}}$ & 0.50$_{\pm\text{0.05}}$ & 0.34$_{\pm\text{0.03}}$ & 0.51$_{\pm\text{0.04}}$ & 0.37$_{\pm\text{0.04}}$ & 0.60$_{\pm\text{0.07}}$ & \textbf{0.87}$_{\pm\text{0.04}}$ \\

& & 50 & 0.62$_{\pm\text{0.03}}$ & 0.59$_{\pm\text{0.02}}$ & 0.56$_{\pm\text{0.05}}$ & 0.63$_{\pm\text{0.05}}$ & 0.28$_{\pm\text{0.02}}$ & 0.65$_{\pm\text{0.03}}$ & 0.41$_{\pm\text{0.03}}$ & 0.75$_{\pm\text{0.05}}$ & \textbf{0.90}$_{\pm\text{0.02}}$ \\

& & 100 & 0.69$_{\pm\text{0.05}}$ & 0.66$_{\pm\text{0.03}}$ & 0.65$_{\pm\text{0.05}}$ & 0.67$_{\pm\text{0.04}}$ & 0.30$_{\pm\text{0.04}}$ & 0.69$_{\pm\text{0.05}}$ & 0.38$_{\pm\text{0.04}}$ & 0.76$_{\pm\text{0.04}}$ & \textbf{0.90}$_{\pm\text{0.04}}$ \\

& & 200 & 0.71$_{\pm\text{0.03}}$ & 0.65$_{\pm\text{0.03}}$ & 0.69$_{\pm\text{0.02}}$ & 0.68$_{\pm\text{0.03}}$ & 0.29$_{\pm\text{0.04}}$ & 0.69$_{\pm\text{0.03}}$ & 0.34$_{\pm\text{0.01}}$ & 0.79$_{\pm\text{0.04}}$ & \textbf{0.91}$_{\pm\text{0.02}}$ \\

& & 500 & 0.80$_{\pm\text{0.02}}$ & 0.68$_{\pm\text{0.02}}$ & 0.73$_{\pm\text{0.02}}$ & 0.75$_{\pm\text{0.02}}$ & 0.26$_{\pm\text{0.02}}$ & 0.73$_{\pm\text{0.02}}$ & 0.37$_{\pm\text{0.02}}$ & 0.81$_{\pm\text{0.04}}$ & \textbf{0.92}$_{\pm\text{0.02}}$ \\

\cmidrule{2-12}

& \multirow[c]{5}{*}{steel} & 20 & 0.45$_{\pm\text{0.05}}$ & 0.37$_{\pm\text{0.03}}$ & 0.40$_{\pm\text{0.04}}$ & 0.47$_{\pm\text{0.04}}$ & 0.17$_{\pm\text{0.03}}$ & 0.43$_{\pm\text{0.05}}$ & 0.29$_{\pm\text{0.05}}$ & 0.53$_{\pm\text{0.08}}$ & \textbf{0.84}$_{\pm\text{0.03}}$ \\

& & 50 & 0.70$_{\pm\text{0.02}}$ & 0.57$_{\pm\text{0.03}}$ & 0.59$_{\pm\text{0.04}}$ & 0.64$_{\pm\text{0.04}}$ & 0.13$_{\pm\text{0.01}}$ & 0.63$_{\pm\text{0.01}}$ & 0.23$_{\pm\text{0.03}}$ & 0.71$_{\pm\text{0.08}}$ & \textbf{0.91}$_{\pm\text{0.02}}$ \\

& & 100 & 0.71$_{\pm\text{0.04}}$ & 0.55$_{\pm\text{0.02}}$ & 0.63$_{\pm\text{0.02}}$ & 0.67$_{\pm\text{0.02}}$ & 0.13$_{\pm\text{0.01}}$ & 0.66$_{\pm\text{0.02}}$ & 0.20$_{\pm\text{0.02}}$ & 0.75$_{\pm\text{0.05}}$ & \textbf{0.92}$_{\pm\text{0.03}}$ \\

& & 200 & 0.75$_{\pm\text{0.01}}$ & 0.50$_{\pm\text{0.04}}$ & 0.65$_{\pm\text{0.03}}$ & 0.70$_{\pm\text{0.01}}$ & 0.13$_{\pm\text{0.02}}$ & 0.67$_{\pm\text{0.02}}$ & 0.17$_{\pm\text{0.01}}$ & 0.77$_{\pm\text{0.04}}$ & \textbf{0.93}$_{\pm\text{0.01}}$ \\

& & 500 & 0.75$_{\pm\text{0.01}}$ & 0.51$_{\pm\text{0.04}}$ & 0.66$_{\pm\text{0.04}}$ & 0.70$_{\pm\text{0.01}}$ & 0.14$_{\pm\text{0.01}}$ & 0.68$_{\pm\text{0.01}}$ & 0.19$_{\pm\text{0.01}}$ & 0.80$_{\pm\text{0.06}}$ & \textbf{0.94}$_{\pm\text{0.02}}$ \\

\cmidrule{2-12}

& \multirow[c]{4}{*}{stock} & 20 & 0.55$_{\pm\text{0.07}}$ & 0.45$_{\pm\text{0.07}}$ & 0.43$_{\pm\text{0.05}}$ & 0.60$_{\pm\text{0.05}}$ & 0.24$_{\pm\text{0.04}}$ & 0.52$_{\pm\text{0.06}}$ & 0.35$_{\pm\text{0.11}}$ & 0.68$_{\pm\text{0.12}}$ & \textbf{0.89}$_{\pm\text{0.02}}$ \\

& & 50 & 0.92$_{\pm\text{0.02}}$ & 0.73$_{\pm\text{0.06}}$ & 0.78$_{\pm\text{0.07}}$ & 0.86$_{\pm\text{0.03}}$ & 0.37$_{\pm\text{0.07}}$ & 0.88$_{\pm\text{0.01}}$ & 0.32$_{\pm\text{0.11}}$ & 0.91$_{\pm\text{0.03}}$ & \textbf{0.95}$_{\pm\text{0.02}}$ \\

& & 100 & 0.96$_{\pm\text{0.02}}$ & 0.67$_{\pm\text{0.07}}$ & 0.83$_{\pm\text{0.05}}$ & 0.91$_{\pm\text{0.03}}$ & 0.49$_{\pm\text{0.10}}$ & 0.93$_{\pm\text{0.01}}$ & 0.17$_{\pm\text{0.04}}$ & 0.95$_{\pm\text{0.02}}$ & \textbf{0.97}$_{\pm\text{0.01}}$ \\

& & 200 & 0.98$_{\pm\text{0.01}}$ & 0.63$_{\pm\text{0.04}}$ & 0.80$_{\pm\text{0.08}}$ & 0.92$_{\pm\text{0.02}}$ & 0.83$_{\pm\text{0.09}}$ & 0.95$_{\pm\text{0.01}}$ & 0.15$_{\pm\text{0.00}}$ & \textbf{0.98}$_{\pm\text{0.01}}$ & 0.98$_{\pm\text{0.00}}$ \\

\midrule

\multirow{9}{*}{\rotatebox{90}{\begin{tabular}{l} \textit{More than 10 classes} \end{tabular}}}& \multirow[c]{3}{*}{energy} & 50 & N/A & 0.25$_{\pm\text{0.06}}$ & 0.34$_{\pm\text{0.06}}$ & 0.42$_{\pm\text{0.06}}$ & 0.24$_{\pm\text{0.05}}$ & 0.49$_{\pm\text{0.10}}$ & 0.24$_{\pm\text{0.09}}$ & N/A & \textbf{0.80}$_{\pm\text{0.04}}$ \\

& & 100 & N/A & 0.28$_{\pm\text{0.07}}$ & 0.42$_{\pm\text{0.09}}$ & 0.44$_{\pm\text{0.05}}$ & 0.22$_{\pm\text{0.07}}$ & 0.41$_{\pm\text{0.08}}$ & 0.16$_{\pm\text{0.04}}$ & N/A & \textbf{0.89}$_{\pm\text{0.01}}$ \\

& & 200 & N/A & 0.30$_{\pm\text{0.08}}$ & 0.43$_{\pm\text{0.08}}$ & 0.47$_{\pm\text{0.09}}$ & 0.25$_{\pm\text{0.06}}$ & 0.40$_{\pm\text{0.08}}$ & 0.12$_{\pm\text{0.04}}$ & N/A & \textbf{0.91}$_{\pm\text{0.01}}$ \\

\cmidrule{2-12}

& \multirow[c]{2}{*}{collins} & 100 & N/A & 0.72$_{\pm\text{0.02}}$ & 0.84$_{\pm\text{0.04}}$ & 0.90$_{\pm\text{0.02}}$ & 0.44$_{\pm\text{0.11}}$ & 0.91$_{\pm\text{0.02}}$ & 0.28$_{\pm\text{0.08}}$ & N/A & \textbf{0.94}$_{\pm\text{0.01}}$ \\

& & 200 & 0.94$_{\pm\text{0.01}}$ & 0.64$_{\pm\text{0.05}}$ & 0.87$_{\pm\text{0.04}}$ & 0.92$_{\pm\text{0.02}}$ & 0.44$_{\pm\text{0.05}}$ & 0.93$_{\pm\text{0.01}}$ & 0.23$_{\pm\text{0.10}}$ & N/A & \textbf{0.96}$_{\pm\text{0.01}}$ \\

\cmidrule{2-12}

& \multirow[c]{4}{*}{texture} & 50 & 0.89$_{\pm\text{0.04}}$ & 0.74$_{\pm\text{0.04}}$ & 0.71$_{\pm\text{0.06}}$ & 0.88$_{\pm\text{0.03}}$ & 0.10$_{\pm\text{0.01}}$ & 0.88$_{\pm\text{0.02}}$ & 0.45$_{\pm\text{0.10}}$ & N/A & \textbf{0.93}$_{\pm\text{0.04}}$ \\

& & 100 & 0.96$_{\pm\text{0.01}}$ & 0.67$_{\pm\text{0.05}}$ & 0.81$_{\pm\text{0.07}}$ & 0.91$_{\pm\text{0.02}}$ & 0.11$_{\pm\text{0.02}}$ & 0.92$_{\pm\text{0.02}}$ & 0.22$_{\pm\text{0.06}}$ & N/A & \textbf{0.97}$_{\pm\text{0.01}}$ \\

& & 200 & 0.96$_{\pm\text{0.02}}$ & 0.56$_{\pm\text{0.04}}$ & 0.80$_{\pm\text{0.07}}$ & 0.93$_{\pm\text{0.01}}$ & 0.12$_{\pm\text{0.01}}$ & 0.95$_{\pm\text{0.01}}$ & 0.08$_{\pm\text{0.01}}$ & N/A & \textbf{0.98}$_{\pm\text{0.01}}$ \\

& & 500 & 0.97$_{\pm\text{0.02}}$ & 0.63$_{\pm\text{0.06}}$ & 0.84$_{\pm\text{0.04}}$ & 0.93$_{\pm\text{0.02}}$ & 0.14$_{\pm\text{0.01}}$ & 0.96$_{\pm\text{0.01}}$ & 0.11$_{\pm\text{0.01}}$ & N/A & \textbf{0.98}$_{\pm\text{0.01}}$ \\

\midrule
\rowcolor{Gainsboro!60}
\multicolumn{3}{l|}{\textbf{Average rank}} & 3.24$_{\pm\text{1.30}}$ & 6.76$_{\pm\text{0.75}}$ & 5.91$_{\pm\text{1.10}}$ & 4.12$_{\pm\text{1.08}}$ & 8.42$_{\pm\text{1.17}}$ & 3.97$_{\pm\text{0.95}}$ & 8.21$_{\pm\text{0.89}}$ & 3.30$_{\pm\text{1.79}}$ & \textbf{1.06}$_{\pm\text{0.24}}$ \\

\bottomrule

\end{tabular}
}
\end{table}
\clearpage

\begin{table}[p]
\centering
\caption{\textbf{KS test between real train data and synthetic data} on eight real-world tabular datasets with varied real data availability. We report the mean $\pm$ std result and average rank across datasets. A higher rank implies higher fidelity. Note that ``N/A'' denotes that a specific generator was not applicable, and the rank is computed with the mean result of other methods. We \textbf{bold} the highest result for each dataset of different sample sizes. TabEBM achieves the best overall performance against benchmark generators.}
\resizebox{\textwidth}{!}{

\setlength{\tabcolsep}{5pt}
\begin{tabular}{m{0.2cm}lr|rrrrrrrr|r}

\toprule

\multicolumn{2}{l}{Datasets}  & $N_{\text{real}}$ & SMOTE & TVAE & CTGAN & NFLOW & TabDDPM & ARF & GOGGLE & TabPFGen & \textbf{TabEBM} \\

\midrule

\multirow{24}{*}{\rotatebox{90}{\begin{tabular}{l} \textit{At most 10 classes} \end{tabular}}}& \multirow[c]{5}{*}{protein} & 20 & N/A & 0.72$_{\pm\text{0.02}}$ & 0.75$_{\pm\text{0.03}}$ & 0.81$_{\pm\text{0.01}}$ & 0.63$_{\pm\text{0.01}}$ & 0.80$_{\pm\text{0.01}}$ & 0.44$_{\pm\text{0.00}}$ & 0.84$_{\pm\text{0.02}}$ & \textbf{0.87}$_{\pm\text{0.01}}$ \\

& & 50 & 0.90$_{\pm\text{0.01}}$ & 0.87$_{\pm\text{0.00}}$ & 0.87$_{\pm\text{0.01}}$ & 0.89$_{\pm\text{0.01}}$ & 0.62$_{\pm\text{0.01}}$ & 0.89$_{\pm\text{0.00}}$ & 0.73$_{\pm\text{0.02}}$ & 0.91$_{\pm\text{0.00}}$ & \textbf{0.93}$_{\pm\text{0.00}}$ \\

& & 100 & 0.92$_{\pm\text{0.00}}$ & 0.88$_{\pm\text{0.01}}$ & 0.91$_{\pm\text{0.00}}$ & 0.91$_{\pm\text{0.00}}$ & 0.60$_{\pm\text{0.01}}$ & 0.91$_{\pm\text{0.00}}$ & 0.66$_{\pm\text{0.02}}$ & 0.94$_{\pm\text{0.00}}$ & \textbf{0.94}$_{\pm\text{0.00}}$ \\

& & 200 & 0.94$_{\pm\text{0.00}}$ & 0.87$_{\pm\text{0.01}}$ & 0.92$_{\pm\text{0.00}}$ & 0.93$_{\pm\text{0.00}}$ & 0.59$_{\pm\text{0.01}}$ & 0.93$_{\pm\text{0.00}}$ & 0.58$_{\pm\text{0.01}}$ & 0.95$_{\pm\text{0.00}}$ & \textbf{0.95}$_{\pm\text{0.00}}$ \\

& & 500 & 0.95$_{\pm\text{0.00}}$ & 0.83$_{\pm\text{0.01}}$ & 0.92$_{\pm\text{0.00}}$ & 0.93$_{\pm\text{0.00}}$ & 0.59$_{\pm\text{0.00}}$ & 0.94$_{\pm\text{0.00}}$ & 0.63$_{\pm\text{0.01}}$ & 0.95$_{\pm\text{0.00}}$ & \textbf{0.96}$_{\pm\text{0.00}}$ \\

\cmidrule{2-12}

& \multirow[c]{5}{*}{fourier} & 20 & N/A & 0.71$_{\pm\text{0.04}}$ & 0.73$_{\pm\text{0.02}}$ & 0.79$_{\pm\text{0.01}}$ & 0.85$_{\pm\text{0.00}}$ & 0.85$_{\pm\text{0.00}}$ & 0.85$_{\pm\text{0.00}}$ & \textbf{0.94}$_{\pm\text{0.00}}$ & 0.90$_{\pm\text{0.00}}$ \\

& & 50 & 0.92$_{\pm\text{0.00}}$ & 0.87$_{\pm\text{0.00}}$ & 0.88$_{\pm\text{0.01}}$ & 0.91$_{\pm\text{0.00}}$ & 0.64$_{\pm\text{0.01}}$ & 0.90$_{\pm\text{0.00}}$ & 0.73$_{\pm\text{0.02}}$ & 0.93$_{\pm\text{0.00}}$ & \textbf{0.94}$_{\pm\text{0.00}}$ \\

& & 100 & 0.94$_{\pm\text{0.00}}$ & 0.87$_{\pm\text{0.01}}$ & 0.92$_{\pm\text{0.00}}$ & 0.93$_{\pm\text{0.00}}$ & 0.63$_{\pm\text{0.01}}$ & 0.93$_{\pm\text{0.00}}$ & 0.62$_{\pm\text{0.02}}$ & 0.95$_{\pm\text{0.00}}$ & \textbf{0.95}$_{\pm\text{0.00}}$ \\

& & 200 & 0.95$_{\pm\text{0.00}}$ & 0.83$_{\pm\text{0.01}}$ & 0.92$_{\pm\text{0.00}}$ & 0.94$_{\pm\text{0.00}}$ & 0.62$_{\pm\text{0.01}}$ & 0.94$_{\pm\text{0.00}}$ & 0.61$_{\pm\text{0.03}}$ & \textbf{0.97}$_{\pm\text{0.00}}$ & 0.96$_{\pm\text{0.00}}$ \\

& & 500 & 0.96$_{\pm\text{0.00}}$ & 0.81$_{\pm\text{0.00}}$ & 0.94$_{\pm\text{0.00}}$ & 0.95$_{\pm\text{0.00}}$ & 0.62$_{\pm\text{0.01}}$ & 0.95$_{\pm\text{0.00}}$ & 0.63$_{\pm\text{0.01}}$ & \textbf{0.97}$_{\pm\text{0.00}}$ & 0.97$_{\pm\text{0.00}}$ \\

\cmidrule{2-12}

& \multirow[c]{5}{*}{biodeg} & 20 & 0.63$_{\pm\text{0.04}}$ & 0.62$_{\pm\text{0.04}}$ & 0.64$_{\pm\text{0.03}}$ & 0.64$_{\pm\text{0.04}}$ & 0.56$_{\pm\text{0.04}}$ & 0.63$_{\pm\text{0.04}}$ & 0.65$_{\pm\text{0.03}}$ & 0.59$_{\pm\text{0.02}}$ & \textbf{0.70}$_{\pm\text{0.01}}$ \\

& & 50 & 0.57$_{\pm\text{0.03}}$ & 0.57$_{\pm\text{0.03}}$ & 0.59$_{\pm\text{0.03}}$ & 0.60$_{\pm\text{0.03}}$ & 0.48$_{\pm\text{0.03}}$ & 0.59$_{\pm\text{0.03}}$ & 0.63$_{\pm\text{0.02}}$ & 0.61$_{\pm\text{0.04}}$ & \textbf{0.73}$_{\pm\text{0.00}}$ \\

& & 100 & 0.56$_{\pm\text{0.03}}$ & 0.55$_{\pm\text{0.03}}$ & 0.57$_{\pm\text{0.03}}$ & 0.59$_{\pm\text{0.03}}$ & 0.46$_{\pm\text{0.03}}$ & 0.58$_{\pm\text{0.03}}$ & 0.59$_{\pm\text{0.03}}$ & 0.59$_{\pm\text{0.02}}$ & \textbf{0.73}$_{\pm\text{0.01}}$ \\

& & 200 & 0.53$_{\pm\text{0.01}}$ & 0.52$_{\pm\text{0.01}}$ & 0.53$_{\pm\text{0.02}}$ & 0.56$_{\pm\text{0.02}}$ & 0.43$_{\pm\text{0.02}}$ & 0.55$_{\pm\text{0.02}}$ & 0.55$_{\pm\text{0.02}}$ & 0.58$_{\pm\text{0.02}}$ & \textbf{0.72}$_{\pm\text{0.01}}$ \\

& & 500 & 0.53$_{\pm\text{0.00}}$ & 0.50$_{\pm\text{0.01}}$ & 0.52$_{\pm\text{0.01}}$ & 0.55$_{\pm\text{0.01}}$ & 0.43$_{\pm\text{0.01}}$ & 0.56$_{\pm\text{0.01}}$ & 0.56$_{\pm\text{0.01}}$ & 0.57$_{\pm\text{0.01}}$ & \textbf{0.72}$_{\pm\text{0.01}}$ \\

\cmidrule{2-12}

& \multirow[c]{5}{*}{steel} & 20 & 0.67$_{\pm\text{0.02}}$ & 0.63$_{\pm\text{0.02}}$ & 0.64$_{\pm\text{0.03}}$ & 0.66$_{\pm\text{0.02}}$ & 0.54$_{\pm\text{0.02}}$ & 0.65$_{\pm\text{0.02}}$ & 0.57$_{\pm\text{0.03}}$ & 0.67$_{\pm\text{0.02}}$ & \textbf{0.76}$_{\pm\text{0.01}}$ \\

& & 50 & 0.69$_{\pm\text{0.02}}$ & 0.64$_{\pm\text{0.03}}$ & 0.65$_{\pm\text{0.03}}$ & 0.67$_{\pm\text{0.02}}$ & 0.51$_{\pm\text{0.02}}$ & 0.67$_{\pm\text{0.03}}$ & 0.56$_{\pm\text{0.02}}$ & 0.72$_{\pm\text{0.02}}$ & \textbf{0.79}$_{\pm\text{0.01}}$ \\

& & 100 & 0.69$_{\pm\text{0.02}}$ & 0.63$_{\pm\text{0.02}}$ & 0.65$_{\pm\text{0.02}}$ & 0.67$_{\pm\text{0.01}}$ & 0.50$_{\pm\text{0.01}}$ & 0.65$_{\pm\text{0.02}}$ & 0.55$_{\pm\text{0.02}}$ & 0.71$_{\pm\text{0.03}}$ & \textbf{0.79}$_{\pm\text{0.01}}$ \\

& & 200 & 0.70$_{\pm\text{0.01}}$ & 0.61$_{\pm\text{0.02}}$ & 0.66$_{\pm\text{0.02}}$ & 0.69$_{\pm\text{0.01}}$ & 0.50$_{\pm\text{0.01}}$ & 0.65$_{\pm\text{0.02}}$ & 0.55$_{\pm\text{0.01}}$ & 0.71$_{\pm\text{0.02}}$ & \textbf{0.79}$_{\pm\text{0.01}}$ \\

& & 500 & 0.70$_{\pm\text{0.01}}$ & 0.62$_{\pm\text{0.02}}$ & 0.66$_{\pm\text{0.02}}$ & 0.68$_{\pm\text{0.01}}$ & 0.49$_{\pm\text{0.01}}$ & 0.65$_{\pm\text{0.01}}$ & 0.58$_{\pm\text{0.01}}$ & 0.74$_{\pm\text{0.02}}$ & \textbf{0.80}$_{\pm\text{0.01}}$ \\

\cmidrule{2-12}

& \multirow[c]{4}{*}{stock} & 20 & 0.86$_{\pm\text{0.02}}$ & 0.82$_{\pm\text{0.01}}$ & 0.82$_{\pm\text{0.02}}$ & 0.86$_{\pm\text{0.01}}$ & 0.74$_{\pm\text{0.03}}$ & 0.86$_{\pm\text{0.02}}$ & 0.63$_{\pm\text{0.07}}$ & 0.89$_{\pm\text{0.01}}$ & \textbf{0.91}$_{\pm\text{0.01}}$ \\

& & 50 & 0.92$_{\pm\text{0.01}}$ & 0.86$_{\pm\text{0.01}}$ & 0.88$_{\pm\text{0.01}}$ & 0.90$_{\pm\text{0.01}}$ & 0.84$_{\pm\text{0.02}}$ & 0.91$_{\pm\text{0.01}}$ & 0.68$_{\pm\text{0.04}}$ & 0.93$_{\pm\text{0.01}}$ & \textbf{0.94}$_{\pm\text{0.00}}$ \\

& & 100 & 0.94$_{\pm\text{0.01}}$ & 0.86$_{\pm\text{0.01}}$ & 0.90$_{\pm\text{0.01}}$ & 0.92$_{\pm\text{0.01}}$ & 0.88$_{\pm\text{0.02}}$ & 0.93$_{\pm\text{0.01}}$ & 0.63$_{\pm\text{0.02}}$ & \textbf{0.95}$_{\pm\text{0.01}}$ & 0.95$_{\pm\text{0.01}}$ \\

& & 200 & 0.95$_{\pm\text{0.01}}$ & 0.86$_{\pm\text{0.01}}$ & 0.90$_{\pm\text{0.01}}$ & 0.93$_{\pm\text{0.01}}$ & 0.92$_{\pm\text{0.01}}$ & 0.94$_{\pm\text{0.00}}$ & 0.63$_{\pm\text{0.01}}$ & \textbf{0.96}$_{\pm\text{0.00}}$ & 0.95$_{\pm\text{0.00}}$ \\

\midrule

\multirow{9}{*}{\rotatebox{90}{\begin{tabular}{l} \textit{More than 10 classes} \end{tabular}}}& \multirow[c]{3}{*}{energy} & 50 & N/A & 0.70$_{\pm\text{0.02}}$ & 0.69$_{\pm\text{0.04}}$ & 0.73$_{\pm\text{0.01}}$ & 0.65$_{\pm\text{0.03}}$ & 0.72$_{\pm\text{0.01}}$ & 0.63$_{\pm\text{0.03}}$ & N/A & \textbf{0.78}$_{\pm\text{0.01}}$ \\

& & 100 & N/A & 0.69$_{\pm\text{0.02}}$ & 0.74$_{\pm\text{0.01}}$ & 0.74$_{\pm\text{0.01}}$ & 0.69$_{\pm\text{0.01}}$ & 0.74$_{\pm\text{0.01}}$ & 0.63$_{\pm\text{0.03}}$ & N/A & \textbf{0.81}$_{\pm\text{0.01}}$ \\

& & 200 & N/A & 0.71$_{\pm\text{0.01}}$ & 0.74$_{\pm\text{0.02}}$ & 0.75$_{\pm\text{0.01}}$ & 0.67$_{\pm\text{0.01}}$ & 0.75$_{\pm\text{0.00}}$ & 0.63$_{\pm\text{0.02}}$ & N/A & \textbf{0.83}$_{\pm\text{0.01}}$ \\

\cmidrule{2-12}

& \multirow[c]{2}{*}{collins} & 100 & N/A & 0.85$_{\pm\text{0.01}}$ & 0.89$_{\pm\text{0.01}}$ & 0.90$_{\pm\text{0.01}}$ & 0.82$_{\pm\text{0.04}}$ & 0.90$_{\pm\text{0.01}}$ & 0.65$_{\pm\text{0.04}}$ & N/A & \textbf{0.93}$_{\pm\text{0.00}}$ \\

& & 200 & 0.93$_{\pm\text{0.00}}$ & 0.83$_{\pm\text{0.02}}$ & 0.91$_{\pm\text{0.01}}$ & 0.92$_{\pm\text{0.00}}$ & 0.80$_{\pm\text{0.03}}$ & 0.92$_{\pm\text{0.00}}$ & 0.63$_{\pm\text{0.05}}$ & N/A & \textbf{0.94}$_{\pm\text{0.00}}$ \\

\cmidrule{2-12}

& \multirow[c]{4}{*}{texture} & 50 & 0.92$_{\pm\text{0.01}}$ & 0.86$_{\pm\text{0.01}}$ & 0.88$_{\pm\text{0.01}}$ & 0.90$_{\pm\text{0.01}}$ & 0.57$_{\pm\text{0.03}}$ & 0.90$_{\pm\text{0.01}}$ & 0.71$_{\pm\text{0.05}}$ & N/A & \textbf{0.93}$_{\pm\text{0.01}}$ \\

& & 100 & 0.94$_{\pm\text{0.01}}$ & 0.86$_{\pm\text{0.01}}$ & 0.91$_{\pm\text{0.01}}$ & 0.92$_{\pm\text{0.00}}$ & 0.61$_{\pm\text{0.02}}$ & 0.92$_{\pm\text{0.01}}$ & 0.63$_{\pm\text{0.03}}$ & N/A & \textbf{0.95}$_{\pm\text{0.00}}$ \\

& & 200 & 0.96$_{\pm\text{0.01}}$ & 0.83$_{\pm\text{0.01}}$ & 0.91$_{\pm\text{0.01}}$ & 0.93$_{\pm\text{0.00}}$ & 0.62$_{\pm\text{0.01}}$ & 0.94$_{\pm\text{0.00}}$ & 0.60$_{\pm\text{0.01}}$ & N/A & \textbf{0.96}$_{\pm\text{0.00}}$ \\

& & 500 & 0.97$_{\pm\text{0.00}}$ & 0.85$_{\pm\text{0.01}}$ & 0.91$_{\pm\text{0.01}}$ & 0.93$_{\pm\text{0.00}}$ & 0.61$_{\pm\text{0.01}}$ & 0.95$_{\pm\text{0.00}}$ & 0.64$_{\pm\text{0.01}}$ & N/A & \textbf{0.97}$_{\pm\text{0.00}}$ \\

\midrule
\rowcolor{Gainsboro!60}
\multicolumn{3}{l|}{\textbf{Average rank}} & 3.91$_{\pm\text{1.65}}$ & 6.97$_{\pm\text{0.92}}$ & 5.76$_{\pm\text{1.00}}$ & 3.97$_{\pm\text{1.07}}$ & 8.33$_{\pm\text{1.05}}$ & 4.15$_{\pm\text{0.94}}$ & 7.55$_{\pm\text{2.22}}$ & 3.21$_{\pm\text{2.16}}$ & \textbf{1.15}$_{\pm\text{0.36}}$ \\

\bottomrule

\end{tabular}
}
\end{table}
\clearpage

\begin{table}[p]
\centering
\caption{\textbf{$\chi^2$ test between real train data and synthetic data} on eight real-world tabular datasets with varied real data availability. We report the mean $\pm$ std result and average rank across datasets. A higher rank implies higher fidelity. Note that ``N/A'' denotes that a specific generator was not applicable, and the rank is computed with the mean result of other methods. We \textbf{bold} the highest result for each dataset of different sample sizes. TabEBM achieves the best overall performance against benchmark generators.}
\resizebox{\textwidth}{!}{

\setlength{\tabcolsep}{5pt}
\begin{tabular}{m{0.2cm}lr|rrrrrrrr|r}

\toprule

\multicolumn{2}{l}{Datasets}  & $N_{\text{real}}$ & SMOTE & TVAE & CTGAN & NFLOW & TabDDPM & ARF & GOGGLE & TabPFGen & \textbf{TabEBM} \\

\midrule

\multirow{24}{*}{\rotatebox{90}{\begin{tabular}{l} \textit{At most 10 classes} \end{tabular}}}& \multirow[c]{5}{*}{protein} & 20 & N/A & 0.02$_{\pm\text{0.01}}$ & 0.08$_{\pm\text{0.03}}$ & 0.19$_{\pm\text{0.07}}$ & 0.02$_{\pm\text{0.01}}$ & 0.05$_{\pm\text{0.03}}$ & 0.02$_{\pm\text{0.01}}$ & 0.33$_{\pm\text{0.23}}$ & \textbf{0.92}$_{\pm\text{0.07}}$ \\

& & 50 & 0.95$_{\pm\text{0.04}}$ & 0.84$_{\pm\text{0.05}}$ & 0.50$_{\pm\text{0.10}}$ & 0.92$_{\pm\text{0.04}}$ & 0.01$_{\pm\text{0.00}}$ & \textbf{0.98}$_{\pm\text{0.02}}$ & 0.53$_{\pm\text{0.05}}$ & 0.63$_{\pm\text{0.14}}$ & 0.96$_{\pm\text{0.04}}$ \\

& & 100 & 0.86$_{\pm\text{0.06}}$ & 0.62$_{\pm\text{0.07}}$ & 0.53$_{\pm\text{0.09}}$ & 0.83$_{\pm\text{0.10}}$ & 0.01$_{\pm\text{0.00}}$ & 0.89$_{\pm\text{0.03}}$ & 0.27$_{\pm\text{0.07}}$ & 0.70$_{\pm\text{0.10}}$ & \textbf{0.91}$_{\pm\text{0.06}}$ \\

& & 200 & 0.80$_{\pm\text{0.05}}$ & 0.46$_{\pm\text{0.07}}$ & 0.46$_{\pm\text{0.09}}$ & 0.77$_{\pm\text{0.05}}$ & 0.01$_{\pm\text{0.00}}$ & 0.76$_{\pm\text{0.06}}$ & 0.02$_{\pm\text{0.02}}$ & 0.66$_{\pm\text{0.08}}$ & \textbf{0.81}$_{\pm\text{0.08}}$ \\

& & 500 & 0.62$_{\pm\text{0.06}}$ & 0.25$_{\pm\text{0.05}}$ & 0.36$_{\pm\text{0.06}}$ & 0.61$_{\pm\text{0.04}}$ & 0.01$_{\pm\text{0.00}}$ & 0.57$_{\pm\text{0.05}}$ & 0.01$_{\pm\text{0.00}}$ & 0.65$_{\pm\text{0.07}}$ & \textbf{0.70}$_{\pm\text{0.05}}$ \\

\cmidrule{2-12}

& \multirow[c]{5}{*}{fourier} & 20 & N/A & 0.02$_{\pm\text{0.02}}$ & 0.05$_{\pm\text{0.02}}$ & 0.15$_{\pm\text{0.05}}$ & 0.37$_{\pm\text{0.04}}$ & 0.35$_{\pm\text{0.05}}$ & 0.36$_{\pm\text{0.06}}$ & \textbf{1.00}$_{\pm\text{0.00}}$ & 1.00$_{\pm\text{0.00}}$ \\

& & 50 & 0.99$_{\pm\text{0.01}}$ & 0.87$_{\pm\text{0.04}}$ & 0.52$_{\pm\text{0.10}}$ & 0.97$_{\pm\text{0.02}}$ & 0.01$_{\pm\text{0.00}}$ & \textbf{0.99}$_{\pm\text{0.01}}$ & 0.49$_{\pm\text{0.06}}$ & 0.74$_{\pm\text{0.05}}$ & 0.99$_{\pm\text{0.01}}$ \\

& & 100 & 0.96$_{\pm\text{0.02}}$ & 0.75$_{\pm\text{0.04}}$ & 0.69$_{\pm\text{0.07}}$ & 0.95$_{\pm\text{0.02}}$ & 0.01$_{\pm\text{0.00}}$ & \textbf{0.98}$_{\pm\text{0.02}}$ & 0.16$_{\pm\text{0.05}}$ & 0.82$_{\pm\text{0.06}}$ & 0.97$_{\pm\text{0.03}}$ \\

& & 200 & 0.92$_{\pm\text{0.03}}$ & 0.41$_{\pm\text{0.07}}$ & 0.59$_{\pm\text{0.04}}$ & 0.92$_{\pm\text{0.03}}$ & 0.01$_{\pm\text{0.00}}$ & \textbf{0.95}$_{\pm\text{0.02}}$ & 0.02$_{\pm\text{0.01}}$ & 0.85$_{\pm\text{0.05}}$ & 0.95$_{\pm\text{0.03}}$ \\

& & 500 & 0.80$_{\pm\text{0.06}}$ & 0.14$_{\pm\text{0.04}}$ & 0.59$_{\pm\text{0.06}}$ & 0.76$_{\pm\text{0.07}}$ & 0.01$_{\pm\text{0.00}}$ & 0.84$_{\pm\text{0.04}}$ & 0.01$_{\pm\text{0.00}}$ & 0.81$_{\pm\text{0.04}}$ & \textbf{0.84}$_{\pm\text{0.02}}$ \\

\cmidrule{2-12}

& \multirow[c]{5}{*}{biodeg} & 20 & 0.23$_{\pm\text{0.06}}$ & 0.21$_{\pm\text{0.06}}$ & 0.16$_{\pm\text{0.04}}$ & 0.28$_{\pm\text{0.06}}$ & 0.08$_{\pm\text{0.04}}$ & 0.28$_{\pm\text{0.07}}$ & 0.12$_{\pm\text{0.06}}$ & 0.29$_{\pm\text{0.10}}$ & \textbf{0.71}$_{\pm\text{0.06}}$ \\

& & 50 & 0.39$_{\pm\text{0.02}}$ & 0.33$_{\pm\text{0.05}}$ & 0.28$_{\pm\text{0.06}}$ & 0.41$_{\pm\text{0.07}}$ & 0.05$_{\pm\text{0.02}}$ & 0.45$_{\pm\text{0.05}}$ & 0.15$_{\pm\text{0.03}}$ & 0.44$_{\pm\text{0.09}}$ & \textbf{0.75}$_{\pm\text{0.06}}$ \\

& & 100 & 0.35$_{\pm\text{0.06}}$ & 0.26$_{\pm\text{0.06}}$ & 0.30$_{\pm\text{0.08}}$ & 0.38$_{\pm\text{0.06}}$ & 0.04$_{\pm\text{0.01}}$ & 0.42$_{\pm\text{0.05}}$ & 0.08$_{\pm\text{0.04}}$ & 0.37$_{\pm\text{0.13}}$ & \textbf{0.67}$_{\pm\text{0.14}}$ \\

& & 200 & 0.25$_{\pm\text{0.06}}$ & 0.19$_{\pm\text{0.05}}$ & 0.24$_{\pm\text{0.05}}$ & 0.30$_{\pm\text{0.07}}$ & 0.03$_{\pm\text{0.01}}$ & 0.31$_{\pm\text{0.06}}$ & 0.02$_{\pm\text{0.00}}$ & 0.37$_{\pm\text{0.06}}$ & \textbf{0.59}$_{\pm\text{0.08}}$ \\

& & 500 & 0.22$_{\pm\text{0.08}}$ & 0.10$_{\pm\text{0.04}}$ & 0.17$_{\pm\text{0.05}}$ & 0.25$_{\pm\text{0.07}}$ & 0.02$_{\pm\text{0.00}}$ & 0.23$_{\pm\text{0.06}}$ & 0.02$_{\pm\text{0.00}}$ & 0.26$_{\pm\text{0.08}}$ & \textbf{0.44}$_{\pm\text{0.09}}$ \\

\cmidrule{2-12}

& \multirow[c]{5}{*}{steel} & 20 & 0.37$_{\pm\text{0.08}}$ & 0.32$_{\pm\text{0.05}}$ & 0.32$_{\pm\text{0.04}}$ & 0.40$_{\pm\text{0.08}}$ & 0.04$_{\pm\text{0.02}}$ & 0.40$_{\pm\text{0.08}}$ & 0.23$_{\pm\text{0.07}}$ & 0.34$_{\pm\text{0.13}}$ & \textbf{0.81}$_{\pm\text{0.05}}$ \\

& & 50 & 0.68$_{\pm\text{0.02}}$ & 0.50$_{\pm\text{0.06}}$ & 0.51$_{\pm\text{0.08}}$ & 0.64$_{\pm\text{0.05}}$ & 0.03$_{\pm\text{0.00}}$ & 0.67$_{\pm\text{0.03}}$ & 0.10$_{\pm\text{0.05}}$ & 0.49$_{\pm\text{0.15}}$ & \textbf{0.85}$_{\pm\text{0.06}}$ \\

& & 100 & 0.61$_{\pm\text{0.09}}$ & 0.37$_{\pm\text{0.07}}$ & 0.47$_{\pm\text{0.08}}$ & 0.61$_{\pm\text{0.04}}$ & 0.03$_{\pm\text{0.00}}$ & 0.61$_{\pm\text{0.07}}$ & 0.06$_{\pm\text{0.02}}$ & 0.51$_{\pm\text{0.07}}$ & \textbf{0.81}$_{\pm\text{0.08}}$ \\

& & 200 & 0.60$_{\pm\text{0.04}}$ & 0.23$_{\pm\text{0.04}}$ & 0.42$_{\pm\text{0.06}}$ & 0.60$_{\pm\text{0.05}}$ & 0.03$_{\pm\text{0.00}}$ & 0.53$_{\pm\text{0.07}}$ & 0.02$_{\pm\text{0.00}}$ & 0.52$_{\pm\text{0.07}}$ & \textbf{0.77}$_{\pm\text{0.06}}$ \\

& & 500 & 0.46$_{\pm\text{0.06}}$ & 0.17$_{\pm\text{0.07}}$ & 0.35$_{\pm\text{0.07}}$ & 0.55$_{\pm\text{0.06}}$ & 0.03$_{\pm\text{0.00}}$ & 0.42$_{\pm\text{0.04}}$ & 0.02$_{\pm\text{0.00}}$ & 0.48$_{\pm\text{0.10}}$ & \textbf{0.69}$_{\pm\text{0.09}}$ \\

\cmidrule{2-12}

& \multirow[c]{4}{*}{stock} & 20 & 0.56$_{\pm\text{0.13}}$ & 0.50$_{\pm\text{0.12}}$ & 0.40$_{\pm\text{0.12}}$ & 0.63$_{\pm\text{0.08}}$ & 0.11$_{\pm\text{0.03}}$ & 0.55$_{\pm\text{0.12}}$ & 0.43$_{\pm\text{0.16}}$ & 0.56$_{\pm\text{0.20}}$ & \textbf{1.00}$_{\pm\text{0.00}}$ \\

& & 50 & 0.99$_{\pm\text{0.03}}$ & 0.85$_{\pm\text{0.10}}$ & 0.88$_{\pm\text{0.13}}$ & 0.99$_{\pm\text{0.03}}$ & 0.12$_{\pm\text{0.06}}$ & \textbf{1.00}$_{\pm\text{0.00}}$ & 0.31$_{\pm\text{0.14}}$ & 0.87$_{\pm\text{0.08}}$ & 0.99$_{\pm\text{0.03}}$ \\

& & 100 & 1.00$_{\pm\text{0.00}}$ & 0.64$_{\pm\text{0.14}}$ & 0.89$_{\pm\text{0.12}}$ & 0.99$_{\pm\text{0.03}}$ & 0.16$_{\pm\text{0.13}}$ & \textbf{1.00}$_{\pm\text{0.00}}$ & 0.13$_{\pm\text{0.05}}$ & 0.91$_{\pm\text{0.10}}$ & 0.99$_{\pm\text{0.03}}$ \\

& & 200 & 0.99$_{\pm\text{0.03}}$ & 0.58$_{\pm\text{0.09}}$ & 0.76$_{\pm\text{0.20}}$ & 0.98$_{\pm\text{0.04}}$ & 0.71$_{\pm\text{0.22}}$ & \textbf{1.00}$_{\pm\text{0.00}}$ & 0.10$_{\pm\text{0.00}}$ & 0.99$_{\pm\text{0.03}}$ & 0.99$_{\pm\text{0.03}}$ \\

\midrule

\multirow{9}{*}{\rotatebox{90}{\begin{tabular}{l} \textit{More than 10 classes} \end{tabular}}}& \multirow[c]{3}{*}{energy} & 50 & N/A & 0.14$_{\pm\text{0.08}}$ & 0.26$_{\pm\text{0.09}}$ & 0.35$_{\pm\text{0.09}}$ & 0.17$_{\pm\text{0.06}}$ & 0.41$_{\pm\text{0.14}}$ & 0.19$_{\pm\text{0.10}}$ & N/A & \textbf{0.80}$_{\pm\text{0.05}}$ \\

& & 100 & N/A & 0.09$_{\pm\text{0.10}}$ & 0.32$_{\pm\text{0.12}}$ & 0.35$_{\pm\text{0.08}}$ & 0.13$_{\pm\text{0.06}}$ & 0.30$_{\pm\text{0.09}}$ & 0.06$_{\pm\text{0.07}}$ & N/A & \textbf{0.92}$_{\pm\text{0.01}}$ \\

& & 200 & N/A & 0.15$_{\pm\text{0.12}}$ & 0.30$_{\pm\text{0.09}}$ & 0.39$_{\pm\text{0.12}}$ & 0.15$_{\pm\text{0.07}}$ & 0.28$_{\pm\text{0.08}}$ & 0.03$_{\pm\text{0.05}}$ & N/A & \textbf{0.96}$_{\pm\text{0.01}}$ \\

\cmidrule{2-12}

& \multirow[c]{2}{*}{collins} & 100 & N/A & 0.61$_{\pm\text{0.08}}$ & 0.73$_{\pm\text{0.12}}$ & 0.87$_{\pm\text{0.08}}$ & 0.10$_{\pm\text{0.07}}$ & \textbf{0.90}$_{\pm\text{0.08}}$ & 0.20$_{\pm\text{0.09}}$ & N/A & 0.89$_{\pm\text{0.04}}$ \\

& & 200 & 0.75$_{\pm\text{0.09}}$ & 0.35$_{\pm\text{0.10}}$ & 0.62$_{\pm\text{0.13}}$ & 0.76$_{\pm\text{0.10}}$ & 0.07$_{\pm\text{0.03}}$ & 0.78$_{\pm\text{0.06}}$ & 0.05$_{\pm\text{0.04}}$ & N/A & \textbf{0.80}$_{\pm\text{0.09}}$ \\

\cmidrule{2-12}

& \multirow[c]{4}{*}{texture} & 50 & 0.90$_{\pm\text{0.12}}$ & 0.79$_{\pm\text{0.11}}$ & 0.63$_{\pm\text{0.14}}$ & 0.94$_{\pm\text{0.09}}$ & 0.02$_{\pm\text{0.00}}$ & \textbf{0.99}$_{\pm\text{0.02}}$ & 0.47$_{\pm\text{0.13}}$ & N/A & 0.93$_{\pm\text{0.12}}$ \\

& & 100 & 0.97$_{\pm\text{0.04}}$ & 0.51$_{\pm\text{0.13}}$ & 0.67$_{\pm\text{0.12}}$ & 0.92$_{\pm\text{0.07}}$ & 0.02$_{\pm\text{0.00}}$ & 0.94$_{\pm\text{0.09}}$ & 0.17$_{\pm\text{0.08}}$ & N/A & \textbf{0.97}$_{\pm\text{0.06}}$ \\

& & 200 & 0.86$_{\pm\text{0.11}}$ & 0.28$_{\pm\text{0.11}}$ & 0.55$_{\pm\text{0.16}}$ & 0.89$_{\pm\text{0.10}}$ & 0.02$_{\pm\text{0.00}}$ & \textbf{0.93}$_{\pm\text{0.07}}$ & 0.00$_{\pm\text{0.01}}$ & N/A & 0.91$_{\pm\text{0.08}}$ \\

& & 500 & 0.74$_{\pm\text{0.17}}$ & 0.21$_{\pm\text{0.09}}$ & 0.51$_{\pm\text{0.08}}$ & 0.73$_{\pm\text{0.07}}$ & 0.02$_{\pm\text{0.00}}$ & 0.73$_{\pm\text{0.07}}$ & 0.00$_{\pm\text{0.00}}$ & N/A & \textbf{0.82}$_{\pm\text{0.12}}$ \\

\midrule
\rowcolor{Gainsboro!60}
\multicolumn{3}{l|}{\textbf{Average rank}} & 3.52$_{\pm\text{1.08}}$ & 6.88$_{\pm\text{1.02}}$ & 6.03$_{\pm\text{1.10}}$ & 3.55$_{\pm\text{1.03}}$ & 8.30$_{\pm\text{1.02}}$ & 2.82$_{\pm\text{1.76}}$ & 8.18$_{\pm\text{0.88}}$ & 4.27$_{\pm\text{1.62}}$ & \textbf{1.45}$_{\pm\text{0.71}}$ \\

\bottomrule

\end{tabular}
}
\end{table}
\clearpage

\FloatBarrier
\subsubsection{Similarity between Real Test Data and Synthetic Data}
\begin{table}[hp]
\centering
\caption{\textbf{Inverse KL between real test data and synthetic data} on eight real-world tabular datasets with varied real data availability. We report the mean $\pm$ std result and average rank across datasets. A higher rank implies higher fidelity. Note that ``N/A'' denotes that a specific generator was not applicable, and the rank is computed with the mean result of other methods. We \textbf{bold} the highest result for each dataset of different sample sizes. TabEBM achieves the best overall performance against benchmark generators.}
\resizebox{\textwidth}{!}{

\setlength{\tabcolsep}{5pt}
\begin{tabular}{m{0.2cm}lr|rrrrrrrr|r}

\toprule

\multicolumn{2}{l}{Datasets}  & $N_{\text{real}}$ & SMOTE & TVAE & CTGAN & NFLOW & TabDDPM & ARF & GOGGLE & TabPFGen & \textbf{TabEBM} \\

\midrule

\multirow{24}{*}{\rotatebox{90}{\begin{tabular}{l} \textit{At most 10 classes} \end{tabular}}}& \multirow[c]{5}{*}{protein} & 20 & N/A & 0.26$_{\pm\text{0.04}}$ & 0.26$_{\pm\text{0.03}}$ & 0.32$_{\pm\text{0.03}}$ & 0.09$_{\pm\text{0.01}}$ & 0.34$_{\pm\text{0.03}}$ & 0.08$_{\pm\text{0.00}}$ & 0.35$_{\pm\text{0.07}}$ & \textbf{0.52}$_{\pm\text{0.06}}$ \\

& & 50 & 0.78$_{\pm\text{0.02}}$ & \textbf{0.82}$_{\pm\text{0.02}}$ & 0.63$_{\pm\text{0.06}}$ & 0.71$_{\pm\text{0.04}}$ & 0.08$_{\pm\text{0.00}}$ & 0.80$_{\pm\text{0.03}}$ & 0.52$_{\pm\text{0.03}}$ & 0.62$_{\pm\text{0.04}}$ & 0.75$_{\pm\text{0.03}}$ \\

& & 100 & \textbf{0.88}$_{\pm\text{0.02}}$ & 0.80$_{\pm\text{0.02}}$ & 0.76$_{\pm\text{0.04}}$ & 0.83$_{\pm\text{0.03}}$ & 0.08$_{\pm\text{0.00}}$ & 0.87$_{\pm\text{0.02}}$ & 0.33$_{\pm\text{0.05}}$ & 0.78$_{\pm\text{0.03}}$ & 0.85$_{\pm\text{0.02}}$ \\

& & 200 & \textbf{0.92}$_{\pm\text{0.01}}$ & 0.75$_{\pm\text{0.03}}$ & 0.81$_{\pm\text{0.03}}$ & 0.91$_{\pm\text{0.01}}$ & 0.08$_{\pm\text{0.00}}$ & 0.91$_{\pm\text{0.01}}$ & 0.12$_{\pm\text{0.01}}$ & 0.89$_{\pm\text{0.02}}$ & 0.92$_{\pm\text{0.01}}$ \\

& & 500 & 0.94$_{\pm\text{0.00}}$ & 0.68$_{\pm\text{0.02}}$ & 0.86$_{\pm\text{0.02}}$ & 0.93$_{\pm\text{0.00}}$ & 0.08$_{\pm\text{0.00}}$ & 0.94$_{\pm\text{0.00}}$ & 0.13$_{\pm\text{0.01}}$ & 0.94$_{\pm\text{0.00}}$ & \textbf{0.95}$_{\pm\text{0.01}}$ \\

\cmidrule{2-12}

& \multirow[c]{5}{*}{fourier} & 20 & N/A & 0.14$_{\pm\text{0.02}}$ & 0.18$_{\pm\text{0.01}}$ & 0.20$_{\pm\text{0.01}}$ & 0.21$_{\pm\text{0.02}}$ & 0.21$_{\pm\text{0.01}}$ & 0.22$_{\pm\text{0.02}}$ & 0.24$_{\pm\text{0.01}}$ & \textbf{0.48}$_{\pm\text{0.03}}$ \\

& & 50 & 0.84$_{\pm\text{0.02}}$ & 0.78$_{\pm\text{0.03}}$ & 0.62$_{\pm\text{0.05}}$ & 0.77$_{\pm\text{0.03}}$ & 0.07$_{\pm\text{0.00}}$ & \textbf{0.85}$_{\pm\text{0.03}}$ & 0.48$_{\pm\text{0.06}}$ & 0.66$_{\pm\text{0.03}}$ & 0.79$_{\pm\text{0.03}}$ \\

& & 100 & \textbf{0.91}$_{\pm\text{0.02}}$ & 0.72$_{\pm\text{0.05}}$ & 0.76$_{\pm\text{0.03}}$ & 0.88$_{\pm\text{0.02}}$ & 0.08$_{\pm\text{0.00}}$ & 0.90$_{\pm\text{0.01}}$ & 0.20$_{\pm\text{0.05}}$ & 0.81$_{\pm\text{0.02}}$ & 0.88$_{\pm\text{0.02}}$ \\

& & 200 & \textbf{0.94}$_{\pm\text{0.01}}$ & 0.58$_{\pm\text{0.02}}$ & 0.79$_{\pm\text{0.03}}$ & 0.93$_{\pm\text{0.01}}$ & 0.09$_{\pm\text{0.00}}$ & 0.93$_{\pm\text{0.01}}$ & 0.08$_{\pm\text{0.01}}$ & 0.90$_{\pm\text{0.02}}$ & 0.93$_{\pm\text{0.01}}$ \\

& & 500 & 0.96$_{\pm\text{0.00}}$ & 0.50$_{\pm\text{0.02}}$ & 0.88$_{\pm\text{0.02}}$ & 0.93$_{\pm\text{0.01}}$ & 0.10$_{\pm\text{0.01}}$ & 0.95$_{\pm\text{0.01}}$ & 0.09$_{\pm\text{0.00}}$ & 0.95$_{\pm\text{0.01}}$ & \textbf{0.96}$_{\pm\text{0.00}}$ \\

\cmidrule{2-12}

& \multirow[c]{5}{*}{biodeg} & 20 & 0.43$_{\pm\text{0.03}}$ & 0.44$_{\pm\text{0.04}}$ & 0.41$_{\pm\text{0.03}}$ & 0.41$_{\pm\text{0.04}}$ & 0.33$_{\pm\text{0.03}}$ & 0.44$_{\pm\text{0.03}}$ & 0.36$_{\pm\text{0.03}}$ & 0.45$_{\pm\text{0.02}}$ & \textbf{0.57}$_{\pm\text{0.03}}$ \\

& & 50 & 0.60$_{\pm\text{0.05}}$ & 0.59$_{\pm\text{0.04}}$ & 0.53$_{\pm\text{0.04}}$ & 0.55$_{\pm\text{0.04}}$ & 0.31$_{\pm\text{0.03}}$ & 0.57$_{\pm\text{0.03}}$ & 0.41$_{\pm\text{0.03}}$ & 0.60$_{\pm\text{0.05}}$ & \textbf{0.71}$_{\pm\text{0.04}}$ \\

& & 100 & 0.65$_{\pm\text{0.04}}$ & 0.65$_{\pm\text{0.02}}$ & 0.62$_{\pm\text{0.03}}$ & 0.63$_{\pm\text{0.03}}$ & 0.31$_{\pm\text{0.03}}$ & 0.64$_{\pm\text{0.03}}$ & 0.37$_{\pm\text{0.03}}$ & 0.65$_{\pm\text{0.03}}$ & \textbf{0.77}$_{\pm\text{0.02}}$ \\

& & 200 & 0.71$_{\pm\text{0.02}}$ & 0.66$_{\pm\text{0.03}}$ & 0.66$_{\pm\text{0.04}}$ & 0.65$_{\pm\text{0.03}}$ & 0.31$_{\pm\text{0.04}}$ & 0.68$_{\pm\text{0.02}}$ & 0.33$_{\pm\text{0.02}}$ & 0.73$_{\pm\text{0.04}}$ & \textbf{0.83}$_{\pm\text{0.02}}$ \\

& & 500 & 0.77$_{\pm\text{0.03}}$ & 0.64$_{\pm\text{0.02}}$ & 0.69$_{\pm\text{0.03}}$ & 0.72$_{\pm\text{0.04}}$ & 0.25$_{\pm\text{0.03}}$ & 0.69$_{\pm\text{0.03}}$ & 0.35$_{\pm\text{0.01}}$ & 0.76$_{\pm\text{0.05}}$ & \textbf{0.88}$_{\pm\text{0.02}}$ \\

\cmidrule{2-12}

& \multirow[c]{5}{*}{steel} & 20 & 0.47$_{\pm\text{0.03}}$ & 0.45$_{\pm\text{0.03}}$ & 0.42$_{\pm\text{0.02}}$ & 0.44$_{\pm\text{0.02}}$ & 0.23$_{\pm\text{0.01}}$ & 0.46$_{\pm\text{0.04}}$ & 0.37$_{\pm\text{0.04}}$ & 0.42$_{\pm\text{0.04}}$ & \textbf{0.70}$_{\pm\text{0.03}}$ \\

& & 50 & 0.65$_{\pm\text{0.03}}$ & 0.59$_{\pm\text{0.03}}$ & 0.58$_{\pm\text{0.03}}$ & 0.60$_{\pm\text{0.04}}$ & 0.21$_{\pm\text{0.01}}$ & 0.63$_{\pm\text{0.03}}$ & 0.30$_{\pm\text{0.03}}$ & 0.62$_{\pm\text{0.05}}$ & \textbf{0.83}$_{\pm\text{0.03}}$ \\

& & 100 & 0.70$_{\pm\text{0.02}}$ & 0.59$_{\pm\text{0.02}}$ & 0.62$_{\pm\text{0.03}}$ & 0.66$_{\pm\text{0.03}}$ & 0.21$_{\pm\text{0.01}}$ & 0.66$_{\pm\text{0.02}}$ & 0.29$_{\pm\text{0.02}}$ & 0.72$_{\pm\text{0.04}}$ & \textbf{0.89}$_{\pm\text{0.02}}$ \\

& & 200 & 0.73$_{\pm\text{0.01}}$ & 0.55$_{\pm\text{0.02}}$ & 0.66$_{\pm\text{0.02}}$ & 0.69$_{\pm\text{0.02}}$ & 0.22$_{\pm\text{0.01}}$ & 0.68$_{\pm\text{0.02}}$ & 0.25$_{\pm\text{0.01}}$ & 0.75$_{\pm\text{0.03}}$ & \textbf{0.91}$_{\pm\text{0.02}}$ \\

& & 500 & 0.74$_{\pm\text{0.01}}$ & 0.55$_{\pm\text{0.04}}$ & 0.66$_{\pm\text{0.02}}$ & 0.69$_{\pm\text{0.01}}$ & 0.22$_{\pm\text{0.01}}$ & 0.70$_{\pm\text{0.01}}$ & 0.27$_{\pm\text{0.01}}$ & 0.78$_{\pm\text{0.05}}$ & \textbf{0.93}$_{\pm\text{0.02}}$ \\

\cmidrule{2-12}

& \multirow[c]{4}{*}{stock} & 20 & 0.50$_{\pm\text{0.09}}$ & 0.50$_{\pm\text{0.09}}$ & 0.41$_{\pm\text{0.05}}$ & 0.47$_{\pm\text{0.08}}$ & 0.25$_{\pm\text{0.04}}$ & 0.54$_{\pm\text{0.09}}$ & 0.40$_{\pm\text{0.13}}$ & 0.35$_{\pm\text{0.05}}$ & \textbf{0.65}$_{\pm\text{0.09}}$ \\

& & 50 & 0.80$_{\pm\text{0.06}}$ & 0.68$_{\pm\text{0.06}}$ & 0.68$_{\pm\text{0.05}}$ & 0.76$_{\pm\text{0.03}}$ & 0.34$_{\pm\text{0.05}}$ & \textbf{0.86}$_{\pm\text{0.04}}$ & 0.33$_{\pm\text{0.11}}$ & 0.69$_{\pm\text{0.09}}$ & 0.85$_{\pm\text{0.05}}$ \\

& & 100 & 0.86$_{\pm\text{0.04}}$ & 0.61$_{\pm\text{0.05}}$ & 0.73$_{\pm\text{0.06}}$ & 0.85$_{\pm\text{0.06}}$ & 0.44$_{\pm\text{0.07}}$ & 0.90$_{\pm\text{0.02}}$ & 0.18$_{\pm\text{0.04}}$ & 0.84$_{\pm\text{0.05}}$ & \textbf{0.91}$_{\pm\text{0.04}}$ \\

& & 200 & 0.92$_{\pm\text{0.02}}$ & 0.59$_{\pm\text{0.05}}$ & 0.75$_{\pm\text{0.07}}$ & 0.90$_{\pm\text{0.03}}$ & 0.67$_{\pm\text{0.09}}$ & 0.94$_{\pm\text{0.01}}$ & 0.15$_{\pm\text{0.00}}$ & 0.94$_{\pm\text{0.03}}$ & \textbf{0.96}$_{\pm\text{0.02}}$ \\

\midrule

\multirow{9}{*}{\rotatebox{90}{\begin{tabular}{l} \textit{More than 10 classes} \end{tabular}}}& \multirow[c]{3}{*}{energy} & 50 & N/A & 0.26$_{\pm\text{0.06}}$ & 0.33$_{\pm\text{0.06}}$ & 0.36$_{\pm\text{0.08}}$ & 0.23$_{\pm\text{0.05}}$ & 0.46$_{\pm\text{0.10}}$ & 0.22$_{\pm\text{0.07}}$ & N/A & \textbf{0.77}$_{\pm\text{0.03}}$ \\

& & 100 & N/A & 0.27$_{\pm\text{0.06}}$ & 0.40$_{\pm\text{0.08}}$ & 0.43$_{\pm\text{0.06}}$ & 0.22$_{\pm\text{0.06}}$ & 0.39$_{\pm\text{0.08}}$ & 0.16$_{\pm\text{0.05}}$ & N/A & \textbf{0.87}$_{\pm\text{0.01}}$ \\

& & 200 & N/A & 0.30$_{\pm\text{0.07}}$ & 0.41$_{\pm\text{0.07}}$ & 0.46$_{\pm\text{0.09}}$ & 0.24$_{\pm\text{0.06}}$ & 0.39$_{\pm\text{0.08}}$ & 0.12$_{\pm\text{0.04}}$ & N/A & \textbf{0.89}$_{\pm\text{0.01}}$ \\

\cmidrule{2-12}

& \multirow[c]{2}{*}{collins} & 100 & N/A & 0.68$_{\pm\text{0.03}}$ & 0.75$_{\pm\text{0.04}}$ & 0.79$_{\pm\text{0.03}}$ & 0.43$_{\pm\text{0.07}}$ & \textbf{0.81}$_{\pm\text{0.02}}$ & 0.28$_{\pm\text{0.07}}$ & N/A & 0.78$_{\pm\text{0.02}}$ \\

& & 200 & 0.87$_{\pm\text{0.02}}$ & 0.62$_{\pm\text{0.03}}$ & 0.81$_{\pm\text{0.03}}$ & 0.88$_{\pm\text{0.02}}$ & 0.44$_{\pm\text{0.04}}$ & \textbf{0.88}$_{\pm\text{0.02}}$ & 0.22$_{\pm\text{0.09}}$ & N/A & 0.87$_{\pm\text{0.01}}$ \\

\cmidrule{2-12}

& \multirow[c]{4}{*}{texture} & 50 & 0.82$_{\pm\text{0.04}}$ & 0.80$_{\pm\text{0.04}}$ & 0.70$_{\pm\text{0.05}}$ & 0.80$_{\pm\text{0.07}}$ & 0.11$_{\pm\text{0.01}}$ & \textbf{0.92}$_{\pm\text{0.01}}$ & 0.48$_{\pm\text{0.11}}$ & N/A & 0.87$_{\pm\text{0.03}}$ \\

& & 100 & 0.89$_{\pm\text{0.02}}$ & 0.69$_{\pm\text{0.03}}$ & 0.79$_{\pm\text{0.06}}$ & 0.89$_{\pm\text{0.02}}$ & 0.13$_{\pm\text{0.01}}$ & \textbf{0.93}$_{\pm\text{0.00}}$ & 0.23$_{\pm\text{0.06}}$ & N/A & 0.92$_{\pm\text{0.02}}$ \\

& & 200 & 0.93$_{\pm\text{0.02}}$ & 0.58$_{\pm\text{0.05}}$ & 0.79$_{\pm\text{0.05}}$ & 0.92$_{\pm\text{0.01}}$ & 0.14$_{\pm\text{0.01}}$ & 0.95$_{\pm\text{0.01}}$ & 0.10$_{\pm\text{0.02}}$ & N/A & \textbf{0.95}$_{\pm\text{0.01}}$ \\

& & 500 & 0.96$_{\pm\text{0.01}}$ & 0.64$_{\pm\text{0.06}}$ & 0.85$_{\pm\text{0.04}}$ & 0.93$_{\pm\text{0.01}}$ & 0.15$_{\pm\text{0.01}}$ & 0.96$_{\pm\text{0.00}}$ & 0.12$_{\pm\text{0.01}}$ & N/A & \textbf{0.97}$_{\pm\text{0.01}}$ \\

\midrule
\rowcolor{Gainsboro!60}
\multicolumn{3}{l|}{\textbf{Average rank}} & 2.94$_{\pm\text{1.31}}$ & 6.03$_{\pm\text{1.69}}$ & 5.82$_{\pm\text{1.07}}$ & 4.39$_{\pm\text{1.30}}$ & 8.48$_{\pm\text{0.71}}$ & 3.00$_{\pm\text{1.52}}$ & 8.24$_{\pm\text{0.94}}$ & 4.45$_{\pm\text{1.91}}$ & \textbf{1.64}$_{\pm\text{1.03}}$ \\

\bottomrule

\end{tabular}
}
\end{table}
\clearpage

\begin{table}[p]
\centering
\caption{\textbf{KS test between real test data and synthetic data} on eight real-world tabular datasets with varied real data availability. We report the mean $\pm$ std result and average rank across datasets. A higher rank implies higher fidelity. Note that ``N/A'' denotes that a specific generator was not applicable, and the rank is computed with the mean result of other methods. We \textbf{bold} the highest result for each dataset of different sample sizes. TabEBM achieves the best overall performance against benchmark generators.}
\resizebox{\textwidth}{!}{

\setlength{\tabcolsep}{5pt}
\begin{tabular}{m{0.2cm}lr|rrrrrrrr|r}

\toprule

\multicolumn{2}{l}{Datasets}  & $N_{\text{real}}$ & SMOTE & TVAE & CTGAN & NFLOW & TabDDPM & ARF & GOGGLE & TabPFGen & \textbf{TabEBM} \\

\midrule

\multirow{24}{*}{\rotatebox{90}{\begin{tabular}{l} \textit{At most 10 classes} \end{tabular}}}& \multirow[c]{5}{*}{protein} & 20 & N/A & 0.69$_{\pm\text{0.01}}$ & 0.72$_{\pm\text{0.02}}$ & 0.75$_{\pm\text{0.01}}$ & 0.61$_{\pm\text{0.01}}$ & 0.74$_{\pm\text{0.01}}$ & 0.41$_{\pm\text{0.02}}$ & 0.77$_{\pm\text{0.02}}$ & \textbf{0.81}$_{\pm\text{0.01}}$ \\

& & 50 & 0.88$_{\pm\text{0.01}}$ & 0.87$_{\pm\text{0.01}}$ & 0.87$_{\pm\text{0.01}}$ & 0.88$_{\pm\text{0.01}}$ & 0.61$_{\pm\text{0.01}}$ & \textbf{0.89}$_{\pm\text{0.01}}$ & 0.73$_{\pm\text{0.02}}$ & 0.86$_{\pm\text{0.01}}$ & 0.88$_{\pm\text{0.01}}$ \\

& & 100 & 0.90$_{\pm\text{0.01}}$ & 0.87$_{\pm\text{0.01}}$ & 0.89$_{\pm\text{0.01}}$ & 0.90$_{\pm\text{0.01}}$ & 0.60$_{\pm\text{0.01}}$ & 0.91$_{\pm\text{0.01}}$ & 0.65$_{\pm\text{0.02}}$ & 0.89$_{\pm\text{0.01}}$ & \textbf{0.91}$_{\pm\text{0.01}}$ \\

& & 200 & 0.92$_{\pm\text{0.01}}$ & 0.86$_{\pm\text{0.01}}$ & 0.90$_{\pm\text{0.01}}$ & 0.91$_{\pm\text{0.01}}$ & 0.59$_{\pm\text{0.01}}$ & 0.92$_{\pm\text{0.01}}$ & 0.58$_{\pm\text{0.01}}$ & 0.91$_{\pm\text{0.01}}$ & \textbf{0.92}$_{\pm\text{0.01}}$ \\

& & 500 & 0.92$_{\pm\text{0.00}}$ & 0.82$_{\pm\text{0.01}}$ & 0.90$_{\pm\text{0.00}}$ & 0.91$_{\pm\text{0.00}}$ & 0.58$_{\pm\text{0.01}}$ & 0.92$_{\pm\text{0.00}}$ & 0.63$_{\pm\text{0.00}}$ & 0.92$_{\pm\text{0.00}}$ & \textbf{0.93}$_{\pm\text{0.00}}$ \\

\cmidrule{2-12}

& \multirow[c]{5}{*}{fourier} & 20 & N/A & 0.67$_{\pm\text{0.03}}$ & 0.69$_{\pm\text{0.02}}$ & 0.73$_{\pm\text{0.01}}$ & 0.75$_{\pm\text{0.01}}$ & 0.75$_{\pm\text{0.01}}$ & 0.75$_{\pm\text{0.01}}$ & 0.76$_{\pm\text{0.01}}$ & \textbf{0.81}$_{\pm\text{0.01}}$ \\

& & 50 & 0.89$_{\pm\text{0.00}}$ & 0.86$_{\pm\text{0.01}}$ & 0.88$_{\pm\text{0.01}}$ & 0.89$_{\pm\text{0.01}}$ & 0.64$_{\pm\text{0.00}}$ & \textbf{0.91}$_{\pm\text{0.00}}$ & 0.73$_{\pm\text{0.02}}$ & 0.87$_{\pm\text{0.01}}$ & 0.89$_{\pm\text{0.01}}$ \\

& & 100 & 0.92$_{\pm\text{0.00}}$ & 0.85$_{\pm\text{0.01}}$ & 0.90$_{\pm\text{0.00}}$ & 0.91$_{\pm\text{0.00}}$ & 0.62$_{\pm\text{0.01}}$ & \textbf{0.92}$_{\pm\text{0.00}}$ & 0.61$_{\pm\text{0.02}}$ & 0.90$_{\pm\text{0.00}}$ & 0.91$_{\pm\text{0.00}}$ \\

& & 200 & 0.93$_{\pm\text{0.00}}$ & 0.82$_{\pm\text{0.01}}$ & 0.91$_{\pm\text{0.01}}$ & 0.93$_{\pm\text{0.00}}$ & 0.62$_{\pm\text{0.01}}$ & \textbf{0.93}$_{\pm\text{0.00}}$ & 0.60$_{\pm\text{0.03}}$ & 0.93$_{\pm\text{0.00}}$ & 0.93$_{\pm\text{0.00}}$ \\

& & 500 & 0.94$_{\pm\text{0.00}}$ & 0.80$_{\pm\text{0.00}}$ & 0.92$_{\pm\text{0.01}}$ & 0.93$_{\pm\text{0.00}}$ & 0.62$_{\pm\text{0.01}}$ & 0.94$_{\pm\text{0.00}}$ & 0.63$_{\pm\text{0.01}}$ & 0.94$_{\pm\text{0.00}}$ & \textbf{0.94}$_{\pm\text{0.00}}$ \\

\cmidrule{2-12}

& \multirow[c]{5}{*}{biodeg} & 20 & 0.61$_{\pm\text{0.03}}$ & 0.61$_{\pm\text{0.03}}$ & 0.63$_{\pm\text{0.03}}$ & 0.63$_{\pm\text{0.03}}$ & 0.55$_{\pm\text{0.04}}$ & 0.62$_{\pm\text{0.03}}$ & 0.64$_{\pm\text{0.03}}$ & 0.56$_{\pm\text{0.03}}$ & \textbf{0.67}$_{\pm\text{0.02}}$ \\

& & 50 & 0.56$_{\pm\text{0.03}}$ & 0.57$_{\pm\text{0.03}}$ & 0.58$_{\pm\text{0.03}}$ & 0.59$_{\pm\text{0.03}}$ & 0.48$_{\pm\text{0.03}}$ & 0.58$_{\pm\text{0.03}}$ & 0.63$_{\pm\text{0.02}}$ & 0.59$_{\pm\text{0.04}}$ & \textbf{0.71}$_{\pm\text{0.01}}$ \\

& & 100 & 0.55$_{\pm\text{0.02}}$ & 0.55$_{\pm\text{0.02}}$ & 0.57$_{\pm\text{0.03}}$ & 0.58$_{\pm\text{0.02}}$ & 0.46$_{\pm\text{0.03}}$ & 0.58$_{\pm\text{0.02}}$ & 0.59$_{\pm\text{0.03}}$ & 0.58$_{\pm\text{0.02}}$ & \textbf{0.71}$_{\pm\text{0.01}}$ \\

& & 200 & 0.53$_{\pm\text{0.01}}$ & 0.51$_{\pm\text{0.02}}$ & 0.53$_{\pm\text{0.02}}$ & 0.55$_{\pm\text{0.02}}$ & 0.43$_{\pm\text{0.02}}$ & 0.55$_{\pm\text{0.02}}$ & 0.55$_{\pm\text{0.01}}$ & 0.57$_{\pm\text{0.02}}$ & \textbf{0.72}$_{\pm\text{0.01}}$ \\

& & 500 & 0.53$_{\pm\text{0.00}}$ & 0.49$_{\pm\text{0.01}}$ & 0.51$_{\pm\text{0.01}}$ & 0.54$_{\pm\text{0.01}}$ & 0.43$_{\pm\text{0.01}}$ & 0.55$_{\pm\text{0.01}}$ & 0.56$_{\pm\text{0.01}}$ & 0.57$_{\pm\text{0.01}}$ & \textbf{0.72}$_{\pm\text{0.00}}$ \\

\cmidrule{2-12}

& \multirow[c]{5}{*}{steel} & 20 & 0.65$_{\pm\text{0.03}}$ & 0.62$_{\pm\text{0.03}}$ & 0.63$_{\pm\text{0.03}}$ & 0.65$_{\pm\text{0.02}}$ & 0.54$_{\pm\text{0.01}}$ & 0.64$_{\pm\text{0.02}}$ & 0.56$_{\pm\text{0.03}}$ & 0.64$_{\pm\text{0.02}}$ & \textbf{0.73}$_{\pm\text{0.02}}$ \\

& & 50 & 0.68$_{\pm\text{0.02}}$ & 0.64$_{\pm\text{0.03}}$ & 0.65$_{\pm\text{0.03}}$ & 0.67$_{\pm\text{0.02}}$ & 0.51$_{\pm\text{0.01}}$ & 0.68$_{\pm\text{0.03}}$ & 0.56$_{\pm\text{0.03}}$ & 0.70$_{\pm\text{0.02}}$ & \textbf{0.77}$_{\pm\text{0.01}}$ \\

& & 100 & 0.68$_{\pm\text{0.01}}$ & 0.63$_{\pm\text{0.02}}$ & 0.65$_{\pm\text{0.01}}$ & 0.66$_{\pm\text{0.01}}$ & 0.50$_{\pm\text{0.01}}$ & 0.66$_{\pm\text{0.02}}$ & 0.55$_{\pm\text{0.02}}$ & 0.70$_{\pm\text{0.03}}$ & \textbf{0.78}$_{\pm\text{0.01}}$ \\

& & 200 & 0.69$_{\pm\text{0.01}}$ & 0.61$_{\pm\text{0.02}}$ & 0.66$_{\pm\text{0.02}}$ & 0.68$_{\pm\text{0.02}}$ & 0.50$_{\pm\text{0.01}}$ & 0.65$_{\pm\text{0.02}}$ & 0.54$_{\pm\text{0.01}}$ & 0.71$_{\pm\text{0.02}}$ & \textbf{0.78}$_{\pm\text{0.01}}$ \\

& & 500 & 0.69$_{\pm\text{0.01}}$ & 0.62$_{\pm\text{0.02}}$ & 0.66$_{\pm\text{0.02}}$ & 0.68$_{\pm\text{0.01}}$ & 0.50$_{\pm\text{0.01}}$ & 0.64$_{\pm\text{0.01}}$ & 0.58$_{\pm\text{0.01}}$ & 0.73$_{\pm\text{0.02}}$ & \textbf{0.79}$_{\pm\text{0.01}}$ \\

\cmidrule{2-12}

& \multirow[c]{4}{*}{stock} & 20 & 0.83$_{\pm\text{0.04}}$ & 0.81$_{\pm\text{0.04}}$ & 0.82$_{\pm\text{0.04}}$ & 0.84$_{\pm\text{0.04}}$ & 0.74$_{\pm\text{0.04}}$ & 0.84$_{\pm\text{0.04}}$ & 0.64$_{\pm\text{0.06}}$ & 0.83$_{\pm\text{0.04}}$ & \textbf{0.86}$_{\pm\text{0.04}}$ \\

& & 50 & 0.89$_{\pm\text{0.02}}$ & 0.84$_{\pm\text{0.02}}$ & 0.87$_{\pm\text{0.03}}$ & 0.88$_{\pm\text{0.02}}$ & 0.83$_{\pm\text{0.02}}$ & 0.89$_{\pm\text{0.02}}$ & 0.67$_{\pm\text{0.05}}$ & 0.88$_{\pm\text{0.02}}$ & \textbf{0.89}$_{\pm\text{0.02}}$ \\

& & 100 & 0.91$_{\pm\text{0.02}}$ & 0.84$_{\pm\text{0.02}}$ & 0.88$_{\pm\text{0.02}}$ & 0.90$_{\pm\text{0.02}}$ & 0.87$_{\pm\text{0.03}}$ & 0.90$_{\pm\text{0.02}}$ & 0.62$_{\pm\text{0.01}}$ & 0.91$_{\pm\text{0.02}}$ & \textbf{0.91}$_{\pm\text{0.02}}$ \\

& & 200 & 0.92$_{\pm\text{0.02}}$ & 0.84$_{\pm\text{0.02}}$ & 0.88$_{\pm\text{0.02}}$ & 0.90$_{\pm\text{0.02}}$ & 0.90$_{\pm\text{0.02}}$ & 0.91$_{\pm\text{0.02}}$ & 0.63$_{\pm\text{0.01}}$ & 0.92$_{\pm\text{0.02}}$ & \textbf{0.92}$_{\pm\text{0.02}}$ \\

\midrule

\multirow{9}{*}{\rotatebox{90}{\begin{tabular}{l} \textit{More than 10 classes} \end{tabular}}}& \multirow[c]{3}{*}{energy} & 50 & N/A & 0.69$_{\pm\text{0.02}}$ & 0.70$_{\pm\text{0.03}}$ & 0.72$_{\pm\text{0.02}}$ & 0.64$_{\pm\text{0.02}}$ & 0.72$_{\pm\text{0.01}}$ & 0.62$_{\pm\text{0.03}}$ & N/A & \textbf{0.76}$_{\pm\text{0.01}}$ \\

& & 100 & N/A & 0.69$_{\pm\text{0.03}}$ & 0.74$_{\pm\text{0.01}}$ & 0.75$_{\pm\text{0.01}}$ & 0.69$_{\pm\text{0.01}}$ & 0.75$_{\pm\text{0.01}}$ & 0.64$_{\pm\text{0.03}}$ & N/A & \textbf{0.81}$_{\pm\text{0.01}}$ \\

& & 200 & N/A & 0.71$_{\pm\text{0.01}}$ & 0.74$_{\pm\text{0.01}}$ & 0.75$_{\pm\text{0.01}}$ & 0.67$_{\pm\text{0.02}}$ & 0.76$_{\pm\text{0.01}}$ & 0.63$_{\pm\text{0.02}}$ & N/A & \textbf{0.83}$_{\pm\text{0.01}}$ \\

\cmidrule{2-12}

& \multirow[c]{2}{*}{collins} & 100 & N/A & 0.83$_{\pm\text{0.01}}$ & 0.88$_{\pm\text{0.01}}$ & 0.90$_{\pm\text{0.01}}$ & 0.81$_{\pm\text{0.04}}$ & \textbf{0.90}$_{\pm\text{0.01}}$ & 0.65$_{\pm\text{0.03}}$ & N/A & 0.90$_{\pm\text{0.01}}$ \\

& & 200 & 0.91$_{\pm\text{0.01}}$ & 0.82$_{\pm\text{0.02}}$ & 0.90$_{\pm\text{0.01}}$ & 0.91$_{\pm\text{0.01}}$ & 0.80$_{\pm\text{0.02}}$ & 0.91$_{\pm\text{0.01}}$ & 0.63$_{\pm\text{0.04}}$ & N/A & \textbf{0.93}$_{\pm\text{0.01}}$ \\

\cmidrule{2-12}

& \multirow[c]{4}{*}{texture} & 50 & 0.90$_{\pm\text{0.01}}$ & 0.87$_{\pm\text{0.02}}$ & 0.88$_{\pm\text{0.01}}$ & 0.91$_{\pm\text{0.01}}$ & 0.55$_{\pm\text{0.04}}$ & 0.92$_{\pm\text{0.01}}$ & 0.73$_{\pm\text{0.05}}$ & N/A & \textbf{0.92}$_{\pm\text{0.01}}$ \\

& & 100 & 0.92$_{\pm\text{0.01}}$ & 0.86$_{\pm\text{0.01}}$ & 0.91$_{\pm\text{0.01}}$ & 0.92$_{\pm\text{0.01}}$ & 0.60$_{\pm\text{0.02}}$ & 0.93$_{\pm\text{0.01}}$ & 0.64$_{\pm\text{0.03}}$ & N/A & \textbf{0.94}$_{\pm\text{0.01}}$ \\

& & 200 & 0.94$_{\pm\text{0.01}}$ & 0.82$_{\pm\text{0.01}}$ & 0.90$_{\pm\text{0.01}}$ & 0.93$_{\pm\text{0.00}}$ & 0.62$_{\pm\text{0.01}}$ & 0.94$_{\pm\text{0.00}}$ & 0.60$_{\pm\text{0.01}}$ & N/A & \textbf{0.94}$_{\pm\text{0.00}}$ \\

& & 500 & 0.95$_{\pm\text{0.00}}$ & 0.84$_{\pm\text{0.02}}$ & 0.91$_{\pm\text{0.01}}$ & 0.93$_{\pm\text{0.01}}$ & 0.61$_{\pm\text{0.02}}$ & 0.95$_{\pm\text{0.00}}$ & 0.64$_{\pm\text{0.01}}$ & N/A & \textbf{0.95}$_{\pm\text{0.00}}$ \\

\midrule
\rowcolor{Gainsboro!60}
\multicolumn{3}{l|}{\textbf{Average rank}} & 3.97$_{\pm\text{1.73}}$ & 7.09$_{\pm\text{0.72}}$ & 5.52$_{\pm\text{0.97}}$ & 3.76$_{\pm\text{1.12}}$ & 8.27$_{\pm\text{1.21}}$ & 3.27$_{\pm\text{1.53}}$ & 7.45$_{\pm\text{2.36}}$ & 4.39$_{\pm\text{1.93}}$ & \textbf{1.27}$_{\pm\text{0.72}}$ \\

\bottomrule

\end{tabular}
}
\end{table}
\clearpage

\begin{table}[p]
\centering
\caption{\textbf{$\chi^2$ test between real test data and synthetic data} on eight real-world tabular datasets with varied real data availability. We report the mean $\pm$ std result and average rank across datasets. A higher rank implies higher fidelity. Note that ``N/A'' denotes that a specific generator was not applicable, and the rank is computed with the mean result of other methods. We \textbf{bold} the highest result for each dataset of different sample sizes. TabEBM achieves the best overall performance against benchmark generators.}
\resizebox{\textwidth}{!}{

\setlength{\tabcolsep}{5pt}
\begin{tabular}{m{0.2cm}lr|rrrrrrrr|r}

\toprule

\multicolumn{2}{l}{Datasets}  & $N_{\text{real}}$ & SMOTE & TVAE & CTGAN & NFLOW & TabDDPM & ARF & GOGGLE & TabPFGen & \textbf{TabEBM} \\

\midrule

\multirow{24}{*}{\rotatebox{90}{\begin{tabular}{l} \textit{At most 10 classes} \end{tabular}}}& \multirow[c]{5}{*}{protein} & 20 & N/A & 0.01$_{\pm\text{0.00}}$ & 0.01$_{\pm\text{0.00}}$ & 0.01$_{\pm\text{0.00}}$ & 0.01$_{\pm\text{0.00}}$ & 0.01$_{\pm\text{0.00}}$ & 0.01$_{\pm\text{0.00}}$ & 0.02$_{\pm\text{0.01}}$ & \textbf{0.06}$_{\pm\text{0.03}}$ \\

& & 50 & 0.26$_{\pm\text{0.05}}$ & \textbf{0.32}$_{\pm\text{0.06}}$ & 0.09$_{\pm\text{0.04}}$ & 0.17$_{\pm\text{0.04}}$ & 0.01$_{\pm\text{0.00}}$ & 0.25$_{\pm\text{0.04}}$ & 0.22$_{\pm\text{0.05}}$ & 0.05$_{\pm\text{0.03}}$ & 0.15$_{\pm\text{0.04}}$ \\

& & 100 & \textbf{0.39}$_{\pm\text{0.04}}$ & 0.33$_{\pm\text{0.05}}$ & 0.18$_{\pm\text{0.05}}$ & 0.29$_{\pm\text{0.09}}$ & 0.01$_{\pm\text{0.00}}$ & 0.34$_{\pm\text{0.04}}$ & 0.13$_{\pm\text{0.04}}$ & 0.12$_{\pm\text{0.04}}$ & 0.26$_{\pm\text{0.05}}$ \\

& & 200 & \textbf{0.48}$_{\pm\text{0.06}}$ & 0.30$_{\pm\text{0.07}}$ & 0.28$_{\pm\text{0.06}}$ & 0.48$_{\pm\text{0.04}}$ & 0.01$_{\pm\text{0.00}}$ & 0.44$_{\pm\text{0.08}}$ & 0.02$_{\pm\text{0.01}}$ & 0.34$_{\pm\text{0.08}}$ & 0.43$_{\pm\text{0.06}}$ \\

& & 500 & 0.54$_{\pm\text{0.05}}$ & 0.24$_{\pm\text{0.07}}$ & 0.33$_{\pm\text{0.06}}$ & \textbf{0.60}$_{\pm\text{0.04}}$ & 0.01$_{\pm\text{0.00}}$ & 0.55$_{\pm\text{0.04}}$ & 0.01$_{\pm\text{0.00}}$ & 0.53$_{\pm\text{0.05}}$ & 0.58$_{\pm\text{0.06}}$ \\

\cmidrule{2-12}

& \multirow[c]{5}{*}{fourier} & 20 & N/A & 0.00$_{\pm\text{0.00}}$ & 0.01$_{\pm\text{0.00}}$ & 0.01$_{\pm\text{0.00}}$ & 0.01$_{\pm\text{0.00}}$ & 0.01$_{\pm\text{0.00}}$ & 0.01$_{\pm\text{0.00}}$ & 0.01$_{\pm\text{0.00}}$ & \textbf{0.10}$_{\pm\text{0.04}}$ \\

& & 50 & 0.42$_{\pm\text{0.08}}$ & \textbf{0.45}$_{\pm\text{0.07}}$ & 0.15$_{\pm\text{0.06}}$ & 0.31$_{\pm\text{0.07}}$ & 0.01$_{\pm\text{0.00}}$ & 0.39$_{\pm\text{0.08}}$ & 0.29$_{\pm\text{0.08}}$ & 0.10$_{\pm\text{0.04}}$ & 0.33$_{\pm\text{0.06}}$ \\

& & 100 & \textbf{0.58}$_{\pm\text{0.08}}$ & 0.44$_{\pm\text{0.07}}$ & 0.32$_{\pm\text{0.07}}$ & 0.52$_{\pm\text{0.09}}$ & 0.01$_{\pm\text{0.00}}$ & 0.48$_{\pm\text{0.07}}$ & 0.08$_{\pm\text{0.04}}$ & 0.26$_{\pm\text{0.07}}$ & 0.49$_{\pm\text{0.07}}$ \\

& & 200 & 0.67$_{\pm\text{0.05}}$ & 0.28$_{\pm\text{0.03}}$ & 0.38$_{\pm\text{0.06}}$ & \textbf{0.68}$_{\pm\text{0.04}}$ & 0.01$_{\pm\text{0.00}}$ & 0.60$_{\pm\text{0.04}}$ & 0.02$_{\pm\text{0.01}}$ & 0.46$_{\pm\text{0.06}}$ & 0.60$_{\pm\text{0.05}}$ \\

& & 500 & \textbf{0.76}$_{\pm\text{0.04}}$ & 0.15$_{\pm\text{0.06}}$ & 0.55$_{\pm\text{0.08}}$ & 0.72$_{\pm\text{0.07}}$ & 0.01$_{\pm\text{0.00}}$ & 0.71$_{\pm\text{0.05}}$ & 0.01$_{\pm\text{0.00}}$ & 0.69$_{\pm\text{0.04}}$ & 0.74$_{\pm\text{0.03}}$ \\

\cmidrule{2-12}

& \multirow[c]{5}{*}{biodeg} & 20 & 0.05$_{\pm\text{0.03}}$ & 0.05$_{\pm\text{0.04}}$ & 0.04$_{\pm\text{0.03}}$ & 0.04$_{\pm\text{0.02}}$ & 0.03$_{\pm\text{0.01}}$ & 0.05$_{\pm\text{0.03}}$ & 0.03$_{\pm\text{0.01}}$ & 0.10$_{\pm\text{0.01}}$ & \textbf{0.13}$_{\pm\text{0.02}}$ \\

& & 50 & 0.13$_{\pm\text{0.06}}$ & 0.12$_{\pm\text{0.04}}$ & 0.07$_{\pm\text{0.02}}$ & 0.08$_{\pm\text{0.04}}$ & 0.02$_{\pm\text{0.00}}$ & 0.08$_{\pm\text{0.05}}$ & 0.06$_{\pm\text{0.03}}$ & 0.08$_{\pm\text{0.04}}$ & \textbf{0.14}$_{\pm\text{0.05}}$ \\

& & 100 & 0.15$_{\pm\text{0.04}}$ & 0.14$_{\pm\text{0.05}}$ & 0.10$_{\pm\text{0.03}}$ & 0.12$_{\pm\text{0.05}}$ & 0.02$_{\pm\text{0.00}}$ & 0.12$_{\pm\text{0.04}}$ & 0.04$_{\pm\text{0.02}}$ & 0.07$_{\pm\text{0.03}}$ & \textbf{0.18}$_{\pm\text{0.04}}$ \\

& & 200 & 0.18$_{\pm\text{0.03}}$ & 0.15$_{\pm\text{0.04}}$ & 0.14$_{\pm\text{0.03}}$ & 0.19$_{\pm\text{0.05}}$ & 0.03$_{\pm\text{0.01}}$ & 0.18$_{\pm\text{0.03}}$ & 0.02$_{\pm\text{0.00}}$ & 0.14$_{\pm\text{0.05}}$ & \textbf{0.28}$_{\pm\text{0.07}}$ \\

& & 500 & 0.25$_{\pm\text{0.07}}$ & 0.13$_{\pm\text{0.05}}$ & 0.21$_{\pm\text{0.07}}$ & 0.28$_{\pm\text{0.05}}$ & 0.03$_{\pm\text{0.01}}$ & 0.28$_{\pm\text{0.07}}$ & 0.02$_{\pm\text{0.00}}$ & 0.23$_{\pm\text{0.07}}$ & \textbf{0.38}$_{\pm\text{0.05}}$ \\

\cmidrule{2-12}

& \multirow[c]{5}{*}{steel} & 20 & 0.13$_{\pm\text{0.04}}$ & 0.16$_{\pm\text{0.04}}$ & 0.09$_{\pm\text{0.03}}$ & 0.10$_{\pm\text{0.04}}$ & 0.03$_{\pm\text{0.00}}$ & 0.14$_{\pm\text{0.03}}$ & 0.12$_{\pm\text{0.05}}$ & 0.07$_{\pm\text{0.02}}$ & \textbf{0.29}$_{\pm\text{0.06}}$ \\

& & 50 & 0.26$_{\pm\text{0.07}}$ & 0.24$_{\pm\text{0.07}}$ & 0.17$_{\pm\text{0.04}}$ & 0.21$_{\pm\text{0.06}}$ & 0.03$_{\pm\text{0.00}}$ & 0.31$_{\pm\text{0.04}}$ & 0.05$_{\pm\text{0.02}}$ & 0.17$_{\pm\text{0.05}}$ & \textbf{0.43}$_{\pm\text{0.07}}$ \\

& & 100 & 0.32$_{\pm\text{0.04}}$ & 0.23$_{\pm\text{0.06}}$ & 0.24$_{\pm\text{0.07}}$ & 0.31$_{\pm\text{0.06}}$ & 0.03$_{\pm\text{0.00}}$ & 0.33$_{\pm\text{0.06}}$ & 0.04$_{\pm\text{0.02}}$ & 0.29$_{\pm\text{0.06}}$ & \textbf{0.49}$_{\pm\text{0.07}}$ \\

& & 200 & 0.36$_{\pm\text{0.04}}$ & 0.17$_{\pm\text{0.03}}$ & 0.29$_{\pm\text{0.06}}$ & 0.34$_{\pm\text{0.05}}$ & 0.03$_{\pm\text{0.00}}$ & 0.36$_{\pm\text{0.06}}$ & 0.02$_{\pm\text{0.00}}$ & 0.35$_{\pm\text{0.06}}$ & \textbf{0.53}$_{\pm\text{0.07}}$ \\

& & 500 & 0.37$_{\pm\text{0.04}}$ & 0.15$_{\pm\text{0.07}}$ & 0.28$_{\pm\text{0.05}}$ & 0.36$_{\pm\text{0.04}}$ & 0.03$_{\pm\text{0.00}}$ & 0.36$_{\pm\text{0.05}}$ & 0.02$_{\pm\text{0.00}}$ & 0.38$_{\pm\text{0.06}}$ & \textbf{0.57}$_{\pm\text{0.06}}$ \\

\cmidrule{2-12}

& \multirow[c]{4}{*}{stock} & 20 & 0.31$_{\pm\text{0.11}}$ & 0.38$_{\pm\text{0.13}}$ & 0.22$_{\pm\text{0.08}}$ & 0.25$_{\pm\text{0.08}}$ & 0.10$_{\pm\text{0.00}}$ & 0.33$_{\pm\text{0.11}}$ & \textbf{0.42}$_{\pm\text{0.18}}$ & 0.11$_{\pm\text{0.03}}$ & 0.41$_{\pm\text{0.19}}$ \\

& & 50 & 0.60$_{\pm\text{0.09}}$ & 0.56$_{\pm\text{0.12}}$ & 0.44$_{\pm\text{0.15}}$ & 0.54$_{\pm\text{0.08}}$ & 0.10$_{\pm\text{0.00}}$ & 0.73$_{\pm\text{0.14}}$ & 0.31$_{\pm\text{0.14}}$ & 0.41$_{\pm\text{0.16}}$ & \textbf{0.76}$_{\pm\text{0.10}}$ \\

& & 100 & 0.71$_{\pm\text{0.11}}$ & 0.40$_{\pm\text{0.11}}$ & 0.53$_{\pm\text{0.15}}$ & 0.68$_{\pm\text{0.12}}$ & 0.12$_{\pm\text{0.06}}$ & \textbf{0.85}$_{\pm\text{0.05}}$ & 0.13$_{\pm\text{0.05}}$ & 0.65$_{\pm\text{0.12}}$ & 0.84$_{\pm\text{0.07}}$ \\

& & 200 & 0.86$_{\pm\text{0.07}}$ & 0.45$_{\pm\text{0.12}}$ & 0.60$_{\pm\text{0.18}}$ & 0.91$_{\pm\text{0.11}}$ & 0.35$_{\pm\text{0.20}}$ & 0.97$_{\pm\text{0.05}}$ & 0.10$_{\pm\text{0.00}}$ & 0.92$_{\pm\text{0.08}}$ & \textbf{0.97}$_{\pm\text{0.05}}$ \\

\midrule

\multirow{9}{*}{\rotatebox{90}{\begin{tabular}{l} \textit{More than 10 classes} \end{tabular}}}& \multirow[c]{3}{*}{energy} & 50 & N/A & 0.15$_{\pm\text{0.08}}$ & 0.25$_{\pm\text{0.09}}$ & 0.28$_{\pm\text{0.10}}$ & 0.17$_{\pm\text{0.06}}$ & 0.40$_{\pm\text{0.14}}$ & 0.15$_{\pm\text{0.10}}$ & N/A & \textbf{0.78}$_{\pm\text{0.05}}$ \\

& & 100 & N/A & 0.10$_{\pm\text{0.09}}$ & 0.31$_{\pm\text{0.12}}$ & 0.34$_{\pm\text{0.08}}$ & 0.14$_{\pm\text{0.06}}$ & 0.30$_{\pm\text{0.09}}$ & 0.06$_{\pm\text{0.07}}$ & N/A & \textbf{0.92}$_{\pm\text{0.02}}$ \\

& & 200 & N/A & 0.16$_{\pm\text{0.11}}$ & 0.30$_{\pm\text{0.09}}$ & 0.38$_{\pm\text{0.12}}$ & 0.15$_{\pm\text{0.07}}$ & 0.28$_{\pm\text{0.08}}$ & 0.03$_{\pm\text{0.05}}$ & N/A & \textbf{0.96}$_{\pm\text{0.01}}$ \\

\cmidrule{2-12}

& \multirow[c]{2}{*}{collins} & 100 & N/A & 0.23$_{\pm\text{0.08}}$ & 0.21$_{\pm\text{0.08}}$ & \textbf{0.25}$_{\pm\text{0.09}}$ & 0.05$_{\pm\text{0.02}}$ & 0.24$_{\pm\text{0.06}}$ & 0.07$_{\pm\text{0.03}}$ & N/A & 0.18$_{\pm\text{0.07}}$ \\

& & 200 & 0.32$_{\pm\text{0.06}}$ & 0.16$_{\pm\text{0.06}}$ & 0.24$_{\pm\text{0.06}}$ & \textbf{0.38}$_{\pm\text{0.05}}$ & 0.06$_{\pm\text{0.02}}$ & 0.33$_{\pm\text{0.07}}$ & 0.02$_{\pm\text{0.04}}$ & N/A & 0.31$_{\pm\text{0.07}}$ \\

\cmidrule{2-12}

& \multirow[c]{4}{*}{texture} & 50 & 0.33$_{\pm\text{0.11}}$ & 0.44$_{\pm\text{0.10}}$ & 0.21$_{\pm\text{0.08}}$ & 0.29$_{\pm\text{0.10}}$ & 0.02$_{\pm\text{0.00}}$ & \textbf{0.55}$_{\pm\text{0.15}}$ & 0.27$_{\pm\text{0.08}}$ & N/A & 0.41$_{\pm\text{0.04}}$ \\

& & 100 & 0.44$_{\pm\text{0.13}}$ & 0.27$_{\pm\text{0.06}}$ & 0.27$_{\pm\text{0.09}}$ & 0.48$_{\pm\text{0.15}}$ & 0.02$_{\pm\text{0.00}}$ & \textbf{0.54}$_{\pm\text{0.06}}$ & 0.07$_{\pm\text{0.05}}$ & N/A & 0.49$_{\pm\text{0.08}}$ \\

& & 200 & 0.51$_{\pm\text{0.11}}$ & 0.19$_{\pm\text{0.10}}$ & 0.34$_{\pm\text{0.10}}$ & 0.58$_{\pm\text{0.11}}$ & 0.02$_{\pm\text{0.00}}$ & \textbf{0.64}$_{\pm\text{0.09}}$ & 0.00$_{\pm\text{0.00}}$ & N/A & 0.59$_{\pm\text{0.09}}$ \\

& & 500 & 0.63$_{\pm\text{0.14}}$ & 0.21$_{\pm\text{0.09}}$ & 0.49$_{\pm\text{0.09}}$ & \textbf{0.67}$_{\pm\text{0.08}}$ & 0.02$_{\pm\text{0.00}}$ & 0.64$_{\pm\text{0.08}}$ & 0.00$_{\pm\text{0.00}}$ & N/A & 0.66$_{\pm\text{0.12}}$ \\

\midrule
\rowcolor{Gainsboro!60}
\multicolumn{3}{l|}{\textbf{Average rank}} & 3.06$_{\pm\text{1.32}}$ & 5.45$_{\pm\text{2.35}}$ & 6.15$_{\pm\text{0.87}}$ & 3.64$_{\pm\text{1.75}}$ & 8.36$_{\pm\text{0.90}}$ & 3.18$_{\pm\text{1.47}}$ & 7.76$_{\pm\text{1.77}}$ & 5.33$_{\pm\text{1.71}}$ & \textbf{2.06}$_{\pm\text{1.43}}$ \\

\bottomrule

\end{tabular}
}
\end{table}
\clearpage

\FloatBarrier
\subsection{Results on Privacy Preservation}
\label{appendix:res_privacy}

\FloatBarrier
\subsubsection{Downstream Accuracy in Data Sharing}

\begin{table}[htbp]
\centering
\caption{\textbf{Classification accuracy} (\%) aggregated over six downstream predictors, comparing data sharing on eight real-world tabular datasets with varied real data availability. We report the mean $\pm$ std balanced accuracy and average accuracy rank across datasets. A higher rank implies higher accuracy. Note that ``N/A'' denotes the inapplicability of a specific generator. Different from \cref{tab:test_acc_per_dataset_augmentation}, ``$-$'' denotes a generator cannot satisfy the requirement of generating 500 stratified samples even after generating 10,000 synthetic samples. The results of these inapplicable or failed generators are computed with the mean results of other methods. We \textbf{bold} the highest result for each dataset of different sample sizes. TVAE learns the joint distribution $p(\rvx, y)$ and fails to maintain the original training label distribution. TabEBM achieves the best overall performance against Baseline and benchmark generators.}
\label{tab:test_acc_per_dataset_share}
\resizebox{\textwidth}{!}{
\begin{tabular}{m{0.2cm}lr|rrrrrrrr|r}

\toprule

\multicolumn{2}{l}{Datasets}  & $N_{\text{real}}$ & SMOTE & TVAE & CTGAN & NFLOW & TabDDPM & ARF & GOGGLE & TabPFGen & \textbf{TabEBM} \\

\midrule

\multirow{24}{*}{\rotatebox{90}{\begin{tabular}{l} \textit{At most 10 classes} \end{tabular}}}& \multirow[c]{5}{*}{protein} & 20 & N/A & 16.38$_{\pm\text{3.34}}$ & 13.06$_{\pm\text{3.57}}$ & 12.99$_{\pm\text{4.16}}$ & 12.64$_{\pm\text{3.06}}$ & 19.42$_{\pm\text{3.96}}$ & 12.88$_{\pm\text{2.32}}$ & 33.11$_{\pm\text{3.64}}$ & \textbf{33.97}$_{\pm\text{3.48}}$ \\

& & 50 & 54.55$_{\pm\text{3.94}}$ & 27.29$_{\pm\text{3.43}}$ & 17.18$_{\pm\text{4.87}}$ & 13.20$_{\pm\text{3.08}}$ & 14.04$_{\pm\text{3.29}}$ & 29.34$_{\pm\text{4.46}}$ & 13.05$_{\pm\text{3.34}}$ & 54.82$_{\pm\text{3.74}}$ & \textbf{55.83}$_{\pm\text{4.35}}$ \\

& & 100 & 72.46$_{\pm\text{3.54}}$ & 40.04$_{\pm\text{4.54}}$ & 26.65$_{\pm\text{3.89}}$ & 12.80$_{\pm\text{3.48}}$ & 17.65$_{\pm\text{3.25}}$ & 35.90$_{\pm\text{5.29}}$ & 13.55$_{\pm\text{3.03}}$ & 71.63$_{\pm\text{3.95}}$ & \textbf{72.99}$_{\pm\text{3.69}}$ \\

& & 200 & 83.12$_{\pm\text{2.33}}$ & 45.33$_{\pm\text{6.01}}$ & 32.52$_{\pm\text{5.60}}$ & 14.07$_{\pm\text{3.94}}$ & 20.79$_{\pm\text{4.42}}$ & 41.63$_{\pm\text{4.24}}$ & 11.68$_{\pm\text{3.60}}$ & 84.18$_{\pm\text{1.90}}$ & \textbf{84.29}$_{\pm\text{2.03}}$ \\

& & 500 & 89.63$_{\pm\text{1.57}}$ & 55.24$_{\pm\text{5.30}}$ & 44.35$_{\pm\text{5.67}}$ & 12.76$_{\pm\text{2.73}}$ & 21.63$_{\pm\text{4.90}}$ & 54.26$_{\pm\text{3.47}}$ & 11.09$_{\pm\text{3.19}}$ & \textbf{91.19}$_{\pm\text{1.48}}$ & 90.99$_{\pm\text{1.53}}$ \\

\cmidrule{2-12}

& \multirow[c]{5}{*}{fourier} & 20 & N/A & $-$ & 13.30$_{\pm\text{3.14}}$ & 10.72$_{\pm\text{3.03}}$ & 10.03$_{\pm\text{2.42}}$ & 17.14$_{\pm\text{4.25}}$ & 10.29$_{\pm\text{2.81}}$ & 33.87$_{\pm\text{4.93}}$ & \textbf{37.06}$_{\pm\text{4.80}}$ \\

& & 50 & 55.11$_{\pm\text{3.56}}$ & $-$ & N/A & 9.88$_{\pm\text{2.51}}$ & 11.96$_{\pm\text{2.30}}$ & 28.63$_{\pm\text{3.71}}$ & 11.52$_{\pm\text{2.46}}$ & 56.11$_{\pm\text{3.28}}$ & \textbf{56.88}$_{\pm\text{2.90}}$ \\

& & 100 & 64.06$_{\pm\text{3.34}}$ & 34.80$_{\pm\text{4.60}}$ & 23.35$_{\pm\text{4.20}}$ & 10.85$_{\pm\text{3.02}}$ & 17.68$_{\pm\text{3.42}}$ & 32.59$_{\pm\text{4.02}}$ & 9.82$_{\pm\text{2.84}}$ & 64.42$_{\pm\text{2.96}}$ & \textbf{64.93}$_{\pm\text{3.06}}$ \\

& & 200 & 70.78$_{\pm\text{2.99}}$ & 40.49$_{\pm\text{3.43}}$ & 32.98$_{\pm\text{5.20}}$ & 10.28$_{\pm\text{3.74}}$ & 28.63$_{\pm\text{4.22}}$ & 39.34$_{\pm\text{3.41}}$ & 9.55$_{\pm\text{2.06}}$ & 71.37$_{\pm\text{2.56}}$ & \textbf{72.01}$_{\pm\text{2.67}}$ \\

& & 500 & 74.84$_{\pm\text{1.47}}$ & 46.83$_{\pm\text{3.74}}$ & 46.13$_{\pm\text{4.33}}$ & 11.13$_{\pm\text{2.47}}$ & 27.25$_{\pm\text{6.04}}$ & 47.19$_{\pm\text{2.89}}$ & 10.31$_{\pm\text{2.19}}$ & 75.86$_{\pm\text{1.91}}$ & \textbf{76.58}$_{\pm\text{1.54}}$ \\

\cmidrule{2-12}

& \multirow[c]{5}{*}{biodeg} & 20 & 68.75$_{\pm\text{4.96}}$ & 64.07$_{\pm\text{7.39}}$ & 53.60$_{\pm\text{7.49}}$ & 54.58$_{\pm\text{8.02}}$ & 48.39$_{\pm\text{2.73}}$ & 58.69$_{\pm\text{6.75}}$ & 48.65$_{\pm\text{9.50}}$ & 69.14$_{\pm\text{5.08}}$ & \textbf{69.66}$_{\pm\text{5.18}}$ \\

& & 50 & 72.08$_{\pm\text{3.01}}$ & 67.11$_{\pm\text{5.03}}$ & 62.21$_{\pm\text{7.04}}$ & 55.12$_{\pm\text{8.81}}$ & 47.86$_{\pm\text{3.58}}$ & 70.14$_{\pm\text{3.92}}$ & 52.14$_{\pm\text{7.45}}$ & 73.53$_{\pm\text{3.39}}$ & \textbf{73.96}$_{\pm\text{3.13}}$ \\

& & 100 & 75.61$_{\pm\text{2.48}}$ & 70.90$_{\pm\text{4.96}}$ & 70.33$_{\pm\text{3.28}}$ & 58.08$_{\pm\text{6.71}}$ & 48.78$_{\pm\text{2.92}}$ & 70.79$_{\pm\text{2.88}}$ & 49.15$_{\pm\text{7.71}}$ & \textbf{76.65}$_{\pm\text{2.03}}$ & 76.56$_{\pm\text{2.29}}$ \\

& & 200 & 78.97$_{\pm\text{1.46}}$ & 71.54$_{\pm\text{4.84}}$ & 71.42$_{\pm\text{4.30}}$ & 55.78$_{\pm\text{7.76}}$ & 47.63$_{\pm\text{3.98}}$ & 72.74$_{\pm\text{3.46}}$ & 50.75$_{\pm\text{6.70}}$ & 79.66$_{\pm\text{2.49}}$ & \textbf{79.80}$_{\pm\text{2.15}}$ \\

& & 500 & 81.26$_{\pm\text{1.42}}$ & 74.57$_{\pm\text{3.49}}$ & 76.32$_{\pm\text{2.93}}$ & 52.78$_{\pm\text{4.17}}$ & 47.31$_{\pm\text{2.99}}$ & 75.67$_{\pm\text{2.27}}$ & 48.05$_{\pm\text{7.50}}$ & \textbf{81.36}$_{\pm\text{1.69}}$ & 81.10$_{\pm\text{1.52}}$ \\

\cmidrule{2-12}

& \multirow[c]{5}{*}{steel} & 20 & 57.55$_{\pm\text{4.83}}$ & 54.39$_{\pm\text{7.65}}$ & 52.13$_{\pm\text{5.78}}$ & 51.21$_{\pm\text{6.68}}$ & 51.28$_{\pm\text{4.58}}$ & 52.03$_{\pm\text{4.88}}$ & 49.25$_{\pm\text{4.36}}$ & 63.27$_{\pm\text{5.66}}$ & \textbf{63.30}$_{\pm\text{5.47}}$ \\

& & 50 & 64.84$_{\pm\text{4.07}}$ & 59.06$_{\pm\text{3.85}}$ & 56.70$_{\pm\text{4.96}}$ & 53.24$_{\pm\text{4.77}}$ & 49.18$_{\pm\text{4.67}}$ & 57.16$_{\pm\text{4.35}}$ & 49.90$_{\pm\text{3.96}}$ & 78.55$_{\pm\text{5.17}}$ & \textbf{79.99}$_{\pm\text{6.37}}$ \\

& & 100 & 72.48$_{\pm\text{4.51}}$ & 61.74$_{\pm\text{4.16}}$ & 59.37$_{\pm\text{5.08}}$ & 52.79$_{\pm\text{4.06}}$ & 45.01$_{\pm\text{5.84}}$ & 57.26$_{\pm\text{4.46}}$ & 49.66$_{\pm\text{3.56}}$ & 90.12$_{\pm\text{4.20}}$ & \textbf{92.33}$_{\pm\text{2.57}}$ \\

& & 200 & 77.85$_{\pm\text{3.50}}$ & 65.45$_{\pm\text{4.07}}$ & 63.31$_{\pm\text{4.63}}$ & 51.02$_{\pm\text{2.94}}$ & 43.71$_{\pm\text{7.51}}$ & 61.41$_{\pm\text{4.21}}$ & 50.20$_{\pm\text{3.70}}$ & 94.61$_{\pm\text{1.75}}$ & \textbf{95.54}$_{\pm\text{1.45}}$ \\

& & 500 & 84.21$_{\pm\text{3.52}}$ & 70.26$_{\pm\text{4.93}}$ & 70.62$_{\pm\text{5.09}}$ & 49.97$_{\pm\text{4.05}}$ & 46.98$_{\pm\text{4.30}}$ & 66.33$_{\pm\text{5.72}}$ & 51.60$_{\pm\text{3.43}}$ & 96.31$_{\pm\text{1.25}}$ & \textbf{97.04}$_{\pm\text{1.17}}$ \\

\cmidrule{2-12}

& \multirow[c]{4}{*}{stock} & 20 & 81.62$_{\pm\text{4.62}}$ & 69.35$_{\pm\text{8.30}}$ & 51.71$_{\pm\text{12.09}}$ & 67.05$_{\pm\text{11.69}}$ & 75.94$_{\pm\text{14.43}}$ & 64.16$_{\pm\text{9.18}}$ & 48.55$_{\pm\text{12.54}}$ & 82.55$_{\pm\text{4.42}}$ & \textbf{83.60}$_{\pm\text{4.09}}$ \\

& & 50 & 87.43$_{\pm\text{2.47}}$ & 76.07$_{\pm\text{5.02}}$ & 69.12$_{\pm\text{5.00}}$ & 70.61$_{\pm\text{8.50}}$ & 85.99$_{\pm\text{2.82}}$ & 78.38$_{\pm\text{3.56}}$ & 49.74$_{\pm\text{9.74}}$ & 88.09$_{\pm\text{2.37}}$ & \textbf{88.49}$_{\pm\text{2.34}}$ \\

& & 100 & 89.63$_{\pm\text{1.30}}$ & 81.18$_{\pm\text{4.24}}$ & 78.44$_{\pm\text{3.91}}$ & 72.05$_{\pm\text{5.43}}$ & 88.27$_{\pm\text{2.34}}$ & 84.02$_{\pm\text{2.86}}$ & 51.07$_{\pm\text{10.98}}$ & 90.13$_{\pm\text{1.57}}$ & \textbf{90.58}$_{\pm\text{1.34}}$ \\

& & 200 & \textbf{91.11}$_{\pm\text{1.29}}$ & 84.05$_{\pm\text{2.63}}$ & 82.12$_{\pm\text{2.93}}$ & 75.44$_{\pm\text{3.54}}$ & 89.92$_{\pm\text{1.56}}$ & 85.66$_{\pm\text{2.26}}$ & 49.74$_{\pm\text{12.06}}$ & 91.09$_{\pm\text{1.52}}$ & 91.07$_{\pm\text{1.07}}$ \\

\midrule

\multirow{9}{*}{\rotatebox{90}{\begin{tabular}{l} \textit{More than 10 classes} \end{tabular}}}& \multirow[c]{3}{*}{energy} & 50 & N/A & 7.17$_{\pm\text{1.81}}$ & 5.20$_{\pm\text{2.04}}$ & 5.33$_{\pm\text{1.61}}$ & 4.21$_{\pm\text{1.69}}$ & 7.26$_{\pm\text{1.89}}$ & 4.52$_{\pm\text{0.57}}$ & N/A & \textbf{23.80}$_{\pm\text{2.60}}$ \\

& & 100 & N/A & $-$ & 7.66$_{\pm\text{1.96}}$ & 5.52$_{\pm\text{1.49}}$ & 4.08$_{\pm\text{1.40}}$ & 9.87$_{\pm\text{2.01}}$ & 4.01$_{\pm\text{1.12}}$ & N/A & \textbf{30.15}$_{\pm\text{3.21}}$ \\

& & 200 & N/A & $-$ & 7.57$_{\pm\text{2.01}}$ & 6.92$_{\pm\text{1.96}}$ & 3.42$_{\pm\text{1.25}}$ & 11.85$_{\pm\text{2.39}}$ & 4.18$_{\pm\text{0.91}}$ & N/A & \textbf{35.74}$_{\pm\text{3.59}}$ \\

\cmidrule{2-12}

& \multirow[c]{2}{*}{collins} & 100 & N/A & $-$ & 5.51$_{\pm\text{0.87}}$ & 4.58$_{\pm\text{0.95}}$ & 11.74$_{\pm\text{1.73}}$ & 5.34$_{\pm\text{1.48}}$ & 3.77$_{\pm\text{0.63}}$ & N/A & \textbf{13.12}$_{\pm\text{1.75}}$ \\

& & 200 & \textbf{17.60}$_{\pm\text{1.83}}$ & $-$ & 5.55$_{\pm\text{1.34}}$ & 4.70$_{\pm\text{0.83}}$ & 13.64$_{\pm\text{1.54}}$ & 5.46$_{\pm\text{0.90}}$ & 3.92$_{\pm\text{0.92}}$ & N/A & 16.80$_{\pm\text{1.55}}$ \\

\cmidrule{2-12}

& \multirow[c]{4}{*}{texture} & 50 & 75.50$_{\pm\text{2.93}}$ & 23.86$_{\pm\text{8.05}}$ & 12.00$_{\pm\text{4.19}}$ & 12.16$_{\pm\text{3.20}}$ & 17.18$_{\pm\text{5.34}}$ & 16.82$_{\pm\text{6.16}}$ & 9.63$_{\pm\text{2.75}}$ & N/A & \textbf{78.84}$_{\pm\text{3.35}}$ \\

& & 100 & 83.77$_{\pm\text{2.52}}$ & 26.87$_{\pm\text{4.19}}$ & 14.05$_{\pm\text{6.20}}$ & 13.95$_{\pm\text{5.32}}$ & 20.08$_{\pm\text{5.52}}$ & 23.82$_{\pm\text{5.96}}$ & 10.21$_{\pm\text{2.69}}$ & N/A & \textbf{85.88}$_{\pm\text{1.87}}$ \\

& & 200 & 87.96$_{\pm\text{1.79}}$ & 21.59$_{\pm\text{8.99}}$ & 15.36$_{\pm\text{4.43}}$ & 12.17$_{\pm\text{4.57}}$ & 20.59$_{\pm\text{5.96}}$ & 42.28$_{\pm\text{5.81}}$ & 9.52$_{\pm\text{2.91}}$ & N/A & \textbf{88.84}$_{\pm\text{1.68}}$ \\

& & 500 & \textbf{91.65}$_{\pm\text{1.14}}$ & 34.48$_{\pm\text{9.28}}$ & 26.45$_{\pm\text{8.77}}$ & 12.37$_{\pm\text{4.63}}$ & 20.07$_{\pm\text{5.56}}$ & 57.94$_{\pm\text{4.77}}$ & 10.22$_{\pm\text{3.69}}$ & N/A & 91.36$_{\pm\text{1.12}}$ \\

\midrule
\rowcolor{Gainsboro!60}
\multicolumn{3}{l|}{\textbf{Average rank}} & 2.76$_{\pm\text{0.76}}$ & 4.52$_{\pm\text{0.77}}$ & 6.18$_{\pm\text{0.95}}$ & 7.48$_{\pm\text{0.76}}$ & 7.03$_{\pm\text{2.07}}$ & 4.91$_{\pm\text{1.16}}$ & 8.52$_{\pm\text{0.57}}$ & 2.36$_{\pm\text{0.89}}$ & \textbf{1.24}$_{\pm\text{0.56}}$ \\

\bottomrule

\end{tabular}
}
\end{table}
\clearpage

\begin{table}[p]
\centering
\caption{\textbf{Classification accuracy} (\%) of LR, comparing data sharing on eight real-world tabular datasets with varied real data availability. We report the mean $\pm$ std balanced accuracy and average accuracy rank across datasets. A higher rank implies higher accuracy. Note that ``N/A'' denotes the inapplicability of a specific generator. Different from \cref{tab:test_acc_per_dataset_augmentation}, ``$-$'' denotes a generator cannot satisfy the requirement of generating 500 stratified samples even after generating 10,000 synthetic samples. The results of these inapplicable or failed generators are computed with the mean results of other methods. We \textbf{bold} the highest result for each dataset of different sample sizes. TVAE learns the joint distribution $p(\rvx, y)$ and fails to maintain the original training label distribution. TabEBM achieves the best overall performance against Baseline and benchmark generators.}
\resizebox{\textwidth}{!}{
\begin{tabular}{m{0.2cm}lr|rrrrrrrr|r}

\toprule

\multicolumn{2}{l}{Datasets}  & $N_{\text{real}}$ & SMOTE & TVAE & CTGAN & NFLOW & TabDDPM & ARF & GOGGLE & TabPFGen & \textbf{TabEBM} \\

\midrule

\multirow{24}{*}{\rotatebox{90}{\begin{tabular}{l} \textit{At most 10 classes} \end{tabular}}}& \multirow[c]{5}{*}{protein} & 20 & N/A & 14.51$_{\pm\text{3.57}}$ & 13.00$_{\pm\text{5.74}}$ & 12.08$_{\pm\text{4.67}}$ & 13.83$_{\pm\text{3.50}}$ & 22.46$_{\pm\text{5.01}}$ & 12.52$_{\pm\text{2.66}}$ & 38.00$_{\pm\text{2.16}}$ & \textbf{38.04}$_{\pm\text{2.34}}$ \\

& & 50 & 61.15$_{\pm\text{4.22}}$ & 24.43$_{\pm\text{3.31}}$ & 16.81$_{\pm\text{5.60}}$ & 12.55$_{\pm\text{2.78}}$ & 14.74$_{\pm\text{4.36}}$ & 33.77$_{\pm\text{5.53}}$ & 12.26$_{\pm\text{4.83}}$ & \textbf{63.01}$_{\pm\text{3.66}}$ & 62.94$_{\pm\text{4.10}}$ \\

& & 100 & 78.16$_{\pm\text{3.53}}$ & 40.22$_{\pm\text{4.23}}$ & 27.14$_{\pm\text{5.20}}$ & 12.40$_{\pm\text{5.31}}$ & 19.34$_{\pm\text{3.12}}$ & 40.99$_{\pm\text{5.14}}$ & 12.33$_{\pm\text{6.06}}$ & \textbf{80.54}$_{\pm\text{3.22}}$ & 80.18$_{\pm\text{2.97}}$ \\

& & 200 & 88.75$_{\pm\text{1.37}}$ & 45.70$_{\pm\text{7.29}}$ & 34.17$_{\pm\text{6.11}}$ & 13.87$_{\pm\text{4.39}}$ & 22.01$_{\pm\text{5.64}}$ & 44.76$_{\pm\text{4.63}}$ & 12.78$_{\pm\text{5.20}}$ & \textbf{90.89}$_{\pm\text{1.53}}$ & 90.09$_{\pm\text{1.86}}$ \\

& & 500 & 94.29$_{\pm\text{1.17}}$ & 60.39$_{\pm\text{2.81}}$ & 47.87$_{\pm\text{7.19}}$ & 12.58$_{\pm\text{3.76}}$ & 24.69$_{\pm\text{5.87}}$ & 58.91$_{\pm\text{2.80}}$ & 9.79$_{\pm\text{3.65}}$ & \textbf{95.86}$_{\pm\text{1.59}}$ & 95.38$_{\pm\text{1.40}}$ \\

\cmidrule{2-12}

& \multirow[c]{5}{*}{fourier} & 20 & N/A & $-$ & 11.64$_{\pm\text{4.35}}$ & 10.16$_{\pm\text{2.99}}$ & 8.48$_{\pm\text{2.86}}$ & 18.75$_{\pm\text{4.29}}$ & 11.38$_{\pm\text{4.85}}$ & 42.98$_{\pm\text{5.14}}$ & \textbf{43.04}$_{\pm\text{5.10}}$ \\

& & 50 & 58.24$_{\pm\text{2.05}}$ & $-$ & N/A & 9.82$_{\pm\text{4.25}}$ & 13.48$_{\pm\text{2.85}}$ & 33.97$_{\pm\text{3.71}}$ & 12.00$_{\pm\text{2.92}}$ & \textbf{60.30}$_{\pm\text{1.58}}$ & 60.28$_{\pm\text{1.63}}$ \\

& & 100 & 65.32$_{\pm\text{2.38}}$ & 28.60$_{\pm\text{2.38}}$ & 21.26$_{\pm\text{2.88}}$ & 11.24$_{\pm\text{3.65}}$ & 19.72$_{\pm\text{3.78}}$ & 37.64$_{\pm\text{3.66}}$ & 11.34$_{\pm\text{2.54}}$ & 67.36$_{\pm\text{2.21}}$ & \textbf{67.38}$_{\pm\text{2.34}}$ \\

& & 200 & 70.74$_{\pm\text{2.13}}$ & 37.50$_{\pm\text{4.11}}$ & 28.86$_{\pm\text{4.32}}$ & 10.12$_{\pm\text{4.56}}$ & 33.58$_{\pm\text{3.65}}$ & 42.75$_{\pm\text{2.57}}$ & 7.52$_{\pm\text{2.05}}$ & \textbf{72.20}$_{\pm\text{3.02}}$ & 72.18$_{\pm\text{2.94}}$ \\

& & 500 & 73.96$_{\pm\text{0.94}}$ & 46.84$_{\pm\text{2.99}}$ & 41.18$_{\pm\text{4.18}}$ & 10.60$_{\pm\text{3.33}}$ & 31.88$_{\pm\text{9.69}}$ & 48.67$_{\pm\text{1.58}}$ & 10.56$_{\pm\text{3.12}}$ & 75.58$_{\pm\text{1.99}}$ & \textbf{76.00}$_{\pm\text{2.93}}$ \\

\cmidrule{2-12}

& \multirow[c]{5}{*}{biodeg} & 20 & 69.93$_{\pm\text{5.59}}$ & 66.92$_{\pm\text{7.16}}$ & 50.10$_{\pm\text{10.26}}$ & 56.08$_{\pm\text{10.54}}$ & 46.15$_{\pm\text{4.22}}$ & 61.15$_{\pm\text{7.27}}$ & 45.12$_{\pm\text{9.23}}$ & 70.78$_{\pm\text{3.95}}$ & \textbf{71.24}$_{\pm\text{4.85}}$ \\

& & 50 & 74.71$_{\pm\text{3.39}}$ & 70.19$_{\pm\text{4.89}}$ & 62.90$_{\pm\text{6.28}}$ & 56.85$_{\pm\text{10.24}}$ & 42.74$_{\pm\text{2.82}}$ & 73.38$_{\pm\text{2.61}}$ & 54.52$_{\pm\text{12.44}}$ & 75.67$_{\pm\text{2.36}}$ & \textbf{76.47}$_{\pm\text{2.99}}$ \\

& & 100 & 77.48$_{\pm\text{1.83}}$ & 71.77$_{\pm\text{5.56}}$ & 71.42$_{\pm\text{3.27}}$ & 59.17$_{\pm\text{7.35}}$ & 46.24$_{\pm\text{3.89}}$ & 74.48$_{\pm\text{2.84}}$ & 48.03$_{\pm\text{12.56}}$ & 77.94$_{\pm\text{2.45}}$ & \textbf{78.34}$_{\pm\text{2.11}}$ \\

& & 200 & 80.56$_{\pm\text{1.77}}$ & 75.37$_{\pm\text{4.47}}$ & 72.61$_{\pm\text{5.99}}$ & 55.71$_{\pm\text{7.21}}$ & 44.16$_{\pm\text{4.71}}$ & 75.09$_{\pm\text{2.93}}$ & 52.94$_{\pm\text{11.50}}$ & 81.21$_{\pm\text{2.06}}$ & \textbf{81.50}$_{\pm\text{1.84}}$ \\

& & 500 & \textbf{82.68}$_{\pm\text{1.01}}$ & 77.92$_{\pm\text{2.73}}$ & 77.55$_{\pm\text{2.55}}$ & 54.90$_{\pm\text{5.52}}$ & 46.53$_{\pm\text{2.91}}$ & 77.87$_{\pm\text{2.10}}$ & 46.40$_{\pm\text{12.51}}$ & 82.38$_{\pm\text{1.61}}$ & 82.14$_{\pm\text{1.30}}$ \\

\cmidrule{2-12}

& \multirow[c]{5}{*}{steel} & 20 & 56.80$_{\pm\text{5.46}}$ & 54.06$_{\pm\text{11.24}}$ & 54.41$_{\pm\text{5.57}}$ & 52.14$_{\pm\text{7.31}}$ & 51.40$_{\pm\text{6.83}}$ & 54.38$_{\pm\text{4.84}}$ & 48.32$_{\pm\text{6.98}}$ & 66.75$_{\pm\text{9.74}}$ & \textbf{67.05}$_{\pm\text{9.39}}$ \\

& & 50 & 66.80$_{\pm\text{6.52}}$ & 61.11$_{\pm\text{3.73}}$ & 58.25$_{\pm\text{6.67}}$ & 52.95$_{\pm\text{6.68}}$ & 49.07$_{\pm\text{7.58}}$ & 61.31$_{\pm\text{4.45}}$ & 48.86$_{\pm\text{7.10}}$ & \textbf{93.51}$_{\pm\text{4.94}}$ & 92.13$_{\pm\text{4.90}}$ \\

& & 100 & 79.38$_{\pm\text{4.02}}$ & 64.64$_{\pm\text{3.99}}$ & 60.57$_{\pm\text{5.90}}$ & 54.40$_{\pm\text{5.67}}$ & 42.51$_{\pm\text{5.11}}$ & 58.86$_{\pm\text{3.66}}$ & 50.95$_{\pm\text{3.77}}$ & \textbf{99.21}$_{\pm\text{0.86}}$ & 99.17$_{\pm\text{0.93}}$ \\

& & 200 & 82.66$_{\pm\text{3.91}}$ & 71.78$_{\pm\text{3.68}}$ & 65.44$_{\pm\text{5.02}}$ & 51.72$_{\pm\text{4.16}}$ & 38.19$_{\pm\text{8.24}}$ & 63.31$_{\pm\text{5.98}}$ & 53.46$_{\pm\text{2.96}}$ & 99.45$_{\pm\text{0.69}}$ & \textbf{99.51}$_{\pm\text{0.69}}$ \\

& & 500 & 90.21$_{\pm\text{3.80}}$ & 76.49$_{\pm\text{5.04}}$ & 75.41$_{\pm\text{6.20}}$ & 48.71$_{\pm\text{6.15}}$ & 46.00$_{\pm\text{4.77}}$ & 72.11$_{\pm\text{7.56}}$ & 52.55$_{\pm\text{3.50}}$ & 99.70$_{\pm\text{0.22}}$ & \textbf{99.72}$_{\pm\text{0.20}}$ \\

\cmidrule{2-12}

& \multirow[c]{4}{*}{stock} & 20 & 80.31$_{\pm\text{4.03}}$ & 71.68$_{\pm\text{8.45}}$ & 54.58$_{\pm\text{13.04}}$ & 70.64$_{\pm\text{9.77}}$ & 71.05$_{\pm\text{11.98}}$ & 68.03$_{\pm\text{7.38}}$ & 49.84$_{\pm\text{19.78}}$ & 79.58$_{\pm\text{4.43}}$ & \textbf{80.33}$_{\pm\text{3.52}}$ \\

& & 50 & 81.28$_{\pm\text{2.87}}$ & 75.75$_{\pm\text{3.84}}$ & 69.67$_{\pm\text{3.14}}$ & 72.99$_{\pm\text{6.89}}$ & 77.72$_{\pm\text{4.99}}$ & 76.64$_{\pm\text{2.54}}$ & 49.01$_{\pm\text{17.32}}$ & \textbf{82.37}$_{\pm\text{3.13}}$ & 82.06$_{\pm\text{2.58}}$ \\

& & 100 & \textbf{83.90}$_{\pm\text{1.88}}$ & 78.79$_{\pm\text{3.25}}$ & 77.37$_{\pm\text{3.85}}$ & 75.64$_{\pm\text{3.63}}$ & 80.43$_{\pm\text{3.70}}$ & 78.22$_{\pm\text{2.48}}$ & 49.60$_{\pm\text{17.93}}$ & 83.65$_{\pm\text{1.65}}$ & 83.56$_{\pm\text{1.81}}$ \\

& & 200 & 83.61$_{\pm\text{1.32}}$ & 79.31$_{\pm\text{2.69}}$ & 80.20$_{\pm\text{2.16}}$ & 76.37$_{\pm\text{1.92}}$ & 79.32$_{\pm\text{2.34}}$ & 78.89$_{\pm\text{2.42}}$ & 48.18$_{\pm\text{17.04}}$ & 83.69$_{\pm\text{1.41}}$ & \textbf{83.82}$_{\pm\text{1.34}}$ \\

\midrule

\multirow{9}{*}{\rotatebox{90}{\begin{tabular}{l} \textit{More than 10 classes} \end{tabular}}}& \multirow[c]{3}{*}{energy} & 50 & N/A & 7.51$_{\pm\text{1.98}}$ & 5.79$_{\pm\text{2.56}}$ & 5.89$_{\pm\text{1.99}}$ & 3.42$_{\pm\text{2.02}}$ & 8.72$_{\pm\text{1.70}}$ & 4.52$_{\pm\text{1.16}}$ & N/A & \textbf{21.56}$_{\pm\text{1.66}}$ \\

& & 100 & N/A & $-$ & 7.16$_{\pm\text{1.69}}$ & 6.15$_{\pm\text{1.09}}$ & 4.88$_{\pm\text{1.99}}$ & 10.67$_{\pm\text{2.27}}$ & 4.03$_{\pm\text{1.33}}$ & N/A & \textbf{27.59}$_{\pm\text{2.14}}$ \\

& & 200 & N/A & $-$ & 7.82$_{\pm\text{2.19}}$ & 6.63$_{\pm\text{2.11}}$ & 3.48$_{\pm\text{1.64}}$ & 12.96$_{\pm\text{2.22}}$ & 4.34$_{\pm\text{1.04}}$ & N/A & \textbf{33.01}$_{\pm\text{2.44}}$ \\

\cmidrule{2-12}

& \multirow[c]{2}{*}{collins} & 100 & N/A & $-$ & 5.33$_{\pm\text{0.98}}$ & 5.49$_{\pm\text{1.10}}$ & 12.30$_{\pm\text{2.14}}$ & 5.98$_{\pm\text{1.61}}$ & 3.61$_{\pm\text{1.35}}$ & N/A & \textbf{14.02}$_{\pm\text{2.45}}$ \\

& & 200 & 18.91$_{\pm\text{1.64}}$ & $-$ & 5.36$_{\pm\text{1.46}}$ & 5.13$_{\pm\text{0.92}}$ & 14.70$_{\pm\text{1.70}}$ & 6.04$_{\pm\text{0.86}}$ & 3.28$_{\pm\text{1.59}}$ & N/A & \textbf{19.11}$_{\pm\text{1.47}}$ \\

\cmidrule{2-12}

& \multirow[c]{4}{*}{texture} & 50 & 86.20$_{\pm\text{2.75}}$ & 25.72$_{\pm\text{6.89}}$ & 13.32$_{\pm\text{4.63}}$ & 13.86$_{\pm\text{4.99}}$ & 23.90$_{\pm\text{6.56}}$ & 18.17$_{\pm\text{8.82}}$ & 10.11$_{\pm\text{3.94}}$ & N/A & \textbf{88.46}$_{\pm\text{2.81}}$ \\

& & 100 & 93.17$_{\pm\text{2.01}}$ & 26.04$_{\pm\text{2.24}}$ & 12.34$_{\pm\text{5.07}}$ & 12.44$_{\pm\text{5.67}}$ & 27.67$_{\pm\text{6.90}}$ & 22.55$_{\pm\text{5.85}}$ & 12.00$_{\pm\text{5.29}}$ & N/A & \textbf{94.23}$_{\pm\text{1.30}}$ \\

& & 200 & 95.70$_{\pm\text{1.24}}$ & 16.65$_{\pm\text{9.20}}$ & 17.15$_{\pm\text{5.04}}$ & 13.14$_{\pm\text{6.40}}$ & 29.67$_{\pm\text{6.59}}$ & 43.54$_{\pm\text{7.07}}$ & 11.12$_{\pm\text{5.24}}$ & N/A & \textbf{95.99}$_{\pm\text{1.14}}$ \\

& & 500 & \textbf{97.17}$_{\pm\text{0.38}}$ & 39.69$_{\pm\text{10.43}}$ & 27.24$_{\pm\text{8.08}}$ & 11.68$_{\pm\text{4.59}}$ & 27.75$_{\pm\text{7.52}}$ & 60.15$_{\pm\text{5.45}}$ & 10.95$_{\pm\text{5.25}}$ & N/A & 96.53$_{\pm\text{0.54}}$ \\

\midrule
\rowcolor{Gainsboro!60}
\multicolumn{3}{l|}{\textbf{Average rank}} & 2.82$_{\pm\text{0.80}}$ & 4.70$_{\pm\text{0.78}}$ & 6.39$_{\pm\text{1.14}}$ & 7.48$_{\pm\text{0.76}}$ & 6.79$_{\pm\text{2.00}}$ & 4.64$_{\pm\text{1.37}}$ & 8.52$_{\pm\text{0.71}}$ & 2.24$_{\pm\text{1.10}}$ & \textbf{1.42}$_{\pm\text{0.61}}$ \\

\bottomrule

\end{tabular}
}
\end{table}
\clearpage

\begin{table}[p]
\centering
\caption{\textbf{Classification accuracy} (\%) of KNN, comparing data sharing on eight real-world tabular datasets with varied real data availability. We report the mean $\pm$ std balanced accuracy and average accuracy rank across datasets. A higher rank implies higher accuracy. Note that ``N/A'' denotes the inapplicability of a specific generator. Different from \cref{tab:test_acc_per_dataset_augmentation}, ``$-$'' denotes a generator cannot satisfy the requirement of generating 500 stratified samples even after generating 10,000 synthetic samples. The results of these inapplicable or failed generators are computed with the mean results of other methods. We \textbf{bold} the highest result for each dataset of different sample sizes. TVAE learns the joint distribution $p(\rvx, y)$ and fails to maintain the original training label distribution. TabEBM achieves the best overall performance against Baseline and benchmark generators.}
\resizebox{\textwidth}{!}{
\begin{tabular}{m{0.2cm}lr|rrrrrrrr|r}

\toprule

\multicolumn{2}{l}{Datasets}  & $N_{\text{real}}$ & SMOTE & TVAE & CTGAN & NFLOW & TabDDPM & ARF & GOGGLE & TabPFGen & \textbf{TabEBM} \\

\midrule

\multirow{24}{*}{\rotatebox{90}{\begin{tabular}{l} \textit{At most 10 classes} \end{tabular}}}& \multirow[c]{5}{*}{protein} & 20 & N/A & 16.84$_{\pm\text{3.19}}$ & 12.84$_{\pm\text{3.39}}$ & 11.26$_{\pm\text{3.06}}$ & 11.63$_{\pm\text{1.92}}$ & 16.53$_{\pm\text{3.52}}$ & 11.76$_{\pm\text{2.51}}$ & 35.75$_{\pm\text{4.48}}$ & \textbf{35.76}$_{\pm\text{4.39}}$ \\

& & 50 & \textbf{55.10}$_{\pm\text{3.65}}$ & 27.34$_{\pm\text{2.43}}$ & 15.55$_{\pm\text{4.09}}$ & 13.25$_{\pm\text{4.00}}$ & 12.03$_{\pm\text{3.72}}$ & 24.09$_{\pm\text{4.43}}$ & 12.53$_{\pm\text{2.50}}$ & 53.42$_{\pm\text{3.59}}$ & 53.44$_{\pm\text{3.32}}$ \\

& & 100 & \textbf{69.44}$_{\pm\text{2.83}}$ & 39.23$_{\pm\text{3.86}}$ & 22.23$_{\pm\text{2.89}}$ & 12.64$_{\pm\text{3.23}}$ & 16.71$_{\pm\text{3.38}}$ & 30.64$_{\pm\text{4.80}}$ & 13.66$_{\pm\text{2.29}}$ & 67.23$_{\pm\text{2.65}}$ & 67.35$_{\pm\text{2.69}}$ \\

& & 200 & \textbf{77.12}$_{\pm\text{2.73}}$ & 41.50$_{\pm\text{4.16}}$ & 27.68$_{\pm\text{5.17}}$ & 14.16$_{\pm\text{4.10}}$ & 20.29$_{\pm\text{3.05}}$ & 36.01$_{\pm\text{2.74}}$ & 11.91$_{\pm\text{4.20}}$ & 75.56$_{\pm\text{2.00}}$ & 76.17$_{\pm\text{2.04}}$ \\

& & 500 & 83.82$_{\pm\text{1.86}}$ & 50.56$_{\pm\text{5.32}}$ & 35.07$_{\pm\text{4.52}}$ & 13.44$_{\pm\text{2.78}}$ & 19.90$_{\pm\text{3.06}}$ & 46.40$_{\pm\text{3.68}}$ & 11.99$_{\pm\text{2.62}}$ & 83.63$_{\pm\text{1.63}}$ & \textbf{84.35}$_{\pm\text{1.33}}$ \\

\cmidrule{2-12}

& \multirow[c]{5}{*}{fourier} & 20 & N/A & $-$ & 19.60$_{\pm\text{4.25}}$ & 11.06$_{\pm\text{3.51}}$ & 9.16$_{\pm\text{1.73}}$ & 15.58$_{\pm\text{3.19}}$ & 9.78$_{\pm\text{2.29}}$ & 42.66$_{\pm\text{5.78}}$ & \textbf{42.78}$_{\pm\text{5.83}}$ \\

& & 50 & \textbf{60.16}$_{\pm\text{1.62}}$ & $-$ & N/A & 9.74$_{\pm\text{2.10}}$ & 11.66$_{\pm\text{2.39}}$ & 24.32$_{\pm\text{4.59}}$ & 12.86$_{\pm\text{2.44}}$ & 58.52$_{\pm\text{1.71}}$ & 58.56$_{\pm\text{1.99}}$ \\

& & 100 & \textbf{66.54}$_{\pm\text{2.75}}$ & 38.18$_{\pm\text{5.70}}$ & 19.86$_{\pm\text{3.82}}$ & 10.68$_{\pm\text{3.49}}$ & 18.42$_{\pm\text{3.40}}$ & 27.64$_{\pm\text{4.26}}$ & 9.48$_{\pm\text{3.68}}$ & 64.64$_{\pm\text{2.26}}$ & 64.98$_{\pm\text{2.58}}$ \\

& & 200 & \textbf{71.08}$_{\pm\text{2.07}}$ & 42.72$_{\pm\text{2.73}}$ & 31.68$_{\pm\text{4.44}}$ & 9.70$_{\pm\text{2.10}}$ & 26.96$_{\pm\text{2.98}}$ & 34.32$_{\pm\text{4.26}}$ & 9.76$_{\pm\text{2.16}}$ & 70.00$_{\pm\text{1.77}}$ & 70.50$_{\pm\text{1.86}}$ \\

& & 500 & \textbf{75.06}$_{\pm\text{1.95}}$ & 47.58$_{\pm\text{3.46}}$ & 42.68$_{\pm\text{3.18}}$ & 10.00$_{\pm\text{1.07}}$ & 26.18$_{\pm\text{3.18}}$ & 41.87$_{\pm\text{3.25}}$ & 10.88$_{\pm\text{1.70}}$ & 73.46$_{\pm\text{1.80}}$ & 73.95$_{\pm\text{1.45}}$ \\

\cmidrule{2-12}

& \multirow[c]{5}{*}{biodeg} & 20 & 69.11$_{\pm\text{3.28}}$ & 65.06$_{\pm\text{6.01}}$ & 54.27$_{\pm\text{6.17}}$ & 52.02$_{\pm\text{5.98}}$ & 47.87$_{\pm\text{4.28}}$ & 55.35$_{\pm\text{6.64}}$ & 47.41$_{\pm\text{12.04}}$ & 67.89$_{\pm\text{4.68}}$ & \textbf{69.62}$_{\pm\text{4.52}}$ \\

& & 50 & 72.84$_{\pm\text{2.33}}$ & 66.10$_{\pm\text{4.38}}$ & 61.89$_{\pm\text{7.87}}$ & 52.26$_{\pm\text{8.81}}$ & 50.78$_{\pm\text{5.21}}$ & 67.72$_{\pm\text{3.53}}$ & 53.48$_{\pm\text{10.48}}$ & 72.01$_{\pm\text{4.14}}$ & \textbf{73.77}$_{\pm\text{3.46}}$ \\

& & 100 & \textbf{75.82}$_{\pm\text{2.03}}$ & 69.86$_{\pm\text{3.93}}$ & 70.02$_{\pm\text{4.13}}$ & 58.95$_{\pm\text{5.48}}$ & 48.67$_{\pm\text{4.05}}$ & 69.39$_{\pm\text{2.95}}$ & 51.11$_{\pm\text{13.28}}$ & 74.77$_{\pm\text{2.03}}$ & 75.15$_{\pm\text{1.93}}$ \\

& & 200 & \textbf{79.67}$_{\pm\text{0.95}}$ & 70.25$_{\pm\text{5.58}}$ & 71.89$_{\pm\text{3.71}}$ & 53.54$_{\pm\text{7.72}}$ & 51.40$_{\pm\text{7.02}}$ & 70.21$_{\pm\text{3.87}}$ & 49.42$_{\pm\text{6.95}}$ & 77.78$_{\pm\text{2.42}}$ & 78.23$_{\pm\text{2.08}}$ \\

& & 500 & \textbf{82.51}$_{\pm\text{1.48}}$ & 72.72$_{\pm\text{3.89}}$ & 75.53$_{\pm\text{3.97}}$ & 52.92$_{\pm\text{5.87}}$ & 46.14$_{\pm\text{4.19}}$ & 73.57$_{\pm\text{3.14}}$ & 50.03$_{\pm\text{3.77}}$ & 80.55$_{\pm\text{1.56}}$ & 81.15$_{\pm\text{1.54}}$ \\

\cmidrule{2-12}

& \multirow[c]{5}{*}{steel} & 20 & 62.72$_{\pm\text{5.46}}$ & 56.75$_{\pm\text{7.62}}$ & 49.84$_{\pm\text{5.79}}$ & 49.97$_{\pm\text{6.59}}$ & 52.29$_{\pm\text{4.71}}$ & 52.67$_{\pm\text{5.56}}$ & 49.23$_{\pm\text{4.34}}$ & \textbf{70.71}$_{\pm\text{3.87}}$ & 69.36$_{\pm\text{4.02}}$ \\

& & 50 & 69.72$_{\pm\text{3.48}}$ & 60.43$_{\pm\text{5.12}}$ & 57.65$_{\pm\text{4.62}}$ & 54.41$_{\pm\text{4.04}}$ & 48.61$_{\pm\text{6.80}}$ & 57.24$_{\pm\text{5.82}}$ & 49.66$_{\pm\text{1.98}}$ & \textbf{82.03}$_{\pm\text{3.08}}$ & 80.57$_{\pm\text{3.35}}$ \\

& & 100 & 76.33$_{\pm\text{4.19}}$ & 63.34$_{\pm\text{2.92}}$ & 60.89$_{\pm\text{5.83}}$ & 52.24$_{\pm\text{4.96}}$ & 42.38$_{\pm\text{6.73}}$ & 58.01$_{\pm\text{6.58}}$ & 51.17$_{\pm\text{1.99}}$ & \textbf{87.79}$_{\pm\text{3.06}}$ & 87.76$_{\pm\text{3.25}}$ \\

& & 200 & 79.90$_{\pm\text{1.97}}$ & 68.00$_{\pm\text{2.85}}$ & 66.31$_{\pm\text{4.27}}$ & 50.19$_{\pm\text{4.02}}$ & 43.82$_{\pm\text{7.44}}$ & 63.29$_{\pm\text{4.68}}$ & 50.99$_{\pm\text{2.53}}$ & 90.90$_{\pm\text{1.80}}$ & \textbf{91.07}$_{\pm\text{1.75}}$ \\

& & 500 & 85.06$_{\pm\text{1.90}}$ & 73.24$_{\pm\text{3.16}}$ & 72.49$_{\pm\text{2.82}}$ & 49.19$_{\pm\text{4.79}}$ & 45.77$_{\pm\text{8.19}}$ & 68.81$_{\pm\text{4.42}}$ & 52.45$_{\pm\text{3.59}}$ & 92.81$_{\pm\text{1.15}}$ & \textbf{92.83}$_{\pm\text{1.35}}$ \\

\cmidrule{2-12}

& \multirow[c]{4}{*}{stock} & 20 & 84.27$_{\pm\text{5.28}}$ & 66.05$_{\pm\text{9.50}}$ & 52.15$_{\pm\text{10.98}}$ & 58.69$_{\pm\text{15.22}}$ & 77.28$_{\pm\text{18.40}}$ & 58.03$_{\pm\text{12.41}}$ & 45.13$_{\pm\text{15.32}}$ & 84.45$_{\pm\text{4.02}}$ & \textbf{84.69}$_{\pm\text{4.16}}$ \\

& & 50 & 89.64$_{\pm\text{2.35}}$ & 75.40$_{\pm\text{3.27}}$ & 67.64$_{\pm\text{6.93}}$ & 64.81$_{\pm\text{9.27}}$ & 88.94$_{\pm\text{1.62}}$ & 77.67$_{\pm\text{4.49}}$ & 48.55$_{\pm\text{11.95}}$ & 89.69$_{\pm\text{1.89}}$ & \textbf{89.70}$_{\pm\text{1.90}}$ \\

& & 100 & 91.93$_{\pm\text{0.77}}$ & 80.87$_{\pm\text{2.69}}$ & 78.37$_{\pm\text{3.93}}$ & 67.33$_{\pm\text{6.34}}$ & 91.03$_{\pm\text{1.15}}$ & 84.53$_{\pm\text{3.34}}$ & 53.19$_{\pm\text{11.60}}$ & 91.98$_{\pm\text{0.63}}$ & \textbf{92.35}$_{\pm\text{0.75}}$ \\

& & 200 & \textbf{93.46}$_{\pm\text{0.93}}$ & 84.65$_{\pm\text{1.61}}$ & 81.64$_{\pm\text{3.69}}$ & 70.56$_{\pm\text{3.41}}$ & 91.81$_{\pm\text{0.69}}$ & 85.58$_{\pm\text{2.64}}$ & 49.98$_{\pm\text{11.39}}$ & 92.53$_{\pm\text{1.09}}$ & 92.94$_{\pm\text{0.99}}$ \\

\midrule

\multirow{9}{*}{\rotatebox{90}{\begin{tabular}{l} \textit{More than 10 classes} \end{tabular}}}& \multirow[c]{3}{*}{energy} & 50 & N/A & 6.75$_{\pm\text{1.25}}$ & 4.88$_{\pm\text{1.00}}$ & 5.21$_{\pm\text{1.39}}$ & 4.71$_{\pm\text{1.44}}$ & 6.06$_{\pm\text{1.55}}$ & 4.63$_{\pm\text{0.47}}$ & N/A & \textbf{25.36}$_{\pm\text{2.29}}$ \\

& & 100 & N/A & $-$ & 7.53$_{\pm\text{1.81}}$ & 5.33$_{\pm\text{1.43}}$ & 3.68$_{\pm\text{1.03}}$ & 9.05$_{\pm\text{2.14}}$ & 4.37$_{\pm\text{1.22}}$ & N/A & \textbf{29.63}$_{\pm\text{2.43}}$ \\

& & 200 & N/A & $-$ & 7.22$_{\pm\text{0.99}}$ & 6.01$_{\pm\text{1.31}}$ & 3.30$_{\pm\text{0.91}}$ & 10.56$_{\pm\text{1.57}}$ & 3.99$_{\pm\text{0.73}}$ & N/A & \textbf{33.85}$_{\pm\text{3.11}}$ \\

\cmidrule{2-12}

& \multirow[c]{2}{*}{collins} & 100 & N/A & $-$ & 5.61$_{\pm\text{0.95}}$ & 4.32$_{\pm\text{0.72}}$ & 13.46$_{\pm\text{1.68}}$ & 4.71$_{\pm\text{1.10}}$ & 3.80$_{\pm\text{0.09}}$ & N/A & \textbf{15.20}$_{\pm\text{2.00}}$ \\

& & 200 & \textbf{19.94}$_{\pm\text{2.15}}$ & $-$ & 5.24$_{\pm\text{0.54}}$ & 4.49$_{\pm\text{0.68}}$ & 15.29$_{\pm\text{1.82}}$ & 4.58$_{\pm\text{0.70}}$ & 3.89$_{\pm\text{0.28}}$ & N/A & 17.66$_{\pm\text{1.80}}$ \\

\cmidrule{2-12}

& \multirow[c]{4}{*}{texture} & 50 & \textbf{78.67}$_{\pm\text{2.72}}$ & 22.15$_{\pm\text{10.01}}$ & 12.36$_{\pm\text{3.58}}$ & 11.19$_{\pm\text{3.12}}$ & 15.85$_{\pm\text{5.14}}$ & 13.67$_{\pm\text{6.44}}$ & 9.05$_{\pm\text{0.13}}$ & N/A & 75.57$_{\pm\text{2.61}}$ \\

& & 100 & \textbf{85.31}$_{\pm\text{2.44}}$ & 25.80$_{\pm\text{5.61}}$ & 13.82$_{\pm\text{7.03}}$ & 11.67$_{\pm\text{4.14}}$ & 19.32$_{\pm\text{4.92}}$ & 25.54$_{\pm\text{5.86}}$ & 9.09$_{\pm\text{0.10}}$ & N/A & 84.69$_{\pm\text{1.70}}$ \\

& & 200 & 88.17$_{\pm\text{1.71}}$ & 18.33$_{\pm\text{7.83}}$ & 13.85$_{\pm\text{3.95}}$ & 10.55$_{\pm\text{4.55}}$ & 23.17$_{\pm\text{7.57}}$ & 42.92$_{\pm\text{6.67}}$ & 9.07$_{\pm\text{0.56}}$ & N/A & \textbf{89.16}$_{\pm\text{1.75}}$ \\

& & 500 & 90.78$_{\pm\text{1.16}}$ & 29.29$_{\pm\text{7.03}}$ & 24.91$_{\pm\text{9.99}}$ & 10.23$_{\pm\text{3.48}}$ & 23.85$_{\pm\text{9.99}}$ & 57.58$_{\pm\text{4.98}}$ & 10.12$_{\pm\text{2.55}}$ & N/A & \textbf{91.23}$_{\pm\text{0.95}}$ \\

\midrule
\rowcolor{Gainsboro!60}
\multicolumn{3}{l|}{\textbf{Average rank}} & 2.03$_{\pm\text{0.99}}$ & 4.39$_{\pm\text{0.84}}$ & 6.03$_{\pm\text{1.07}}$ & 7.70$_{\pm\text{0.81}}$ & 6.97$_{\pm\text{2.05}}$ & 5.30$_{\pm\text{0.92}}$ & 8.33$_{\pm\text{0.82}}$ & 2.73$_{\pm\text{0.83}}$ & \textbf{1.52}$_{\pm\text{0.51}}$ \\

\bottomrule

\end{tabular}
}
\end{table}
\clearpage

\begin{table}[p]
\centering
\caption{\textbf{Classification accuracy} (\%) of MLP, comparing data sharing on eight real-world tabular datasets with varied real data availability. We report the mean $\pm$ std balanced accuracy and average accuracy rank across datasets. A higher rank implies higher accuracy. Note that ``N/A'' denotes the inapplicability of a specific generator. Different from \cref{tab:test_acc_per_dataset_augmentation}, ``$-$'' denotes a generator cannot satisfy the requirement of generating 500 stratified samples even after generating 10,000 synthetic samples. The results of these inapplicable or failed generators are computed with the mean results of other methods. We \textbf{bold} the highest result for each dataset of different sample sizes. TVAE learns the joint distribution $p(\rvx, y)$ and fails to maintain the original training label distribution. TabEBM achieves the best overall performance against Baseline and benchmark generators.}
\resizebox{\textwidth}{!}{
\begin{tabular}{m{0.2cm}lr|rrrrrrrr|r}

\toprule

\multicolumn{2}{l}{Datasets}  & $N_{\text{real}}$ & SMOTE & TVAE & CTGAN & NFLOW & TabDDPM & ARF & GOGGLE & TabPFGen & \textbf{TabEBM} \\

\midrule

\multirow{24}{*}{\rotatebox{90}{\begin{tabular}{l} \textit{At most 10 classes} \end{tabular}}}& \multirow[c]{5}{*}{protein} & 20 & N/A & 15.81$_{\pm\text{3.26}}$ & 12.08$_{\pm\text{4.27}}$ & 13.06$_{\pm\text{5.22}}$ & 12.93$_{\pm\text{3.80}}$ & 20.17$_{\pm\text{4.44}}$ & 12.98$_{\pm\text{2.50}}$ & 36.01$_{\pm\text{2.68}}$ & \textbf{36.23}$_{\pm\text{2.49}}$ \\

& & 50 & 56.85$_{\pm\text{4.56}}$ & 26.83$_{\pm\text{4.86}}$ & 14.19$_{\pm\text{4.81}}$ & 13.68$_{\pm\text{1.80}}$ & 14.92$_{\pm\text{3.38}}$ & 31.36$_{\pm\text{4.84}}$ & 13.78$_{\pm\text{3.91}}$ & 58.69$_{\pm\text{4.22}}$ & \textbf{58.83}$_{\pm\text{4.59}}$ \\

& & 100 & 75.86$_{\pm\text{3.03}}$ & 41.60$_{\pm\text{4.43}}$ & 25.78$_{\pm\text{4.77}}$ & 11.34$_{\pm\text{3.38}}$ & 20.82$_{\pm\text{4.55}}$ & 40.31$_{\pm\text{5.63}}$ & 14.54$_{\pm\text{3.04}}$ & 77.47$_{\pm\text{3.39}}$ & \textbf{77.56}$_{\pm\text{3.65}}$ \\

& & 200 & 87.85$_{\pm\text{1.99}}$ & 50.87$_{\pm\text{6.26}}$ & 32.49$_{\pm\text{6.76}}$ & 12.90$_{\pm\text{3.77}}$ & 23.85$_{\pm\text{5.06}}$ & 45.40$_{\pm\text{5.31}}$ & 10.24$_{\pm\text{3.54}}$ & \textbf{90.01}$_{\pm\text{2.00}}$ & 89.48$_{\pm\text{2.00}}$ \\

& & 500 & 94.37$_{\pm\text{1.66}}$ & 62.81$_{\pm\text{4.29}}$ & 47.22$_{\pm\text{6.43}}$ & 12.06$_{\pm\text{2.49}}$ & 24.51$_{\pm\text{7.67}}$ & 60.31$_{\pm\text{4.02}}$ & 10.41$_{\pm\text{2.60}}$ & \textbf{96.26}$_{\pm\text{1.35}}$ & 96.03$_{\pm\text{1.56}}$ \\

\cmidrule{2-12}

& \multirow[c]{5}{*}{fourier} & 20 & N/A & $-$ & 12.16$_{\pm\text{2.77}}$ & 10.56$_{\pm\text{3.94}}$ & 10.38$_{\pm\text{2.81}}$ & 19.66$_{\pm\text{4.97}}$ & 11.04$_{\pm\text{3.61}}$ & 34.40$_{\pm\text{3.85}}$ & \textbf{35.04}$_{\pm\text{3.63}}$ \\

& & 50 & 52.72$_{\pm\text{2.11}}$ & $-$ & N/A & 9.42$_{\pm\text{3.23}}$ & 13.78$_{\pm\text{3.07}}$ & 33.94$_{\pm\text{2.46}}$ & 10.88$_{\pm\text{3.14}}$ & 55.12$_{\pm\text{1.81}}$ & \textbf{55.30}$_{\pm\text{1.65}}$ \\

& & 100 & 60.82$_{\pm\text{2.85}}$ & 33.16$_{\pm\text{3.91}}$ & 21.44$_{\pm\text{3.15}}$ & 10.88$_{\pm\text{4.21}}$ & 21.96$_{\pm\text{2.34}}$ & 39.68$_{\pm\text{4.04}}$ & 8.12$_{\pm\text{3.25}}$ & 63.54$_{\pm\text{1.81}}$ & \textbf{64.08}$_{\pm\text{1.87}}$ \\

& & 200 & 68.64$_{\pm\text{2.78}}$ & 42.18$_{\pm\text{3.95}}$ & 30.56$_{\pm\text{5.91}}$ & 10.94$_{\pm\text{4.87}}$ & 33.80$_{\pm\text{5.10}}$ & 43.92$_{\pm\text{2.17}}$ & 8.46$_{\pm\text{2.68}}$ & \textbf{71.76}$_{\pm\text{1.62}}$ & 71.36$_{\pm\text{1.94}}$ \\

& & 500 & 73.78$_{\pm\text{1.12}}$ & 50.48$_{\pm\text{4.03}}$ & 45.94$_{\pm\text{3.55}}$ & 13.10$_{\pm\text{4.09}}$ & 35.18$_{\pm\text{5.84}}$ & 51.10$_{\pm\text{3.53}}$ & 9.66$_{\pm\text{2.62}}$ & 76.50$_{\pm\text{1.87}}$ & \textbf{77.65}$_{\pm\text{1.81}}$ \\

\cmidrule{2-12}

& \multirow[c]{5}{*}{biodeg} & 20 & 68.59$_{\pm\text{4.24}}$ & 62.24$_{\pm\text{8.67}}$ & 53.56$_{\pm\text{5.65}}$ & 55.15$_{\pm\text{9.09}}$ & 45.87$_{\pm\text{3.29}}$ & 57.76$_{\pm\text{4.42}}$ & 45.98$_{\pm\text{11.23}}$ & 72.04$_{\pm\text{4.95}}$ & \textbf{72.10}$_{\pm\text{4.69}}$ \\

& & 50 & 73.48$_{\pm\text{2.67}}$ & 65.96$_{\pm\text{6.24}}$ & 62.15$_{\pm\text{7.11}}$ & 55.84$_{\pm\text{11.19}}$ & 45.55$_{\pm\text{8.63}}$ & 72.40$_{\pm\text{3.40}}$ & 49.95$_{\pm\text{6.48}}$ & 77.20$_{\pm\text{2.84}}$ & \textbf{77.24}$_{\pm\text{3.18}}$ \\

& & 100 & 77.20$_{\pm\text{1.72}}$ & 70.20$_{\pm\text{5.67}}$ & 71.90$_{\pm\text{3.27}}$ & 57.99$_{\pm\text{6.21}}$ & 49.10$_{\pm\text{5.76}}$ & 71.42$_{\pm\text{2.59}}$ & 45.52$_{\pm\text{6.61}}$ & 78.58$_{\pm\text{2.25}}$ & \textbf{78.93}$_{\pm\text{2.11}}$ \\

& & 200 & 81.30$_{\pm\text{0.91}}$ & 72.26$_{\pm\text{5.86}}$ & 71.59$_{\pm\text{4.85}}$ & 55.75$_{\pm\text{7.93}}$ & 44.37$_{\pm\text{7.27}}$ & 73.82$_{\pm\text{4.38}}$ & 48.45$_{\pm\text{8.83}}$ & 82.06$_{\pm\text{1.67}}$ & \textbf{82.25}$_{\pm\text{1.69}}$ \\

& & 500 & 83.55$_{\pm\text{0.62}}$ & 74.36$_{\pm\text{4.32}}$ & 77.55$_{\pm\text{2.54}}$ & 53.37$_{\pm\text{4.84}}$ & 43.97$_{\pm\text{4.60}}$ & 76.51$_{\pm\text{2.60}}$ & 47.04$_{\pm\text{9.41}}$ & \textbf{83.63}$_{\pm\text{1.19}}$ & 83.33$_{\pm\text{0.86}}$ \\

\cmidrule{2-12}

& \multirow[c]{5}{*}{steel} & 20 & 57.49$_{\pm\text{5.60}}$ & 53.28$_{\pm\text{11.47}}$ & 50.57$_{\pm\text{7.01}}$ & 52.01$_{\pm\text{8.53}}$ & 52.76$_{\pm\text{9.11}}$ & 51.25$_{\pm\text{4.91}}$ & 47.98$_{\pm\text{5.99}}$ & \textbf{64.34}$_{\pm\text{6.21}}$ & 64.33$_{\pm\text{5.70}}$ \\

& & 50 & 66.70$_{\pm\text{3.60}}$ & 62.02$_{\pm\text{5.08}}$ & 57.41$_{\pm\text{5.18}}$ & 54.51$_{\pm\text{6.50}}$ & 50.58$_{\pm\text{7.20}}$ & 59.85$_{\pm\text{5.16}}$ & 50.78$_{\pm\text{6.19}}$ & \textbf{83.06}$_{\pm\text{6.05}}$ & 82.38$_{\pm\text{5.79}}$ \\

& & 100 & 76.17$_{\pm\text{4.30}}$ & 63.05$_{\pm\text{4.11}}$ & 60.62$_{\pm\text{4.71}}$ & 54.81$_{\pm\text{5.06}}$ & 44.64$_{\pm\text{9.29}}$ & 59.60$_{\pm\text{4.83}}$ & 48.93$_{\pm\text{2.99}}$ & 95.36$_{\pm\text{3.13}}$ & \textbf{95.45}$_{\pm\text{3.29}}$ \\

& & 200 & 80.23$_{\pm\text{2.17}}$ & 68.45$_{\pm\text{3.46}}$ & 64.73$_{\pm\text{5.71}}$ & 50.83$_{\pm\text{4.76}}$ & 39.73$_{\pm\text{12.23}}$ & 62.36$_{\pm\text{3.96}}$ & 47.89$_{\pm\text{7.71}}$ & 98.34$_{\pm\text{1.12}}$ & \textbf{98.37}$_{\pm\text{0.81}}$ \\

& & 500 & 85.76$_{\pm\text{3.61}}$ & 73.89$_{\pm\text{4.48}}$ & 71.48$_{\pm\text{4.21}}$ & 49.32$_{\pm\text{6.51}}$ & 43.65$_{\pm\text{5.66}}$ & 68.12$_{\pm\text{4.63}}$ & 53.56$_{\pm\text{4.70}}$ & \textbf{99.56}$_{\pm\text{0.34}}$ & 99.45$_{\pm\text{0.44}}$ \\

\cmidrule{2-12}

& \multirow[c]{4}{*}{stock} & 20 & 82.93$_{\pm\text{4.66}}$ & 71.78$_{\pm\text{7.33}}$ & 55.09$_{\pm\text{13.94}}$ & 68.46$_{\pm\text{8.87}}$ & 76.71$_{\pm\text{15.45}}$ & 62.13$_{\pm\text{10.43}}$ & 52.68$_{\pm\text{12.66}}$ & 83.83$_{\pm\text{3.86}}$ & \textbf{83.91}$_{\pm\text{3.99}}$ \\

& & 50 & 89.48$_{\pm\text{2.12}}$ & 75.93$_{\pm\text{3.72}}$ & 67.04$_{\pm\text{4.78}}$ & 73.18$_{\pm\text{9.70}}$ & 87.47$_{\pm\text{2.35}}$ & 78.81$_{\pm\text{3.68}}$ & 48.79$_{\pm\text{5.17}}$ & 90.25$_{\pm\text{1.96}}$ & \textbf{90.36}$_{\pm\text{2.20}}$ \\

& & 100 & 90.85$_{\pm\text{0.72}}$ & 81.61$_{\pm\text{3.37}}$ & 79.76$_{\pm\text{3.51}}$ & 71.03$_{\pm\text{5.88}}$ & 90.17$_{\pm\text{1.66}}$ & 85.83$_{\pm\text{2.70}}$ & 50.41$_{\pm\text{14.72}}$ & 91.60$_{\pm\text{1.00}}$ & \textbf{91.72}$_{\pm\text{0.97}}$ \\

& & 200 & 92.03$_{\pm\text{0.79}}$ & 84.67$_{\pm\text{2.45}}$ & 82.92$_{\pm\text{2.01}}$ & 74.97$_{\pm\text{4.41}}$ & 91.21$_{\pm\text{1.00}}$ & 87.06$_{\pm\text{2.18}}$ & 44.90$_{\pm\text{17.08}}$ & \textbf{92.24}$_{\pm\text{0.91}}$ & 92.00$_{\pm\text{0.93}}$ \\

\midrule

\multirow{9}{*}{\rotatebox{90}{\begin{tabular}{l} \textit{More than 10 classes} \end{tabular}}}& \multirow[c]{3}{*}{energy} & 50 & N/A & 7.77$_{\pm\text{1.39}}$ & 5.64$_{\pm\text{2.40}}$ & 5.75$_{\pm\text{1.72}}$ & 4.30$_{\pm\text{2.07}}$ & 8.28$_{\pm\text{1.91}}$ & 4.46$_{\pm\text{0.25}}$ & N/A & \textbf{23.91}$_{\pm\text{1.49}}$ \\

& & 100 & N/A & $-$ & 7.44$_{\pm\text{2.38}}$ & 5.34$_{\pm\text{1.60}}$ & 3.68$_{\pm\text{1.68}}$ & 10.38$_{\pm\text{1.67}}$ & 4.29$_{\pm\text{0.71}}$ & N/A & \textbf{29.24}$_{\pm\text{2.45}}$ \\

& & 200 & N/A & $-$ & 8.74$_{\pm\text{2.17}}$ & 7.38$_{\pm\text{2.33}}$ & 3.84$_{\pm\text{1.29}}$ & 12.88$_{\pm\text{2.09}}$ & 4.08$_{\pm\text{1.12}}$ & N/A & \textbf{38.27}$_{\pm\text{3.50}}$ \\

\cmidrule{2-12}

& \multirow[c]{2}{*}{collins} & 100 & N/A & $-$ & 6.06$_{\pm\text{0.96}}$ & 5.04$_{\pm\text{1.41}}$ & 12.80$_{\pm\text{1.48}}$ & 5.69$_{\pm\text{1.80}}$ & 3.67$_{\pm\text{0.43}}$ & N/A & \textbf{13.66}$_{\pm\text{1.86}}$ \\

& & 200 & 18.91$_{\pm\text{1.62}}$ & $-$ & 5.78$_{\pm\text{1.75}}$ & 4.90$_{\pm\text{0.94}}$ & 15.94$_{\pm\text{1.74}}$ & 5.84$_{\pm\text{1.19}}$ & 4.42$_{\pm\text{1.18}}$ & N/A & \textbf{19.49}$_{\pm\text{1.56}}$ \\

\cmidrule{2-12}

& \multirow[c]{4}{*}{texture} & 50 & 83.45$_{\pm\text{3.05}}$ & 28.11$_{\pm\text{7.00}}$ & 14.57$_{\pm\text{6.49}}$ & 12.56$_{\pm\text{2.92}}$ & 22.50$_{\pm\text{6.71}}$ & 18.07$_{\pm\text{4.88}}$ & 9.88$_{\pm\text{4.02}}$ & N/A & \textbf{85.41}$_{\pm\text{2.73}}$ \\

& & 100 & 90.73$_{\pm\text{1.80}}$ & 33.20$_{\pm\text{4.80}}$ & 15.18$_{\pm\text{7.46}}$ & 13.72$_{\pm\text{6.03}}$ & 29.36$_{\pm\text{8.98}}$ & 25.80$_{\pm\text{5.24}}$ & 10.77$_{\pm\text{5.02}}$ & N/A & \textbf{91.87}$_{\pm\text{1.42}}$ \\

& & 200 & 93.31$_{\pm\text{1.19}}$ & 26.95$_{\pm\text{15.32}}$ & 18.10$_{\pm\text{5.37}}$ & 10.39$_{\pm\text{4.68}}$ & 28.79$_{\pm\text{10.29}}$ & 44.44$_{\pm\text{5.46}}$ & 7.70$_{\pm\text{2.74}}$ & N/A & \textbf{93.78}$_{\pm\text{1.41}}$ \\

& & 500 & \textbf{95.61}$_{\pm\text{1.10}}$ & 39.39$_{\pm\text{9.19}}$ & 29.60$_{\pm\text{8.79}}$ & 13.15$_{\pm\text{4.59}}$ & 27.70$_{\pm\text{6.21}}$ & 62.88$_{\pm\text{5.56}}$ & 10.30$_{\pm\text{3.10}}$ & N/A & 95.06$_{\pm\text{0.96}}$ \\

\midrule
\rowcolor{Gainsboro!60}
\multicolumn{3}{l|}{\textbf{Average rank}} & 2.91$_{\pm\text{0.71}}$ & 4.73$_{\pm\text{0.75}}$ & 6.45$_{\pm\text{1.12}}$ & 7.48$_{\pm\text{0.87}}$ & 6.79$_{\pm\text{2.13}}$ & 4.70$_{\pm\text{1.33}}$ & 8.39$_{\pm\text{0.70}}$ & 2.21$_{\pm\text{1.05}}$ & \textbf{1.33}$_{\pm\text{0.60}}$ \\

\bottomrule

\end{tabular}
}
\end{table}
\clearpage

\begin{table}[p]
\centering
\caption{\textbf{Classification accuracy} (\%) of RF, comparing data sharing on eight real-world tabular datasets with varied real data availability. We report the mean $\pm$ std balanced accuracy and average accuracy rank across datasets. A higher rank implies higher accuracy. Note that ``N/A'' denotes the inapplicability of a specific generator. Different from \cref{tab:test_acc_per_dataset_augmentation}, ``$-$'' denotes a generator cannot satisfy the requirement of generating 500 stratified samples even after generating 10,000 synthetic samples. The results of these inapplicable or failed generators are computed with the mean results of other methods. We \textbf{bold} the highest result for each dataset of different sample sizes. TVAE learns the joint distribution $p(\rvx, y)$ and fails to maintain the original training label distribution. TabEBM achieves the best overall performance against Baseline and benchmark generators.}
\resizebox{\textwidth}{!}{
\begin{tabular}{m{0.2cm}lr|rrrrrrrr|r}

\toprule

\multicolumn{2}{l}{Datasets}  & $N_{\text{real}}$ & SMOTE & TVAE & CTGAN & NFLOW & TabDDPM & ARF & GOGGLE & TabPFGen & \textbf{TabEBM} \\

\midrule

\multirow{24}{*}{\rotatebox{90}{\begin{tabular}{l} \textit{At most 10 classes} \end{tabular}}}& \multirow[c]{5}{*}{protein} & 20 & N/A & 19.29$_{\pm\text{3.85}}$ & 13.63$_{\pm\text{2.95}}$ & 12.71$_{\pm\text{2.84}}$ & 12.75$_{\pm\text{2.09}}$ & 20.47$_{\pm\text{4.29}}$ & 13.25$_{\pm\text{2.90}}$ & 31.86$_{\pm\text{2.24}}$ & \textbf{34.05}$_{\pm\text{2.24}}$ \\

& & 50 & 56.09$_{\pm\text{2.77}}$ & 32.29$_{\pm\text{3.05}}$ & 19.37$_{\pm\text{5.89}}$ & 12.62$_{\pm\text{2.05}}$ & 14.87$_{\pm\text{2.75}}$ & 32.60$_{\pm\text{4.15}}$ & 13.29$_{\pm\text{2.50}}$ & 54.25$_{\pm\text{1.99}}$ & \textbf{56.96}$_{\pm\text{3.32}}$ \\

& & 100 & 71.14$_{\pm\text{2.65}}$ & 44.50$_{\pm\text{3.08}}$ & 29.72$_{\pm\text{3.69}}$ & 12.37$_{\pm\text{3.12}}$ & 17.71$_{\pm\text{3.51}}$ & 38.22$_{\pm\text{4.85}}$ & 13.67$_{\pm\text{1.00}}$ & 70.68$_{\pm\text{3.24}}$ & \textbf{72.74}$_{\pm\text{2.97}}$ \\

& & 200 & 81.22$_{\pm\text{2.51}}$ & 50.10$_{\pm\text{3.89}}$ & 37.08$_{\pm\text{5.17}}$ & 14.34$_{\pm\text{3.15}}$ & 22.23$_{\pm\text{3.43}}$ & 43.48$_{\pm\text{3.74}}$ & 11.76$_{\pm\text{3.27}}$ & 81.93$_{\pm\text{2.47}}$ & \textbf{82.35}$_{\pm\text{2.09}}$ \\

& & 500 & 87.93$_{\pm\text{1.52}}$ & 58.17$_{\pm\text{5.02}}$ & 49.44$_{\pm\text{5.57}}$ & 12.03$_{\pm\text{2.40}}$ & 22.88$_{\pm\text{2.67}}$ & 53.67$_{\pm\text{2.94}}$ & 10.59$_{\pm\text{3.14}}$ & \textbf{90.81}$_{\pm\text{1.37}}$ & 89.78$_{\pm\text{1.73}}$ \\

\cmidrule{2-12}

& \multirow[c]{5}{*}{fourier} & 20 & N/A & $-$ & 11.28$_{\pm\text{1.63}}$ & 11.52$_{\pm\text{2.65}}$ & 9.24$_{\pm\text{2.16}}$ & 17.76$_{\pm\text{3.96}}$ & 9.78$_{\pm\text{1.33}}$ & 31.94$_{\pm\text{3.25}}$ & \textbf{38.16}$_{\pm\text{4.48}}$ \\

& & 50 & 65.32$_{\pm\text{3.86}}$ & $-$ & N/A & 9.98$_{\pm\text{1.98}}$ & 12.12$_{\pm\text{2.56}}$ & 31.76$_{\pm\text{3.66}}$ & 11.24$_{\pm\text{2.70}}$ & 65.64$_{\pm\text{3.73}}$ & \textbf{66.04}$_{\pm\text{2.91}}$ \\

& & 100 & 73.82$_{\pm\text{2.70}}$ & 44.84$_{\pm\text{5.88}}$ & 29.68$_{\pm\text{5.97}}$ & 11.04$_{\pm\text{2.51}}$ & 21.90$_{\pm\text{4.68}}$ & 35.60$_{\pm\text{4.77}}$ & 9.30$_{\pm\text{2.87}}$ & 74.28$_{\pm\text{2.81}}$ & \textbf{74.92}$_{\pm\text{3.12}}$ \\

& & 200 & 78.66$_{\pm\text{1.64}}$ & 48.56$_{\pm\text{3.52}}$ & 41.94$_{\pm\text{5.54}}$ & 10.24$_{\pm\text{2.93}}$ & 31.36$_{\pm\text{4.10}}$ & 44.80$_{\pm\text{3.88}}$ & 9.88$_{\pm\text{2.32}}$ & 79.48$_{\pm\text{2.18}}$ & \textbf{79.52}$_{\pm\text{2.03}}$ \\

& & 500 & 79.64$_{\pm\text{1.39}}$ & 54.80$_{\pm\text{4.51}}$ & 57.06$_{\pm\text{4.59}}$ & 11.94$_{\pm\text{2.26}}$ & 25.08$_{\pm\text{4.65}}$ & 54.04$_{\pm\text{3.10}}$ & 10.66$_{\pm\text{2.71}}$ & 81.64$_{\pm\text{1.61}}$ & \textbf{81.85}$_{\pm\text{0.53}}$ \\

\cmidrule{2-12}

& \multirow[c]{5}{*}{biodeg} & 20 & \textbf{69.34}$_{\pm\text{6.13}}$ & 64.52$_{\pm\text{7.42}}$ & 52.83$_{\pm\text{7.85}}$ & 53.88$_{\pm\text{7.31}}$ & 49.80$_{\pm\text{0.45}}$ & 60.95$_{\pm\text{6.69}}$ & 50.47$_{\pm\text{9.67}}$ & 67.37$_{\pm\text{5.45}}$ & 67.06$_{\pm\text{4.88}}$ \\

& & 50 & 70.17$_{\pm\text{3.69}}$ & 68.14$_{\pm\text{3.98}}$ & 62.50$_{\pm\text{6.74}}$ & 54.84$_{\pm\text{7.67}}$ & 49.94$_{\pm\text{1.03}}$ & 67.21$_{\pm\text{4.25}}$ & 51.99$_{\pm\text{6.12}}$ & \textbf{71.94}$_{\pm\text{4.16}}$ & 71.91$_{\pm\text{2.77}}$ \\

& & 100 & 74.80$_{\pm\text{2.96}}$ & 71.82$_{\pm\text{3.70}}$ & 68.29$_{\pm\text{2.67}}$ & 56.44$_{\pm\text{5.66}}$ & 49.62$_{\pm\text{0.68}}$ & 67.68$_{\pm\text{2.88}}$ & 50.71$_{\pm\text{5.41}}$ & \textbf{75.68}$_{\pm\text{2.01}}$ & 74.93$_{\pm\text{2.08}}$ \\

& & 200 & 77.49$_{\pm\text{2.13}}$ & 69.22$_{\pm\text{3.80}}$ & 69.63$_{\pm\text{2.86}}$ & 55.47$_{\pm\text{7.86}}$ & 48.70$_{\pm\text{1.53}}$ & 70.40$_{\pm\text{3.01}}$ & 48.04$_{\pm\text{5.01}}$ & \textbf{78.64}$_{\pm\text{2.29}}$ & 76.96$_{\pm\text{3.02}}$ \\

& & 500 & \textbf{80.31}$_{\pm\text{1.32}}$ & 74.83$_{\pm\text{1.76}}$ & 74.44$_{\pm\text{2.90}}$ & 51.30$_{\pm\text{3.00}}$ & 49.68$_{\pm\text{0.61}}$ & 73.88$_{\pm\text{1.59}}$ & 45.73$_{\pm\text{8.13}}$ & 79.56$_{\pm\text{1.95}}$ & 78.66$_{\pm\text{1.76}}$ \\

\cmidrule{2-12}

& \multirow[c]{5}{*}{steel} & 20 & 55.71$_{\pm\text{3.47}}$ & 55.23$_{\pm\text{4.31}}$ & 52.14$_{\pm\text{3.64}}$ & 50.93$_{\pm\text{6.29}}$ & 49.93$_{\pm\text{0.56}}$ & 52.02$_{\pm\text{5.80}}$ & 48.88$_{\pm\text{4.70}}$ & 57.20$_{\pm\text{2.92}}$ & \textbf{58.27}$_{\pm\text{2.40}}$ \\

& & 50 & 62.30$_{\pm\text{2.94}}$ & 56.05$_{\pm\text{3.07}}$ & 55.40$_{\pm\text{4.01}}$ & 52.16$_{\pm\text{3.88}}$ & 49.98$_{\pm\text{0.29}}$ & 54.55$_{\pm\text{2.89}}$ & 50.65$_{\pm\text{1.92}}$ & 64.26$_{\pm\text{2.55}}$ & \textbf{66.98}$_{\pm\text{3.30}}$ \\

& & 100 & 67.63$_{\pm\text{3.75}}$ & 58.93$_{\pm\text{4.82}}$ & 56.65$_{\pm\text{4.04}}$ & 51.86$_{\pm\text{2.52}}$ & 47.84$_{\pm\text{2.76}}$ & 56.13$_{\pm\text{2.99}}$ & 49.33$_{\pm\text{2.96}}$ & 72.47$_{\pm\text{4.01}}$ & \textbf{77.42}$_{\pm\text{3.85}}$ \\

& & 200 & 71.41$_{\pm\text{3.46}}$ & 62.15$_{\pm\text{4.45}}$ & 59.89$_{\pm\text{4.05}}$ & 51.44$_{\pm\text{1.39}}$ & 48.94$_{\pm\text{1.59}}$ & 58.23$_{\pm\text{2.57}}$ & 49.25$_{\pm\text{3.05}}$ & 81.56$_{\pm\text{4.46}}$ & \textbf{86.14}$_{\pm\text{4.04}}$ \\

& & 500 & 77.17$_{\pm\text{2.01}}$ & 67.42$_{\pm\text{3.15}}$ & 65.34$_{\pm\text{4.17}}$ & 50.86$_{\pm\text{3.85}}$ & 49.83$_{\pm\text{0.48}}$ & 60.96$_{\pm\text{4.42}}$ & 51.04$_{\pm\text{3.62}}$ & 86.49$_{\pm\text{4.38}}$ & \textbf{90.60}$_{\pm\text{4.65}}$ \\

\cmidrule{2-12}

& \multirow[c]{4}{*}{stock} & 20 & 80.81$_{\pm\text{4.71}}$ & 71.58$_{\pm\text{6.48}}$ & 48.00$_{\pm\text{11.14}}$ & 66.07$_{\pm\text{12.69}}$ & 77.42$_{\pm\text{15.77}}$ & 61.80$_{\pm\text{9.34}}$ & 46.62$_{\pm\text{11.62}}$ & 84.30$_{\pm\text{4.96}}$ & \textbf{84.65}$_{\pm\text{3.38}}$ \\

& & 50 & 89.45$_{\pm\text{2.09}}$ & 74.30$_{\pm\text{8.64}}$ & 66.75$_{\pm\text{7.07}}$ & 69.95$_{\pm\text{10.51}}$ & 87.73$_{\pm\text{2.21}}$ & 79.90$_{\pm\text{4.82}}$ & 47.28$_{\pm\text{12.21}}$ & 89.27$_{\pm\text{1.89}}$ & \textbf{90.04}$_{\pm\text{2.09}}$ \\

& & 100 & 91.38$_{\pm\text{1.80}}$ & 84.09$_{\pm\text{3.75}}$ & 79.16$_{\pm\text{4.14}}$ & 72.00$_{\pm\text{4.77}}$ & 90.96$_{\pm\text{2.04}}$ & 86.38$_{\pm\text{2.97}}$ & 47.78$_{\pm\text{10.15}}$ & 92.07$_{\pm\text{1.56}}$ & \textbf{92.22}$_{\pm\text{1.36}}$ \\

& & 200 & \textbf{93.46}$_{\pm\text{0.79}}$ & 86.85$_{\pm\text{2.22}}$ & 83.47$_{\pm\text{3.47}}$ & 78.62$_{\pm\text{3.19}}$ & 92.72$_{\pm\text{0.81}}$ & 87.66$_{\pm\text{0.89}}$ & 49.91$_{\pm\text{13.88}}$ & 93.35$_{\pm\text{1.32}}$ & 93.07$_{\pm\text{0.99}}$ \\

\midrule

\multirow{9}{*}{\rotatebox{90}{\begin{tabular}{l} \textit{More than 10 classes} \end{tabular}}}& \multirow[c]{3}{*}{energy} & 50 & N/A & 6.42$_{\pm\text{1.49}}$ & 4.52$_{\pm\text{2.22}}$ & 4.88$_{\pm\text{1.49}}$ & 4.67$_{\pm\text{1.31}}$ & 6.62$_{\pm\text{2.47}}$ & 4.47$_{\pm\text{0.38}}$ & N/A & \textbf{28.79}$_{\pm\text{3.70}}$ \\

& & 100 & N/A & $-$ & 7.92$_{\pm\text{1.14}}$ & 5.40$_{\pm\text{1.11}}$ & 4.21$_{\pm\text{0.94}}$ & 9.03$_{\pm\text{1.53}}$ & 3.37$_{\pm\text{1.21}}$ & N/A & \textbf{39.20}$_{\pm\text{3.05}}$ \\

& & 200 & N/A & $-$ & 7.12$_{\pm\text{1.77}}$ & 6.01$_{\pm\text{1.38}}$ & 3.26$_{\pm\text{0.91}}$ & 11.78$_{\pm\text{3.10}}$ & 4.32$_{\pm\text{0.75}}$ & N/A & \textbf{42.55}$_{\pm\text{3.93}}$ \\

\cmidrule{2-12}

& \multirow[c]{2}{*}{collins} & 100 & N/A & $-$ & 6.10$_{\pm\text{0.63}}$ & 4.09$_{\pm\text{0.82}}$ & 11.95$_{\pm\text{1.92}}$ & 4.95$_{\pm\text{1.23}}$ & 3.99$_{\pm\text{0.64}}$ & N/A & \textbf{13.67}$_{\pm\text{1.31}}$ \\

& & 200 & \textbf{17.74}$_{\pm\text{1.75}}$ & $-$ & 5.86$_{\pm\text{1.39}}$ & 4.56$_{\pm\text{0.98}}$ & 13.22$_{\pm\text{1.22}}$ & 5.23$_{\pm\text{0.63}}$ & 4.08$_{\pm\text{0.62}}$ & N/A & 16.24$_{\pm\text{1.45}}$ \\

\cmidrule{2-12}

& \multirow[c]{4}{*}{texture} & 50 & 71.18$_{\pm\text{2.80}}$ & 22.41$_{\pm\text{7.88}}$ & 10.33$_{\pm\text{1.47}}$ & 10.55$_{\pm\text{1.87}}$ & 12.35$_{\pm\text{4.09}}$ & 19.83$_{\pm\text{5.72}}$ & 9.50$_{\pm\text{2.91}}$ & N/A & \textbf{75.77}$_{\pm\text{3.60}}$ \\

& & 100 & 79.85$_{\pm\text{2.02}}$ & 26.34$_{\pm\text{1.17}}$ & 14.84$_{\pm\text{6.52}}$ & 15.77$_{\pm\text{5.11}}$ & 14.16$_{\pm\text{4.42}}$ & 26.07$_{\pm\text{6.44}}$ & 8.97$_{\pm\text{0.33}}$ & N/A & \textbf{81.99}$_{\pm\text{2.39}}$ \\

& & 200 & 84.17$_{\pm\text{2.07}}$ & 22.70$_{\pm\text{5.57}}$ & 12.59$_{\pm\text{4.81}}$ & 11.61$_{\pm\text{3.40}}$ & 11.68$_{\pm\text{3.73}}$ & 46.83$_{\pm\text{4.78}}$ & 10.17$_{\pm\text{3.08}}$ & N/A & \textbf{84.90}$_{\pm\text{1.98}}$ \\

& & 500 & \textbf{88.31}$_{\pm\text{1.36}}$ & 32.92$_{\pm\text{10.58}}$ & 27.62$_{\pm\text{9.73}}$ & 13.33$_{\pm\text{4.79}}$ & 10.84$_{\pm\text{2.37}}$ & 62.04$_{\pm\text{4.01}}$ & 9.50$_{\pm\text{3.84}}$ & N/A & 88.14$_{\pm\text{1.50}}$ \\

\midrule
\rowcolor{Gainsboro!60}
\multicolumn{3}{l|}{\textbf{Average rank}} & 2.55$_{\pm\text{0.84}}$ & 4.39$_{\pm\text{0.80}}$ & 6.09$_{\pm\text{1.01}}$ & 7.42$_{\pm\text{0.90}}$ & 7.09$_{\pm\text{1.94}}$ & 5.15$_{\pm\text{1.03}}$ & 8.55$_{\pm\text{0.62}}$ & 2.36$_{\pm\text{0.89}}$ & \textbf{1.39}$_{\pm\text{0.70}}$ \\

\bottomrule

\end{tabular}
}
\end{table}
\clearpage

\begin{table}[p]
\centering
\caption{\textbf{Classification accuracy} (\%) of XGBoost, comparing data sharing on eight real-world tabular datasets with varied real data availability. We report the mean $\pm$ std balanced accuracy and average accuracy rank across datasets. A higher rank implies higher accuracy. Note that ``N/A'' denotes the inapplicability of a specific generator. Different from \cref{tab:test_acc_per_dataset_augmentation}, ``$-$'' denotes a generator cannot satisfy the requirement of generating 500 stratified samples even after generating 10,000 synthetic samples. The results of these inapplicable or failed generators are computed with the mean results of other methods. We \textbf{bold} the highest result for each dataset of different sample sizes. TVAE learns the joint distribution $p(\rvx, y)$ and fails to maintain the original training label distribution. TabEBM achieves the best overall performance against Baseline and benchmark generators.}
\resizebox{\textwidth}{!}{
\begin{tabular}{m{0.2cm}lr|rrrrrrrr|r}

\toprule

\multicolumn{2}{l}{Datasets}  & $N_{\text{real}}$ & SMOTE & TVAE & CTGAN & NFLOW & TabDDPM & ARF & GOGGLE & TabPFGen & \textbf{TabEBM} \\

\midrule

\multirow{24}{*}{\rotatebox{90}{\begin{tabular}{l} \textit{At most 10 classes} \end{tabular}}}& \multirow[c]{5}{*}{protein} & 20 & N/A & 15.47$_{\pm\text{2.77}}$ & 13.60$_{\pm\text{3.62}}$ & 15.03$_{\pm\text{4.60}}$ & 11.77$_{\pm\text{3.36}}$ & 18.32$_{\pm\text{3.84}}$ & 14.61$_{\pm\text{2.21}}$ & 23.54$_{\pm\text{4.33}}$ & \textbf{25.22}$_{\pm\text{3.56}}$ \\

& & 50 & 38.51$_{\pm\text{4.83}}$ & 25.36$_{\pm\text{3.19}}$ & 18.88$_{\pm\text{4.31}}$ & 13.84$_{\pm\text{4.50}}$ & 12.72$_{\pm\text{1.99}}$ & 24.39$_{\pm\text{4.09}}$ & 13.13$_{\pm\text{4.44}}$ & 41.88$_{\pm\text{6.20}}$ & \textbf{43.79}$_{\pm\text{6.81}}$ \\

& & 100 & 60.71$_{\pm\text{5.70}}$ & 33.23$_{\pm\text{4.65}}$ & 23.54$_{\pm\text{3.74}}$ & 14.36$_{\pm\text{3.09}}$ & 14.08$_{\pm\text{1.84}}$ & 28.49$_{\pm\text{6.26}}$ & 14.48$_{\pm\text{4.96}}$ & 54.91$_{\pm\text{6.96}}$ & \textbf{61.11}$_{\pm\text{6.11}}$ \\

& & 200 & 73.31$_{\pm\text{3.15}}$ & 35.50$_{\pm\text{6.72}}$ & 26.74$_{\pm\text{5.01}}$ & 15.45$_{\pm\text{4.45}}$ & 14.71$_{\pm\text{3.38}}$ & 36.61$_{\pm\text{3.61}}$ & 10.41$_{\pm\text{2.52}}$ & 75.20$_{\pm\text{2.13}}$ & \textbf{76.96}$_{\pm\text{2.90}}$ \\

& & 500 & 83.11$_{\pm\text{1.82}}$ & 41.18$_{\pm\text{6.31}}$ & 37.95$_{\pm\text{5.62}}$ & 13.56$_{\pm\text{2.40}}$ & 14.47$_{\pm\text{2.40}}$ & 48.92$_{\pm\text{4.09}}$ & 11.25$_{\pm\text{3.92}}$ & 85.30$_{\pm\text{1.56}}$ & \textbf{85.39}$_{\pm\text{1.90}}$ \\

\cmidrule{2-12}

& \multirow[c]{5}{*}{fourier} & 20 & N/A & $-$ & 11.16$_{\pm\text{2.78}}$ & 10.36$_{\pm\text{2.58}}$ & 12.38$_{\pm\text{3.08}}$ & 14.64$_{\pm\text{4.90}}$ & 9.18$_{\pm\text{1.60}}$ & 21.80$_{\pm\text{4.86}}$ & \textbf{26.80}$_{\pm\text{4.82}}$ \\

& & 50 & 41.70$_{\pm\text{8.00}}$ & $-$ & N/A & 10.06$_{\pm\text{1.96}}$ & 10.00$_{\pm\text{1.41}}$ & 19.96$_{\pm\text{4.44}}$ & 12.24$_{\pm\text{3.23}}$ & 43.10$_{\pm\text{6.37}}$ & \textbf{47.16}$_{\pm\text{5.42}}$ \\

& & 100 & \textbf{54.52}$_{\pm\text{5.80}}$ & 27.78$_{\pm\text{5.49}}$ & 19.40$_{\pm\text{4.93}}$ & 10.98$_{\pm\text{2.29}}$ & 12.22$_{\pm\text{3.24}}$ & 24.24$_{\pm\text{4.70}}$ & 10.34$_{\pm\text{2.64}}$ & 51.40$_{\pm\text{5.42}}$ & 52.62$_{\pm\text{5.05}}$ \\

& & 200 & 65.96$_{\pm\text{4.60}}$ & 31.60$_{\pm\text{2.49}}$ & 24.76$_{\pm\text{5.83}}$ & 11.84$_{\pm\text{3.99}}$ & 13.12$_{\pm\text{3.24}}$ & 31.74$_{\pm\text{3.75}}$ & 11.36$_{\pm\text{2.39}}$ & 62.68$_{\pm\text{4.32}}$ & \textbf{66.26}$_{\pm\text{4.52}}$ \\

& & 500 & 71.44$_{\pm\text{1.57}}$ & 34.72$_{\pm\text{1.86}}$ & 36.56$_{\pm\text{5.98}}$ & 11.64$_{\pm\text{2.58}}$ & 12.72$_{\pm\text{3.83}}$ & 37.86$_{\pm\text{3.31}}$ & 9.42$_{\pm\text{1.66}}$ & 72.00$_{\pm\text{2.94}}$ & \textbf{74.45}$_{\pm\text{1.66}}$ \\

\cmidrule{2-12}

& \multirow[c]{5}{*}{biodeg} & 20 & 66.52$_{\pm\text{5.96}}$ & 60.35$_{\pm\text{7.63}}$ & 58.55$_{\pm\text{6.57}}$ & 54.65$_{\pm\text{8.42}}$ & 50.67$_{\pm\text{4.16}}$ & 59.80$_{\pm\text{7.09}}$ & 52.58$_{\pm\text{13.73}}$ & 65.99$_{\pm\text{6.54}}$ & \textbf{66.60}$_{\pm\text{6.87}}$ \\

& & 50 & 68.01$_{\pm\text{2.67}}$ & 63.62$_{\pm\text{4.85}}$ & 60.53$_{\pm\text{7.91}}$ & 56.10$_{\pm\text{6.94}}$ & 48.13$_{\pm\text{3.77}}$ & \textbf{68.90}$_{\pm\text{4.66}}$ & 52.89$_{\pm\text{9.18}}$ & 68.60$_{\pm\text{4.27}}$ & 68.82$_{\pm\text{3.18}}$ \\

& & 100 & 71.98$_{\pm\text{4.21}}$ & 68.71$_{\pm\text{4.56}}$ & 68.58$_{\pm\text{3.57}}$ & 60.67$_{\pm\text{9.00}}$ & 49.08$_{\pm\text{3.11}}$ & 68.74$_{\pm\text{3.58}}$ & 49.51$_{\pm\text{8.42}}$ & \textbf{74.73}$_{\pm\text{1.88}}$ & 72.69$_{\pm\text{3.36}}$ \\

& & 200 & 74.29$_{\pm\text{2.14}}$ & 68.94$_{\pm\text{4.03}}$ & 69.87$_{\pm\text{3.15}}$ & 59.26$_{\pm\text{7.34}}$ & 47.26$_{\pm\text{2.99}}$ & 70.89$_{\pm\text{3.66}}$ & 55.64$_{\pm\text{7.90}}$ & 76.03$_{\pm\text{4.77}}$ & \textbf{77.15}$_{\pm\text{2.53}}$ \\

& & 500 & 76.04$_{\pm\text{3.09}}$ & 70.53$_{\pm\text{5.19}}$ & 73.65$_{\pm\text{3.54}}$ & 53.73$_{\pm\text{4.78}}$ & 47.86$_{\pm\text{4.57}}$ & 74.39$_{\pm\text{2.84}}$ & 49.08$_{\pm\text{11.19}}$ & \textbf{78.99}$_{\pm\text{2.89}}$ & 77.61$_{\pm\text{2.83}}$ \\

\cmidrule{2-12}

& \multirow[c]{5}{*}{steel} & 20 & \textbf{55.76}$_{\pm\text{4.40}}$ & 53.17$_{\pm\text{6.28}}$ & 53.71$_{\pm\text{9.38}}$ & 51.36$_{\pm\text{5.69}}$ & 51.32$_{\pm\text{6.29}}$ & 51.36$_{\pm\text{6.40}}$ & 51.11$_{\pm\text{4.18}}$ & 55.71$_{\pm\text{5.46}}$ & 54.95$_{\pm\text{5.04}}$ \\

& & 50 & 61.07$_{\pm\text{3.58}}$ & 56.97$_{\pm\text{3.93}}$ & 54.38$_{\pm\text{4.00}}$ & 53.54$_{\pm\text{3.82}}$ & 46.83$_{\pm\text{6.18}}$ & 53.60$_{\pm\text{4.17}}$ & 49.46$_{\pm\text{6.57}}$ & 63.64$_{\pm\text{6.52}}$ & \textbf{71.58}$_{\pm\text{14.28}}$ \\

& & 100 & 64.12$_{\pm\text{5.76}}$ & 59.03$_{\pm\text{5.74}}$ & 57.01$_{\pm\text{5.40}}$ & 52.86$_{\pm\text{3.06}}$ & 44.22$_{\pm\text{7.97}}$ & 55.21$_{\pm\text{4.00}}$ & 46.92$_{\pm\text{7.56}}$ & 88.61$_{\pm\text{12.72}}$ & \textbf{96.34}$_{\pm\text{2.73}}$ \\

& & 200 & 74.96$_{\pm\text{6.42}}$ & 58.42$_{\pm\text{6.66}}$ & 59.25$_{\pm\text{3.48}}$ & 51.94$_{\pm\text{3.30}}$ & 43.96$_{\pm\text{10.64}}$ & 58.92$_{\pm\text{3.56}}$ & 49.58$_{\pm\text{5.96}}$ & 98.85$_{\pm\text{1.71}}$ & \textbf{99.28}$_{\pm\text{0.72}}$ \\

& & 500 & 81.79$_{\pm\text{5.79}}$ & 61.11$_{\pm\text{8.40}}$ & 65.70$_{\pm\text{7.71}}$ & 51.68$_{\pm\text{2.71}}$ & 46.63$_{\pm\text{6.69}}$ & 59.19$_{\pm\text{7.43}}$ & 49.98$_{\pm\text{5.15}}$ & 99.62$_{\pm\text{1.11}}$ & \textbf{99.92}$_{\pm\text{0.13}}$ \\

\cmidrule{2-12}

& \multirow[c]{4}{*}{stock} & 20 & 77.64$_{\pm\text{6.07}}$ & 64.61$_{\pm\text{10.04}}$ & 50.93$_{\pm\text{6.99}}$ & 67.70$_{\pm\text{12.94}}$ & 74.63$_{\pm\text{13.41}}$ & 68.29$_{\pm\text{7.28}}$ & 45.59$_{\pm\text{10.96}}$ & 80.14$_{\pm\text{4.76}}$ & \textbf{84.22}$_{\pm\text{4.50}}$ \\

& & 50 & 84.83$_{\pm\text{2.98}}$ & 77.89$_{\pm\text{7.53}}$ & 73.57$_{\pm\text{5.76}}$ & 70.29$_{\pm\text{6.08}}$ & 84.75$_{\pm\text{4.01}}$ & 79.19$_{\pm\text{2.51}}$ & 54.82$_{\pm\text{11.81}}$ & 86.97$_{\pm\text{3.25}}$ & \textbf{88.59}$_{\pm\text{3.54}}$ \\

& & 100 & 87.69$_{\pm\text{1.26}}$ & 80.12$_{\pm\text{8.75}}$ & 77.23$_{\pm\text{4.37}}$ & 71.03$_{\pm\text{8.08}}$ & 85.72$_{\pm\text{4.00}}$ & 83.91$_{\pm\text{3.27}}$ & 55.60$_{\pm\text{10.94}}$ & 89.21$_{\pm\text{3.39}}$ & \textbf{91.14}$_{\pm\text{1.97}}$ \\

& & 200 & 90.58$_{\pm\text{3.02}}$ & 83.34$_{\pm\text{3.80}}$ & 81.04$_{\pm\text{3.24}}$ & 76.60$_{\pm\text{5.86}}$ & \textbf{91.52}$_{\pm\text{3.49}}$ & 87.49$_{\pm\text{3.41}}$ & 55.43$_{\pm\text{12.95}}$ & 90.78$_{\pm\text{3.56}}$ & 90.94$_{\pm\text{1.18}}$ \\

\midrule

\multirow{9}{*}{\rotatebox{90}{\begin{tabular}{l} \textit{More than 10 classes} \end{tabular}}}& \multirow[c]{3}{*}{energy} & 50 & N/A & 7.37$_{\pm\text{2.92}}$ & 5.16$_{\pm\text{2.03}}$ & 4.90$_{\pm\text{1.44}}$ & 3.95$_{\pm\text{1.60}}$ & 6.62$_{\pm\text{1.84}}$ & N/A & N/A & \textbf{19.38}$_{\pm\text{3.84}}$ \\

& & 100 & N/A & $-$ & 8.25$_{\pm\text{2.78}}$ & 5.40$_{\pm\text{2.24}}$ & 3.95$_{\pm\text{1.37}}$ & 10.21$_{\pm\text{2.45}}$ & N/A & N/A & \textbf{25.08}$_{\pm\text{5.99}}$ \\

& & 200 & N/A & $-$ & 6.97$_{\pm\text{2.92}}$ & 8.56$_{\pm\text{2.68}}$ & 3.21$_{\pm\text{1.53}}$ & 11.08$_{\pm\text{2.96}}$ & N/A & N/A & \textbf{31.03}$_{\pm\text{4.94}}$ \\

\cmidrule{2-12}

& \multirow[c]{2}{*}{collins} & 100 & N/A & $-$ & 4.45$_{\pm\text{0.82}}$ & 3.98$_{\pm\text{0.70}}$ & 8.18$_{\pm\text{1.43}}$ & 5.39$_{\pm\text{1.68}}$ & N/A & N/A & \textbf{9.03}$_{\pm\text{1.14}}$ \\

& & 200 & \textbf{12.52}$_{\pm\text{1.98}}$ & $-$ & 5.52$_{\pm\text{1.57}}$ & 4.43$_{\pm\text{0.63}}$ & 9.06$_{\pm\text{1.21}}$ & 5.61$_{\pm\text{1.10}}$ & N/A & N/A & 11.49$_{\pm\text{1.45}}$ \\

\cmidrule{2-12}

& \multirow[c]{4}{*}{texture} & 50 & 57.99$_{\pm\text{3.35}}$ & 20.94$_{\pm\text{8.45}}$ & 9.44$_{\pm\text{4.78}}$ & 12.63$_{\pm\text{3.07}}$ & 11.32$_{\pm\text{4.19}}$ & 14.35$_{\pm\text{4.95}}$ & N/A & N/A & \textbf{68.97}$_{\pm\text{4.98}}$ \\

& & 100 & 69.80$_{\pm\text{4.34}}$ & 23.00$_{\pm\text{7.10}}$ & 14.06$_{\pm\text{4.92}}$ & 16.18$_{\pm\text{5.64}}$ & 9.90$_{\pm\text{2.39}}$ & 19.13$_{\pm\text{6.41}}$ & N/A & N/A & \textbf{76.62}$_{\pm\text{2.55}}$ \\

& & 200 & 78.46$_{\pm\text{2.72}}$ & 23.32$_{\pm\text{7.04}}$ & 15.09$_{\pm\text{2.96}}$ & 15.15$_{\pm\text{3.79}}$ & 9.66$_{\pm\text{1.64}}$ & 33.67$_{\pm\text{5.07}}$ & N/A & N/A & \textbf{80.37}$_{\pm\text{2.13}}$ \\

& & 500 & \textbf{86.39}$_{\pm\text{1.70}}$ & 31.13$_{\pm\text{9.16}}$ & 22.88$_{\pm\text{7.24}}$ & 13.46$_{\pm\text{5.72}}$ & 10.22$_{\pm\text{1.70}}$ & 47.06$_{\pm\text{3.83}}$ & N/A & N/A & 85.83$_{\pm\text{1.68}}$ \\

\midrule
\rowcolor{Gainsboro!60}
\multicolumn{3}{l|}{\textbf{Average rank}} & 2.76$_{\pm\text{0.90}}$ & 4.97$_{\pm\text{0.92}}$ & 6.39$_{\pm\text{1.32}}$ & 7.42$_{\pm\text{0.75}}$ & 7.48$_{\pm\text{2.35}}$ & 4.97$_{\pm\text{1.26}}$ & 7.11$_{\pm\text{2.15}}$ & 2.62$_{\pm\text{0.95}}$ & \textbf{1.27}$_{\pm\text{0.52}}$ \\

\bottomrule

\end{tabular}
}
\end{table}
\clearpage

\begin{table}[p]
\centering
\caption{\textbf{Classification accuracy} (\%) of TabPFN, comparing data sharing on eight real-world tabular datasets with varied real data availability. We report the mean $\pm$ std balanced accuracy and average accuracy rank across datasets. A higher rank implies higher accuracy. Note that ``N/A'' denotes the inapplicability of a specific generator. Different from \cref{tab:test_acc_per_dataset_augmentation}, ``$-$'' denotes a generator cannot satisfy the requirement of generating 500 stratified samples even after generating 10,000 synthetic samples. The results of these inapplicable or failed generators are computed with the mean results of other methods. We \textbf{bold} the highest result for each dataset of different sample sizes. TVAE learns the joint distribution $p(\rvx, y)$ and fails to maintain the original training label distribution. TabEBM achieves the best overall performance against Baseline and benchmark generators.}
\resizebox{\textwidth}{!}{
\begin{tabular}{m{0.2cm}lr|rrrrrrrr|r}

\toprule

\multicolumn{2}{l}{Datasets}  & $N_{\text{real}}$ & SMOTE & TVAE & CTGAN & NFLOW & TabDDPM & ARF & GOGGLE & TabPFGen & \textbf{TabEBM} \\

\midrule

\multirow{24}{*}{\rotatebox{90}{\begin{tabular}{l} \textit{At most 10 classes} \end{tabular}}}& \multirow[c]{5}{*}{protein} & 20 & N/A & 16.36$_{\pm\text{3.39}}$ & 13.22$_{\pm\text{1.42}}$ & 13.80$_{\pm\text{4.56}}$ & 12.93$_{\pm\text{3.67}}$ & 18.55$_{\pm\text{2.64}}$ & 12.14$_{\pm\text{1.16}}$ & 33.46$_{\pm\text{5.96}}$ & \textbf{34.50}$_{\pm\text{5.86}}$ \\

& & 50 & \textbf{59.60}$_{\pm\text{3.63}}$ & 27.51$_{\pm\text{3.72}}$ & 18.26$_{\pm\text{4.54}}$ & 13.28$_{\pm\text{3.34}}$ & 14.95$_{\pm\text{3.52}}$ & 29.82$_{\pm\text{3.75}}$ & 13.30$_{\pm\text{1.88}}$ & 57.68$_{\pm\text{2.79}}$ & 59.00$_{\pm\text{3.97}}$ \\

& & 100 & \textbf{79.47}$_{\pm\text{3.48}}$ & 41.49$_{\pm\text{6.98}}$ & 31.50$_{\pm\text{3.05}}$ & 13.69$_{\pm\text{2.75}}$ & 17.21$_{\pm\text{3.12}}$ & 36.73$_{\pm\text{5.05}}$ & 12.61$_{\pm\text{0.81}}$ & 78.93$_{\pm\text{4.23}}$ & 79.02$_{\pm\text{3.74}}$ \\

& & 200 & 90.49$_{\pm\text{2.23}}$ & 48.30$_{\pm\text{7.73}}$ & 36.99$_{\pm\text{5.39}}$ & 13.71$_{\pm\text{3.79}}$ & 21.67$_{\pm\text{5.92}}$ & 43.53$_{\pm\text{5.42}}$ & 12.96$_{\pm\text{2.86}}$ & \textbf{91.47}$_{\pm\text{1.28}}$ & 90.71$_{\pm\text{1.30}}$ \\

& & 500 & 94.28$_{\pm\text{1.36}}$ & 58.31$_{\pm\text{8.05}}$ & 48.52$_{\pm\text{4.70}}$ & 12.91$_{\pm\text{2.56}}$ & 23.35$_{\pm\text{7.76}}$ & 57.37$_{\pm\text{3.28}}$ & 12.51$_{\pm\text{3.23}}$ & \textbf{95.26}$_{\pm\text{1.37}}$ & 95.00$_{\pm\text{1.28}}$ \\

\cmidrule{2-12}

& \multirow[c]{5}{*}{fourier} & 20 & N/A & $-$ & 13.98$_{\pm\text{3.05}}$ & 10.64$_{\pm\text{2.52}}$ & 10.56$_{\pm\text{1.86}}$ & 16.44$_{\pm\text{4.20}}$ & 10.56$_{\pm\text{3.20}}$ & 29.42$_{\pm\text{6.72}}$ & \textbf{36.56}$_{\pm\text{4.95}}$ \\

& & 50 & 52.52$_{\pm\text{3.71}}$ & $-$ & N/A & 10.24$_{\pm\text{1.53}}$ & 10.72$_{\pm\text{1.53}}$ & 27.82$_{\pm\text{3.41}}$ & 9.90$_{\pm\text{0.33}}$ & \textbf{53.98}$_{\pm\text{4.48}}$ & 53.92$_{\pm\text{3.82}}$ \\

& & 100 & 63.36$_{\pm\text{3.56}}$ & 36.24$_{\pm\text{4.25}}$ & 28.44$_{\pm\text{4.45}}$ & 10.30$_{\pm\text{1.94}}$ & 11.84$_{\pm\text{3.10}}$ & 30.76$_{\pm\text{2.70}}$ & 10.32$_{\pm\text{2.04}}$ & 65.32$_{\pm\text{3.26}}$ & \textbf{65.58}$_{\pm\text{3.38}}$ \\

& & 200 & 69.62$_{\pm\text{4.71}}$ & 40.36$_{\pm\text{3.77}}$ & 40.08$_{\pm\text{5.17}}$ & 8.82$_{\pm\text{3.98}}$ & 32.98$_{\pm\text{6.25}}$ & 38.50$_{\pm\text{3.82}}$ & 10.30$_{\pm\text{0.76}}$ & 72.08$_{\pm\text{2.45}}$ & \textbf{72.26}$_{\pm\text{2.70}}$ \\

& & 500 & 75.14$_{\pm\text{1.83}}$ & 46.56$_{\pm\text{5.61}}$ & 53.38$_{\pm\text{4.52}}$ & 9.52$_{\pm\text{1.46}}$ & 32.46$_{\pm\text{9.02}}$ & 49.62$_{\pm\text{2.60}}$ & 10.68$_{\pm\text{1.36}}$ & \textbf{75.98}$_{\pm\text{1.25}}$ & 75.60$_{\pm\text{0.85}}$ \\

\cmidrule{2-12}

& \multirow[c]{5}{*}{biodeg} & 20 & 69.01$_{\pm\text{4.56}}$ & 65.35$_{\pm\text{7.46}}$ & 52.29$_{\pm\text{8.45}}$ & 55.71$_{\pm\text{6.77}}$ & 50.00$_{\pm\text{0.00}}$ & 57.11$_{\pm\text{8.37}}$ & 50.35$_{\pm\text{1.11}}$ & 70.77$_{\pm\text{4.90}}$ & \textbf{71.36}$_{\pm\text{5.30}}$ \\

& & 50 & 73.27$_{\pm\text{3.31}}$ & 68.69$_{\pm\text{5.88}}$ & 63.31$_{\pm\text{6.35}}$ & 54.85$_{\pm\text{8.00}}$ & 50.00$_{\pm\text{0.00}}$ & 71.23$_{\pm\text{5.04}}$ & 50.00$_{\pm\text{0.00}}$ & \textbf{75.78}$_{\pm\text{2.59}}$ & 75.56$_{\pm\text{3.19}}$ \\

& & 100 & 76.39$_{\pm\text{2.12}}$ & 73.06$_{\pm\text{6.33}}$ & 71.75$_{\pm\text{2.75}}$ & 55.26$_{\pm\text{6.54}}$ & 50.00$_{\pm\text{0.00}}$ & 73.02$_{\pm\text{2.42}}$ & 50.00$_{\pm\text{0.00}}$ & 78.22$_{\pm\text{1.55}}$ & \textbf{79.28}$_{\pm\text{2.16}}$ \\

& & 200 & 80.52$_{\pm\text{0.85}}$ & 73.19$_{\pm\text{5.28}}$ & 72.93$_{\pm\text{5.24}}$ & 54.94$_{\pm\text{8.51}}$ & 49.88$_{\pm\text{0.38}}$ & 76.03$_{\pm\text{2.88}}$ & 50.00$_{\pm\text{0.00}}$ & 82.22$_{\pm\text{1.72}}$ & \textbf{82.70}$_{\pm\text{1.73}}$ \\

& & 500 & 82.47$_{\pm\text{1.02}}$ & 77.06$_{\pm\text{3.03}}$ & 79.18$_{\pm\text{2.08}}$ & 50.47$_{\pm\text{1.03}}$ & 49.67$_{\pm\text{1.05}}$ & 77.79$_{\pm\text{1.36}}$ & 50.00$_{\pm\text{0.00}}$ & 83.03$_{\pm\text{0.96}}$ & \textbf{83.67}$_{\pm\text{0.83}}$ \\

\cmidrule{2-12}

& \multirow[c]{5}{*}{steel} & 20 & 56.80$_{\pm\text{4.60}}$ & 53.88$_{\pm\text{5.00}}$ & 52.12$_{\pm\text{3.28}}$ & 50.86$_{\pm\text{5.69}}$ & 50.00$_{\pm\text{0.00}}$ & 50.47$_{\pm\text{1.75}}$ & 50.00$_{\pm\text{0.00}}$ & 64.92$_{\pm\text{5.75}}$ & \textbf{65.82}$_{\pm\text{6.29}}$ \\

& & 50 & 62.45$_{\pm\text{4.31}}$ & 57.80$_{\pm\text{2.19}}$ & 57.11$_{\pm\text{5.27}}$ & 51.84$_{\pm\text{3.72}}$ & 50.00$_{\pm\text{0.00}}$ & 56.40$_{\pm\text{3.61}}$ & 50.00$_{\pm\text{0.00}}$ & 84.81$_{\pm\text{7.86}}$ & \textbf{86.32}$_{\pm\text{6.58}}$ \\

& & 100 & 71.25$_{\pm\text{5.06}}$ & 61.44$_{\pm\text{3.38}}$ & 60.45$_{\pm\text{4.62}}$ & 50.59$_{\pm\text{3.10}}$ & 48.48$_{\pm\text{3.17}}$ & 55.74$_{\pm\text{4.69}}$ & 50.66$_{\pm\text{2.08}}$ & 97.29$_{\pm\text{1.40}}$ & \textbf{97.85}$_{\pm\text{1.38}}$ \\

& & 200 & 77.96$_{\pm\text{3.06}}$ & 63.92$_{\pm\text{3.33}}$ & 64.22$_{\pm\text{5.27}}$ & 50.00$_{\pm\text{0.00}}$ & 47.65$_{\pm\text{4.95}}$ & 62.33$_{\pm\text{4.50}}$ & 50.00$_{\pm\text{0.01}}$ & 98.58$_{\pm\text{0.72}}$ & \textbf{98.89}$_{\pm\text{0.69}}$ \\

& & 500 & 85.29$_{\pm\text{4.02}}$ & 69.41$_{\pm\text{5.33}}$ & 73.34$_{\pm\text{5.43}}$ & 50.09$_{\pm\text{0.27}}$ & 50.00$_{\pm\text{0.00}}$ & 68.78$_{\pm\text{5.84}}$ & 50.00$_{\pm\text{0.00}}$ & 99.71$_{\pm\text{0.30}}$ & 99.71$_{\pm\text{0.27}}$ \\

\cmidrule{2-12}

& \multirow[c]{4}{*}{stock} & 20 & \textbf{83.79}$_{\pm\text{3.01}}$ & 70.41$_{\pm\text{7.99}}$ & 49.52$_{\pm\text{16.46}}$ & 70.75$_{\pm\text{10.65}}$ & 78.58$_{\pm\text{11.59}}$ & 66.66$_{\pm\text{8.26}}$ & 51.45$_{\pm\text{4.91}}$ & 82.98$_{\pm\text{4.47}}$ & 83.79$_{\pm\text{4.96}}$ \\

& & 50 & 89.91$_{\pm\text{2.44}}$ & 77.18$_{\pm\text{3.13}}$ & 70.04$_{\pm\text{2.32}}$ & 72.42$_{\pm\text{8.52}}$ & 89.35$_{\pm\text{1.75}}$ & 78.06$_{\pm\text{3.32}}$ & 50.00$_{\pm\text{0.00}}$ & 89.97$_{\pm\text{2.12}}$ & \textbf{90.17}$_{\pm\text{1.73}}$ \\

& & 100 & 92.05$_{\pm\text{1.34}}$ & 81.60$_{\pm\text{3.66}}$ & 78.73$_{\pm\text{3.67}}$ & 75.24$_{\pm\text{3.91}}$ & 91.32$_{\pm\text{1.47}}$ & 85.24$_{\pm\text{2.39}}$ & 49.83$_{\pm\text{0.54}}$ & 92.25$_{\pm\text{1.21}}$ & \textbf{92.46}$_{\pm\text{1.18}}$ \\

& & 200 & 93.50$_{\pm\text{0.91}}$ & 85.46$_{\pm\text{3.04}}$ & 83.47$_{\pm\text{3.02}}$ & 75.51$_{\pm\text{2.44}}$ & 92.95$_{\pm\text{1.04}}$ & 87.25$_{\pm\text{2.03}}$ & 50.00$_{\pm\text{0.00}}$ & \textbf{93.93}$_{\pm\text{0.85}}$ & 93.65$_{\pm\text{1.01}}$ \\

\midrule
\rowcolor{Gainsboro!60}
\multicolumn{3}{l|}{\textbf{Average rank}} & 2.77$_{\pm\text{0.69}}$ & 4.73$_{\pm\text{0.87}}$ & 5.88$_{\pm\text{1.26}}$ & 7.42$_{\pm\text{1.06}}$ & 7.33$_{\pm\text{1.72}}$ & 5.21$_{\pm\text{0.83}}$ & 8.42$_{\pm\text{0.52}}$ & 1.85$_{\pm\text{0.62}}$ & \textbf{1.40}$_{\pm\text{0.49}}$ \\

\bottomrule

\end{tabular}
}
\end{table}
\clearpage

\FloatBarrier
\subsubsection{DCR Evaluation}

\begin{table}[hp]
\centering
\caption{\textbf{DCR between real train data and synthetic data} on eight real-world tabular datasets with varied real data availability. We report the mean $\pm$ std result and average rank across datasets. A higher rank implies better privacy preservation. Note that ``N/A'' denotes that a specific generator was not applicable, and the rank is computed with the mean result of other methods. We \textbf{bold} the highest result for each dataset of different sample sizes. Even though ARF and NFLOW show high DCR, our experiments demonstrate that they do not learn the data distribution well, leading to poor downstream accuracy. TabEBM achieves competitive overall DCR against benchmark generators.}
\resizebox{\textwidth}{!}{

\setlength{\tabcolsep}{5pt}
\begin{tabular}{m{0.2cm}lr|rrrrrrr|r}

\toprule

\multicolumn{2}{l}{Datasets}  & $N_{\text{real}}$ & SMOTE & TVAE & CTGAN & NFLOW & TabDDPM & ARF & TabPFGen & \textbf{TabEBM} \\

\midrule

\multirow{24}{*}{\rotatebox{90}{\begin{tabular}{l} \textit{At most 10 classes} \end{tabular}}}& \multirow[c]{5}{*}{protein} & 20 & N/A & 0.24$_{\pm\text{0.06}}$ & 0.36$_{\pm\text{0.10}}$ & 0.29$_{\pm\text{0.07}}$ & \textbf{0.60}$_{\pm\text{0.06}}$ & 0.49$_{\pm\text{0.03}}$ & 0.24$_{\pm\text{0.11}}$ & 0.39$_{\pm\text{0.05}}$ \\

& & 50 & 0.20$_{\pm\text{0.03}}$ & 0.34$_{\pm\text{0.09}}$ & 0.42$_{\pm\text{0.06}}$ & 0.21$_{\pm\text{0.06}}$ & \textbf{0.62}$_{\pm\text{0.01}}$ & 0.47$_{\pm\text{0.08}}$ & 0.21$_{\pm\text{0.11}}$ & 0.37$_{\pm\text{0.11}}$ \\

& & 100 & 0.20$_{\pm\text{0.03}}$ & 0.33$_{\pm\text{0.07}}$ & 0.37$_{\pm\text{0.05}}$ & 0.26$_{\pm\text{0.10}}$ & \textbf{0.54}$_{\pm\text{0.03}}$ & 0.46$_{\pm\text{0.06}}$ & 0.12$_{\pm\text{0.07}}$ & 0.27$_{\pm\text{0.15}}$ \\

& & 200 & 0.19$_{\pm\text{0.03}}$ & 0.31$_{\pm\text{0.05}}$ & 0.35$_{\pm\text{0.04}}$ & 0.31$_{\pm\text{0.06}}$ & \textbf{0.51}$_{\pm\text{0.04}}$ & 0.44$_{\pm\text{0.05}}$ & 0.10$_{\pm\text{0.03}}$ & 0.26$_{\pm\text{0.06}}$ \\

& & 500 & 0.19$_{\pm\text{0.02}}$ & 0.32$_{\pm\text{0.06}}$ & 0.30$_{\pm\text{0.05}}$ & 0.33$_{\pm\text{0.02}}$ & \textbf{0.48}$_{\pm\text{0.03}}$ & 0.43$_{\pm\text{0.05}}$ & 0.09$_{\pm\text{0.06}}$ & 0.23$_{\pm\text{0.13}}$ \\

\cmidrule{2-11}

& \multirow[c]{5}{*}{fourier} & 20 & N/A & 0.19$_{\pm\text{0.17}}$ & \textbf{0.61}$_{\pm\text{0.04}}$ & 0.52$_{\pm\text{0.08}}$ & 0.56$_{\pm\text{0.05}}$ & 0.57$_{\pm\text{0.04}}$ & 0.48$_{\pm\text{0.00}}$ & 0.40$_{\pm\text{0.00}}$ \\

& & 50 & 0.20$_{\pm\text{0.02}}$ & 0.29$_{\pm\text{0.26}}$ & 0.48$_{\pm\text{0.17}}$ & 0.33$_{\pm\text{0.10}}$ & \textbf{0.67}$_{\pm\text{0.05}}$ & 0.54$_{\pm\text{0.07}}$ & 0.31$_{\pm\text{0.07}}$ & 0.43$_{\pm\text{0.03}}$ \\

& & 100 & 0.23$_{\pm\text{0.02}}$ & 0.53$_{\pm\text{0.06}}$ & 0.50$_{\pm\text{0.04}}$ & 0.37$_{\pm\text{0.09}}$ & \textbf{0.60}$_{\pm\text{0.08}}$ & 0.59$_{\pm\text{0.06}}$ & 0.31$_{\pm\text{0.07}}$ & 0.44$_{\pm\text{0.04}}$ \\

& & 200 & 0.22$_{\pm\text{0.02}}$ & 0.56$_{\pm\text{0.06}}$ & 0.53$_{\pm\text{0.05}}$ & 0.37$_{\pm\text{0.08}}$ & \textbf{0.58}$_{\pm\text{0.02}}$ & 0.56$_{\pm\text{0.04}}$ & 0.30$_{\pm\text{0.03}}$ & 0.46$_{\pm\text{0.04}}$ \\

& & 500 & 0.25$_{\pm\text{0.03}}$ & \textbf{0.67}$_{\pm\text{0.04}}$ & 0.54$_{\pm\text{0.06}}$ & 0.40$_{\pm\text{0.06}}$ & 0.62$_{\pm\text{0.04}}$ & 0.61$_{\pm\text{0.05}}$ & 0.28$_{\pm\text{0.03}}$ & 0.45$_{\pm\text{0.05}}$ \\

\cmidrule{2-11}

& \multirow[c]{5}{*}{biodeg} & 20 & 0.29$_{\pm\text{0.05}}$ & 0.19$_{\pm\text{0.08}}$ & 0.26$_{\pm\text{0.07}}$ & 0.33$_{\pm\text{0.08}}$ & 0.26$_{\pm\text{0.15}}$ & \textbf{0.46}$_{\pm\text{0.05}}$ & 0.38$_{\pm\text{0.03}}$ & 0.39$_{\pm\text{0.04}}$ \\

& & 50 & 0.18$_{\pm\text{0.05}}$ & 0.17$_{\pm\text{0.06}}$ & 0.16$_{\pm\text{0.04}}$ & 0.24$_{\pm\text{0.04}}$ & 0.14$_{\pm\text{0.05}}$ & \textbf{0.31}$_{\pm\text{0.05}}$ & 0.31$_{\pm\text{0.07}}$ & 0.30$_{\pm\text{0.07}}$ \\

& & 100 & 0.11$_{\pm\text{0.04}}$ & 0.17$_{\pm\text{0.04}}$ & 0.17$_{\pm\text{0.04}}$ & 0.22$_{\pm\text{0.06}}$ & 0.10$_{\pm\text{0.02}}$ & 0.21$_{\pm\text{0.02}}$ & \textbf{0.24}$_{\pm\text{0.07}}$ & 0.21$_{\pm\text{0.08}}$ \\

& & 200 & 0.08$_{\pm\text{0.02}}$ & 0.14$_{\pm\text{0.02}}$ & 0.14$_{\pm\text{0.03}}$ & 0.20$_{\pm\text{0.04}}$ & 0.11$_{\pm\text{0.04}}$ & \textbf{0.20}$_{\pm\text{0.04}}$ & 0.13$_{\pm\text{0.08}}$ & 0.15$_{\pm\text{0.07}}$ \\

& & 500 & 0.08$_{\pm\text{0.03}}$ & 0.16$_{\pm\text{0.03}}$ & 0.13$_{\pm\text{0.03}}$ & 0.18$_{\pm\text{0.05}}$ & 0.10$_{\pm\text{0.04}}$ & \textbf{0.18}$_{\pm\text{0.03}}$ & 0.05$_{\pm\text{0.03}}$ & 0.09$_{\pm\text{0.04}}$ \\

\cmidrule{2-11}

& \multirow[c]{5}{*}{steel} & 20 & 0.38$_{\pm\text{0.09}}$ & 0.21$_{\pm\text{0.10}}$ & 0.25$_{\pm\text{0.12}}$ & 0.33$_{\pm\text{0.07}}$ & \textbf{0.48}$_{\pm\text{0.15}}$ & 0.43$_{\pm\text{0.08}}$ & 0.23$_{\pm\text{0.05}}$ & 0.21$_{\pm\text{0.03}}$ \\

& & 50 & 0.27$_{\pm\text{0.11}}$ & 0.27$_{\pm\text{0.09}}$ & 0.22$_{\pm\text{0.09}}$ & 0.34$_{\pm\text{0.05}}$ & \textbf{0.45}$_{\pm\text{0.15}}$ & 0.40$_{\pm\text{0.06}}$ & 0.15$_{\pm\text{0.06}}$ & 0.24$_{\pm\text{0.08}}$ \\

& & 100 & 0.20$_{\pm\text{0.11}}$ & 0.28$_{\pm\text{0.07}}$ & 0.22$_{\pm\text{0.08}}$ & 0.30$_{\pm\text{0.08}}$ & 0.37$_{\pm\text{0.19}}$ & \textbf{0.41}$_{\pm\text{0.07}}$ & 0.15$_{\pm\text{0.09}}$ & 0.25$_{\pm\text{0.10}}$ \\

& & 200 & 0.19$_{\pm\text{0.08}}$ & 0.28$_{\pm\text{0.04}}$ & 0.22$_{\pm\text{0.04}}$ & 0.32$_{\pm\text{0.06}}$ & 0.31$_{\pm\text{0.11}}$ & \textbf{0.40}$_{\pm\text{0.04}}$ & 0.14$_{\pm\text{0.06}}$ & 0.30$_{\pm\text{0.09}}$ \\

& & 500 & 0.17$_{\pm\text{0.04}}$ & 0.29$_{\pm\text{0.07}}$ & 0.24$_{\pm\text{0.07}}$ & 0.32$_{\pm\text{0.05}}$ & 0.21$_{\pm\text{0.07}}$ & \textbf{0.37}$_{\pm\text{0.05}}$ & 0.10$_{\pm\text{0.05}}$ & 0.25$_{\pm\text{0.09}}$ \\

\cmidrule{2-11}

& \multirow[c]{4}{*}{stock} & 20 & 0.24$_{\pm\text{0.05}}$ & 0.37$_{\pm\text{0.08}}$ & 0.42$_{\pm\text{0.07}}$ & 0.46$_{\pm\text{0.05}}$ & 0.45$_{\pm\text{0.12}}$ & \textbf{0.50}$_{\pm\text{0.05}}$ & 0.41$_{\pm\text{0.06}}$ & 0.46$_{\pm\text{0.03}}$ \\

& & 50 & 0.16$_{\pm\text{0.03}}$ & 0.41$_{\pm\text{0.08}}$ & 0.34$_{\pm\text{0.04}}$ & 0.43$_{\pm\text{0.06}}$ & 0.28$_{\pm\text{0.07}}$ & 0.39$_{\pm\text{0.03}}$ & 0.37$_{\pm\text{0.14}}$ & \textbf{0.46}$_{\pm\text{0.02}}$ \\

& & 100 & 0.15$_{\pm\text{0.04}}$ & 0.39$_{\pm\text{0.04}}$ & 0.33$_{\pm\text{0.05}}$ & \textbf{0.46}$_{\pm\text{0.04}}$ & 0.17$_{\pm\text{0.02}}$ & 0.33$_{\pm\text{0.03}}$ & 0.34$_{\pm\text{0.09}}$ & 0.44$_{\pm\text{0.03}}$ \\

& & 200 & 0.14$_{\pm\text{0.02}}$ & 0.38$_{\pm\text{0.04}}$ & 0.28$_{\pm\text{0.03}}$ & 0.45$_{\pm\text{0.05}}$ & 0.11$_{\pm\text{0.01}}$ & 0.32$_{\pm\text{0.03}}$ & 0.39$_{\pm\text{0.11}}$ & \textbf{0.46}$_{\pm\text{0.04}}$ \\

\midrule

\multirow{9}{*}{\rotatebox{90}{\begin{tabular}{l} \textit{More than 10 classes} \end{tabular}}} & \multirow[c]{3}{*}{energy} & 50 & N/A & 0.18$_{\pm\text{0.20}}$ & 0.36$_{\pm\text{0.08}}$ & \textbf{0.48}$_{\pm\text{0.07}}$ & 0.00$_{\pm\text{0.00}}$ & 0.46$_{\pm\text{0.06}}$ & N/A & 0.44$_{\pm\text{0.02}}$ \\

& & 100 & N/A & 0.08$_{\pm\text{0.17}}$ & 0.40$_{\pm\text{0.03}}$ & \textbf{0.46}$_{\pm\text{0.05}}$ & 0.00$_{\pm\text{0.00}}$ & 0.40$_{\pm\text{0.04}}$ & N/A & 0.40$_{\pm\text{0.04}}$ \\

& & 200 & N/A & 0.04$_{\pm\text{0.11}}$ & 0.30$_{\pm\text{0.16}}$ & \textbf{0.45}$_{\pm\text{0.05}}$ & 0.00$_{\pm\text{0.00}}$ & 0.38$_{\pm\text{0.03}}$ & N/A & 0.42$_{\pm\text{0.04}}$ \\

\cmidrule{2-11}

& \multirow[c]{2}{*}{collins} & 100 & N/A & 0.00$_{\pm\text{0.00}}$ & 0.30$_{\pm\text{0.05}}$ & 0.33$_{\pm\text{0.05}}$ & 0.26$_{\pm\text{0.05}}$ & 0.33$_{\pm\text{0.06}}$ & N/A & \textbf{0.38}$_{\pm\text{0.11}}$ \\

& & 200 & 0.18$_{\pm\text{0.03}}$ & 0.00$_{\pm\text{0.00}}$ & 0.23$_{\pm\text{0.06}}$ & 0.29$_{\pm\text{0.08}}$ & 0.19$_{\pm\text{0.03}}$ & 0.33$_{\pm\text{0.09}}$ & N/A & \textbf{0.36}$_{\pm\text{0.12}}$ \\

\cmidrule{2-11}

& \multirow[c]{4}{*}{texture} & 50 & 0.21$_{\pm\text{0.05}}$ & 0.00$_{\pm\text{0.00}}$ & 0.17$_{\pm\text{0.19}}$ & 0.26$_{\pm\text{0.08}}$ & 0.26$_{\pm\text{0.13}}$ & \textbf{0.42}$_{\pm\text{0.07}}$ & N/A & 0.40$_{\pm\text{0.13}}$ \\

& & 100 & 0.16$_{\pm\text{0.04}}$ & 0.08$_{\pm\text{0.14}}$ & 0.35$_{\pm\text{0.06}}$ & 0.26$_{\pm\text{0.03}}$ & 0.37$_{\pm\text{0.09}}$ & \textbf{0.46}$_{\pm\text{0.08}}$ & N/A & 0.42$_{\pm\text{0.10}}$ \\

& & 200 & 0.13$_{\pm\text{0.03}}$ & 0.24$_{\pm\text{0.18}}$ & 0.29$_{\pm\text{0.12}}$ & 0.32$_{\pm\text{0.05}}$ & 0.40$_{\pm\text{0.08}}$ & 0.37$_{\pm\text{0.05}}$ & N/A & \textbf{0.42}$_{\pm\text{0.07}}$ \\

& & 500 & 0.13$_{\pm\text{0.02}}$ & 0.04$_{\pm\text{0.11}}$ & 0.15$_{\pm\text{0.11}}$ & 0.34$_{\pm\text{0.05}}$ & 0.37$_{\pm\text{0.07}}$ & 0.31$_{\pm\text{0.03}}$ & N/A & \textbf{0.44}$_{\pm\text{0.05}}$ \\

\midrule
\rowcolor{Gainsboro!60}
\multicolumn{3}{l|}{\textbf{Average rank}} & 6.64$_{\pm\text{1.35}}$ & 5.48$_{\pm\text{2.08}}$ & 4.82$_{\pm\text{1.42}}$ & 3.45$_{\pm\text{1.75}}$ & 4.03$_{\pm\text{2.82}}$ & \textbf{2.21}$_{\pm\text{1.17}}$ & 5.85$_{\pm\text{1.85}}$ & 3.52$_{\pm\text{1.95}}$ \\

\bottomrule

\end{tabular}
}
\end{table}
\clearpage

\FloatBarrier
\subsubsection{Delta-presence Evaluation}

\begin{table}[hp]
\centering
\caption{\textbf{$\delta$-presence between real train data and synthetic data} on eight real-world tabular datasets with varied real data availability. We report the mean $\pm$ std result and average rank across datasets. A higher rank implies better privacy preservation. Note that ``N/A'' denotes that a specific generator was not applicable, and the rank is computed with the mean result of other methods. We \textbf{bold} the best result for each dataset of different sample sizes. TabEBM achieves the best overall performance against benchmark generators.}
\resizebox{\textwidth}{!}{

\setlength{\tabcolsep}{5pt}
\begin{tabular}{m{0.2cm}lr|rrrrrrr|r}

\toprule

\multicolumn{2}{l}{Datasets}  & $N_{\text{real}}$ & SMOTE & TVAE & CTGAN & NFLOW & TabDDPM & ARF & TabPFGen & \textbf{TabEBM} \\

\midrule

\multirow{24}{*}{\rotatebox{90}{\begin{tabular}{l} \textit{At most 10 classes} \end{tabular}}}& \multirow[c]{5}{*}{protein} & 20 & N/A & \textbf{0.03}$_{\pm\text{0.00}}$ & 0.33$_{\pm\text{0.94}}$ & 0.13$_{\pm\text{0.23}}$ & 0.07$_{\pm\text{0.11}}$ & 0.05$_{\pm\text{0.04}}$ & 0.03$_{\pm\text{0.00}}$ & 0.03$_{\pm\text{0.00}}$ \\

& & 50 & \textbf{0.07}$_{\pm\text{0.00}}$ & 0.07$_{\pm\text{0.00}}$ & 0.07$_{\pm\text{0.00}}$ & 0.39$_{\pm\text{0.92}}$ & 0.11$_{\pm\text{0.10}}$ & 0.09$_{\pm\text{0.03}}$ & 0.07$_{\pm\text{0.00}}$ & 0.07$_{\pm\text{0.00}}$ \\

& & 100 & 0.19$_{\pm\text{0.02}}$ & 0.36$_{\pm\text{0.19}}$ & 0.50$_{\pm\text{0.88}}$ & 2.45$_{\pm\text{3.23}}$ & 0.23$_{\pm\text{0.03}}$ & 0.36$_{\pm\text{0.14}}$ & \textbf{0.17}$_{\pm\text{0.01}}$ & 0.17$_{\pm\text{0.01}}$ \\

& & 200 & 0.62$_{\pm\text{0.23}}$ & 2.84$_{\pm\text{2.07}}$ & 2.03$_{\pm\text{1.05}}$ & 11.55$_{\pm\text{7.74}}$ & 1.71$_{\pm\text{0.72}}$ & 2.73$_{\pm\text{1.46}}$ & 0.57$_{\pm\text{0.17}}$ & 0.57$_{\pm\text{0.17}}$ \\

& & 500 & \textbf{1.15}$_{\pm\text{0.12}}$ & 4.62$_{\pm\text{2.61}}$ & 2.41$_{\pm\text{1.32}}$ & 26.85$_{\pm\text{9.98}}$ & 6.51$_{\pm\text{1.41}}$ & 2.26$_{\pm\text{0.53}}$ & 1.20$_{\pm\text{0.32}}$ & 1.20$_{\pm\text{0.32}}$ \\

\cmidrule{2-11}

& \multirow[c]{5}{*}{fourier} & 20 & N/A & 0.02$_{\pm\text{0.00}}$ & 0.02$_{\pm\text{0.00}}$ & 0.06$_{\pm\text{0.07}}$ & 0.04$_{\pm\text{0.03}}$ & 0.04$_{\pm\text{0.04}}$ & 0.02$_{\pm\text{0.00}}$ & 0.02$_{\pm\text{0.00}}$ \\

& & 50 & 0.08$_{\pm\text{0.00}}$ & 0.10$_{\pm\text{0.01}}$ & 0.09$_{\pm\text{0.01}}$ & 0.09$_{\pm\text{0.01}}$ & 0.08$_{\pm\text{0.00}}$ & 0.08$_{\pm\text{0.00}}$ & 0.08$_{\pm\text{0.00}}$ & 0.08$_{\pm\text{0.00}}$ \\

& & 100 & 0.18$_{\pm\text{0.01}}$ & 0.31$_{\pm\text{0.10}}$ & 0.26$_{\pm\text{0.12}}$ & 2.74$_{\pm\text{2.76}}$ & 0.30$_{\pm\text{0.07}}$ & 0.53$_{\pm\text{0.16}}$ & 0.17$_{\pm\text{0.01}}$ & \textbf{0.17}$_{\pm\text{0.01}}$ \\

& & 200 & 0.73$_{\pm\text{0.48}}$ & 1.56$_{\pm\text{0.63}}$ & 3.52$_{\pm\text{1.67}}$ & 10.60$_{\pm\text{4.58}}$ & 1.52$_{\pm\text{0.67}}$ & 2.52$_{\pm\text{1.14}}$ & 0.45$_{\pm\text{0.05}}$ & \textbf{0.44}$_{\pm\text{0.05}}$ \\

& & 500 & 1.42$_{\pm\text{0.31}}$ & 6.06$_{\pm\text{3.50}}$ & 5.65$_{\pm\text{4.16}}$ & 25.63$_{\pm\text{12.53}}$ & 3.39$_{\pm\text{1.53}}$ & 2.99$_{\pm\text{0.80}}$ & 1.18$_{\pm\text{0.20}}$ & \textbf{1.16}$_{\pm\text{0.18}}$ \\

\cmidrule{2-11}

& \multirow[c]{5}{*}{biodeg} & 20 & 0.03$_{\pm\text{0.00}}$ & 0.09$_{\pm\text{0.10}}$ & 0.12$_{\pm\text{0.15}}$ & 0.22$_{\pm\text{0.32}}$ & \textbf{0.03}$_{\pm\text{0.00}}$ & 0.14$_{\pm\text{0.30}}$ & 0.03$_{\pm\text{0.00}}$ & 0.03$_{\pm\text{0.00}}$ \\

& & 50 & 0.08$_{\pm\text{0.01}}$ & 0.10$_{\pm\text{0.04}}$ & 0.08$_{\pm\text{0.01}}$ & 0.12$_{\pm\text{0.04}}$ & 0.10$_{\pm\text{0.04}}$ & 0.10$_{\pm\text{0.04}}$ & 0.08$_{\pm\text{0.01}}$ & \textbf{0.08}$_{\pm\text{0.00}}$ \\

& & 100 & 0.35$_{\pm\text{0.25}}$ & 0.52$_{\pm\text{0.29}}$ & 0.98$_{\pm\text{0.92}}$ & 1.14$_{\pm\text{0.89}}$ & 0.59$_{\pm\text{0.41}}$ & 0.51$_{\pm\text{0.30}}$ & 0.24$_{\pm\text{0.06}}$ & 0.24$_{\pm\text{0.06}}$ \\

& & 200 & 2.35$_{\pm\text{1.76}}$ & 2.27$_{\pm\text{1.01}}$ & 2.80$_{\pm\text{1.74}}$ & 4.18$_{\pm\text{1.74}}$ & 5.10$_{\pm\text{4.44}}$ & 1.70$_{\pm\text{0.82}}$ & 0.74$_{\pm\text{0.23}}$ & 0.74$_{\pm\text{0.23}}$ \\

& & 500 & 2.91$_{\pm\text{1.43}}$ & 3.95$_{\pm\text{1.08}}$ & 3.58$_{\pm\text{1.71}}$ & 8.85$_{\pm\text{3.07}}$ & 11.92$_{\pm\text{8.19}}$ & 2.37$_{\pm\text{1.04}}$ & \textbf{1.57}$_{\pm\text{0.40}}$ & 1.61$_{\pm\text{0.42}}$ \\

\cmidrule{2-11}

& \multirow[c]{5}{*}{steel} & 20 & 0.04$_{\pm\text{0.01}}$ & 0.08$_{\pm\text{0.08}}$ & 0.10$_{\pm\text{0.12}}$ & 0.08$_{\pm\text{0.06}}$ & 0.06$_{\pm\text{0.02}}$ & 0.15$_{\pm\text{0.30}}$ & 0.03$_{\pm\text{0.00}}$ & 0.03$_{\pm\text{0.00}}$ \\

& & 50 & 0.08$_{\pm\text{0.01}}$ & 0.12$_{\pm\text{0.03}}$ & 0.09$_{\pm\text{0.02}}$ & 0.17$_{\pm\text{0.08}}$ & 1.04$_{\pm\text{0.98}}$ & 0.09$_{\pm\text{0.01}}$ & 0.08$_{\pm\text{0.01}}$ & 0.08$_{\pm\text{0.01}}$ \\

& & 100 & 0.22$_{\pm\text{0.03}}$ & 0.39$_{\pm\text{0.12}}$ & 0.44$_{\pm\text{0.40}}$ & 0.73$_{\pm\text{0.30}}$ & 2.17$_{\pm\text{2.47}}$ & 0.27$_{\pm\text{0.06}}$ & \textbf{0.20}$_{\pm\text{0.03}}$ & 0.20$_{\pm\text{0.03}}$ \\

& & 200 & 1.39$_{\pm\text{0.78}}$ & 2.05$_{\pm\text{0.99}}$ & 3.39$_{\pm\text{2.50}}$ & 5.55$_{\pm\text{3.58}}$ & 6.68$_{\pm\text{4.55}}$ & 1.29$_{\pm\text{0.69}}$ & 0.49$_{\pm\text{0.07}}$ & \textbf{0.48}$_{\pm\text{0.06}}$ \\

& & 500 & 1.93$_{\pm\text{0.60}}$ & 4.39$_{\pm\text{3.30}}$ & 5.23$_{\pm\text{3.90}}$ & 10.52$_{\pm\text{7.22}}$ & 33.92$_{\pm\text{25.31}}$ & 2.39$_{\pm\text{0.88}}$ & \textbf{1.68}$_{\pm\text{0.88}}$ & 1.70$_{\pm\text{0.89}}$ \\

\cmidrule{2-11}

& \multirow[c]{4}{*}{stock} & 20 & 0.03$_{\pm\text{0.00}}$ & 0.03$_{\pm\text{0.00}}$ & 0.03$_{\pm\text{0.00}}$ & 0.03$_{\pm\text{0.00}}$ & 0.04$_{\pm\text{0.00}}$ & \textbf{0.03}$_{\pm\text{0.00}}$ & 0.03$_{\pm\text{0.00}}$ & 0.03$_{\pm\text{0.00}}$ \\

& & 50 & 0.08$_{\pm\text{0.00}}$ & 0.09$_{\pm\text{0.00}}$ & 0.09$_{\pm\text{0.00}}$ & 0.08$_{\pm\text{0.00}}$ & 0.09$_{\pm\text{0.01}}$ & 0.08$_{\pm\text{0.00}}$ & 0.08$_{\pm\text{0.00}}$ & \textbf{0.08}$_{\pm\text{0.00}}$ \\

& & 100 & 0.19$_{\pm\text{0.02}}$ & 0.23$_{\pm\text{0.04}}$ & 0.26$_{\pm\text{0.12}}$ & 0.23$_{\pm\text{0.04}}$ & 0.22$_{\pm\text{0.04}}$ & 0.20$_{\pm\text{0.02}}$ & 0.18$_{\pm\text{0.01}}$ & \textbf{0.18}$_{\pm\text{0.01}}$ \\

& & 200 & 0.51$_{\pm\text{0.08}}$ & 0.97$_{\pm\text{0.30}}$ & 1.83$_{\pm\text{1.14}}$ & 0.90$_{\pm\text{0.32}}$ & 0.57$_{\pm\text{0.07}}$ & 0.60$_{\pm\text{0.16}}$ & 0.49$_{\pm\text{0.07}}$ & \textbf{0.48}$_{\pm\text{0.08}}$ \\

\midrule
\multirow{9}{*}{\rotatebox{90}{\begin{tabular}{l} \textit{More than 10 classes} \end{tabular}}} & \multirow[c]{3}{*}{energy} & 50 & N/A & 0.09$_{\pm\text{0.02}}$ & 0.08$_{\pm\text{0.00}}$ & 0.08$_{\pm\text{0.00}}$ & \textbf{0.06}$_{\pm\text{0.00}}$ & 0.08$_{\pm\text{0.00}}$ & N/A & 0.08$_{\pm\text{0.00}}$ \\

& & 100 & N/A & 1.56$_{\pm\text{1.51}}$ & 0.19$_{\pm\text{0.03}}$ & 0.19$_{\pm\text{0.03}}$ & 0.99$_{\pm\text{2.46}}$ & 0.16$_{\pm\text{0.01}}$ & N/A & \textbf{0.16}$_{\pm\text{0.00}}$ \\

& & 200 & N/A & 4.15$_{\pm\text{3.04}}$ & 1.67$_{\pm\text{0.84}}$ & 0.72$_{\pm\text{0.16}}$ & 13.28$_{\pm\text{9.21}}$ & 0.44$_{\pm\text{0.04}}$ & N/A & \textbf{0.38}$_{\pm\text{0.03}}$ \\

\cmidrule{2-11}

& \multirow[c]{2}{*}{collins} & 100 & N/A & 1.79$_{\pm\text{1.23}}$ & 0.18$_{\pm\text{0.02}}$ & 0.18$_{\pm\text{0.03}}$ & 0.17$_{\pm\text{0.02}}$ & 0.17$_{\pm\text{0.01}}$ & N/A & \textbf{0.16}$_{\pm\text{0.01}}$ \\

& & 200 & 0.43$_{\pm\text{0.08}}$ & 4.20$_{\pm\text{2.55}}$ & 0.76$_{\pm\text{0.26}}$ & 1.73$_{\pm\text{1.09}}$ & 0.94$_{\pm\text{0.59}}$ & 0.95$_{\pm\text{0.47}}$ & N/A & \textbf{0.39}$_{\pm\text{0.05}}$ \\

\cmidrule{2-11}

& \multirow[c]{4}{*}{texture} & 50 & 0.08$_{\pm\text{0.00}}$ & 0.08$_{\pm\text{0.00}}$ & 0.08$_{\pm\text{0.00}}$ & \textbf{0.08}$_{\pm\text{0.00}}$ & 0.10$_{\pm\text{0.02}}$ & 0.08$_{\pm\text{0.00}}$ & N/A & 0.08$_{\pm\text{0.00}}$ \\

& & 100 & 0.17$_{\pm\text{0.01}}$ & 0.42$_{\pm\text{0.22}}$ & 0.19$_{\pm\text{0.02}}$ & 0.41$_{\pm\text{0.16}}$ & 0.44$_{\pm\text{0.32}}$ & 0.25$_{\pm\text{0.05}}$ & N/A & \textbf{0.17}$_{\pm\text{0.01}}$ \\

& & 200 & 0.57$_{\pm\text{0.18}}$ & 2.34$_{\pm\text{1.14}}$ & 1.76$_{\pm\text{1.05}}$ & 5.78$_{\pm\text{2.82}}$ & 7.02$_{\pm\text{4.14}}$ & 1.54$_{\pm\text{0.94}}$ & N/A & \textbf{0.45}$_{\pm\text{0.08}}$ \\

& & 500 & 1.33$_{\pm\text{0.61}}$ & 2.88$_{\pm\text{1.09}}$ & 2.42$_{\pm\text{1.35}}$ & 23.40$_{\pm\text{10.74}}$ & 16.64$_{\pm\text{12.35}}$ & 2.36$_{\pm\text{1.06}}$ & N/A & \textbf{1.03}$_{\pm\text{0.14}}$ \\

\midrule
\rowcolor{Gainsboro!60}
\multicolumn{3}{l|}{\textbf{Average rank}} & 3.30$_{\pm\text{1.37}}$ & 5.91$_{\pm\text{1.68}}$ & 5.30$_{\pm\text{1.74}}$ & 6.45$_{\pm\text{1.86}}$ & 5.82$_{\pm\text{2.11}}$ & 4.45$_{\pm\text{1.66}}$ & 3.00$_{\pm\text{1.90}}$ & \textbf{1.76}$_{\pm\text{1.05}}$ \\

\bottomrule

\end{tabular}
}
\end{table}
\clearpage

%% file: checklist.tex
\section*{NeurIPS Paper Checklist}

\begin{enumerate}

\item {\bf Claims}
    \item[] Question: Do the main claims made in the abstract and introduction accurately reflect the paper's contributions and scope?
    \item[] Answer: \answerYes{} 
    \item[] Justification: \cref{sec:intro} details our research objectives and highlights our contributions.

\item {\bf Limitations}
    \item[] Question: Does the paper discuss the limitations of the work performed by the authors?
    \item[] Answer: \answerYes{} 
    \item[] Justification: Presented in \cref{sec:related-work}.

\item {\bf Theory Assumptions and Proofs}
    \item[] Question: For each theoretical result, does the paper provide the full set of assumptions and a complete (and correct) proof?
    \item[] Answer: \answerYes{} 
    \item[] Justification: \cref{sec:method} presents the theoretical results of our proposed method.

    \item {\bf Experimental Result Reproducibility}
    \item[] Question: Does the paper fully disclose all the information needed to reproduce the main experimental results of the paper to the extent that it affects the main claims and/or conclusions of the paper (regardless of whether the code and data are provided or not)?
    \item[] Answer: \answerYes{} 
    \item[] Justification: Refer to \cref{appendix:reproducibility}, where we provide full details on reproducing the results in the paper. We provide an open-source library of the proposed method.

\item {\bf Open access to data and code}
    \item[] Question: Does the paper provide open access to the data and code, with sufficient instructions to faithfully reproduce the main experimental results, as described in supplemental material?
    \item[] Answer: \answerYes{} 
    \item[] Justification: Refer to \cref{appendix:reproducibility}. All datasets used in this paper are publicly available, and the implementations of benchmark generators are open-source. We also provide an open-source library \url{https://github.com/andreimargeloiu/TabEBM} . 

\item {\bf Experimental Setting/Details}
    \item[] Question: Does the paper specify all the training and test details (e.g., data splits, hyperparameters, how they were chosen, type of optimizer, etc.) necessary to understand the results?
    \item[] Answer: \answerYes{} 
    \item[] Justification: \cref{appendix:reproducibility} provides full descriptions of the experimental setup. \looseness-1

\item {\bf Experiment Statistical Significance}
    \item[] Question: Does the paper report error bars suitably and correctly defined or other appropriate information about the statistical significance of the experiments?
    \item[] Answer: \answerYes{} 
    \item[] Justification: Refer to \cref{sec:exp}, where we provide standard deviations for all tables. \cref{fig:acc_vs_time} and \cref{fig:acc_vs_sample_size_and_classes_and_time} (Right) contain error bars. Due to the page limit, the error bars for all other figures are available in \cref{appendix:numerical_results}. In \cref{sec:exp_fidelity} and \cref{appendix:res_fidelity}, we show statistical significance tests of the similarity between real data and synthetic data.    

\item {\bf Experiments Compute Resources}
    \item[] Question: For each experiment, does the paper provide sufficient information on the computer resources (type of compute workers, memory, time of execution) needed to reproduce the experiments?
    \item[] Answer: \answerYes{} 
    \item[] Justification: Refer to \cref{appendix:software-and-computing}, where we provide full details on the computation resources used in the paper.
    
\item {\bf Code Of Ethics}
    \item[] Question: Does the research conducted in the paper conform, in every respect, with the NeurIPS Code of Ethics \url{https://neurips.cc/public/EthicsGuidelines}?
    \item[] Answer: \answerYes{} 
    \item[] Justification: We carefully check the NeurIPS Code of Ethics, and we confirm that our work follows the Code in every respect.

\item {\bf Broader Impacts}
    \item[] Question: Does the paper discuss both potential positive societal impacts and negative societal impacts of the work performed?
    \item[] Answer: \answerYes{} 
    \item[] Justification: Refer to \cref{sec:border_impacts}, where we include the societal impacts of our work.

\item {\bf Safeguards}
    \item[] Question: Does the paper describe safeguards that have been put in place for responsible release of data or models that have a high risk for misuse (e.g., pretrained language models, image generators, or scraped datasets)?
    \item[] Answer: \answerNA{} 
    \item[] Justification: \answerNA{}

\item {\bf Licenses for existing assets}
    \item[] Question: Are the creators or original owners of assets (e.g., code, data, models), used in the paper, properly credited and are the license and terms of use explicitly mentioned and properly respected?
    \item[] Answer: \answerYes{} 
    \item[] Justification: Refer to \cref{appendix:reproducibility}, where we provide the open-source licenses followed by the creators or original owners of assets.

\item {\bf New Assets}
    \item[] Question: Are new assets introduced in the paper well documented and is the documentation provided alongside the assets?
    \item[] Answer: \answerYes{} 
    \item[] Justification: We provide the implementation of our method as a python library attached to this submission. We will make it publicly available post-publication.

\item {\bf Crowdsourcing and Research with Human Subjects}
    \item[] Question: For crowdsourcing experiments and research with human subjects, does the paper include the full text of instructions given to participants and screenshots, if applicable, as well as details about compensation (if any)? 
    \item[] Answer: \answerNA{} 
    \item[] Justification: \answerNA{}
\item {\bf Institutional Review Board (IRB) Approvals or Equivalent for Research with Human Subjects}
    \item[] Question: Does the paper describe potential risks incurred by study participants, whether such risks were disclosed to the subjects, and whether Institutional Review Board (IRB) approvals (or an equivalent approval/review based on the requirements of your country or institution) were obtained?
    \item[] Answer: \answerNo{} 
    \item[] Justification:\answerNA{}

\end{enumerate}

%% file: neurips_2024.bbl
\begin{thebibliography}{10}

\bibitem{alami2020artificial}
Hassane Alami, Lysanne Rivard, Pascale Lehoux, Steven~J Hoffman, Stephanie Bernadette~Mafalda Cadeddu, Mathilde Savoldelli, Mamane~Abdoulaye Samri, Mohamed~Ali Ag~Ahmed, Richard Fleet, and Jean-Paul Fortin.
\newblock Artificial intelligence in health care: laying the foundation for responsible, sustainable, and inclusive innovation in low-and middle-income countries.
\newblock {\em Globalization and Health}, 16:1--6, 2020.

\bibitem{antoniou2017data}
Antreas Antoniou, Amos Storkey, and Harrison Edwards.
\newblock Data augmentation generative adversarial networks.
\newblock {\em arXiv preprint arXiv:1711.04340}, 2017.

\bibitem{balestriero2022effects}
Randall Balestriero, Leon Bottou, and Yann LeCun.
\newblock The effects of regularization and data augmentation are class dependent.
\newblock {\em Advances in Neural Information Processing Systems}, 35:37878--37891, 2022.

\bibitem{bansal2022systematic}
Ms~Aayushi Bansal, Dr~Rewa Sharma, and Dr~Mamta Kathuria.
\newblock A systematic review on data scarcity problem in deep learning: solution and applications.
\newblock {\em ACM Computing Surveys (CSUR)}, 54(10s):1--29, 2022.

\bibitem{baxevanis2020bioinformatics}
Andreas~D Baxevanis, Gary~D Bader, and David~S Wishart.
\newblock {\em Bioinformatics}.
\newblock John Wiley \& Sons, 2020.

\bibitem{bespalov2016failed}
Anton Bespalov, Thomas Steckler, Bruce Altevogt, Elena Koustova, Phil Skolnick, Daniel Deaver, Mark~J Millan, Jesper~F Bastlund, Dario Doller, Jeffrey Witkin, et~al.
\newblock Failed trials for central nervous system disorders do not necessarily invalidate preclinical models and drug targets.
\newblock {\em Nature Reviews Drug Discovery}, 15(7):516--516, 2016.

\bibitem{OpenML}
B.~Bischl, Giuseppe Casalicchio, Matthias Feurer, Frank Hutter, Michel Lang, Rafael~Gomes Mantovani, Jan~N. van Rijn, and Joaquin Vanschoren.
\newblock Openml benchmarking suites.
\newblock {\em Proceedings of the Neural Information Processing Systems Track on Datasets and Benchmarks (NeurIPS 2021)}, 2021.

\bibitem{borisov2022deep}
Vadim Borisov, Tobias Leemann, Kathrin Se{\ss}ler, Johannes Haug, Martin Pawelczyk, and Gjergji Kasneci.
\newblock Deep neural networks and tabular data: A survey.
\newblock {\em IEEE Transactions on Neural Networks and Learning Systems}, 2022.

\bibitem{borisov2022language}
Vadim Borisov, Kathrin Sessler, Tobias Leemann, Martin Pawelczyk, and Gjergji Kasneci.
\newblock Language models are realistic tabular data generators.
\newblock In {\em The Eleventh International Conference on Learning Representations}, 2022.

\bibitem{breiman2001random}
Leo Breiman.
\newblock Random forests.
\newblock {\em Machine learning}, 45(1):5--32, 2001.

\bibitem{cai2020transfer}
Chenjing Cai, Shiwei Wang, Youjun Xu, Weilin Zhang, Ke~Tang, Qi~Ouyang, Luhua Lai, and Jianfeng Pei.
\newblock Transfer learning for drug discovery.
\newblock {\em Journal of Medicinal Chemistry}, 63(16):8683--8694, 2020.

\bibitem{chang2022towards}
Rees Chang, Yu-Xiong Wang, and Elif Ertekin.
\newblock Towards overcoming data scarcity in materials science: unifying models and datasets with a mixture of experts framework.
\newblock {\em npj Computational Materials}, 8(1):242, 2022.

\bibitem{chawla2002smote}
Nitesh~V Chawla, Kevin~W Bowyer, Lawrence~O Hall, and W~Philip Kegelmeyer.
\newblock Smote: synthetic minority over-sampling technique.
\newblock {\em Journal of artificial intelligence research}, 16:321--357, 2002.

\bibitem{chen2016xgboost}
Tianqi Chen and Carlos Guestrin.
\newblock Xgboost: A scalable tree boosting system.
\newblock In {\em Proceedings of the 22nd acm sigkdd international conference on knowledge discovery and data mining}, pages 785--794, 2016.

\bibitem{ciecierski2022artificial}
Tadeusz Ciecierski-Holmes, Ritvij Singh, Miriam Axt, Stephan Brenner, and Sandra Barteit.
\newblock Artificial intelligence for strengthening healthcare systems in low-and middle-income countries: a systematic scoping review.
\newblock {\em npj Digital Medicine}, 5(1):162, 2022.

\bibitem{cox1958regression}
David~R Cox.
\newblock The regression analysis of binary sequences.
\newblock {\em Journal of the Royal Statistical Society Series B: Statistical Methodology}, 20(2):215--232, 1958.

\bibitem{csiszar1975divergence}
Imre Csisz{\'a}r.
\newblock I-divergence geometry of probability distributions and minimization problems.
\newblock {\em The annals of probability}, pages 146--158, 1975.

\bibitem{dekoninck2024evading}
Jasper Dekoninck, Mark~Niklas M{\"u}ller, Maximilian Baader, Marc Fischer, and Martin Vechev.
\newblock Evading data contamination detection for language models is (too) easy.
\newblock {\em arXiv preprint arXiv:2402.02823}, 2024.

\bibitem{deng2024investigating}
Chunyuan Deng, Yilun Zhao, Xiangru Tang, Mark~B. Gerstein, and Arman Cohan.
\newblock Investigating data contamination in modern benchmarks for large language models.
\newblock In {\em North American Chapter of the Association for Computational Linguistics}, 2024.

\bibitem{dooley2024forecastpfn}
Samuel Dooley, Gurnoor~Singh Khurana, Chirag Mohapatra, Siddartha~V Naidu, and Colin White.
\newblock Forecastpfn: Synthetically-trained zero-shot forecasting.
\newblock {\em Advances in Neural Information Processing Systems}, 36, 2024.

\bibitem{douzas2018effective}
Georgios Douzas and Fernando Bacao.
\newblock Effective data generation for imbalanced learning using conditional generative adversarial networks.
\newblock {\em Expert Systems with applications}, 91:464--471, 2018.

\bibitem{dua2019uci}
Dheeru Dua and Casey Graff.
\newblock Uci machine learning repository, 2017.

\bibitem{dunning2012privacy}
Larry~A Dunning and Ray Kresman.
\newblock Privacy preserving data sharing with anonymous id assignment.
\newblock {\em IEEE transactions on information forensics and security}, 8(2):402--413, 2012.

\bibitem{durkan2019neural}
Conor Durkan, Artur Bekasov, Iain Murray, and George Papamakarios.
\newblock Neural spline flows.
\newblock {\em Advances in neural information processing systems}, 32, 2019.

\bibitem{fang2024large}
Xi~Fang, Weijie Xu, Fiona~Anting Tan, Ziqing Hu, Jiani Zhang, Yanjun Qi, Srinivasan~H. Sengamedu, and Christos Faloutsos.
\newblock Large language models (llms) on tabular data: Prediction, generation, and understanding - a survey.
\newblock {\em Transactions on Machine Learning Research}, 2024.

\bibitem{feng2021survey}
Steven~Y Feng, Varun Gangal, Jason Wei, Sarath Chandar, Soroush Vosoughi, Teruko Mitamura, and Eduard Hovy.
\newblock A survey of data augmentation approaches for nlp.
\newblock In {\em Findings of the Association for Computational Linguistics: ACL-IJCNLP 2021}, pages 968--988, 2021.

\bibitem{fix1985discriminatory}
Evelyn Fix.
\newblock {\em Discriminatory analysis: nonparametric discrimination, consistency properties}, volume~1.
\newblock USAF school of Aviation Medicine, 1985.

\bibitem{gorishniy2021revisiting}
Yury Gorishniy, Ivan Rubachev, Valentin Khrulkov, and Artem Babenko.
\newblock Revisiting deep learning models for tabular data.
\newblock {\em Advances in Neural Information Processing Systems}, 34:18932--18943, 2021.

\bibitem{JEM}
Will Grathwohl, Kuan-Chieh Wang, Joern-Henrik Jacobsen, David Duvenaud, Mohammad Norouzi, and Kevin Swersky.
\newblock Your classifier is secretly an energy based model and you should treat it like one.
\newblock In {\em International Conference on Learning Representations}, 2019.

\bibitem{grinsztajn2022tree}
L{\'e}o Grinsztajn, Edouard Oyallon, and Ga{\"e}l Varoquaux.
\newblock Why do tree-based models still outperform deep learning on typical tabular data?
\newblock {\em Advances in neural information processing systems}, 35:507--520, 2022.

\bibitem{hansen2023reimagining}
Lasse Hansen, Nabeel Seedat, Mihaela van~der Schaar, and Andrija Petrovic.
\newblock Reimagining synthetic tabular data generation through data-centric ai: A comprehensive benchmark.
\newblock {\em Advances in Neural Information Processing Systems}, 36:33781--33823, 2023.

\bibitem{hernandez2022synthetic}
Mikel Hernandez, Gorka Epelde, Ane Alberdi, Rodrigo Cilla, and Debbie Rankin.
\newblock Synthetic data generation for tabular health records: A systematic review.
\newblock {\em Neurocomputing}, 493:28--45, 2022.

\bibitem{TabPFN}
Noah Hollmann, Samuel M{\"u}ller, Katharina Eggensperger, and Frank Hutter.
\newblock Tab{PFN}: A transformer that solves small tabular classification problems in a second.
\newblock In {\em The Eleventh International Conference on Learning Representations}, 2023.

\bibitem{matplotlib}
J.~D. Hunter.
\newblock Matplotlib: A 2d graphics environment.
\newblock {\em Computing in Science \& Engineering}, 9(3):90--95, 2007.

\bibitem{jha2019enhancing}
Dipendra Jha, Kamal Choudhary, Francesca Tavazza, Wei-keng Liao, Alok Choudhary, Carelyn Campbell, and Ankit Agrawal.
\newblock Enhancing materials property prediction by leveraging computational and experimental data using deep transfer learning.
\newblock {\em Nature communications}, 10(1):5316, 2019.

\bibitem{jiang2024investigating}
Minhao Jiang, Ken~Ziyu Liu, Ming Zhong, Rylan Schaeffer, Siru Ouyang, Jiawei Han, and Sanmi Koyejo.
\newblock Investigating data contamination for pre-training language models.
\newblock {\em arXiv preprint arXiv:2401.06059}, 2024.

\bibitem{jiang2024protogate}
Xiangjian Jiang, Andrei Margeloiu, Nikola Simidjievski, and Mateja Jamnik.
\newblock Protogate: Prototype-based neural networks with global-to-local feature selection for tabular biomedical data.
\newblock In {\em Proceedings of the 41st International Conference on Machine Learning (ICML)}, 2024.

\bibitem{jin2019review}
Hao Jin, Yan Luo, Peilong Li, and Jomol Mathew.
\newblock A review of secure and privacy-preserving medical data sharing.
\newblock {\em IEEE access}, 7:61656--61669, 2019.

\bibitem{karson1968handbook}
Marvin Karson.
\newblock Handbook of methods of applied statistics. volume i: Techniques of computation descriptive methods, and statistical inference. volume ii: Planning of surveys and experiments. im chakravarti, rg laha, and j. roy, new york, john wiley; 1967., 1968.

\bibitem{kim2022stasy}
Jayoung Kim, Chaejeong Lee, and Noseong Park.
\newblock Stasy: Score-based tabular data synthesis.
\newblock In {\em The Eleventh International Conference on Learning Representations}, 2022.

\bibitem{kirichenko2024understanding}
Polina Kirichenko, Mark Ibrahim, Randall Balestriero, Diane Bouchacourt, Shanmukha~Ramakrishna Vedantam, Hamed Firooz, and Andrew~G Wilson.
\newblock Understanding the detrimental class-level effects of data augmentation.
\newblock {\em Advances in Neural Information Processing Systems}, 36, 2024.

\bibitem{kotelnikov2023tabddpm}
Akim Kotelnikov, Dmitry Baranchuk, Ivan Rubachev, and Artem Babenko.
\newblock Tabddpm: Modelling tabular data with diffusion models.
\newblock In {\em International Conference on Machine Learning}, pages 17564--17579. PMLR, 2023.

\bibitem{EBM2006LeCun}
Yann LeCun, Sumit Chopra, Raia Hadsell, M~Ranzato, and Fujie Huang.
\newblock A tutorial on energy-based learning.
\newblock {\em Predicting structured data}, 1(0), 2006.

\bibitem{lee2023codi}
Chaejeong Lee, Jayoung Kim, and Noseong Park.
\newblock Codi: Co-evolving contrastive diffusion models for mixed-type tabular synthesis.
\newblock In {\em International Conference on Machine Learning}, pages 18940--18956. PMLR, 2023.

\bibitem{imbalanced-learn}
Guillaume Lema{{\^i}}tre, Fernando Nogueira, and Christos~K. Aridas.
\newblock Imbalanced-learn: A python toolbox to tackle the curse of imbalanced datasets in machine learning.
\newblock {\em Journal of Machine Learning Research}, 18(17):1--5, 2017.

\bibitem{levin2022transfer}
Roman Levin, Valeriia Cherepanova, Avi Schwarzschild, Arpit Bansal, C~Bayan Bruss, Tom Goldstein, Andrew~Gordon Wilson, and Micah Goldblum.
\newblock Transfer learning with deep tabular models.
\newblock In {\em The Eleventh International Conference on Learning Representations}, 2022.

\bibitem{liu2023goggle}
Tennison Liu, Zhaozhi Qian, Jeroen Berrevoets, and Mihaela van~der Schaar.
\newblock Goggle: Generative modelling for tabular data by learning relational structure.
\newblock In {\em The Eleventh International Conference on Learning Representations}, 2023.

\bibitem{TabPFGen}
Junwei Ma, Apoorv Dankar, George Stein, Guangwei Yu, and Anthony Caterini.
\newblock Tabpfgen--tabular data generation with tabpfn.
\newblock In {\em NeurIPS 2023 Second Table Representation Learning Workshop}, 2023.

\bibitem{magar2022data}
Inbal Magar and Roy Schwartz.
\newblock Data contamination: From memorization to exploitation.
\newblock In {\em Proceedings of the 60th Annual Meeting of the Association for Computational Linguistics (Volume 2: Short Papers)}, pages 157--165, 2022.

\bibitem{malin2013biomedical}
Bradley~A Malin, Khaled~El Emam, and Christine~M O'Keefe.
\newblock Biomedical data privacy: problems, perspectives, and recent advances.
\newblock {\em Journal of the American medical informatics association}, 20(1):2--6, 2013.

\bibitem{manousakas2023usefulness}
Dionysis Manousakas and Serg{\"u}l Ayd{\"o}re.
\newblock On the usefulness of synthetic tabular data generation.
\newblock In {\em Data-centric Machine Learning Research (DMLR) Workshop at the 40th International Conference on Machine Learning (ICML)}, 2023.

\bibitem{margeloiu2022weight}
Andrei Margeloiu, Nikola Simidjievski, Pietro Lio, and Mateja Jamnik.
\newblock Weight predictor network with feature selection for small sample tabular biomedical data.
\newblock {\em AAAI Conference on Artificial Intelligence}, 2023.

\bibitem{mccarter2024whatexactlyhas}
Calvin McCarter.
\newblock What exactly has tabpfn learned to do?
\newblock In {\em ICLR Blogposts 2024}, 2024.

\bibitem{mcelfresh2024neural}
Duncan McElfresh, Sujay Khandagale, Jonathan Valverde, Vishak Prasad~C, Ganesh Ramakrishnan, Micah Goldblum, and Colin White.
\newblock When do neural nets outperform boosted trees on tabular data?
\newblock {\em Advances in Neural Information Processing Systems}, 36, 2024.

\bibitem{mchugh2013chi}
Mary~L McHugh.
\newblock The chi-square test of independence.
\newblock {\em Biochemia medica}, 23(2):143--149, 2013.

\bibitem{micci2001preprocessing}
Daniele Micci-Barreca.
\newblock A preprocessing scheme for high-cardinality categorical attributes in classification and prediction problems.
\newblock {\em ACM SIGKDD explorations newsletter}, 3(1):27--32, 2001.

\bibitem{mollura2020artificial}
Daniel~J Mollura, Melissa~P Culp, Erica Pollack, Gillian Battino, John~R Scheel, Victoria~L Mango, Ameena Elahi, Alan Schweitzer, and Farouk Dako.
\newblock Artificial intelligence in low-and middle-income countries: innovating global health radiology.
\newblock {\em Radiology}, 297(3):513--520, 2020.

\bibitem{morford2011preclinical}
LaRonda~L Morford, Christopher~J Bowman, Diann~L Blanset, Ingrid~B B{\o}gh, Gary~J Chellman, Wendy~G Halpern, Gerhard~F Weinbauer, and Timothy~P Coogan.
\newblock Preclinical safety evaluations supporting pediatric drug development with biopharmaceuticals: strategy, challenges, current practices.
\newblock {\em Birth Defects Research Part B: Developmental and Reproductive Toxicology}, 92(4):359--380, 2011.

\bibitem{PFN}
Samuel M{\"u}ller, Noah Hollmann, Sebastian~Pineda Arango, Josif Grabocka, and Frank Hutter.
\newblock Transformers can do bayesian inference.
\newblock In {\em International Conference on Learning Representations}, 2022.

\bibitem{mumuni2022data}
Alhassan Mumuni and Fuseini Mumuni.
\newblock Data augmentation: A comprehensive survey of modern approaches.
\newblock {\em Array}, 16:100258, 2022.

\bibitem{nagler2023statistical}
Thomas Nagler.
\newblock Statistical foundations of prior-data fitted networks.
\newblock In {\em International Conference on Machine Learning}, pages 25660--25676. PMLR, 2023.

\bibitem{nergiz2019delta}
Mehmet~Ercan Nergiz, Maurizio Atzori, and Christopher~W Clifton.
\newblock \(\delta\)-presence.
\newblock {\em Encyclopedia of Cryptography, Security and Privacy}, pages 1--5, 2019.

\bibitem{park2018data}
Noseong Park, Mahmoud Mohammadi, Kshitij Gorde, Sushil Jajodia, Hongkyu Park, and Youngmin Kim.
\newblock Data synthesis based on generative adversarial networks.
\newblock {\em Proceedings of the VLDB Endowment}, 11(10), 2018.

\bibitem{paszke2019pytorch}
Adam Paszke, Sam Gross, Francisco Massa, Adam Lerer, James Bradbury, Gregory Chanan, Trevor Killeen, Zeming Lin, Natalia Gimelshein, Luca Antiga, et~al.
\newblock Pytorch: An imperative style, high-performance deep learning library.
\newblock {\em Advances in Neural Information Processing Systems}, 32, 2019.

\bibitem{scikit-learn}
F.~Pedregosa, G.~Varoquaux, A.~Gramfort, V.~Michel, B.~Thirion, O.~Grisel, M.~Blondel, P.~Prettenhofer, R.~Weiss, V.~Dubourg, J.~Vanderplas, A.~Passos, D.~Cournapeau, M.~Brucher, M.~Perrot, and E.~Duchesnay.
\newblock Scikit-learn: Machine learning in {P}ython.
\newblock {\em Journal of Machine Learning Research}, 12:2825--2830, 2011.

\bibitem{prokhorenkova2017catboost}
Liudmila Prokhorenkova, Gleb Gusev, Aleksandr Vorobev, Anna~Veronika Dorogush, and Andrey Gulin.
\newblock Catboost: unbiased boosting with categorical features.
\newblock {\em Advances in neural information processing systems}, 31, 2018.

\bibitem{qian2024synthcity}
Zhaozhi Qian, Rob Davis, and Mihaela van~der Schaar.
\newblock Synthcity: a benchmark framework for diverse use cases of tabular synthetic data.
\newblock {\em Advances in Neural Information Processing Systems}, 36, 2024.

\bibitem{qin2023class}
Yiming Qin, Huangjie Zheng, Jiangchao Yao, Mingyuan Zhou, and Ya~Zhang.
\newblock Class-balancing diffusion models.
\newblock In {\em Proceedings of the IEEE/CVF Conference on Computer Vision and Pattern Recognition}, pages 18434--18443, 2023.

\bibitem{raschka2018model}
Sebastian Raschka.
\newblock Model evaluation, model selection, and algorithm selection in machine learning.
\newblock {\em arXiv preprint arXiv:1811.12808}, 2018.

\bibitem{sampath2021survey}
Vignesh Sampath, I{\~n}aki Maurtua, Juan~Jose Aguilar~Martin, and Aitor Gutierrez.
\newblock A survey on generative adversarial networks for imbalance problems in computer vision tasks.
\newblock {\em Journal of big Data}, 8:1--59, 2021.

\bibitem{seedat2024curated}
Nabeel Seedat, Nicolas Huynh, Boris van Breugel, and Mihaela van~der Schaar.
\newblock Curated llm: Synergy of llms and data curation for tabular augmentation in low-data regimes.
\newblock In {\em Forty-first International Conference on Machine Learning}, 2024.

\bibitem{shorten2019survey}
Connor Shorten and Taghi~M Khoshgoftaar.
\newblock A survey on image data augmentation for deep learning.
\newblock {\em Journal of big data}, 6(1):1--48, 2019.

\bibitem{shorten2021text}
Connor Shorten, Taghi~M Khoshgoftaar, and Borko Furht.
\newblock Text data augmentation for deep learning.
\newblock {\em Journal of big Data}, 8(1):101, 2021.

\bibitem{shwartz2022tabular}
Ravid Shwartz-Ziv and Amitai Armon.
\newblock Tabular data: Deep learning is not all you need.
\newblock {\em Information Fusion}, 81:84--90, 2022.

\bibitem{stadler2022search}
Theresa Stadler and Carmela Troncoso.
\newblock Why the search for a privacy-preserving data sharing mechanism is failing.
\newblock {\em Nature Computational Science}, 2(4):208--210, 2022.

\bibitem{sufi2024addressing}
Fahim Sufi.
\newblock Addressing data scarcity in the medical domain: A gpt-based approach for synthetic data generation and feature extraction.
\newblock {\em Information}, 15(5):264, 2024.

\bibitem{sun2023private}
Haoyuan Sun, Navid Azizan, Akash Srivastava, and Hao Wang.
\newblock Private synthetic data meets ensemble learning.
\newblock {\em arXiv preprint arXiv:2310.09729}, 2023.

\bibitem{tu2024causality}
Ruibo Tu, Zineb Senane, Lele Cao, Cheng Zhang, Hedvig Kjellstr{\"o}m, and Gustav~Eje Henter.
\newblock Causality for tabular data synthesis: A high-order structure causal benchmark framework.
\newblock {\em arXiv preprint arXiv:2406.08311}, 2024.

\bibitem{ubbens2023gpfn}
Jordan Ubbens, Ian Stavness, and Andrew~G Sharpe.
\newblock Gpfn: Prior-data fitted networks for genomic prediction.
\newblock {\em bioRxiv}, pages 2023--09, 2023.

\bibitem{van2023synthetic}
Boris Van~Breugel, Zhaozhi Qian, and Mihaela Van Der~Schaar.
\newblock Synthetic data, real errors: how (not) to publish and use synthetic data.
\newblock In {\em International Conference on Machine Learning}, pages 34793--34808. PMLR, 2023.

\bibitem{van2024tabular}
Boris van Breugel and Mihaela van~der Schaar.
\newblock Why tabular foundation models should be a research priority.
\newblock In {\em Forty-first International Conference on Machine Learning}, 2024.

\bibitem{van2001art}
David~A Van~Dyk and Xiao-Li Meng.
\newblock The art of data augmentation.
\newblock {\em Journal of Computational and Graphical Statistics}, 10(1):1--50, 2001.

\bibitem{watson2023adversarial}
David~S Watson, Kristin Blesch, Jan Kapar, and Marvin~N Wright.
\newblock Adversarial random forests for density estimation and generative modeling.
\newblock In {\em International Conference on Artificial Intelligence and Statistics}, pages 5357--5375. PMLR, 2023.

\bibitem{SGLD}
Max Welling and Yee~W Teh.
\newblock Bayesian learning via stochastic gradient langevin dynamics.
\newblock In {\em Proceedings of the 28th international conference on machine learning (ICML-11)}, pages 681--688. Citeseer, 2011.

\bibitem{xu2019modeling}
Lei Xu, Maria Skoularidou, Alfredo Cuesta-Infante, and Kalyan Veeramachaneni.
\newblock Modeling tabular data using conditional gan.
\newblock {\em Advances in neural information processing systems}, 32, 2019.

\bibitem{zhang2018towards}
Aiqing Zhang and Xiaodong Lin.
\newblock Towards secure and privacy-preserving data sharing in e-health systems via consortium blockchain.
\newblock {\em Journal of medical systems}, 42(8):140, 2018.

\bibitem{zhang2023mixed}
Hengrui Zhang, Jiani Zhang, Zhengyuan Shen, Balasubramaniam Srinivasan, Xiao Qin, Christos Faloutsos, Huzefa Rangwala, and George Karypis.
\newblock Mixed-type tabular data synthesis with score-based diffusion in latent space.
\newblock In {\em The Twelfth International Conference on Learning Representations}, 2023.

\bibitem{zhao2021ctab}
Zilong Zhao, Aditya Kunar, Robert Birke, and Lydia~Y Chen.
\newblock Ctab-gan: Effective table data synthesizing.
\newblock In {\em Asian Conference on Machine Learning}, pages 97--112. PMLR, 2021.

\bibitem{zheng2020privacy}
Xu~Zheng and Zhipeng Cai.
\newblock Privacy-preserved data sharing towards multiple parties in industrial iots.
\newblock {\em IEEE Journal on Selected Areas in Communications}, 38(5):968--979, 2020.

\end{thebibliography}
